\documentclass{article}

\PassOptionsToPackage{numbers, compress, sort}{natbib}
\usepackage[eandd,final]{neurips_2026}

\usepackage[utf8]{inputenc}
\usepackage[T1]{fontenc}
\usepackage[hypertexnames=false]{hyperref}
\hypersetup{citebordercolor={0.53 0.72 0.88}}
\usepackage{url}
\usepackage{booktabs}
\usepackage{amsfonts}
\usepackage{nicefrac}
\usepackage{microtype}
\usepackage{xcolor}
\usepackage[normalem]{ulem}
\usepackage{amsmath,amssymb,mathtools}
\usepackage{graphicx}
\usepackage{tabularx}
\usepackage{multirow}
\usepackage{array}
\newcolumntype{L}[1]{>{\raggedright\arraybackslash}p{#1}}
\newcolumntype{C}[1]{>{\centering\arraybackslash}p{#1}}
\newcolumntype{M}[1]{>{\raggedright\arraybackslash}m{#1}}
\newcolumntype{Y}{>{\raggedright\arraybackslash}X}
\newcolumntype{P}[1]{>{\raggedright\arraybackslash}p{#1}}
\usepackage{enumitem}
\usepackage{tikz}
\usetikzlibrary{arrows.meta,positioning,calc,fit}
\usepackage{pgfplots}
\usepackage{float}
\usepackage{wrapfig}
\pgfplotsset{compat=1.18}

\usepackage{fvextra}
\usepackage{capt-of}
\usepackage{needspace}
\usepackage{xcolor}

\title{DriveStreamBench: Evaluating User-Conditioned Watch-and-Notify in Streaming Driving Video}

\author{%
\textbf{Yi Wang\thanks{Equal contribution.} \quad Xin Zhao\footnotemark[1] \quad Tian Meng\footnotemark[1] \quad Rui Dai\thanks{Project lead.}} \\
\textbf{Chi Zhang \quad Cui Tang \quad Bingzheng Liu \quad Fucheng Wang} \\
\textbf{Haopeng Zhang \quad Ruidong Ding \quad Yang Li \quad Kaikui Liu \quad Xiangxiang Chu\thanks{Corresponding author.}} \\
AMAP, Alibaba Group
}

\date{}

\DefineVerbatimEnvironment{PromptBlock}{Verbatim}{
  fontsize=\footnotesize,
  baselinestretch=0.95,
  breaklines=true,
  breakanywhere=true,
  breaksymbolleft={},
  breaksymbolright={},
  frame=single,
  framerule=0.25pt,
  framesep=3pt
}

\providecommand{\DSBExampleFigDir}{figures/examples}

\newcommand{\alertf}{\(\mathrm{AlertScore}@2\mathrm{s}\)}

\begin{document}

\maketitle

\begin{abstract}

Vision-language models (VLMs) have made steady progress as in-vehicle driving assistants, yet existing evaluations largely remain offline, focusing on perception and short-term reasoning over pre-recorded driving scenes. A complementary capability is still underexplored: whether a model can continuously monitor a causal driving video stream and issue a calibrated alert when a user-specified condition is met. Existing driving benchmarks mainly assess reactive understanding, while recent streaming video benchmarks are developed largely outside the driving domain. To this end, we introduce DriveStreamBench, a driving and streaming video benchmark that organizes this capability spectrum into four cognitive levels constructed from a shared pool of driving events, to support level-wise capability profiling and gated evaluation. We evaluate various vision-language models along with DriveStream-SFT, a supervised reference baseline trained on the benchmark's training split, under a unified streaming protocol. 
DriveStreamBench tests a basic but necessary capability, and all evaluated models fall short of it; current vision-language models are not yet ready for user-conditioned alerting in driving.
Resources are available at \href{https://github.com/AMAP-ML/DriveStreamBench}{\nolinkurl{https://github.com/AMAP-ML/DriveStreamBench}} and \href{https://github.com/AMAP-ML/DriveStreamBench/releases}{\nolinkurl{https://github.com/AMAP-ML/DriveStreamBench/releases}}.

\end{abstract}

\section{Introduction}

Vision-language models (VLMs) are increasingly studied for multimodal tasks~\citep{zong2026l2dir,ding2026llatisa,yuan2026videostar,wang2026socioreasoner,xie2026qhawkeye,su2026videocoe,huang2026taminghallucinations}, including in-vehicle driving assistance~\citep{yuan2026autodriver}, but existing driving benchmarks such as LingoQA, DriveLM, and NuScenes-QA~\citep{lingoqa,drivelm,nuscenesqa} mostly evaluate reactive understanding over pre-recorded scenes.
A complementary capability remains under-evaluated: user-conditioned watch-and-notify, where a model observes a causal driving video stream and decides when, if at all, to alert for a specified condition while avoiding false alarms.

We introduce DriveStreamBench, a front-camera causal-video benchmark organized into four cognitive levels: L1 recognition, L2 ego-centric relation understanding, L3 temporal event comprehension, and L4 streaming watch-and-notify. 
L1--L3 are reactive QA tasks, while L4 is the primary alerting task. 
In each L4 session, the model receives video frames in temporal order, maintains a watch instruction such as \emph{Alert me when the lead vehicle starts moving}, and decides at each step whether to issue \texttt{ALERT} or remain at \texttt{NO REPLY}. 
Success therefore requires not only video understanding, but also calibrated alert timing and false-alert control under causal input.

DriveStreamBench differs from classical proactive video interaction~\citep{streamingbench,ovobench,proactivebench,omnimmi} in that the alert trigger is specified by the user and evaluated against event-aligned timing in a causal driving stream. 
Thus, it complements driving QA benchmarks, which do not measure watch-and-notify timing, and general streaming benchmarks, which are largely domain-agnostic.

DriveStreamBench draws from nuPlan~\citep{nuplan} and web-collected dashcam footage, covering five in-domain L4-Core event families and one held-out L4-OOD stress split. 
Beyond L4, L1/L2 serve as foundation-level controls, and paired L3/L4 sessions support temporal grounding analysis through L3-Gated \(\mathrm{AlertScore}@\tau\). 

For L4, we score each session under causal video time. 
A model receives positive credit only when its first alert is temporally aligned with the annotated reference time, and it is penalized for alerting on negative sessions. 
\(\mathrm{AlertScore}@\tau\) summarizes these requirements as the harmonic mean of \(\mathrm{Hit}@\tau\) and Negative Silence Rate (\(\mathrm{NSR}\)). 
Across evaluated systems, performance remains far below the human calibration reference on L4, even when L1/L2 foundation controls are strong, exposing the challenge of turning visual understanding into a calibrated streaming alert policy.

Our main contributions are summarized as follows:
\begin{itemize}[
    leftmargin=1.35em,
    labelsep=0.45em,
    itemsep=0.35em,
    topsep=0.35em,
    parsep=0pt,
    partopsep=0pt
]
\item \textbf{A driver-centric benchmark suite centered on watch-and-notify.}
We introduce DriveStreamBench, a front-camera benchmark suite for driver-centered video understanding. DriveStreamBench spans four cognitive levels: L1 (foundational recognition), L2 (ego-centric relation understanding), L3 (temporal event comprehension), and L4 (streaming alert decisions). The primary L4 task formalizes user-conditioned watch-and-notify in an online front-camera driving setting, where models observe video frames sequentially and decide whether to issue an alert or remain silent at each step.

\item \textbf{An online alert evaluation protocol with temporal grounding.}
We define a causal streaming protocol and \(\mathrm{AlertScore}@\tau\), which jointly measures event-aligned alerting and false-alert restraint. L1/L2 serve as foundational controls, while paired L3/L4 sessions support \emph{L3-Gated \(\mathrm{AlertScore}@\tau\)}, a secondary metric that conditions L4 alert credit on the correctness of the paired L3 temporal question and distinguishes event-grounded alerts from lucky or shortcut-driven alerts.

\item \textbf{A systematic model study and reproducible reference package.}
We evaluate diverse VLM families under a unified protocol and find that strong L1/L2 performance does not reliably translate to L3 temporal comprehension or L4 streaming alerting. We release DriveStream-SFT, annotations, metadata, evaluation code, and baseline training scripts.
\end{itemize}

\section{Related Work}

\paragraph{Driving-domain question answering and reasoning.}
Driving VQA and reasoning benchmarks evaluate VLM understanding of autonomous-driving scenes in largely reactive settings. 
LingoQA, DriveLM, and NuScenes-QA study visual grounding, graph-style reasoning, and multimodal QA~\citep{lingoqa,drivelm,nuscenesqa}; NuPlanQA and STRIDE-QA further use structured driving data for multi-view, ego-centric, spatial, and spatiotemporal reasoning~\citep{nuplanqa,strideqa}. 
Recent benchmarks extend this direction toward safety-critical, temporal, and first-person driving-video reasoning, including DVBench, STSBench, and ScenePilot-Bench~\citep{dvbench,stsbench,scenepilotbench}. 
These resources are valuable for scene understanding, but the model typically answers at a fixed frame, after observing a clip, or after the relevant evidence is available. 
DriveStreamBench instead evaluates whether a model can maintain a user-specified watch condition over causal video time and decide when, if at all, to alert.

\paragraph{Streaming and online video benchmarks.}
StreamingBench evaluates real-time, omni-source, and contextual streaming comprehension and reports a sizable gap between offline-trained MLLMs and human-level online video understanding~\citep{streamingbench}. OVO-Bench studies temporal awareness in online video understanding through backward tracing, real-time perception, and forward active responding~\citep{ovobench}. ProactiveVideoQA and OmniMMI focus on proactive or multimodal interaction settings where timing and turn-taking matter~\citep{proactivebench,omnimmi}. RTV-Bench evaluates continuous perception, understanding, and reasoning through multi-timestamp and hierarchical question structures in real-time video~\citep{rtvbench}. SVBench studies temporal multi-turn dialogues for streaming video understanding~\citep{svbench}. ODV-Bench is particularly relevant because it evaluates real-time streaming video understanding in autonomous-driving scenarios~\citep{streamforest}. Closely related in spirit, QEVD studies situated asynchronous interaction, where a model must decide what feedback to provide and when to provide it during an unfolding activity~\citep{qevd}. These benchmarks establish streaming evaluation as a research target, but they do not evaluate driver-centered, user-conditioned watch-and-notify with explicit event reference times and false-alert restraint.

\paragraph{Streaming video models.}
Recent models explore different routes to online video understanding. VideoLLM-online adapts VLMs to streaming through online training and inference~\citep{videollmonline}. Dispider disentangles perception, decision, and reaction for active real-time interaction~\citep{dispider}. StreamForest introduces persistent event memory for efficient online video understanding~\citep{streamforest}, while SimpleStream shows that a sliding recent-frame window can be a strong streaming baseline~\citep{simplestream}. MMDuet2 and M4 study proactive response timing and multimodal interaction~\citep{mmduet2,omnimmi}. These works contribute model architectures, memory mechanisms, or training strategies for streaming interaction; DriveStreamBench complements them with a driving-specific, user-conditioned alerting evaluation target, and we include representative systems as baselines in Section~\ref{sec:experiments}.

\paragraph{Positioning.}
DriveStreamBench is a driver-centered streaming-video benchmark for user-conditioned watch-and-notify. Unlike driving QA benchmarks, it evaluates whether models can monitor a causal driving stream for a user-specified condition and issue event-aligned alerts while satisfying no-alert restraint. Unlike general streaming and proactive-video benchmarks, it anchors the alert target to driving events and user-specified watch conditions. Beyond alert timing, DriveStreamBench supports level-wise capability profiling and gated alert evaluation: L1/L2 provide foundation-level controls, while paired L3/L4 sessions support L3-gated alert evaluation. 
A detailed feature comparison is provided in Appendix Table~\ref{tab:benchmark_comparison}.
We focus on instruction-conditioned VLM systems; fixed detector-rule cascades require event-specific detectors and thresholds, and therefore address a different, more closed alerting setting.

\section{DriveStreamBench: Task and Data}

\begin{figure}[t]
    \centering
    \includegraphics[width=\linewidth]{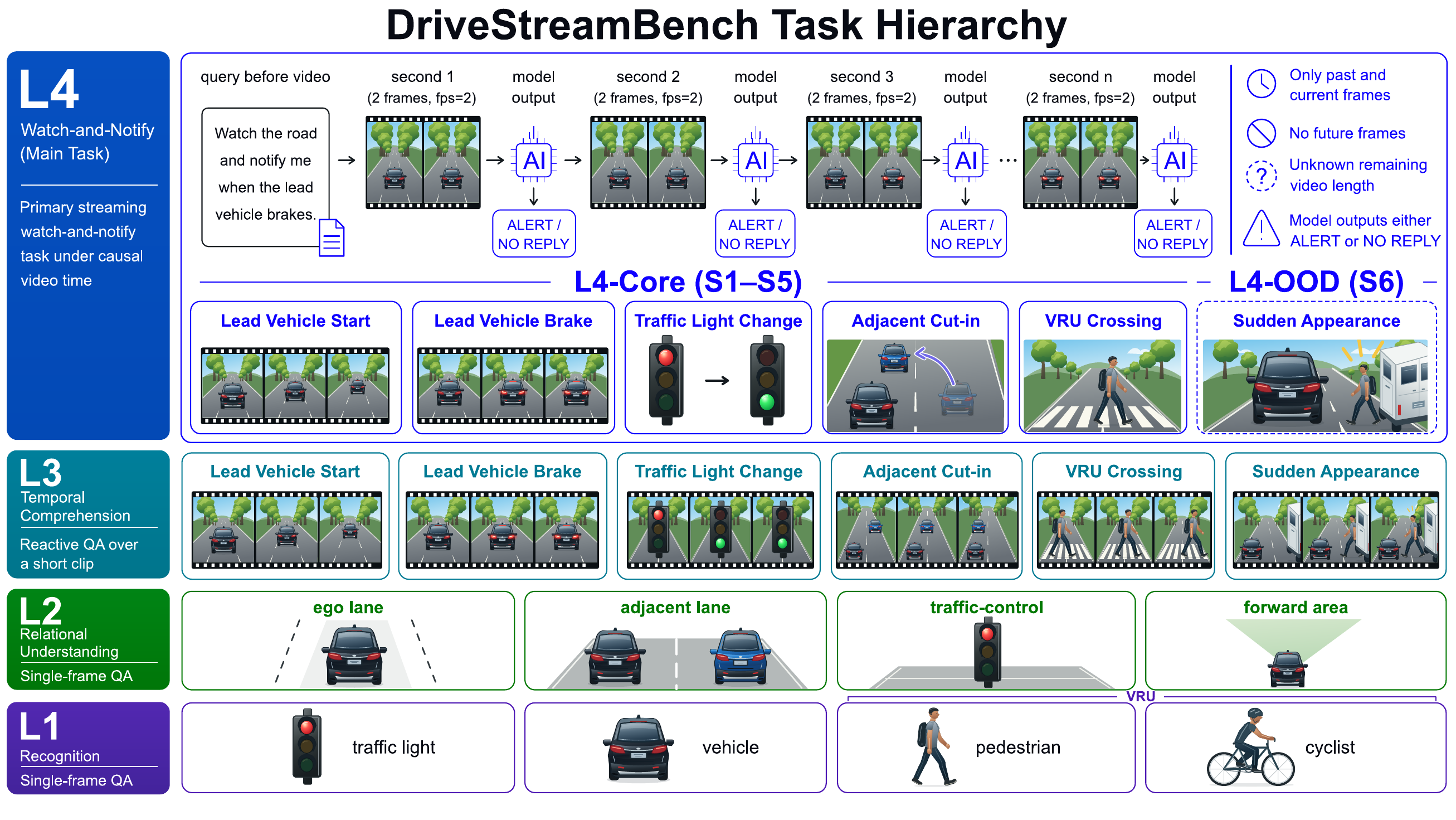}
    \caption{The four-level taxonomy of DriveStreamBench. L1--L3 are reactive QA tasks in binary or multiple-choice formats, while L4 is the primary streaming watch-and-notify task. L1/L2 serve as foundation-level probes, and paired L3/L4 sessions support temporal grounding analysis for alert decisions under causal video time.}
    \label{fig:taxonomy}
\end{figure}

\subsection{Four-Level Cognitive Taxonomy}
\label{sec:four_taxo}

Figure~\ref{fig:taxonomy} shows the four cognitive levels of DriveStreamBench. L1 and L2 are single-frame reactive QA levels: L1 evaluates basic recognition of driving-relevant scene elements, while L2 evaluates ego-centric spatial relations and traffic-control constraints. L3 evaluates short-term temporal understanding, such as whether a relevant state change has occurred, and L4 evaluates online watch-and-notify behavior. This L1 to L4 progression reflects increasing cognitive demand and is grounded in Bloom's revised taxonomy, Endsley's situation-awareness model, and signal-detection theory (Appendix~\ref{app:taxonomy-grounding}) \citep{anderson2001bloom,endsley1995sa,macmillan2004detection}.

The hierarchy supports level-wise capability profiling. L1 and L2 act as foundation-level controls: their aggregate performance indicates whether models can reliably perceive driving-relevant entities and relations in reactive settings. Paired L3/L4 sessions further enable a stricter check for alert decisions: an event-aligned L4 alert receives credit only if the model also answers the paired L3 temporal question correctly. We capture this with L3-Gated \(\mathrm{AlertScore}@\tau\), a secondary metric detailed in Appendix~\ref{app:l3g-results}. L4 is evaluated under a causal streaming protocol with a fixed decision cadence, allowing us to measure event-aligned alerting and false-alert restraint under a common online setting. We additionally report runtime latency to assess whether models can keep pace with the incoming stream.

\subsection{L4 Task Definition: Watch-and-Notify}
\label{sec:task_def}

The primary L4 task in DriveStreamBench is watch-and-notify. Each L4 example is a monitoring session over a front-facing driving video. Before the session begins, the user provides a watch instruction \(Q \in \mathcal{Q}\), such as \emph{Alert me when the lead vehicle starts moving}. The model then observes the video stream in decision turns and decides at each turn whether to issue an alert or remain silent.

Let \(X_i \in \mathcal{X}\) denote the visual observation revealed at turn \(i\), e.g., the frames sampled during the \(i\)-th one-second interval. DriveStreamBench clips are stored at 10~FPS, but the default L4 inference setting samples the observation stream at 2~FPS. Thus, \(X_i\) contains two frames from the \(i\)-th one-second interval, while the decision cadence remains 1~Hz. Let \(A_i \in \mathcal{A}\) denote the model response at turn \(i\). Let \(\mathcal{H}\) denote the space of finite interaction histories, with \(H_i = (X_0, A_0, \ldots, X_{i-1}, A_{i-1}) \in \mathcal{H}\) denoting the history before turn \(i\). Before producing \(A_i\), the model may access the watch instruction \(Q\), the previous history \(H_i\), and the current observation \(X_i\), but not future frames or the remaining video duration.
We formalize the model as a causal policy
\[
f_\theta : \mathcal{Q} \times \mathcal{H} \times \mathcal{X} \rightarrow \mathcal{A},
\qquad
A_i = f_\theta(Q, H_i, X_i).
\]

For L4 evaluation, the prompted valid response set is
\(
\mathcal{A}_{\mathrm{valid}} = \{\texttt{NO REPLY}, \texttt{ALERT}\}.
\)
\texttt{NO REPLY} means that the model continues monitoring without alerting, while \texttt{ALERT} means that the model emits an alert at the current video time. 
During evaluation, each model output is mapped to the common alerting interface by a fixed model-family adapter, as described in Section~\ref{sec:l4_protocol}. 
For instruction-following VLMs, outputs outside the literal \(\{\texttt{NO REPLY}, \texttt{ALERT}\}\) set are treated as \texttt{INVALID}. 
For third-party streaming transfer baselines, non-empty non-\texttt{NO REPLY} outputs are treated as alert actions to avoid requiring a literal \texttt{ALERT} token from models with heterogeneous native interfaces.

\subsection{L4 Subcategories and Alert Modes}
\label{sec:subcat}

Table~\ref{tab:l4-task-defs} summarizes the six L4 subcategories, their alert modes, operational definitions, and representative watch instructions. S1--S5 form L4-Core, the primary in-domain L4 benchmark. These categories cover lead-vehicle start, lead-vehicle brake, traffic-light change, adjacent cut-in, and vulnerable road user (VRU) crossing. Among them, S4 (adjacent cut-in) is a low-resource long-tail category: adjacent cut-in events are scarce in nuPlan, so its training split is substantially smaller than the other L4-Core categories. We therefore treat S4 as a low-resource L4-Core stress slice. S6 forms L4-OOD, a stress split for sudden-appearance alerting. It is out-of-distribution (OOD) in both alert semantics and video domain: the alert category is held out from training, and the source videos differ from the L4-Core distribution. We report L4-OOD separately from the primary L4-Core ranking.

Each subcategory has an alert mode determined by the temporal semantics of the event. In \emph{event-triggered} categories, the event has a discrete observable onset, so alerting before the event occurs is strictly incorrect. For example, if the instruction asks the model to alert when a traffic light changes, an alert before the observed state change is a false anticipation. In \emph{target-timing} categories, the relevant driving condition develops gradually, so a slightly early alert can still be appropriate. For example, when an adjacent vehicle is beginning to cut in, alerting shortly before the annotated reference alert time can still be acceptable.

\begin{table}[t]
    \centering
    \small
    \setlength{\tabcolsep}{3.2pt}
    \renewcommand{\arraystretch}{1.04}
    \caption{
    Alert modes and definitions for L4 subcategories.
    S1--S5 form L4-Core, the primary in-domain L4 benchmark.
    S6 forms L4-OOD, an optional stress split that combines a held-out sudden-appearance alert category with external dashcam video domain shift. User questions' details are provided in Appendix~\ref{app:l1l4-examples}.
    }
    \label{tab:l4-task-defs}
    \begin{tabularx}{\textwidth}{L{0.25\textwidth} L{0.15\textwidth} X}
        \toprule
        \textbf{Subcategory} & \textbf{Alert mode} & \textbf{Definition} \\
        \midrule

        S1: lead-vehicle start &
        Event-triggered &
        Alert when the lead vehicle in the ego lane starts moving from a stopped or waiting state.  \\
        \specialrule{0.025em}{0.28ex}{0.28ex}

        S2: lead-vehicle brake &
        Event-triggered &
        Alert when the lead vehicle in the ego lane brakes and its brake lights turn on. \\
        \specialrule{0.025em}{0.28ex}{0.28ex}

        S3: traffic-light change &
        Event-triggered &
        Alert when the driving-relevant traffic signal state changes. \\
        \specialrule{0.025em}{0.28ex}{0.28ex}

        S4: adjacent cut-in &
        Target-timing &
        Alert when a vehicle in an adjacent lane merges into or cuts into the ego lane.  \\
        \specialrule{0.025em}{0.28ex}{0.28ex}

        S5: VRU crossing &
        Target-timing &
        Alert when a pedestrian or cyclist enters or starts crossing the road ahead. \\
        \specialrule{0.025em}{0.28ex}{0.28ex}

        S6: sudden appearance &
        Event-triggered &
        Alert when a pedestrian, vehicle, or other relevant target suddenly appears from an occluded or hidden area.  \\

        \bottomrule
    \end{tabularx}
\end{table}

\subsection{Dataset Construction and Statistics}

DriveStreamBench is built primarily from nuPlan front-camera videos~\citep{nuplan}.
All L1/L2 instances and the L3/L4-Core subcategories S1--S5 are nuPlan-based.
S6 sudden appearance is collected from external dashcam videos and used only as a
separate L4-OOD stress split, as detailed in
Appendix~\ref{app:sudden_appearance_privacy}.

We construct the Core benchmark in an L4-first manner: candidate watch-and-notify
sessions are mined from structured driving signals and then finalized through
human verification and senior adjudication. For accepted positive L4 sessions,
the reference alert time \(t_{\mathrm{ref}}\) is annotated on 10~FPS clips.

For Core L3/L4, train/test splits are defined at the extracted-clip level within
each evaluated subcategory. Broader nuPlan logs, routes, scenes, and geographic regions may be shared across splits. Paired L3 questions inherit the split of their
corresponding L4 sessions. When L1/L2 examples are derived from the same source
sessions or clips, they follow the corresponding split assignment before
task-specific filtering. This preserves cross-level split consistency and avoids
direct train/test frame overlap within each evaluated subcategory. L4-Core is used as the primary in-distribution, category-conditioned setting for
streaming watch-and-notify. S6 is reported separately as an OOD stress split with
external dashcam videos and a held-out sudden-appearance alert category.

For L4-Core, each subcategory uses 25 training and 20 test natural-language
watch-instruction paraphrases, with no full-instruction overlap. L4 negatives are
category-matched non-events in which the watch instruction remains meaningful but
the target event does not occur. Their construction descriptors are used only for
auditing in Appendix~\ref{app:data_stats}, not as scoring labels. The
low-resource S4 adjacent-cut-in slice is retained as a long-tail stress category,
and we additionally report an S4-excluded macro as a sensitivity view.

\begin{table}[t]
    \centering
    \small
    \setlength{\tabcolsep}{3pt}
    \renewcommand{\arraystretch}{1.05}
    \caption{
    Summary of the DriveStreamBench data split used in this paper.
    The \textbf{Train} column denotes the released training pool, and the \textbf{Test} column denotes the held-out evaluation split.
    }
    \label{tab:data}
    \begin{tabular*}{\textwidth}{@{\extracolsep{\fill}} l r r  l r r  l r r @{}}
        \toprule
        \multicolumn{3}{c}{\textbf{L1 recognition}} &
        \multicolumn{3}{c}{\textbf{L2 relational understanding}} &
        \multicolumn{3}{c}{\textbf{Paired L3/L4 event levels}} \\
        \cmidrule(lr){1-3} \cmidrule(lr){4-6} \cmidrule(lr){7-9}
        \textbf{Subcategory} & \textbf{Train} & \textbf{Test} &
        \textbf{Subcategory} & \textbf{Train} & \textbf{Test} &
        \textbf{Subcategory} & \textbf{Train} & \textbf{Test} \\
        \midrule
        vehicle       & 2{,}527 & 510   & ego lane              & 1{,}806 & 364   & S1: lead-vehicle start    & 1{,}229 & 401   \\
        VRU           & 1{,}424 & 308   & adjacent lane         &    638  & 151   & S2: lead-vehicle brake    & 1{,}187 & 399   \\
        traffic light &    892  & 206   & traffic-light control & 1{,}096 & 194   & S3: traffic-light change  & 1{,}195 & 403   \\
                      &         &       & forward area          & 1{,}671 & 364   & S4: adjacent cut-in       &    208  & 341   \\
                      &         &       &                       &         &       & S5: VRU crossing          & 1{,}184 & 405   \\
                      &         &       &                       &         &       & \textbf{L3/L4-Core total} & \textbf{5{,}003} & \textbf{1{,}949} \\
                      &         &       &                       &         &       & S6: sudden appearance     &      -- & 399   \\
        \midrule
        \textbf{Total} & \textbf{4{,}843} & \textbf{1{,}024} &
        \textbf{Total} & \textbf{5{,}211} & \textbf{1{,}073} &
        \textbf{L3/L4 total} & \textbf{5{,}003} & \textbf{2{,}348} \\
        \bottomrule
    \end{tabular*}
\end{table}

Table~\ref{tab:data} summarizes the train/test composition across all four levels. The
L3/L4 block reports paired event subcategories because each accepted L4 session has a
corresponding L3 temporal question. L4-Core contains 5{,}003 training sessions and
1{,}949 test sessions, while L4-OOD contains 399 test sessions and is held out from
training. L1/L2 are summarized by question subcategory. Appendices~\ref{app:data_construction}
and~\ref{app:data_stats} provide the full construction pipeline, human verification
process, split-wise statistics, duration and reference-time analyses, and auxiliary
construction audits.

We provide a public GitHub release and a validated Croissant metadata file
with core and Responsible AI fields; Appendix~\ref{app:ethics_release} summarizes the
release boundary, licensing assumptions, privacy safeguards, limitations, and misuse risks.

\section{Evaluation Protocol and Metrics}

\subsection{Reactive Metrics on L1--L3}
\label{sec:l13-metric}

L1 and L2 are evaluated by standard accuracy, and outputs that fail to parse to a valid option are counted as incorrect. 
L3 is evaluated by standard binary F1-score for event occurrence rather than accuracy, since its labels are imbalanced. Details are provided in Appendix~\ref{app:protocol}.
We additionally report the L1--L3 Macro Score:
\(\mathrm{Macro}_{\mathrm{L1\text{-}L3}}=\left(\mathrm{Acc}_{\mathrm{L1}}+\mathrm{Acc}_{\mathrm{L2}}+\mathrm{F1}_{\mathrm{L3}}
\right)/3\),
which summarizes overall reactive QA performance across L1--L3 and is used in the model comparisons in Section~\ref{sec:bottleneck}.

\subsection{L4 Inference and Video-Time Scoring Protocol}
\label{sec:l4_protocol}

L4 sessions are evaluated as streaming chat interactions, as detailed in Appendix~\ref{app:prompt-l4}. DriveStreamBench clips are stored at 10~FPS and are uniformly downsampled to 2~FPS for the default L4 model input. The model makes one decision per second, so each decision turn receives the two downsampled frames within that one-second interval. 
At each decision turn, the model only receives frames whose timestamps are no later than the current decision time, so no future frames are leaked.
The assistant is instructed to respond at every turn with either \texttt{NO REPLY} or \texttt{ALERT}.  
We use fixed model-family adapters to map heterogeneous outputs to the common DriveStreamBench alerting interface. 
For instruction-following VLMs, including the Qwen3-VL ladder, DriveStream-SFT variants, and proprietary API VLMs, we use strict literal parsing: exact or normalized \texttt{NO REPLY} is mapped to \texttt{NO REPLY}, exact or normalized \texttt{ALERT} is mapped to \texttt{ALERT}, and other non-empty outputs are mapped to \texttt{INVALID}. 
For third-party streaming transfer baselines whose native output interfaces are not literal \texttt{ALERT}/\texttt{NO REPLY}, we use a more permissive action-level adapter: empty output, native silence, or normalized \texttt{NO REPLY} is mapped to \texttt{NO REPLY}, while any other non-empty output is mapped to \texttt{ALERT}. 
M4 keeps its native interface trigger adapter, which extracts the alert time from the model's native requirement-met string. 
The full per-baseline action mapping is fixed before evaluation and summarized in Table~\ref{tab:adapter-mapping}.

Evaluation follows a unified first-decision rule. 
For positive sessions, the first turn not mapped to \texttt{NO REPLY} determines the model's alert time \(t_{\mathrm{alert}}\). 
If this first non-silent decision is mapped to \texttt{INVALID}, the session receives no hit credit. 
If a positive session remains \texttt{NO REPLY} throughout, it is counted as a miss. 
For negative sessions, the session contributes to the Negative Silence Rate only if every turn is mapped to \texttt{NO REPLY}; any \texttt{ALERT} or \texttt{INVALID} breaks negative silence. 
This first-decision rule reflects the interruptive nature of alerting: once the model leaves silence, the user has already been interrupted, so later corrections do not erase the earlier alert decision.

The alert time \(t_{\mathrm{alert}}\) is the video time of the decision step, and reference times \(t_{\mathrm{ref}}\) are manually annotated on the released 10~FPS clips. The timing windows and hit definition used to determine whether \(t_{\mathrm{alert}}\) is aligned with \(t_{\mathrm{ref}}\) are detailed in Section~\ref{sec:metrics}. 
This protocol fixes the DriveStreamBench interaction interface while making native silence handling explicit. 
For streaming and proactive baselines with native silence mechanisms, the model-specific action adapters are fixed before evaluation and documented in Appendix~\ref{app:adapter-details}. 
These rows evaluate direct protocol transfer: whether existing streaming or proactive systems can maintain silence when no alert is needed and produce a timely non-silent alert action when the monitored event occurs.

\subsection{Alert Score for L4}
\label{sec:metrics}

We define \(\mathrm{AlertScore}@\tau\) on an arbitrary L4 evaluation slice, such as one subcategory. 
Let \(\mathcal{P}\) and \(\mathcal{N}\) denote the positive and negative sessions in the slice. 
Following the model-family action mapping and first-decision rule in Section~\ref{sec:l4_protocol}, each decision turn is mapped to one of \(\{\texttt{NO REPLY}, \texttt{ALERT}, \texttt{INVALID}\}\). 
For action-level streaming adapters, the \texttt{INVALID} class is not used because any non-empty non-\texttt{NO REPLY} output is treated as \texttt{ALERT}; for strict-literal VLM parsers, non-literal non-empty outputs are mapped to \texttt{INVALID}.

For each positive session \(i\in\mathcal{P}\), let \(t_{\mathrm{alert}}^{(i)}\) be the video time of the first turn not mapped to \texttt{NO REPLY}, if such a turn exists. 
Let \(z_{\mathrm{first}}^{(i)}\in\{\texttt{ALERT},\texttt{INVALID}\}\) denote the mapped action at this first non-silent turn. 
If the session remains \texttt{NO REPLY} throughout, then \(t_{\mathrm{alert}}^{(i)}\) and \(z_{\mathrm{first}}^{(i)}\) are undefined and the positive session receives no hit credit.

The correctness window \(W_\tau^{(i)}\) is determined by the alert mode:
\[
W_\tau^{(i)} =
\begin{cases}
[t_{\mathrm{ref}}^{(i)},\, t_{\mathrm{ref}}^{(i)}+\tau],
& \text{event-triggered},\\
[t_{\mathrm{ref}}^{(i)}-\tau/2,\, t_{\mathrm{ref}}^{(i)}+\tau/2],
& \text{target-timing}.
\end{cases}
\]
Event-triggered alerts use a one-sided post-event window because early alerts are factually incorrect. 
Target-timing alerts use a symmetric window because useful anticipation is allowed. 
Both windows have total width \(\tau\), so \(\mathrm{Hit}@\tau\) remains comparable across modes.
For each positive session \(i\in\mathcal{P}\), the per-session timely-hit indicator is \(\mathrm{hit}_i^\tau =
\mathbf{1}\!\left[
z_{\mathrm{first}}^{(i)}=\texttt{ALERT}
\,\wedge\,
t_{\mathrm{alert}}^{(i)}\in W_\tau^{(i)}
\right]\),
The indicator is defined as zero when no first non-silent turn exists. 
Thus a positive session receives hit credit only when the first non-silent action is \texttt{ALERT} and its video time falls inside the correct window.

The positive-side hit rate is \(\mathrm{Hit}@\tau=\frac{1}{|\mathcal{P}|}\sum_{i\in\mathcal{P}}\mathrm{hit}_i^\tau\).
For negative sessions, the Negative Silence Rate is \(\mathrm{NSR}=\frac{1}{|\mathcal{N}|}\sum_{j\in\mathcal{N}}\mathbf{1}\!\left[\text{every turn in session } j \text{ is mapped to } \texttt{NO REPLY}\right]\).
Any \texttt{ALERT} or \texttt{INVALID} action on a negative session breaks negative silence. 
Positive-session alerts do not enter \(\mathrm{NSR}\); early, late, missing, or invalid first decisions on positives are handled by \(\mathrm{Hit}@\tau\).

Finally, \(\mathrm{AlertScore}@\tau\) is the harmonic mean of event-aligned alerting and negative-session restraint:
\begin{equation}
\mathrm{AlertScore}@\tau =
\begin{cases}
2\,\mathrm{Hit}@\tau\cdot\mathrm{NSR}
/(\mathrm{Hit}@\tau+\mathrm{NSR}),
& \mathrm{Hit}@\tau+\mathrm{NSR}>0,\\[1.0em]
0,
& \mathrm{Hit}@\tau+\mathrm{NSR}=0.
\end{cases}
\end{equation}
\(\mathrm{AlertScore}@\tau\) is task-specific: a model must both alert at the event-aligned video time on positives and remain mapped to \texttt{NO REPLY} throughout negative sessions. 
We also report \(\mathrm{Hit}@\tau\) and \(\mathrm{NSR}\) separately to show whether gains come from better timing, better restraint, or both. Unless otherwise stated, the quantities above are computed per L4 subcategory. 
The primary ranking metric is \emph{Core Macro}, \(\mathrm{CoreMacro}@2\mathrm{s}\), obtained by averaging \(\mathrm{AlertScore}@2\mathrm{s}\) over the five L4-Core subcategories S1--S5. 
We report AlertScore at
\(\tau\in\{1,2,3,5\}\,\mathrm{s}\), using \(\tau=2\,\mathrm{s}\) for the main
leaderboard and the remaining tolerances for timing-sensitivity analysis.
The L4-OOD stress split S6 is reported separately in Appendix~\ref{app:s6}.

We additionally report L3-Gated \(\mathrm{AlertScore}@\tau\) as a stricter secondary metric. 
It gives positive-side L4 hit credit only when the paired L3 temporal question is also answered correctly, while keeping the negative-side \(\mathrm{NSR}\) term unchanged. 
The full definition and results are provided in Appendix~\ref{app:l3g-results}.

\section{Experiments}
\label{sec:experiments}

\subsection{Experimental Setup}

We evaluate four groups of systems on the full L4-Core benchmark: an open-weight Qwen3-VL Instruct scale ladder~\citep{qwen3vl}, general streaming video models~\citep{streamforest,simplestream,dispider,videollmonline}, proactive-trained streaming models~\citep{mmduet2,omnimmi}, and our supervised reference baseline DriveStream-SFT. All full-test models are evaluated under the L4 inference and video-time scoring protocol in Section~\ref{sec:l4_protocol}. The primary ranking metric is \emph{Core Macro}, \(\mathrm{CoreMacro}@2\mathrm{s}\). S6 is held out as an L4-OOD stress split and reported separately in Appendix~\ref{app:s6}. Detailed L1--L3 results, latency measurements, and frame-rate sensitivity analyses are reported in Appendix Tables~\ref{tab:l13-main} and~\ref{tab:latency}, and Appendix~\ref{app:fps-ablation}.

Proprietary API VLMs (GPT-5.4~\citep{openai_gpt54}, Gemini 3.1 Pro~\citep{gemini31pro}) are evaluated separately on a short-version protocol due to API image/context limits and evaluation cost. These results are reported in Appendix~\ref{app:short_version} and are not used for the primary full-test ranking. We do not include proprietary API VLMs in wall-clock latency comparisons because measured API latency depends on network conditions, provider-side scheduling, and serving load in addition to model computation.

\subsection{Main Results}
\label{sec:gap}

\begin{table}[t]
    \centering
    \small
    \setlength{\tabcolsep}{3pt}
    \renewcommand{\arraystretch}{0.95}
    \caption{
    Main L4 leaderboard.
    \(\mathrm{CoreMacro}@2\mathrm{s}\) is the average of \(\mathrm{AlertScore}@2\mathrm{s}\) over S1--S5; S4 is a low-resource long-tail slice and an S4-excluded ID Macro is reported in Appendix~\ref{app:s4-excluded}.
    Qwen3-VL, DriveStream-SFT, proprietary API VLMs, and human inputs use strict literal parsing; third-party streaming transfer baselines use the action-level adapters in Table~\ref{tab:adapter-mapping}. In each metric column, \textbf{bold} marks the best result and \underline{underlining} marks the second best. The human row is the primary 2-FPS calibration; its Hit and NSR components are not reported. Appendix~\ref{app:human_baseline} retains the original 10-FPS reference.
    }
    \label{tab:leaderboard}
    \begin{tabular*}{\textwidth}{@{\extracolsep{\fill}}lccccccccc@{}}
        \toprule
        & & \multicolumn{5}{c}{\(\mathrm{AlertScore}@2\mathrm{s}\) on L4-Core (\%)} & \multicolumn{2}{c}{Components (\%)} & \\
        \cmidrule(lr){3-7} \cmidrule(lr){8-9}
        Model & fps & S1 & S2 & S3 & S4 & S5 & \(\mathrm{Hit}@2\mathrm{s}\) & \(\mathrm{NSR}\) & \textbf{Core Macro (\%)} \\
        \midrule
        \multicolumn{10}{@{}l}{\textit{Open-source VLMs, zero-shot; strict literal parser}} \\
        Qwen3-VL-2B~\citep{qwen3vl}  & 2 &  0.0 &  0.0 &  3.8 &  0.0 & 21.8 &  2.8 & 98.7 &  5.1 \\
        Qwen3-VL-4B~\citep{qwen3vl}  & 2 & 10.7 &  0.0 & 38.3 &  2.4 & 61.7 & 20.1 & 63.6 & 22.6 \\
        Qwen3-VL-8B~\citep{qwen3vl}  & 2 & 23.0 & 17.2 & 52.2 &  4.8 & 54.3 & 29.4 & 47.9 & 30.3 \\
        Qwen3-VL-32B~\citep{qwen3vl} & 2 & 36.0 & 20.1 & 77.6 &  8.9 & 65.9 & 36.3 & 63.1 & 41.7 \\
        \midrule
        \multicolumn{10}{@{}l}{\textit{General streaming transfer baselines; action-level adapter}} \\
        StreamForest~\citep{streamforest}       & 1 &  0.0 &  0.0 &  0.0 &  0.0 &  0.0 &  0.0 & \textbf{100.0} &  0.0 \\
        SimpleStream~\citep{simplestream}       & 1 &  9.3 &  0.0 & 55.7 &  0.0 & 24.6 & 11.8 &  92.4 & 17.9 \\
        Dispider~\citep{dispider}               & 1 &  1.8 &  2.0 &  9.8 & 15.2 & 20.7 &  5.8 &  37.9 &  9.9 \\
        VideoLLM-online~\citep{videollmonline}  & 2 &  0.0 &  0.0 &  0.0 &  0.0 &  0.0 &  5.3 &   0.0 &  0.0 \\
        \midrule
        \multicolumn{10}{@{}l}{\textit{Proactive-trained streaming transfer baselines; action-level or native adapter}} \\
        MMDuet2~\citep{mmduet2}       & 1 &  0.0 &  0.0 & 26.9 &  0.0 & 26.2 &  6.2 &  97.2 & 10.6 \\
        M4~\citep{omnimmi}  & 1 & 13.6 &  0.0 &  1.9 &  9.3 &  4.0 &  3.2 &  69.9 &  5.8 \\
        \midrule
        \multicolumn{10}{@{}l}{\textit{Reference baselines trained on DriveStreamBench; strict literal parser}} \\
        DriveStream-SFT-L4 & 2 & \textbf{91.7} & 50.5 & 86.7 &  0.0 & 81.6 & \underline{53.3} & \underline{98.6} & 62.1 \\
        DriveStream-SFT    & 2 & \underline{91.4} & \underline{67.4} & \underline{88.6} & \underline{31.0} & \underline{83.2} & \textbf{62.0} & 96.5 & \underline{72.3} \\
        \midrule
        Human & 2 & 85.9 & \textbf{92.1} & \textbf{98.0} & \textbf{94.4} & \textbf{85.7} & -- & -- & \textbf{91.2} \\
        \bottomrule
    \end{tabular*}
\end{table}

Table~\ref{tab:leaderboard} reports the full-test L4-Core leaderboard. 
The leaderboard shows that watch-and-notify remains difficult under the L4 protocol. 
Scaling the Qwen3-VL ladder improves Core Macro from 5.1\% to 41.7\%, but the best zero-shot model remains far below the human calibration row.
The streaming and proactive transfer rows should be read as direct protocol-transfer baselines. 
To avoid penalizing these models merely for not emitting the literal \texttt{ALERT} token, we evaluate the third-party streaming transfer baselines with fixed action-level adapters from their native silence or trigger interfaces to the common alert/no-alert interface (Appendix~\ref{app:adapter-details}). 
This is permissive: any non-empty non-\texttt{NO REPLY} output is treated as an alert action for these baselines. 
Even under this relaxed transfer interface, the strongest third-party streaming/proactive transfer baseline reaches only 17.9\% Core Macro, far below DriveStream-SFT at 72.3\%. 
Among baselines whose scores are materially changed by the action-level adapter, MMDuet2 reaches 10.6\% and Dispider reaches 9.9\%. 
The component metrics reveal why this recovered signal is still not a calibrated alerting policy: Dispider recovers some positive-side hits under the action adapter, but its \(\mathrm{NSR}\) drops to 37.9\%, while VideoLLM-online obtains nonzero Hit@2 but collapses to 0.0\% \(\mathrm{NSR}\). 
Thus, free-form event reports do not by themselves form a calibrated watch-and-notify policy; Appendix~\ref{app:streaming-failure-cases} shows turn-level examples of descriptive substitution, first-turn acknowledgements, and native-trigger timing mismatch.

DriveStream-SFT serves as a benchmark-trained, in-distribution supervised reference model trained on the benchmark training split. 
Starting from Qwen3-VL-8B, L4-only supervision raises Core Macro from 30.3\% to 62.1\%. 
Under the current extracted-clip, in-distribution L4-Core protocol, adding L1--L3 multi-task supervision improves benchmark-aligned alert policy performance from 62.1\% to 72.3\%. 
This result should be interpreted as benchmark-aligned policy learning under the current extracted-clip L4-Core protocol, rather than as evidence of log-level, route-level, or geography-level generalization. 
DriveStream-SFT is an in-domain supervised reference for the seen L4-Core alert families. 
Leave-one-out results in Appendix~\ref{app:leaveoneout-ablation} show weak transfer to held-out alert families, indicating that direct event-family supervision remains important under the Qwen3-VL-8B L4-only recipe. 
A notable residual weakness remains on S4 adjacent cut-in, the low-resource L4-Core stress slice analyzed in Appendix~\ref{app:lowres-s4}. 
The full training-signal ablation is reported in Appendix Table~\ref{tab:training}. 
The strongest timing-only reference in Appendix~\ref{app:temporal-prior} reaches only 9.9\% Core Macro, far below DriveStream-SFT at 72.3\%, suggesting that the leaderboard is not explained by simple non-visual alert-position priors. 
Additional robustness checks show that the main ranking is not driven by the first-decision rule, strict literal compliance, or the low-resource S4 slice (Appendix~\ref{app:robustness-sensitivity}). 
These analyses address protocol-level and slice-weighting artifacts within L4-Core, but do not replace future log-disjoint or geography-disjoint evaluations.
In particular, excluding S4 from the macro increases DriveStream-SFT from 72.3\% to 82.7\% and Qwen3-VL-32B from 41.7\% to 49.9\%, so the reference baseline's lead is not caused by assigning one fifth of Core Macro to the low-resource S4 slice.
Stratified bootstrap 95\% confidence intervals for Core Macro are provided in Appendix~\ref{app:ci}. 
Frame-rate sensitivity is reported in Appendix~\ref{app:fps-ablation}. 
The short-version API evaluation in Appendix~\ref{app:short_version} shows the same pattern: GPT-5.4 improves over zero-shot open-weight VLMs but still trails DriveStream-SFT, while Gemini 3.1 Pro is limited by weak no-alert restraint.

The human row is included as a calibration reference for task feasibility and remaining headroom, not as a rank-comparable model baseline. 
On independently sampled test subsets, human Core Macro is 91.2 at both 1 and 2 FPS, compared with 92.8 for the original 10-FPS reference. The 2-FPS condition matches the default model input rate and serves as the primary human calibration. Human responses are continuous, while model decisions follow a 1-Hz cadence. Appendix~\ref{app:human_baseline} reports the sampling protocol and per-subcategory results.
We therefore use the human result to contextualize the scale of the remaining gap rather than to claim protocol-matched human--model comparison.

\subsection{Capability Profiling Beyond the Main Leaderboard}
\label{sec:bottleneck}

Table~\ref{tab:bottleneck} reports level-wise performance for the Qwen3-VL ladder and DriveStream-SFT. L1 and L2 are measured by accuracy, L3 is measured by F1-score, and L4 is measured by \(\mathrm{CoreMacro}@2\mathrm{s}\). The Qwen3-VL ladder performs strongly on L1 and improves on L2 with scale, but L3 F1 remains much lower and L4 \(\mathrm{CoreMacro}@2\mathrm{s}\) drops further. 
We interpret this pattern as a profiling signal. High L1/L2 scores show that single-frame recognition and relation probes do not automatically translate into temporal event understanding or calibrated alert decisions, but they do not rule out stream-level tracking, relevance-maintenance, or attention-allocation errors.

\begin{wraptable}[9]{r}{0.5\textwidth}
    \vspace{-0.7\baselineskip}
    \centering
    \small
    \setlength{\abovecaptionskip}{0pt}
    \setlength{\belowcaptionskip}{3pt}
    \setlength{\tabcolsep}{2.2pt}
    \renewcommand{\arraystretch}{0.92}
    \caption{
    Level-wise results. 
    }
    \label{tab:bottleneck}
    \begin{tabular}{l c c c c}
        \toprule
        Model & L1 & L2 & L3 & L4 \\
        \midrule
        Qwen3-VL-2B   & 92.5 & 77.4 & 29.3 &  5.1 \\
        Qwen3-VL-4B   & 92.4 & 82.5 & 46.9 & 22.6 \\
        Qwen3-VL-8B   & \underline{93.1} & 84.4 & 43.5 & 30.3 \\
        Qwen3-VL-32B  & 91.2 & \underline{88.1} & \underline{57.6} & \underline{41.7} \\
        DriveStream-SFT & \textbf{96.6} & \textbf{94.1} & \textbf{90.0} & \textbf{72.3} \\
        \bottomrule
    \end{tabular}
    \vspace{0.15em}
\end{wraptable}

Appendix analyses further qualify the L4 results along two axes. 
First, L3-Gated AlertScore (Appendix~\ref{app:l3g-results}) checks whether an in-window L4 alert is also supported by the paired L3 temporal question. On the Qwen3-VL ladder, this gate reduces \(\mathrm{AlertScore}@2\mathrm{s}\) by 0.4--6.7 points, corresponding to a 7.8--22.1\% relative gap. In contrast, the final multi-task DriveStream-SFT row drops by only 1.4 points, with the smallest gap ratio of 1.9\%. This suggests that its high L4 performance is accompanied by much stronger consistency with the paired L3 temporal probe.
Second, the S4 breakdown detailed in Appendix~\ref{app:lowres-s4} shows a genuine L4 failure mode: adjacent cut-in remains weak even after L4 supervision because 
it combines limited positive supervision with a gradual target-timing boundary.
Additional leave-one-out experiments are detailed in Appendix~\ref{app:leaveoneout-ablation}. 
To separate positive-side timing from negative-session restraint, Appendix~\ref{app:hit-only} reports a hit-only analysis under the same model-family adapters used in the main leaderboard. 
The results show that the action-level transfer adapter recovers limited positive-side signal for several third-party streaming baselines, but their CoreHit@2 remains far below DriveStream-SFT. 
Together with the main-table \(\mathrm{NSR}\) values, this indicates that their low AlertScore is not merely a consequence of the negative-restraint term: they still lack a calibrated combination of timely positive alerts and silence on negatives.

\paragraph{Scope and limitations.} DriveStreamBench studies front-camera watch-and-notify for a non-actuating driver assistant, with six event families and a default 2-FPS/1-Hz protocol. Explicit localization, planning, vehicle control, and driver workload are outside the evaluation scope. Broader event coverage and finer-grained recognition are directions for extension.

\section{Conclusion}

DriveStreamBench evaluates user-conditioned watch-and-notify in causal driving video. 
It combines level-wise capability profiling and gated alert evaluation with a streaming L4 protocol in which models must issue event-aligned alerts on positives and remain silent on negatives.
Across open-source VLMs, streaming video models, proactive-trained models, and supervised reference baselines, 
the results show a large gap between evaluated VLMs and the human calibration reference under our benchmark's protocol.
Benchmark-aligned supervision improves the ALERT/NO REPLY policy, under the current in-distribution, category-conditioned L4-Core split, but substantial headroom remains, especially for temporal grounding and low-resource event categories.

\begin{ack}
This work was supported by Alibaba Group. The authors declare no additional competing interests.
\end{ack}

\clearpage

\bibliographystyle{plainnat}
\bibliography{references}

\appendix

\clearpage

\section{Additional Related Work Comparison}
\label{app:related-comparison}

\begin{table}[H]
    \centering
    \small
    \setlength{\tabcolsep}{3.2pt}
    \renewcommand{\arraystretch}{0.98}
    \caption{
    Feature comparison with representative driving and streaming-video benchmarks.
    Symbols indicate support: \(\checkmark\) = supported, \(\triangle\) = partial or not primary, and \(\times\) = not supported.
    }
    \label{tab:benchmark_comparison}
    \begin{tabular*}{\textwidth}{@{\extracolsep{\fill}}lcccccc@{}}
        \toprule
        \textbf{Benchmark} &
        \textbf{Driving} &
        \textbf{Streaming} &
        \textbf{\mbox{User-cond.}} &
        \textbf{\mbox{Alert timing}} &
        \textbf{\mbox{No-alert}} &
        \textbf{\mbox{Level-wise}} \\
        \midrule

        LingoQA~\citep{lingoqa} &
        \(\checkmark\) & \(\times\) & \(\triangle\) & \(\times\) & \(\times\) & \(\times\) \\

        DriveLM~\citep{drivelm} &
        \(\checkmark\) & \(\times\) & \(\triangle\) & \(\times\) & \(\times\) & \(\triangle\) \\

        NuScenes-QA~\citep{nuscenesqa} &
        \(\checkmark\) & \(\times\) & \(\triangle\) & \(\times\) & \(\times\) & \(\times\) \\

        NuPlanQA~\citep{nuplanqa} &
        \(\checkmark\) & \(\times\) & \(\triangle\) & \(\times\) & \(\times\) & \(\triangle\) \\

        DVBench~\citep{dvbench} &
        \(\checkmark\) & \(\times\) & \(\triangle\) & \(\times\) & \(\times\) & \(\triangle\) \\

        STSBench~\citep{stsbench} &
        \(\checkmark\) & \(\times\) & \(\triangle\) & \(\times\) & \(\times\) & \(\triangle\) \\

        StreamingBench~\citep{streamingbench} &
        \(\times\) & \(\checkmark\) & \(\triangle\) & \(\triangle\) & \(\times\) & \(\times\) \\

        OVO-Bench~\citep{ovobench} &
        \(\times\) & \(\checkmark\) & \(\triangle\) & \(\triangle\) & \(\times\) & \(\times\) \\

        ODV-Bench~\citep{streamforest} &
        \(\checkmark\) & \(\checkmark\) & \(\triangle\) & \(\times\) & \(\times\) & \(\triangle\) \\

        ProactiveVideoQA~\citep{proactivebench} &
        \(\times\) & \(\checkmark\) & \(\checkmark\) & \(\checkmark\) & \(\triangle\) & \(\times\) \\

        OmniMMI~\citep{omnimmi} &
        \(\times\) & \(\checkmark\) & \(\triangle\) & \(\checkmark\) & \(\checkmark\) & \(\times\) \\

        QEVD~\citep{qevd} &
        \(\times\) & \(\checkmark\) & \(\checkmark\) & \(\checkmark\) & \(\triangle\) & \(\times\) \\

        \midrule

        \textbf{DriveStreamBench (ours)} &
        \(\checkmark\) & \(\checkmark\) & \(\checkmark\) & \(\checkmark\) & \(\checkmark\) & \(\checkmark\) \\

        \bottomrule
    \end{tabular*}
\end{table}

\section{Dataset construction details}
\label{app:data_construction}
\label{app:data} %

We first provide additional details on the construction, annotation, split design, and statistical auditing of DriveStreamBench. Section~\ref{app:data_construction} describes the L4-anchored construction pipeline, candidate mining and negative construction, L4 annotation and senior-final reference-time assignment, reactive QA construction for L1--L3, and split consistency. Section~\ref{app:data_stats} then reports split-wise dataset statistics, duration and reference-time audits, and auxiliary negative construction descriptor audits.

\subsection{L4-anchored construction overview}
\label{app:construction_overview}

DriveStreamBench is constructed as an L4-anchored foundation-to-alert evaluation. We first mine candidate L4 watch-and-notify sessions, including positive event candidates and category-conditioned negative candidates. Positive candidates are proposed from event-specific structured cues, while negative candidates are plausible non-event sessions in which the watch instruction remains meaningful but the target event does not occur. All temporal annotation is performed on videos sampled at 10 FPS, so annotators inspect ten sequential frames per second of session time when judging event occurrence and timing.

Each L4 candidate is independently reviewed by three annotators, who provide candidate-level valid/invalid decisions according to the corresponding positive- or negative-candidate validity criteria. For mined positive candidates marked valid, annotators additionally provide a candidate reference-time anchor \(t_{\mathrm{ref}}^{(a)}\), while mined negative candidates receive no \(t_{\mathrm{ref}}^{(a)}\) because no alert is warranted. A senior annotator then reviews the video, the three valid/invalid decisions, and candidate timestamps when available, and assigns the final accept/reject decision, the final positive/negative label, and, for accepted positives, the senior-final \(t_{\mathrm{ref}}\). Accepted positives evaluate event-aligned alert timing, accepted negatives evaluate false-alert restraint, and together they define the primary L4 streaming task.

\begin{figure}[t]
    \centering
    \includegraphics[width=\linewidth]{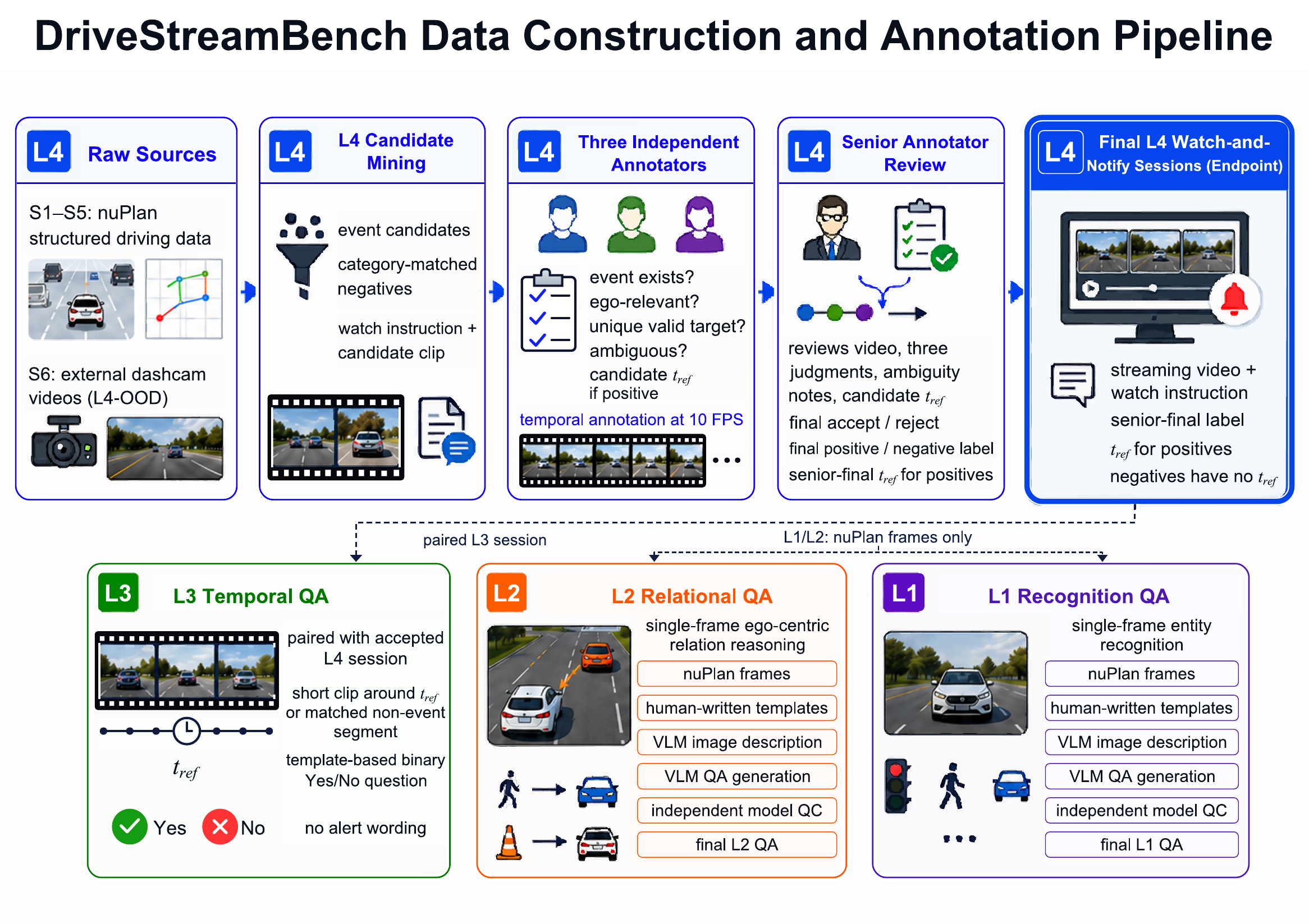}
    \caption{Dataset construction pipeline. DriveStreamBench is built in an L4-anchored manner: L4 candidate sessions are first mined from structured driving data and external dashcam videos, then reviewed by three independent annotators and finalized by a senior annotator. Accepted L4 sessions define the primary streaming task; paired L3 questions provide temporal grounding, while L1/L2 serve as foundation-level controls.}
    \label{fig:app_construction_pipeline}
\end{figure}

Each accepted L4 session is paired with one canonical L3 temporal question over the same monitored event, while L1 and L2 are constructed as foundation-level controls for recognition and ego-centric relation reasoning rather than as per-session gates for every L4 sample. Figure~\ref{fig:app_construction_pipeline} summarizes the end-to-end construction and verification pipeline.

\subsection{Candidate mining and negative construction}
\label{app:candidate_mining}

For S1--S5, candidate sessions are mined from nuPlan using structured driving signals, including object boxes, object trajectories, ego pose, map and lane geometry, and traffic-light states. These signals are used only to propose candidate sessions; final L4 labels and, for accepted positives, senior-final \(t_{\mathrm{ref}}\) values are determined by human review and senior adjudication. S6 sudden appearance is constructed from external dashcam videos, held out from training, and reported separately as an L4-OOD stress split rather than as part of the primary L4-Core ranking. Table~\ref{tab:l4_candidate_mining} summarizes the category-wise proposal-mining heuristics and representative negative-construction patterns. 

\begin{table}[t]
\centering
\small
\setlength{\tabcolsep}{4pt}
\renewcommand{\arraystretch}{1.15}
\caption{
Category-wise candidate mining and negative construction strategy for L4.
Structured signals are used only to propose candidate sessions; final labels and, for accepted positives,
senior-final \(t_{\mathrm{ref}}\) are assigned by human review and adjudication.
}
\label{tab:l4_candidate_mining}
\begin{tabularx}{\textwidth}{L{0.13\textwidth} p{0.22\textwidth} X X}
\toprule
\textbf{Subcategory} &
\textbf{Signals for proposal mining} &
\textbf{Positive-candidate proposal heuristic} &
\textbf{Representative negative construction patterns} \\
\midrule

S1: lead-vehicle start &
Lead-vehicle track in ego lane; object trajectory; speed profile &
Search for lead-vehicle tracks in the ego lane whose speed changes from near zero to a clearly moving state. &
No lead vehicle; stationary lead vehicle; low-speed creeping or slight motion that does not satisfy the start definition. \\

S2: lead-vehicle brake &
Lead-vehicle track in ego lane; acceleration profile &
Search for intervals in which the lead vehicle has sustained negative acceleration. &
Lead vehicle maintains speed; only slight slowing without sustained braking; motion that is not clearly ego-relevant. \\

S3: traffic-light change &
Traffic-light state; ego pose; lane and map geometry; ego-relevance matching &
Search for ego-relevant traffic-light state transitions. &
No visible traffic light; stable ego-relevant light; non-ego-relevant light changes state. \\

S4: adjacent cut-in &
Adjacent-lane vehicle trajectory; lane geometry; ego pose &
Search for adjacent-vehicle trajectories that approach the ego lane, cross the adjacent/ego-lane boundary, or intersect the ego-vehicle path. &
Adjacent-lane traffic without a completed cut-in; nearby vehicle approaches the lane boundary but does not enter the ego lane. \\

S5: VRU crossing &
VRU trajectory; ego trajectory; forward conflict region &
Search for pedestrian or cyclist trajectories that intersect the ego path or its forward conflict region. &
Pedestrian or cyclist visible near the road but not entering the ego-relevant path; roadside motion without a valid crossing event. \\

S6: sudden appearance &
External dashcam video review; occlusion context; target emergence from hidden area &
Constructed from external dashcam videos by collecting cases in which a pedestrian, vehicle, or other relevant target suddenly appears from an occluded or hidden area. &
Occlusions or roadside targets are present, but the target is already visible, not ego-relevant, or does not suddenly emerge from a hidden area. \\

\bottomrule
\end{tabularx}
\end{table}

Negative sessions are category-conditioned plausible non-events rather than generic idle clips. Their purpose is to support false-alert analysis under watch instructions that remain semantically plausible but unsatisfied. During construction, we seek coverage over three coarse patterns: simple absence cases, structured non-events where the relevant entity is present but no valid trigger occurs, and trigger-like confounders where visual or temporal cues resemble the target event but fail the operational event definition.

These patterns are used only to guide construction and auditing. They are not adjudicated benchmark labels, training targets, leaderboard strata, or scoring fields. The official L4 evaluation uses only the senior-final positive/negative label, the senior-final \(t_{\mathrm{ref}}\) for positives, and the model's first alert decision under the streaming protocol.

\subsection{L4 annotation and senior-final reference time}
\label{app:l4_verification}

Each L4 candidate session is independently annotated by three annotators on the 10 FPS clips. The annotation interface allows pausing, rewinding, and frame stepping, so annotators can inspect event occurrence and timing at frame-level granularity. This construction-time interface is used only for benchmark annotation and is distinct from the causal L4 model-evaluation protocol.

For each candidate, annotators provide a candidate-level valid/invalid decision. For a mined positive candidate, valid means that the instructed event occurs, is ego-relevant, has a unique valid target event, and is sufficiently visible under the watch instruction. For a mined negative candidate, valid means that the watch instruction remains meaningful for the clip but the target event does not occur. Invalid candidates include ambiguous cases, unclear ego relevance, multiple plausible target events, insufficient visibility, or mismatches with the intended event definition.

For mined positive candidates marked valid, annotators additionally provide a candidate reference-time anchor \(t_{\mathrm{ref}}^{(a)}\). For event-triggered categories, this anchor corresponds to the observable onset or state change that makes the event valid. For target-timing categories, it corresponds to the nominal point at which the developing condition becomes sufficiently clear and ego-relevant for the alert instruction. Mined negative candidates receive no \(t_{\mathrm{ref}}^{(a)}\), because no alert is warranted.

After the three independent annotations are collected, a senior annotator reviews each candidate session, the three valid/invalid decisions, and candidate \(t_{\mathrm{ref}}^{(a)}\) values when available. The senior annotator assigns the final accept/reject decision, the final positive/negative label, and, for accepted positives, the senior-final \(t_{\mathrm{ref}}\). Accepted negatives receive a senior-final negative label and have no \(t_{\mathrm{ref}}\). The released L4 scoring metadata contains only the senior-final label for each accepted session and, for positives, the senior-final \(t_{\mathrm{ref}}\).

\subsection{Reactive QA construction and split consistency}
\label{app:l13_construction_splits}

Each accepted L4 session is paired with one canonical L3 temporal question over the same monitored event. L3 is a reactive temporal QA task rather than a streaming alerting task: it asks whether the monitored event occurred in a short clip, without using alert or notification wording. L3 questions are generated from subcategory-specific templates, and the answer space is binary,
\[
\mathcal{Y}_{\mathrm{L3}}=\{\texttt{Yes}, \texttt{No}\}.
\]
For positive L4 sessions, the L3 clip is sampled around the senior-final \(t_{\mathrm{ref}}\): the clip starts \(3\)--\(6\) seconds before \(t_{\mathrm{ref}}\) and ends \(1\)--\(3\) seconds after \(t_{\mathrm{ref}}\), subject to the available session boundary. The gold answer is \texttt{Yes}. For negative L4 sessions, the L3 input is sampled once by the fixed preprocessing pipeline as a \(5\)--\(10\)-second segment from the matched non-event session, constrained to remain within the same event subcategory and not to contain the target event. The gold answer is \texttt{No}. This canonical L3--L4 pairing supports the L3-to-L4 grounding analysis in the main evaluation.

L1 and L2 are single-frame foundation-level controls derived from nuPlan frames and are not treated as per-session prerequisite gates for every L4 example. L1 Recognition tests driving-relevant entity recognition, while L2 Relational Understanding tests ego-centric relation reasoning. Table~\ref{tab:l3_l2_l1_categories} summarizes the reactive QA categories, definitions, and example question forms for L3--L1. L1 questions exclude spatial relation reasoning, temporal change, and alert decisions; L2 questions exclude temporal-change cues and multi-frame reasoning.

\begin{table}[t]
\centering
\small
\setlength{\tabcolsep}{2pt}
\renewcommand{\arraystretch}{1.08}
\caption{Reactive QA question-category definitions for L3--L1. L3 Temporal Comprehension follows the L4 event subcategories as paired short-clip temporal probes. L2 Relational Understanding covers ego-lane, adjacent-lane, traffic-light-control, and forward-area relation reasoning from a single frame. L1 Recognition covers vehicle, VRU, and traffic-light recognition from a single frame.}
\label{tab:l3_l2_l1_categories}
\begin{tabularx}{\textwidth}{L{0.13\textwidth} L{0.18\textwidth} X L{0.31\textwidth}}
\toprule
\textbf{Level} & \textbf{Subcategory} & \textbf{Scope} & \textbf{Example question} \\
\midrule

\multirow{6}{0.13\textwidth}{\raggedright L3 Temporal Comprehen\-sion} &
S1: lead-vehicle start &
Determine from a short clip whether the lead vehicle in the ego lane started moving. &
Did the lead vehicle in the ego lane start moving? \\

&
S2: lead-vehicle brake &
Determine from a short clip whether the lead vehicle in the ego lane braked. &
Did the lead vehicle in the ego lane brake? \\

&
S3: traffic-light change &
Determine from a short clip whether the ego-relevant traffic light changed state. &
Did the ego-relevant traffic light change state? \\

&
S4: adjacent cut-in &
Determine from a short clip whether a vehicle in an adjacent lane cut into the ego lane. &
Did a vehicle in an adjacent lane cut into the ego lane? \\

&
S5: VRU crossing &
Determine from a short clip whether a pedestrian or cyclist entered or started crossing the road ahead. &
Did a pedestrian or cyclist enter or start crossing the road ahead? \\

&
S6: sudden appearance &
Determine from a short clip whether a relevant target suddenly appeared from an occluded or hidden area. &
Did a pedestrian, vehicle, or other relevant target suddenly appear from a hidden area? \\

\midrule

\multirow{4}{0.13\textwidth}{\raggedright L2 Relational Understanding} &
Ego lane &
Determine whether a vehicle, participant, or region is located in or relevant to the ego lane. &
Is the vehicle ahead in the ego lane? \\

&
Adjacent lane &
Determine whether a vehicle or participant is in an adjacent lane or spatially related to adjacent-lane traffic. &
Is there a vehicle in the adjacent lane to the right of the ego vehicle? \\

&
Traffic-light control &
Determine whether a traffic light or traffic-control element is relevant to the ego vehicle. &
Is the visible traffic light controlling the ego vehicle's movement? \\

&
Forward area &
Determine whether a participant, obstacle, conflict region, or occlusion area lies in the ego vehicle's forward area. &
Is there a pedestrian in the ego vehicle's forward area? \\

\midrule

\multirow{3}{0.13\textwidth}{\raggedright L1 Recognition} &
Vehicle &
Recognize vehicles in a single driving frame. &
Is there a vehicle visible in the scene? \\

&
VRU &
Recognize vulnerable road users, including pedestrians and cyclists, in a single driving frame. &
Is there a pedestrian or cyclist visible in the scene? \\

&
Traffic light &
Recognize traffic lights and their visible states in a single driving frame. &
Is there a traffic light visible in the scene? \\

\bottomrule
\end{tabularx}
\end{table}

For L1 and L2, we use a human-template-guided, VLM-assisted QA construction pipeline. Human-written templates define the intended visual concept, allowed question form, expected answer format, and level-purity constraints for each subcategory. Given a candidate frame and the corresponding template, Qwen3.6 Plus generates construction-time visual descriptions and candidate QA pairs. GPT-5.5 is then used as an independent multimodal audit model: it receives the frame, candidate question, candidate answer, and template constraints, and checks single-frame answerability, answer uniqueness, category consistency, level purity, and distractor quality. Candidates that fail these checks are rejected or regenerated. The construction-time descriptions and quality-control outputs are not provided to evaluated models. These VLMs are used only for construction-time proposal and filtering; the resulting QA pool is separately audited in Section~\ref{app:data_quality}.

Train/test split consistency is enforced at the evaluated-subcategory level.
For paired L3/L4 samples, each canonical L3 temporal question inherits the split
of its corresponding L4 session, and samples derived from the same extracted
temporal segment within the same L3/L4 subcategory are assigned to the same
split. This prevents direct reuse of the same clip or frame sequence across train
and test for the evaluated event subcategory. L1/L2 examples follow the corresponding side of the source assignment when they
are derived from the same source sessions or clips. They are then further
deduplicated and filtered within each L1/L2 subcategory, so that directly reused
source frames are not shared across train and test for the same evaluated
question type. This design preserves cross-level alignment while preventing
direct train/test frame overlap within each evaluated subcategory.

Because L4-Core is constructed from a shared nuPlan-derived source distribution,
we interpret it as an in-distribution, category-conditioned benchmark for causal
watch-and-notify behavior, rather than as a cross-domain generalization
benchmark. To probe robustness beyond the Core source distribution and seen alert
families, we include S6 as a separate L4-OOD stress split, using external dashcam
videos and a held-out sudden-appearance alert category. S6 is reported separately
from the primary L4-Core leaderboard.

\subsection{Sudden-Appearance Data Construction and Privacy Redaction}
\label{app:sudden_appearance_privacy}

This appendix describes how we construct the sudden-appearance subset and how
we support local privacy redaction without redistributing source visual media.

\paragraph{Data construction.}
The sudden-appearance subset focuses on cases where a pedestrian, cyclist or vehicle appears from an occluded or hard-to-notice region
and require an immediate driver alert. We collect candidate dashcam videos
from publicly accessible video-sharing websites, primarily
Bilibili\footnote{\url{https://www.bilibili.com/}}, using event-related keyword
searches. Search terms cover sudden pedestrian appearance, cyclist or e-bike
crossing, vehicles emerging from blind spots, near-miss situations, and related
driving events.

For each source video, we extract a short session around the relevant
driving segment. Preprocessing for local reconstruction include frame extraction, frame-rate conversion, resolution normalization, and
privacy-redaction mask generation. Annotators then manually mark the key frame
at which the target first becomes visible enough to justify an alert. This key
frame is treated as the event onset for streaming evaluation. Frames before the
onset provide negative context, while the turn containing the onset is labeled
as the first positive alert turn. Clips are discarded if the event onset is
ambiguous, the viewpoint is unsuitable, or the target
cannot be reliably localized in time.

\paragraph{Privacy targets.}
For locally reconstructed S6 inputs, three types of regions are redacted:
\begin{itemize}[leftmargin=1.3em,itemsep=2pt,topsep=3pt]
    \item \textbf{Human faces}, including pedestrians, riders, and drivers visible through vehicle windows.
    \item \textbf{Vehicle license plates}, including plates on nearby vehicles.
    \item \textbf{Watermarks and overlay text}, including platform watermarks,
    uploader names, dashcam logos, timestamps, GPS coordinates, and
    other embedded on-screen text.
\end{itemize}

\paragraph{Redaction procedure.}
We use off-the-shelf vision models to produce the initial redaction regions. A
face detector is used to localize visible human faces, and a license-plate
detector together with OCR-based text matching is used to identify vehicle
plates. For watermarks and persistent overlay text, we apply an OCR/text
detector to sampled frames from the same source video and keep text regions
that consistently appear in fixed screen locations. The detected regions from
the three sources are merged into a binary mask for each indexed frame. These
masks are intended to guide local blurring or obscuring during reconstruction;
they contain no RGB content from the source video.

\paragraph{Released artifacts.}
We do not release raw videos redacted frames or any other source visual pixels from the web-sourced dashcam videos. For the sudden-appearance subset, the public release
contains only reproducibility metadata and local-processing artifacts, including
source URLs, key-frame annotations, watch instructions, positive/negative labels,
preprocessing scripts, and binary redaction masks.

Users who wish to reproduce the visual inputs must obtain the source videos
from the original platforms, comply with the corresponding licenses, platform
terms, and local laws, and apply the provided preprocessing and redaction
pipeline locally.

\clearpage
\section{Dataset statistics and audits}
\label{app:data_stats}

This appendix reports detailed statistics beyond the compact split summary in the main paper. Appendix~\ref{app:split_stats} reports split-wise L4 and L1--L3 counts. Appendix~\ref{app:duration_tref_audit} reports duration and reference-time audits. Appendix~\ref{app:negative_descriptor_audit} reports the test-split composition of auxiliary negative construction descriptors. These statistics are provided for transparency and reproducibility; the official L4 scoring protocol depends only on the final positive/negative event label, the senior-final \(t_{\mathrm{ref}}\) for positives, and the model's first alert decision.

\subsection{Paired L3/L4 split composition}
\label{app:split_stats}

Table~\ref{tab:datastats} complements the compact split summary in Table~\ref{tab:data} by decomposing the paired L3/L4 event-level counts into positive and negative instances. Each accepted L4 session is paired with one canonical L3 temporal question over the same monitored event, and the paired L3 question follows the positive/negative label of its corresponding L4 session. Therefore, the counts and positive/negative composition in Table~\ref{tab:datastats} apply to both L3 and L4. This decomposition is useful for interpreting L4 AlertScore@\(\tau\), since positive L4 sessions determine event-aligned Hit@\(\tau\), while negative L4 sessions determine false-alert restraint through NSR. L4-Core (S1--S5) defines the primary ranking. S6 is an L4-OOD stress split, held out from training and excluded from the primary ranking. Split-wise L1 and L2 total counts are reported in Table~\ref{tab:data}; the reactive QA category definitions are summarized in Table~\ref{tab:l3_l2_l1_categories}.

\begin{table}[H]
\centering
\small
\setlength{\tabcolsep}{4pt}
\caption{Paired L3/L4 split composition. Counts are reported as positive + negative instances and apply to both L3 and L4, since each accepted L4 session has one canonical L3 temporal question with the same positive/negative label. L4-Core (S1--S5) defines the primary ranking, while S6 is an L4-OOD stress split excluded from the primary ranking.}
\label{tab:datastats}
\begin{tabular}{l c c c}
\toprule
\textbf{Subcategory} & \textbf{Train (pos+neg)} & \textbf{Test (pos+neg)} & \textbf{Alert mode} \\
\midrule
S1: lead-vehicle start & 300 + 929 & 101 + 300 & Event-triggered \\
S2: lead-vehicle brake & 294 + 893 & 94 + 305 & Event-triggered \\
S3: traffic-light change & 296 + 899 & 103 + 300 & Event-triggered \\
S4: adjacent cut-in & 54 + 154 & 81 + 260 & Target-timing \\
S5: VRU crossing & 293 + 891 & 98 + 307 & Target-timing \\
S6: sudden appearance & -- & 99 + 300 & Event-triggered \\
\midrule
\textbf{L3/L4-Core total (S1--S5)} & \textbf{1{,}237 + 3{,}766} & \textbf{477 + 1{,}472} & -- \\
\textbf{L3/L4 total (S1--S6)} & \textbf{1{,}237 + 3{,}766} & \textbf{576 + 1{,}772} & -- \\
\bottomrule
\end{tabular}
\end{table}

\subsection{Reactive L1/L2 QA format composition}
\label{app:reactive_qa_format_stats}

Table~\ref{tab:l1_l2_format_stats} reports the answer-format composition of L1 and L2. These statistics complement the split-wise counts in Table~\ref{tab:data} by showing the proportion of multiple-choice and binary questions within each L1/L2 subcategory. L3 uses template-based binary questions paired with the corresponding L4 sessions, as described in Appendix~\ref{app:l13_construction_splits}.

\begin{table}[H]
\centering
\small
\setlength{\tabcolsep}{4pt}
\renewcommand{\arraystretch}{1.05}
\caption{Answer-format composition of L1 and L2 reactive QA instances. Counts are aggregated over train and test splits. MCQ denotes multiple-choice questions, and Binary denotes two-choice yes/no questions.}
\label{tab:l1_l2_format_stats}
\begin{tabular}{l l r r r r r}
\toprule
\textbf{Level} & \textbf{Subcategory} & \textbf{Total} & \textbf{MCQ} & \textbf{Binary} & \textbf{MCQ \%} & \textbf{Binary \%} \\
\midrule
L1 & Vehicle & 3{,}037 & 26 & 3{,}011 & 0.86\% & 99.14\% \\
L1 & VRU & 1{,}732 & 0 & 1{,}732 & 0.00\% & 100.00\% \\
L1 & Traffic light & 1{,}098 & 0 & 1{,}098 & 0.00\% & 100.00\% \\
\midrule
L2 & Ego lane & 2{,}170 & 1{,}014 & 1{,}156 & 46.73\% & 53.27\% \\
L2 & Adjacent lane & 789 & 15 & 774 & 1.90\% & 98.10\% \\
L2 & Traffic-light control & 1{,}290 & 359 & 931 & 27.83\% & 72.17\% \\
L2 & Forward area & 2{,}035 & 13 & 2{,}022 & 0.64\% & 99.36\% \\
\bottomrule
\end{tabular}
\end{table}

\subsection{Duration and reference-time audits}
\label{app:duration_tref_audit}

\begin{figure*}[t]
    \centering
    \setlength{\tabcolsep}{2pt}
    \begin{tabular}{ccc}
        \includegraphics[width=0.32\textwidth]{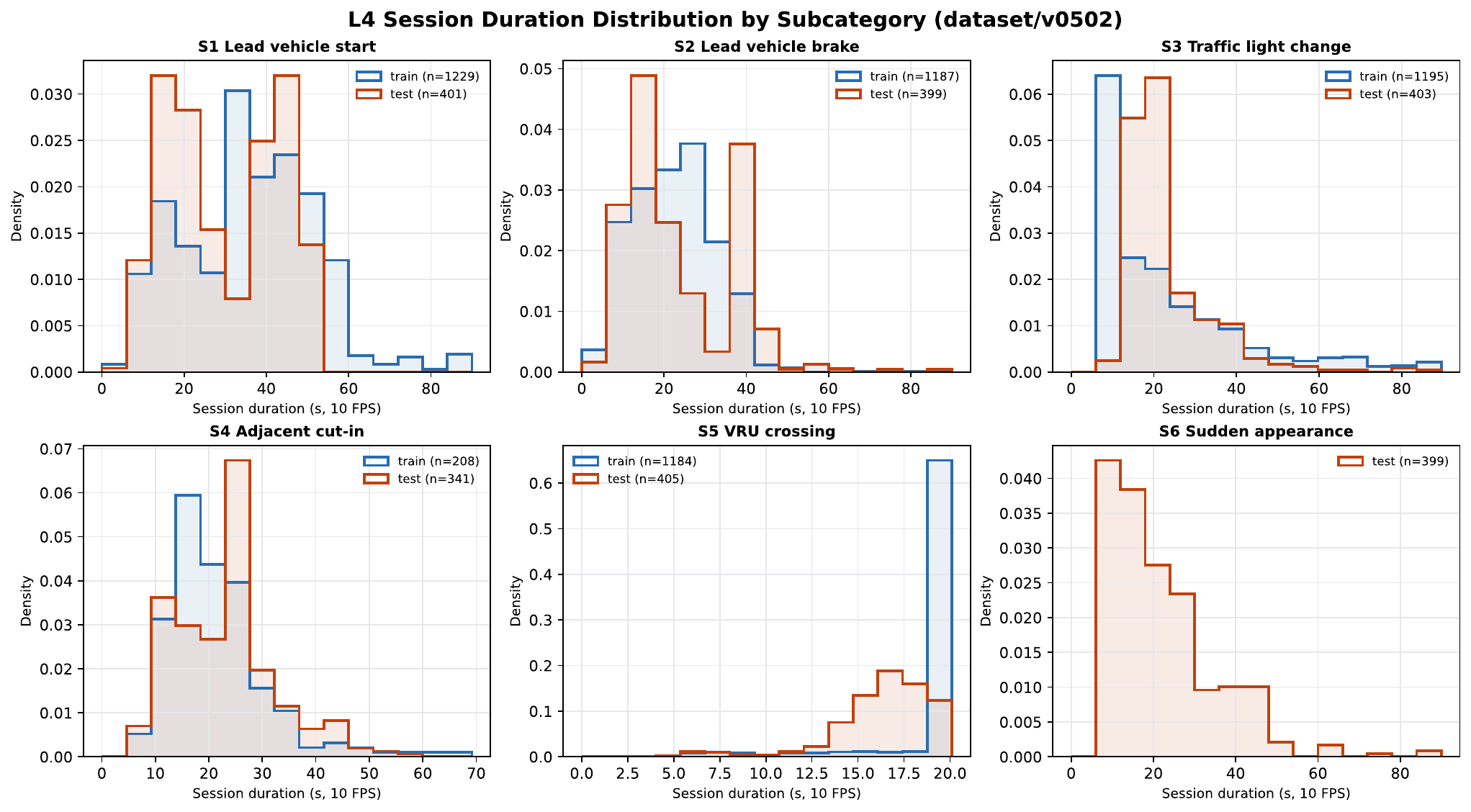} &
        \includegraphics[width=0.32\textwidth]{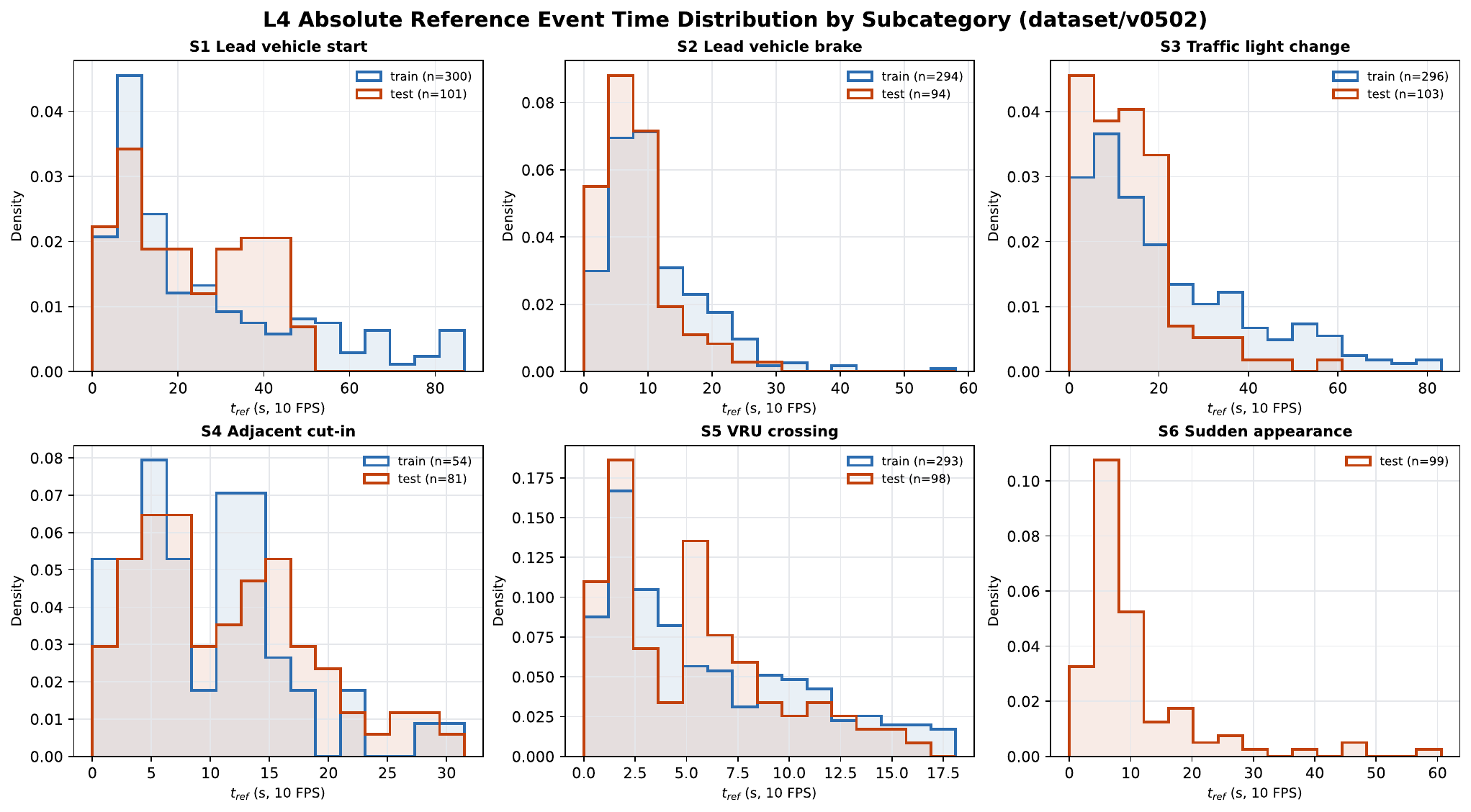} &
        \includegraphics[width=0.32\textwidth]{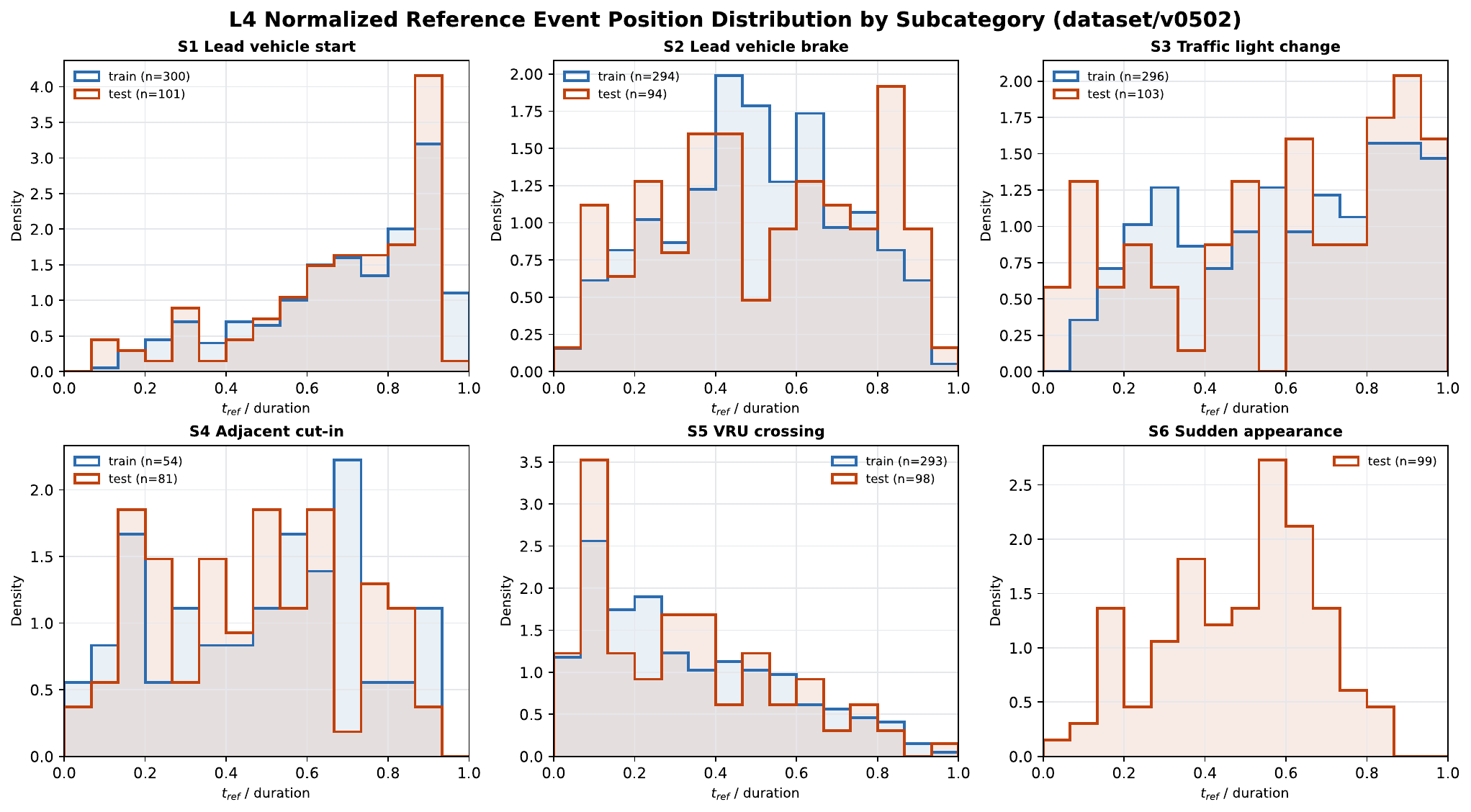} \\
        {\scriptsize (a) Duration distribution} &
        {\scriptsize (b) Absolute \(t_{\mathrm{ref}}\) distribution} &
        {\scriptsize (c) Normalized \(t_{\mathrm{ref}}/\mathrm{duration}\)} \\
    \end{tabular}
    \caption{Temporal audit statistics for the paired L3/L4 event levels. (a) Distribution of session duration by subcategory and split. Duration is reported because session length affects false-alert opportunities and may interact with temporal-prior baselines. (b) Distribution of absolute reference alert time \(t_{\mathrm{ref}}\) for positive sessions, measured in seconds from session start. This is the temporal coordinate observable to an online model under the streaming protocol. (c) Distribution of normalized event position \(t_{\mathrm{ref}}/\mathrm{duration}\) for positive sessions. Unlike absolute \(t_{\mathrm{ref}}\), normalized position is not observable to an online model during inference; it is reported only as a dataset-construction transparency statistic.}
    \label{fig:app_l4_temporal_audits}
\end{figure*}

Figure~\ref{fig:app_l4_temporal_audits} summarizes three temporal audit statistics for the paired L3/L4 event levels. The left panel reports L4 session duration distributions by subcategory and split. The middle panel reports absolute \(t_{\mathrm{ref}}\) distributions for positive sessions, measured in video seconds from session start. The right panel reports normalized event position \(t_{\mathrm{ref}}/\mathrm{duration}\) as a dataset-construction transparency statistic. Absolute \(t_{\mathrm{ref}}\) is the temporal coordinate observable to an online model under the streaming protocol. By contrast, normalized event position is not observable during inference because the online model does not know the future session duration. Table~\ref{tab:l4_duration_tref_stats} reports the corresponding median and interquartile-range statistics.

\begin{table}[t]
\centering
\scriptsize
\setlength{\tabcolsep}{3pt}
\renewcommand{\arraystretch}{1.04}
\caption{
Session-duration and event-time statistics for L4.
Absolute \(t_{\mathrm{ref}}\) is measured in seconds from session start and corresponds to the temporal variable observable under the streaming protocol.
Normalized event position \(t_{\mathrm{ref}}/\mathrm{duration}\) is reported only as a dataset-construction transparency statistic, since the online model does not know future session duration.
}
\label{tab:l4_duration_tref_stats}
\resizebox{\textwidth}{!}{%
\begin{tabular}{l l l r r r r r r r}
\toprule
\textbf{Subcategory} & \textbf{Split} & \textbf{Label} & \textbf{\#} &
\textbf{Duration med.} & \textbf{Duration IQR} &
\textbf{\(t_{\mathrm{ref}}\) med.} & \textbf{\(t_{\mathrm{ref}}\) IQR} &
\textbf{Norm. med.} & \textbf{Norm. IQR} \\
\midrule

\multirow{4}{*}{S1: lead-vehicle start}
& train & positive & 300 & 23.9 & 29.3 & 16.0 & 31.1 & 0.7 & 0.3 \\
& train & negative & 929 & 37.8 & 16.7 & -- & -- & -- & -- \\
& test & positive & 101 & 29.0 & 26.6 & 19.7 & 27.0 & 0.7 & 0.3 \\
& test & negative & 300 & 29.0 & 24.8 & -- & -- & -- & -- \\
\specialrule{0.025em}{0.28ex}{0.28ex}

\multirow{4}{*}{S2: lead-vehicle brake}
& train & positive & 294 & 21.4 & 14.3 & 9.3 & 8.7 & 0.5 & 0.3 \\
& train & negative & 893 & 22.3 & 14.4 & -- & -- & -- & -- \\
& test & positive & 94 & 14.8 & 10.9 & 7.2 & 6.1 & 0.5 & 0.4 \\
& test & negative & 305 & 20.5 & 24.3 & -- & -- & -- & -- \\
\specialrule{0.025em}{0.28ex}{0.28ex}

\multirow{4}{*}{S3: traffic-light change}
& train & positive & 296 & 28.8 & 27.9 & 16.2 & 24.5 & 0.6 & 0.5 \\
& train & negative & 899 & 12.5 & 14.1 & -- & -- & -- & -- \\
& test & positive & 103 & 20.0 & 8.3 & 12.5 & 12.8 & 0.7 & 0.5 \\
& test & negative & 300 & 20.0 & 8.1 & -- & -- & -- & -- \\
\specialrule{0.025em}{0.28ex}{0.28ex}

\multirow{4}{*}{S4: adjacent cut-in}
& train & positive & 54 & 18.5 & 9.3 & 9.0 & 9.5 & 0.5 & 0.4 \\
& train & negative & 154 & 20.8 & 9.9 & -- & -- & -- & -- \\
& test & positive & 81 & 25.4 & 10.7 & 10.2 & 10.5 & 0.5 & 0.4 \\
& test & negative & 260 & 23.7 & 11.8 & -- & -- & -- & -- \\
\specialrule{0.025em}{0.28ex}{0.28ex}

\multirow{4}{*}{S5: VRU crossing}
& train & positive & 293 & 20.0 & 3.3 & 4.4 & 7.1 & 0.3 & 0.4 \\
& train & negative & 891 & 20.0 & 0.0 & -- & -- & -- & -- \\
& test & positive & 98 & 16.6 & 2.9 & 5.0 & 5.8 & 0.3 & 0.4 \\
& test & negative & 307 & 16.9 & 2.8 & -- & -- & -- & -- \\
\specialrule{0.025em}{0.28ex}{0.28ex}

\multirow{2}{*}{S6: sudden appearance}
& test & positive & 99 & 17.8 & 17.3 & 6.7 & 6.2 & 0.5 & 0.3 \\
& test & negative & 300 & 19.0 & 16.2 & -- & -- & -- & -- \\

\bottomrule
\end{tabular}%
}
\end{table}

\subsection{L4 annotation agreement audit}
\label{app:l4_annotation_agreement}

We audit pre-adjudication agreement using only the three independent annotators before senior review. The senior annotator is not treated as an independent annotator for this audit, because senior review uses the three annotations and candidate timestamps to assign the final benchmark metadata. Since positive and negative candidates are mined from separate candidate pools, we report valid/invalid agreement separately for positive-candidate and negative-candidate pools.

For each candidate \(i\), annotator \(a\)'s pre-adjudication decision is denoted as
\[
v_i^{(a)} \in \{\mathrm{Valid}, \mathrm{Invalid}\}.
\]
For a mined positive candidate, \(\mathrm{Valid}\) means that the instructed event occurs, is ego-relevant, has a unique valid target event, and is sufficiently visible under the watch instruction. For a mined negative candidate, \(\mathrm{Valid}\) means that the watch instruction remains meaningful for the clip but the target event does not occur.

For each candidate pool, we report three valid/invalid agreement diagnostics. Unanimous agreement measures the fraction of candidates for which all three annotators give the same valid/invalid decision:
\[
\mathrm{Unanimous}
=
\frac{1}{N}
\sum_i
\mathbf{1}\!\left[
v_i^{(1)} = v_i^{(2)} = v_i^{(3)}
\right].
\]
Majority-valid rate measures the fraction of candidates for which at least two annotators mark the candidate as valid:
\[
\mathrm{MajorityValid}
=
\frac{1}{N}
\sum_i
\mathbf{1}\!\left[
\sum_a \mathbf{1}\!\left[v_i^{(a)}=\mathrm{Valid}\right] \ge 2
\right].
\]
We report this quantity as a qualification-consensus diagnostic rather than as a chance-corrected reliability coefficient. Pairwise valid/invalid agreement averages agreement over the three annotator pairs:
\[
\mathrm{PairwiseAgreement}
=
\frac{
\sum_i \sum_{a<b}
\mathbf{1}\!\left[
v_i^{(a)} = v_i^{(b)}
\right]
}{
\sum_i \sum_{a<b} 1
}.
\]

\begin{table}[t]
\centering
\small
\setlength{\tabcolsep}{5pt}
\renewcommand{\arraystretch}{1.08}
\caption{Pre-adjudication valid/invalid agreement among independent annotators. Positive and negative candidates are reported separately because they are mined from separate candidate pools. Majority-valid denotes candidates marked valid by at least two annotators.}
\label{tab:l4_validity_iaa}
\begin{tabularx}{\textwidth}{@{}>{\raggedright\arraybackslash}X l r c c c@{}}
\toprule
\textbf{Subcategory} &
\textbf{Pool} &
\textbf{\# Cand.} &
\textbf{Unanim.} &
\textbf{Maj.-valid} &
\textbf{Pairwise} \\
\midrule

\multirow{2}{*}{S1: lead-vehicle start}
& Pos. & 516  & 94.4\% & 78.7\% & 96.3\% \\
& Neg. & 1580 & 94.3\% & 79.1\% & 96.2\% \\

\addlinespace[2pt]
\multirow{2}{*}{S2: lead-vehicle brake}
& Pos. & 542  & 95.6\% & 72.1\% & 97.1\% \\
& Neg. & 1672 & 96.0\% & 72.1\% & 97.3\% \\

\addlinespace[2pt]
\multirow{2}{*}{S3: traffic-light change}
& Pos. & 461  & 96.1\% & 87.0\% & 97.4\% \\
& Neg. & 1384 & 95.5\% & 87.0\% & 97.0\% \\

\addlinespace[2pt]
\multirow{2}{*}{S4: adjacent cut-in}
& Pos. & 167 & 94.6\% & 82.0\% & 96.4\% \\
& Neg. & 466 & 94.6\% & 89.7\% & 96.4\% \\

\addlinespace[2pt]
\multirow{2}{*}{S5: VRU crossing}
& Pos. & 507  & 94.5\% & 78.1\% & 96.3\% \\
& Neg. & 1654 & 94.3\% & 73.8\% & 96.2\% \\

\addlinespace[2pt]
\multirow{2}{*}{S6: sudden appearance}
& Pos. & 108 & 98.1\% & 92.6\%  & 98.8\% \\
& Neg. & 310 & 97.7\% & 100.0\% & 98.5\% \\

\bottomrule
\end{tabularx}
\end{table}

We further audit reference-time agreement on majority-valid positive L4 candidates. A positive candidate is included in this timing audit only when at least two independent annotators mark it as valid and provide candidate reference-time anchors. The timestamps are recorded as frame indices on 10 FPS videos and converted to seconds when reporting timing error. For sample \(i\) and annotator pair \((a,b)\), let \(t_i^{(a)}\) and \(t_i^{(b)}\) denote the annotated reference frames. Annotator pair \((a,b)\) is included for sample \(i\) only when both annotators mark the candidate as valid and both provide \(t_{\mathrm{ref}}\). The pairwise timing error in seconds is
\[
e_i^{(a,b)}
=
\frac{
\left|t_i^{(a)} - t_i^{(b)}\right|
}{10}.
\]

We report the median pairwise timing error, the 75th and 90th percentiles of the pairwise timing error, and thresholded pairwise time agreement. For a tolerance \(\delta\), pairwise reference-time agreement is defined as
\[
\mathrm{TimeIAA@}\delta
=
\frac{
\sum_i \sum_{a<b}
\mathbf{1}\!\left[
v_i^{(a)}=v_i^{(b)}=\mathrm{Valid}
\right]
\mathbf{1}\!\left[
\frac{\left|t_i^{(a)} - t_i^{(b)}\right|}{10} \le \delta
\right]
}{
\sum_i \sum_{a<b}
\mathbf{1}\!\left[
v_i^{(a)}=v_i^{(b)}=\mathrm{Valid}
\right]
}.
\]
We report TimeIAA@1s and TimeIAA@2s. In implementation, these are equivalent to frame thresholds of 10 frames and 20 frames, respectively. Negative candidates are excluded from reference-time agreement because no alert is warranted and no \(t_{\mathrm{ref}}\) is assigned.

\begin{table}[t]
\centering
\small
\caption{Reference-time agreement among independent annotators on majority-valid positive L4 candidates.}
\label{tab:l4_time_iaa}
\begin{tabular}{lrrrrr}
\toprule
Subcategory &
Median err. (s) &
P75 err. (s) &
P90 err. (s) &
TimeIAA@1s &
TimeIAA@2s \\
\midrule
S1: lead-vehicle start  & 0.20 & 0.30 & 0.60 & 96.7\% & 100.0\% \\
S2: lead-vehicle brake  & 0.10 & 0.10 & 0.20 & 100.0\% & 100.0\% \\
S3: traffic-light change  & 0.00 & 0.00 & 0.01 & 100.0\% & 100.0\% \\
S4: adjacent cut-in  & 0.30 & 0.42 & 0.51 & 100.0\% & 100.0\% \\
S5: VRU crossing  & 0.10 & 0.10 & 0.32 & 100.0\% & 100.0\% \\
S6: sudden appearance  & 0.10 & 0.12 & 0.41 & 100.0\% & 100.0\% \\

\bottomrule
\end{tabular}
\end{table}

\subsection{Negative construction descriptor audit}
\label{app:negative_descriptor_audit}

Negative sessions are category-matched plausible non-events rather than generic idle clips.
During construction and auditing, we use three coarse negative construction descriptors to inspect
negative-session diversity: \emph{Simple Non-event}, \emph{Structured Non-event}, and
\emph{Trigger-like Confounder}. These descriptors document construction intent, such as absent
monitored entities, relevant entities without the target event, or cues that resemble the target event
but fail the operational event definition.

\begin{figure}[t]
\centering
\includegraphics[width=\textwidth]{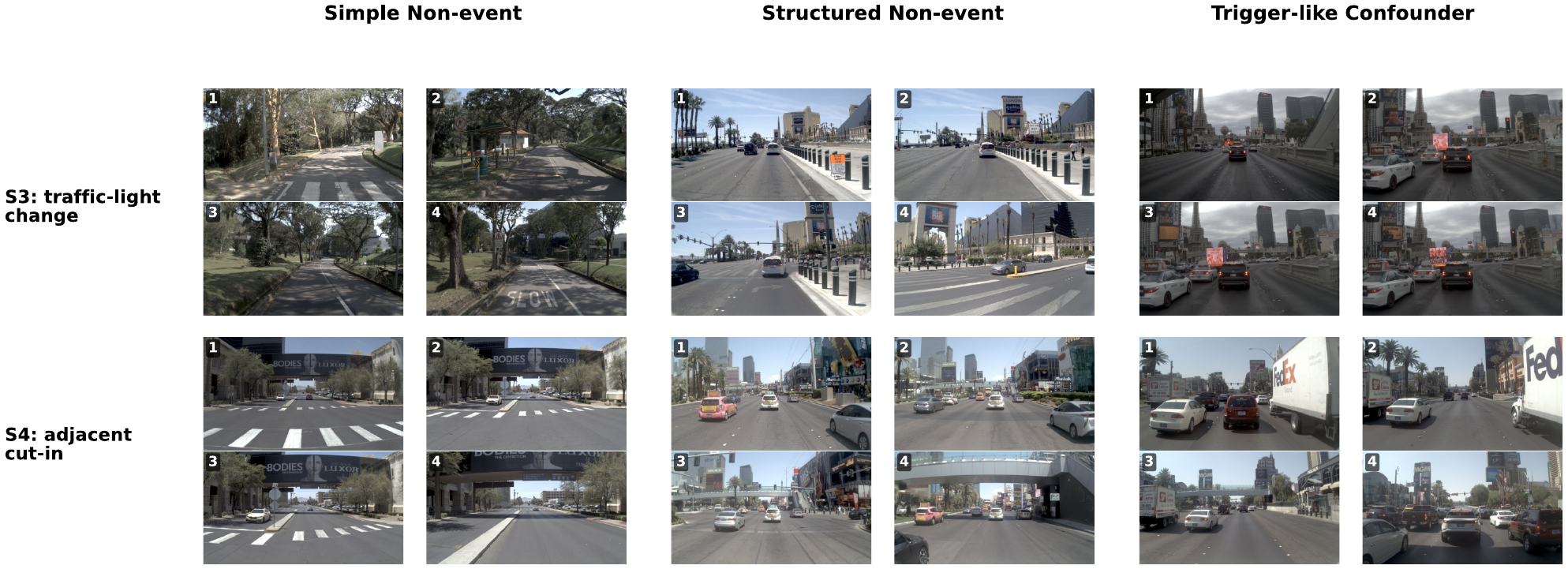}
\caption{
Representative negative-construction examples for S3 traffic-light change and S4 adjacent cut-in.
Columns correspond to the three coarse descriptor families used during construction and auditing:
Simple Non-event, Structured Non-event, and Trigger-like Confounder.
These examples are illustrative only; the descriptor families are construction-audit annotations
rather than benchmark labels or evaluation strata.
}
\label{fig:negative_descriptor_s3_s4_grid}
\end{figure}

Importantly, these descriptors are not adjudicated benchmark labels. They are not used for scoring,
training targets, leaderboard stratification, or quantitative model-performance analysis. The official
L4 evaluation uses only the senior-final positive/negative event label, the senior-final
\(t_{\mathrm{ref}}\) for positives, and the model's first alert decision. We therefore report qualitative
construction patterns and representative examples rather than treating negative-subtype counts as
benchmark statistics. Table~\ref{tab:negative_descriptor_examples} summarizes the category-specific
patterns, while Figure~\ref{fig:negative_descriptor_s3_s4_grid} illustrates representative S3
traffic-light-change and S4 adjacent-cut-in negatives under the three coarse descriptor families.

\begin{table}[t]
\centering
\scriptsize
\setlength{\tabcolsep}{3pt}
\renewcommand{\arraystretch}{1.12}
\caption{
Category-specific negative construction patterns.
The examples are representative construction and audit patterns rather than exhaustive subtype labels.
They are provided to make negative-session diversity transparent without treating construction
descriptors as benchmark statistics.
}
\label{tab:negative_descriptor_examples}
\begin{tabularx}{\textwidth}{
    L{0.17\textwidth}
    L{0.22\textwidth}
    L{0.27\textwidth}
    X
}
\toprule
\textbf{L4 subcategory} &
\textbf{Simple Non-event} &
\textbf{Structured Non-event} &
\textbf{Trigger-like Confounder} \\
\midrule

S1: lead-vehicle start &
No lead vehicle is present in the scene. &
A lead vehicle is present but remains stationary, or no trigger-like condition is observed. &
The lead vehicle exhibits low-speed creeping or stop-and-go motion without satisfying the start definition; an adjacent vehicle starts moving while the lead vehicle does not. \\

\midrule

S2: lead-vehicle brake &
No lead vehicle is present in the scene. &
The lead vehicle maintains stable speed or accelerates. &
The lead vehicle mildly decelerates, but not to a degree that satisfies the braking criterion. \\

\midrule

S3: traffic-light change &
No traffic light is present in the scene. &
The ego-relevant traffic light remains stable throughout the clip. &
Traffic lights at the same intersection but not relevant to the ego lane change state; the ego vehicle slows or stops for other interactions while the ego-relevant light remains unchanged. \\

\midrule

S4: adjacent cut-in &
No relevant adjacent-lane interaction occurs, or adjacent vehicles are absent, stationary, or nearly stationary. &
A vehicle in an adjacent lane moves in the same direction as the ego vehicle but shows no cut-in behavior. &
A vehicle in an adjacent lane approaches the ego lane or exhibits a lane-approaching trajectory without completing a valid cut-in. \\

\midrule

S5: VRU crossing &
No vulnerable road user is present, or no crossing scenario exists. &
VRUs are present but remain stationary or move along the roadside without entering the ego-relevant path. &
VRUs exhibit lateral motion near a crosswalk area, or multiple VRUs create apparent crossing trajectories without satisfying the crossing definition. \\

\midrule

S6: sudden appearance &
Open-road or night-time driving with no occlusion or roadside target. &
Traffic participants are visible in open-road or urban traffic scenes, but no occlusion-triggered sudden appearance occurs. &
Limited visibility or occlusion is present due to road geometry, vegetation, construction barriers, parked vehicles, building corners, or adverse weather, but no qualifying target suddenly appears. \\

\bottomrule
\end{tabularx}
\end{table}

\subsection{Human Validation and Quality Audit}
\label{app:data_quality}

We conducted a manual quality audit before public release. After the
audit, all sampled items marked as rejected were corrected before public
release. The audit is a spot check: it remediates the failures found in the
sampled audit pool, but it does not certify that every unsampled item is
error-free.

\paragraph{L4 audit setup and checklist.}
The audit was carried out by an independent 15-member quality-control team that
was not involved in the initial annotation process. Each audited item was
reviewed by one quality-control reviewer using the same checklist. A sample was
accepted only if the reviewer verified that: (i) the video clip, watch
instruction, and annotation were mutually consistent; (ii) for positive
samples, the clip contained a single qualifying target event and the annotated
\(t_{\mathrm{ref}}\) matched the event onset or reference alert point defined
for that subcategory; for negative samples, no qualifying target event occurred
in the clip; and (iii) the visual evidence was sufficient for a human reviewer
to verify the judgment. A sample failing any check was marked as rejected and
sent to the correction step.

\paragraph{L4 sampling.}
For each L4 subcategory, we randomly sampled a roughly balanced set of positive
and negative items for audit. The resulting audited sample sizes are reported
in Tables~\ref{tab:data_quality_train} and~\ref{tab:data_quality_test}.

\paragraph{Reported statistics.}
For each subcategory, we report the audited size \(n\), accepted count \(a\),
and the pre-remediation audit acceptance rate
\[
\hat{p}=\frac{a}{n}.
\]
The
95\% confidence interval is computed with the Wilson score interval on the
audited sample:
\[
\mathrm{CI}_{95}^{\mathrm{Wilson}}
=
\frac{
\hat{p}+\frac{z^2}{2n}
\;\pm\;
z\sqrt{
\frac{\hat{p}(1-\hat{p})}{n}
+
\frac{z^2}{4n^2}
}
}{
1+\frac{z^2}{n}
},
\qquad z=1.96.
\]
Wilson intervals are used instead of the normal
approximation because several audit cells have acceptance rates close to
100\%, where the normal interval can be unstable or overly optimistic. For
example, even when all \(100\) audited items are accepted, the Wilson interval
gives a lower bound below 100\% rather than a degenerate interval.

\begin{table}[H]
\centering
\small
\setlength{\tabcolsep}{6pt}
\renewcommand{\arraystretch}{1.04}
\caption{L4 train-split audit.}
\label{tab:data_quality_train}
\begin{tabular}{l c c c c}
\toprule
\textbf{Subcategory} & \(\mathbf{n}\) & \(\mathbf{a}\) &
\(\boldsymbol{\hat{p}}\) &
\(\mathbf{CI}_{95}^{\mathrm{Wilson}}\) \\
\midrule

S1: lead-vehicle start   & 100 & 100 & 100.00\% & [96.30,\,100.00] \\
\specialrule{0.025em}{0.28ex}{0.28ex}

S2: lead-vehicle brake & 100 &  96 &  96.00\% & [90.16,\, 98.43] \\
\specialrule{0.025em}{0.28ex}{0.28ex}

S3: traffic-light change & 100 &  93 &  93.00\% & [86.25,\, 96.57] \\
\specialrule{0.025em}{0.28ex}{0.28ex}

S4: adjacent cut-in      &  79 &  73 &  92.41\% & [84.40,\, 96.47] \\
\specialrule{0.025em}{0.28ex}{0.28ex}

S5: VRU crossing         & 100 & 100 & 100.00\% & [96.30,\,100.00] \\
\midrule

\textbf{Train total} & \textbf{479} & \textbf{462} &
\textbf{96.45\%} & \textbf{[94.39,\,97.77]} \\
\bottomrule
\end{tabular}
\end{table}

\begin{table}[H]
\centering
\small
\setlength{\tabcolsep}{6pt}
\renewcommand{\arraystretch}{1.04}
\caption{L4 test-split audit.}
\label{tab:data_quality_test}
\begin{tabular}{l c c c c}
\toprule
\textbf{Subcategory} & \(\mathbf{n}\) & \(\mathbf{a}\) &
\(\boldsymbol{\hat{p}}\) &
\(\mathbf{CI}_{95}^{\mathrm{Wilson}}\) \\
\midrule

S1: lead-vehicle start   & 100 &  99 &  99.00\% & [94.55,\,99.82] \\
\specialrule{0.025em}{0.28ex}{0.28ex}

S2: lead-vehicle brake & 100 &  95 &  95.00\% & [88.82,\,97.85] \\
\specialrule{0.025em}{0.28ex}{0.28ex}

S3: traffic-light change & 100 &  99 &  99.00\% & [94.55,\,99.82] \\
\specialrule{0.025em}{0.28ex}{0.28ex}

S4: adjacent cut-in      & 100 &  99 &  99.00\% & [94.55,\,99.82] \\
\specialrule{0.025em}{0.28ex}{0.28ex}

S5: VRU crossing         & 100 &  97 &  97.00\% & [91.55,\,98.97] \\
\specialrule{0.025em}{0.28ex}{0.28ex}

S6: sudden appearance    & 100 & 100 & 100.00\% & [96.30,\,100.00] \\
\midrule

\textbf{Test total} & \textbf{600} & \textbf{589} &
\textbf{98.17\%} & \textbf{[96.75,\,98.97]} \\
\bottomrule
\end{tabular}
\end{table}

\paragraph{L4 failure modes.}
For each rejected audit item, we recorded the main error type before correction.
These error types were used only for audit correction, not as benchmark labels
or evaluation strata.

\begin{table}[H]
\centering
\small
\setlength{\tabcolsep}{6pt}
\caption{L4 audit failure categories.}
\label{tab:data_quality_failure_types}
\begin{tabular}{l p{0.66\linewidth}}
\toprule
Failure category & Definition \\
\midrule

Timing error &
The target event is present, but \(t_{\mathrm{ref}}\) is placed too early or
too late relative to the intended alert point. \\

Event-definition mismatch &
The observed event does not meet the subcategory definition. \\

Insufficient visual evidence &
The clip is too occluded, truncated, or ambiguous for a reliable judgment. \\

\bottomrule
\end{tabular}
\end{table}

\paragraph{L3 temporal QA audit.}
L3 was not audited separately because it is directly paired with L4.
Each L3 item is derived from the corresponding accepted L4 session: positive
clips are cropped around \(t_{\mathrm{ref}}\), while negative clips use the
matched non-event segment. Thus, once the L4 audit verifies event validity,
subcategory consistency, reference-time alignment, and visual sufficiency, the
evidence needed for the paired L3 label is already covered.

\paragraph{L1--L2 reactive QA audit.}
We also audited the L1--L2 reactive QA test split. This audit follows the same
release-candidate convention as the L4 audit, but the checklist is adapted to
single-frame QA: there are no streaming decisions or trigger timestamps. An
L1--L2 item was accepted only if the reviewer verified that: (i) the image and
question were mutually consistent and answerable; (ii) for L1, the queried
entity was clearly present or absent, and for L2, the queried ego-centric
relation was visually supported; and (iii) for multiple-choice items, the gold
option was the best-supported answer among the four provided options. Items
failing any check were corrected before release.

\begin{table}[H]
\centering
\small
\setlength{\tabcolsep}{6pt}
\renewcommand{\arraystretch}{1.04}
\caption{L1 test-split audit.}
\label{tab:data_quality_l1}
\begin{tabular}{l c c c c}
\toprule
\textbf{Category} & \(\mathbf{n}\) & \(\mathbf{a}\) &
\(\boldsymbol{\hat{p}}\) &
\(\mathbf{CI}_{95}^{\mathrm{Wilson}}\) \\
\midrule

traffic light & 50 & 50 & 100.00\% & [92.86,\,100.00] \\
\specialrule{0.025em}{0.28ex}{0.28ex}

vehicle       & 50 & 49 &  98.00\% & [89.50,\,\,99.65] \\
\specialrule{0.025em}{0.28ex}{0.28ex}

VRU           & 50 & 49 &  98.00\% & [89.50,\,\,99.65] \\
\midrule

\textbf{L1 total} & \textbf{150} & \textbf{148} &
\textbf{98.67\%} & \textbf{[95.27,\,\,99.63]} \\
\bottomrule
\end{tabular}
\end{table}

\begin{table}[H]
\centering
\small
\setlength{\tabcolsep}{6pt}
\renewcommand{\arraystretch}{1.04}
\caption{L2 test-split audit.}
\label{tab:data_quality_l2}
\begin{tabular}{l c c c c}
\toprule
\textbf{Category} & \(\mathbf{n}\) & \(\mathbf{a}\) &
\(\boldsymbol{\hat{p}}\) &
\(\mathbf{CI}_{95}^{\mathrm{Wilson}}\) \\
\midrule

adjacent lane          & 50 & 47 & 94.00\% & [83.78,\,\,97.94] \\
\specialrule{0.025em}{0.28ex}{0.28ex}

ego lane               & 50 & 46 & 92.00\% & [81.16,\,\,96.85] \\
\specialrule{0.025em}{0.28ex}{0.28ex}

forward area           & 50 & 46 & 92.00\% & [81.16,\,\,96.85] \\
\specialrule{0.025em}{0.28ex}{0.28ex}

traffic-light control  & 50 & 49 & 98.00\% & [89.50,\,\,99.65] \\
\midrule

\textbf{L2 total} & \textbf{200} & \textbf{188} &
\textbf{94.00\%} & \textbf{[89.81,\,\,96.53]} \\
\bottomrule
\end{tabular}
\end{table}

\paragraph{L1--L2 failure modes.}
For each rejected L1--L2 audit item, we recorded the main error type before
correction. These error types were used only for audit correction, not as
benchmark labels or evaluation groups.

\begin{table}[H]
\centering
\small
\setlength{\tabcolsep}{6pt}
\caption{L1--L2 audit failure categories.}
\label{tab:data_quality_l12_failure_types}
\begin{tabularx}{\linewidth}{@{}p{0.30\linewidth} X@{}}
\toprule
Failure category & Definition \\
\midrule

Existence or polarity mismatch &
The reviewer disagreed with whether the queried entity, relation, or condition
was present in the image. \\

Fine-grained attribute mismatch &
The item depended on a fine-grained attribute, such as vehicle color or type,
and the candidate gold answer was not the best-supported choice. \\

Insufficient visual evidence &
The image was too small, low quality, occluded, or ambiguous for a reliable
judgment. \\

\bottomrule
\end{tabularx}
\end{table}

\paragraph{Limitations.}
This release audit is intended as a sampled quality-control check. Each audited item was
reviewed by one quality-control reviewer, so the audit does not measure
inter-reviewer variation on the sampled release pool. All sampled items that
failed the checklist were corrected before release; nevertheless, as with any
sampled audit, the procedure cannot rule out residual annotation errors outside
the audited subset. The Wilson intervals therefore quantify sampling
uncertainty for the observed audit acceptance rates under the checklist.
We report these statistics as transparent release-audit diagnostics.

\subsection{Extending the watch conditions}
\label{app:watch-condition-extensions}
The same causal monitoring and first-alert protocol can support three extensions:
context-dependent relational conditions, such as following distance under changing
traffic speed, visibility, and road conditions; multi-agent interaction conditions,
such as a potential conflict during merging or at an intersection; and anticipatory
conditions under partial occlusion, such as a road user likely to enter the ego path.
These extensions require explicit alert criteria and annotations of relative motion
and relevant scene context. Building them calls for broader video sources covering
both routine and adverse driving, together with dedicated sampling and annotation.
Rare adverse scenarios and context-dependent trigger criteria make this a separate
curation effort.

\clearpage
\section{Prompt and Conversation Format}
\label{app:prompts}

\newcommand{\promptheading}[1]{%
  \vspace{0.65em}
  \noindent\textbf{#1}\par
  \vspace{0.25em}
}

This appendix lists the prompt templates and conversation format used in
DriveStreamBench. The placeholders \texttt{<image>} denote image inputs supplied
by the evaluation harness, and brace-marked fields such as
\texttt{\{question\}} denote text fields filled from the corresponding sample
metadata. L1--L3 use reactive question answering prompts. L4 uses a causal
streaming conversation format for watch-and-notify decisions.

\subsection{L1--L3 Reactive QA Prompts}
\label{app:prompt-l1l3}

For L1--L3, the model answers a question from the current visual input and,
when applicable, previously observed frames. Binary questions and
multiple-choice questions use separate system instructions so that the output
parser can remain deterministic.

\promptheading{Binary system instruction.}
\begin{PromptBlock}
You are a helpful assistant. Your task is to answer questions based on
continuously incoming video frames.

You are observing a driving scene over time. When asked a reactive question
about the scene, use the available frames, including current and prior frames
when provided, to determine the answer.

Output policy:
1) If the answer is yes, output exactly: YES
2) If the answer is no, output exactly: NO
3) Any other output is invalid.

Do not use XML tags or any extra text.
\end{PromptBlock}

\promptheading{Multiple-choice system instruction.}
\begin{PromptBlock}
You observe a front-facing driving scene over time. When asked a reactive
question, answer using only the available current and previous frames.

Output policy:
1) Output exactly one uppercase option letter from the listed choices,
   such as A, B, C, or D.
2) Any other output is invalid.

Do not output the option text, lowercase letters, numbers, XML tags, or any
extra text.
\end{PromptBlock}

\promptheading{Single-image user template.}
\begin{PromptBlock}
{question}
<image>
\end{PromptBlock}

\promptheading{Multiple-image user template.}
\begin{PromptBlock}
{question}
<image>
<image>
...
<image>
\end{PromptBlock}

\promptheading{Multiple-choice user template.}
\begin{PromptBlock}
<image>
{question}
A. {choice_a}
B. {choice_b}
C. {choice_c}
D. {choice_d}
\end{PromptBlock}

\begin{table}[h]
\centering
\small
\setlength{\tabcolsep}{4pt}
\renewcommand{\arraystretch}{1.08}
\caption{Prompt-template usage for L1--L3 reactive QA.}
\label{tab:app_prompt_l1l3_usage}
\begin{tabularx}{\linewidth}{@{}l
    >{\raggedright\arraybackslash}X
    >{\raggedright\arraybackslash}X
    >{\raggedright\arraybackslash}X@{}}
\toprule
\textbf{Level} & \textbf{Image input} & \textbf{Question format} & \textbf{Valid output} \\
\midrule
L1 & Single image & Binary or multiple-choice &
\texttt{YES}/\texttt{NO} or one option letter \\
L2 & Single image & Binary or multiple-choice &
\texttt{YES}/\texttt{NO} or one option letter \\
L3 & Multiple images & Binary temporal question &
\texttt{YES} or \texttt{NO} \\
\bottomrule
\end{tabularx}
\end{table}

\subsection{L4 Alert Decision Prompt and Conversation Format}
\label{app:prompt-l4}

L4 uses the same causal input convention but changes the response space from
question answering to alert decisions. The watch instruction is provided at the
first decision turn. In later turns, the model receives only newly sampled
frames; the prior instruction and earlier frames remain available through the
conversation history for streaming models.

\promptheading{System instruction.}
\begin{PromptBlock}
You monitor a front-facing driving video as a causal stream. At each decision
turn, decide whether the user-specified event is happening now, using only the
current frames and the previous history.

Output policy:
1) If the event is happening now, output exactly: ALERT
2) If the event is not happening now, output exactly: NO REPLY
3) Any other output is invalid.

Do not output XML tags or any extra text.
\end{PromptBlock}

\promptheading{Initial streaming user input.}
At the first L4 decision turn, the model receives the current sampled frames
together with the watch instruction:

\begin{PromptBlock}
{question}
<image>
<image>
\end{PromptBlock}

\promptheading{Subsequent streaming user input.}
At each later L4 decision turn, the model receives only the newly sampled
frames:

\begin{PromptBlock}
<image>
<image>
\end{PromptBlock}

\promptheading{Valid model outputs.}
At every L4 decision turn, the model must output exactly one of the following
two strings:

\begin{PromptBlock}
ALERT
NO REPLY
\end{PromptBlock}

\promptheading{Conversation example.}
The following example illustrates an L4 sample as a causal multi-turn conversation. The watch instruction appears only in the first user turn, together with the frames from the first decision interval. In each subsequent user turn, only the newly revealed frames are appended, while previous turns remain in the conversation history. At every turn, the model must decide whether to output \texttt{NO REPLY} or issue \texttt{ALERT}.

\begin{PromptBlock}
------------------------------------------------------------------------------------------
SYSTEM PROMPT:
------------------------------------------------------------------------------------------
You monitor a front-facing driving video as a causal stream. At each decision
turn, decide whether the user-specified event is happening now, using only the
current frames and the previous history.

Output policy:
1) If the event is happening now, output exactly: ALERT
2) If the event is not happening now, output exactly: NO REPLY
3) Any other output is invalid.

Do not output XML tags or any extra text.
------------------------------------------------------------------------------------------
User Turn1:
------------------------------------------------------------------------------------------
Alert me when the car in front starts.
<frame1>
<frame2>

Asst A1:
NO REPLY
------------------------------------------------------------------------------------------
User Turn2:
------------------------------------------------------------------------------------------
<frame3>
<frame4>

Asst A2:
NO REPLY
------------------------------------------------------------------------------------------
User Turn3:
------------------------------------------------------------------------------------------
<frame5>
<frame6>

Asst A3:
NO REPLY
------------------------------------------------------------------------------------------
User Turn4:
------------------------------------------------------------------------------------------
<frame7>
<frame8>

Asst A4:
ALERT
\end{PromptBlock}

\clearpage
\section{L4 Watch-Instruction Paraphrase Pools}
\label{app:l4-paraphrases}

This appendix lists the natural-language watch-instruction paraphrase pools used
for L4. The section lists separate train and test pools with no full-instruction overlap. For L4-Core,
the training sessions use the train pool and the test sessions use the test
pool. S6 remains an L4-OOD stress split with no training sessions in the
benchmark split; its train-side paraphrase pool is listed here for completeness. For traffic-light-change
instructions, the placeholders \(\{\mathrm{from\_color}\}\) and
\(\{\mathrm{to\_color}\}\) are instantiated by the corresponding signal
transition for each sample.

\subsection{S1: Lead-Vehicle Start}

\paragraph{Training paraphrases.}
\begin{enumerate}[leftmargin=1.8em,itemsep=0.15em,topsep=0.2em]
\item Please monitor the front vehicle and alert me when it starts moving.
\item Let me know when the car ahead begins to move.
\item Warn me when the leading vehicle starts moving.
\item Tell me as soon as the front car starts moving forward.
\item Alert me the moment the vehicle in front begins moving.
\item Give me a heads-up once the car ahead begins moving.
\item Notify me when the vehicle in front pulls forward.
\item Watch the car in front and tell me when it starts moving.
\item I need to know when the lead vehicle begins to move.
\item Keep an eye on the front car and let me know once it starts moving.
\item Flag it when the vehicle ahead starts rolling.
\item Please inform me the moment the car in front starts moving.
\item Signal me when the leading vehicle begins moving from a stop.
\item Let me know right away when the car ahead starts moving.
\item Alert me if the front vehicle begins moving forward.
\item Tell me when the stopped vehicle ahead starts to move.
\item Watch for the lead car starting to move and warn me.
\item Give me a warning when the vehicle in front gets moving.
\item Monitor the car ahead and notify me when it starts moving.
\item Heads up when the front vehicle starts to move forward.
\item I want to be alerted when the car in front begins moving.
\item Let me know whenever the lead vehicle starts moving.
\item Warn me as soon as the vehicle ahead moves forward.
\item Please watch the front car and tell me when it starts moving.
\item Can you alert me when the car ahead starts moving?
\end{enumerate}

\paragraph{Test paraphrases.}
\begin{enumerate}[leftmargin=1.8em,itemsep=0.15em,topsep=0.2em]
\item Notify me when the vehicle ahead starts moving forward.
\item Give me a heads-up the moment the front car begins to move.
\item Warn me when the car in front starts moving.
\item Tell me if the leading vehicle begins to move.
\item Let me know as soon as the car ahead starts rolling.
\item Flag it when the stopped vehicle in front starts moving.
\item I would like a warning when the front car starts to move.
\item Alert me when the lead car moves forward from a stop.
\item Watch the vehicle in front and inform me when it begins moving.
\item Give me a heads-up if the car ahead pulls away.
\item Inform me when the front vehicle begins moving.
\item Please notify me when the car in front starts moving from a stop.
\item Keep me updated if the lead vehicle starts moving.
\item Signal me the moment the vehicle ahead starts moving.
\item Can you let me know when the front car begins to move?
\item I need a notification when the car ahead starts rolling forward.
\item Monitor the vehicle in front and alert me when it begins to move.
\item Tell me immediately when the leading car begins to move.
\item Heads up if the front vehicle starts moving.
\item Warn me when the car directly ahead begins moving.
\end{enumerate}

\subsection{S2: Lead-Vehicle Brake}

\paragraph{Training paraphrases.}
\begin{enumerate}[leftmargin=1.8em,itemsep=0.15em,topsep=0.2em]
\item Alert me when the lead vehicle brakes.
\item Warn me if the car ahead starts braking.
\item Let me know when the vehicle in front hits the brakes.
\item Notify me when the leading car begins to brake.
\item Tell me when the car ahead brakes.
\item Give me a heads-up if the vehicle in front starts to brake.
\item Please watch the car ahead and alert me when it brakes.
\item Flag it if the lead vehicle brakes.
\item Monitor the vehicle ahead and warn me when it starts braking.
\item Keep an eye on the car in front and let me know when it brakes.
\item I need to know when the front vehicle brakes.
\item Alert me the moment the car ahead starts to brake.
\item Watch the lead car and tell me when it begins braking.
\item Give me a warning when the vehicle ahead brakes.
\item Give me a quick heads-up when the front car brakes.
\item Please inform me whenever the leading vehicle brakes.
\item Signal me when the front car starts braking.
\item Let me know right away when the vehicle ahead brakes.
\item Warn me as soon as the lead vehicle begins to brake.
\item Alert me if the car directly ahead starts braking.
\item Tell me whenever the front car brakes.
\item If you notice the vehicle ahead braking, let me know.
\item Can you alert me when the lead vehicle brakes?
\item I want to know when the car ahead starts to brake.
\item Please monitor the vehicle in front for braking and alert me.
\end{enumerate}

\paragraph{Test paraphrases.}
\begin{enumerate}[leftmargin=1.8em,itemsep=0.15em,topsep=0.2em]
\item Notify me when the car ahead begins braking.
\item As soon as the front vehicle starts braking, alert me.
\item Warn me whenever the leading vehicle brakes.
\item Tell me if the car in front begins to brake.
\item Let me know the moment the lead car brakes.
\item Please let me know if the vehicle ahead brakes.
\item I would like a warning when the front car starts braking.
\item Alert me if the car ahead applies the brakes.
\item Keep watch on the vehicle in front and notify me once it starts braking.
\item Give me a heads-up when the leading car brakes.
\item Inform me right away if the vehicle in front begins to brake.
\item Keep me posted if the car ahead starts braking.
\item Can you warn me when the front vehicle brakes?
\item Signal me if the lead car begins braking.
\item I need an alert whenever the vehicle ahead starts to brake.
\item Please tell me when the car directly in front begins braking.
\item Monitor the lead vehicle and flag when it starts braking.
\item Heads up when the front car brakes.
\item Warn me if you see the vehicle ahead start to brake.
\item Let me know if the lead car begins to brake.
\end{enumerate}

\subsection{S3: Traffic-Light Change}

\paragraph{Training paraphrases.}
\begin{enumerate}[leftmargin=1.8em,itemsep=0.15em,topsep=0.2em]
\item Alert me when the traffic light changes from \(\{\mathrm{from\_color}\}\) to \(\{\mathrm{to\_color}\}\).
\item Warn me if the traffic signal changes from \(\{\mathrm{from\_color}\}\) to \(\{\mathrm{to\_color}\}\).
\item Let me know when the light changes from \(\{\mathrm{from\_color}\}\) to \(\{\mathrm{to\_color}\}\).
\item Notify me when the traffic light switches from \(\{\mathrm{from\_color}\}\) to \(\{\mathrm{to\_color}\}\).
\item Tell me as soon as the signal changes from \(\{\mathrm{from\_color}\}\) to \(\{\mathrm{to\_color}\}\).
\item Give me a heads-up when the traffic light turns from \(\{\mathrm{from\_color}\}\) to \(\{\mathrm{to\_color}\}\).
\item Please watch the traffic light and alert me when it goes from \(\{\mathrm{from\_color}\}\) to \(\{\mathrm{to\_color}\}\).
\item Flag it when the traffic signal changes from \(\{\mathrm{from\_color}\}\) to \(\{\mathrm{to\_color}\}\).
\item Monitor the traffic light and warn me when it changes from \(\{\mathrm{from\_color}\}\) to \(\{\mathrm{to\_color}\}\).
\item Keep an eye on the signal and let me know when it changes from \(\{\mathrm{from\_color}\}\) to \(\{\mathrm{to\_color}\}\).
\item I need to know when the traffic light changes from \(\{\mathrm{from\_color}\}\) to \(\{\mathrm{to\_color}\}\).
\item Alert me the moment the traffic signal switches from \(\{\mathrm{from\_color}\}\) to \(\{\mathrm{to\_color}\}\).
\item Watch the traffic light and tell me when it turns from \(\{\mathrm{from\_color}\}\) to \(\{\mathrm{to\_color}\}\).
\item Warn me as soon as the traffic light changes from \(\{\mathrm{from\_color}\}\) to \(\{\mathrm{to\_color}\}\).
\item Alert me when the light at the intersection changes from \(\{\mathrm{from\_color}\}\) to \(\{\mathrm{to\_color}\}\).
\item Tell me whenever the traffic signal shifts from \(\{\mathrm{from\_color}\}\) to \(\{\mathrm{to\_color}\}\).
\item Watch for the traffic light changing from \(\{\mathrm{from\_color}\}\) to \(\{\mathrm{to\_color}\}\) and notify me.
\item Can you alert me when the traffic light changes from \(\{\mathrm{from\_color}\}\) to \(\{\mathrm{to\_color}\}\)?
\item I want to be warned when the signal ahead changes from \(\{\mathrm{from\_color}\}\) to \(\{\mathrm{to\_color}\}\).
\item Please monitor the traffic light for a change from \(\{\mathrm{from\_color}\}\) to \(\{\mathrm{to\_color}\}\).
\item Give me a warning when the traffic light goes from \(\{\mathrm{from\_color}\}\) to \(\{\mathrm{to\_color}\}\).
\item Let me know right away when the signal changes from \(\{\mathrm{from\_color}\}\) to \(\{\mathrm{to\_color}\}\).
\item Inform me if the traffic light transitions from \(\{\mathrm{from\_color}\}\) to \(\{\mathrm{to\_color}\}\).
\item Signal me when the traffic light changes from \(\{\mathrm{from\_color}\}\) to \(\{\mathrm{to\_color}\}\).
\item Please tell me when the light ahead changes from \(\{\mathrm{from\_color}\}\) to \(\{\mathrm{to\_color}\}\).
\end{enumerate}

\paragraph{Test paraphrases.}
\begin{enumerate}[leftmargin=1.8em,itemsep=0.15em,topsep=0.2em]
\item Notify me when the traffic signal turns from \(\{\mathrm{from\_color}\}\) to \(\{\mathrm{to\_color}\}\).
\item Give me a heads-up once the traffic light changes from \(\{\mathrm{from\_color}\}\) to \(\{\mathrm{to\_color}\}\).
\item Warn me whenever the signal goes from \(\{\mathrm{from\_color}\}\) to \(\{\mathrm{to\_color}\}\).
\item Tell me if the traffic light ahead changes from \(\{\mathrm{from\_color}\}\) to \(\{\mathrm{to\_color}\}\).
\item Let me know the moment the traffic signal changes from \(\{\mathrm{from\_color}\}\) to \(\{\mathrm{to\_color}\}\).
\item Please flag it when the traffic light changes from \(\{\mathrm{from\_color}\}\) to \(\{\mathrm{to\_color}\}\).
\item I would like a warning when the traffic light switches from \(\{\mathrm{from\_color}\}\) to \(\{\mathrm{to\_color}\}\).
\item Alert me if the signal at the intersection changes from \(\{\mathrm{from\_color}\}\) to \(\{\mathrm{to\_color}\}\).
\item Watch the traffic light and inform me when it changes from \(\{\mathrm{from\_color}\}\) to \(\{\mathrm{to\_color}\}\).
\item Give me a heads-up when the signal changes from \(\{\mathrm{from\_color}\}\) to \(\{\mathrm{to\_color}\}\).
\item Inform me when the traffic signal shifts from \(\{\mathrm{from\_color}\}\) to \(\{\mathrm{to\_color}\}\).
\item Keep me posted if the traffic light changes from \(\{\mathrm{from\_color}\}\) to \(\{\mathrm{to\_color}\}\).
\item Can you warn me when the traffic signal changes from \(\{\mathrm{from\_color}\}\) to \(\{\mathrm{to\_color}\}\)?
\item Signal me if the traffic light turns from \(\{\mathrm{from\_color}\}\) to \(\{\mathrm{to\_color}\}\).
\item I need an alert whenever the traffic light changes from \(\{\mathrm{from\_color}\}\) to \(\{\mathrm{to\_color}\}\).
\item Let me know when the light ahead goes from \(\{\mathrm{from\_color}\}\) to \(\{\mathrm{to\_color}\}\).
\item Warn me once the signal changes from \(\{\mathrm{from\_color}\}\) to \(\{\mathrm{to\_color}\}\).
\item Tell me when the signal turns from \(\{\mathrm{from\_color}\}\) to \(\{\mathrm{to\_color}\}\).
\item Give me a heads-up if the light changes from \(\{\mathrm{from\_color}\}\) to \(\{\mathrm{to\_color}\}\).
\item I'd like to know when the traffic light goes from \(\{\mathrm{from\_color}\}\) to \(\{\mathrm{to\_color}\}\).
\end{enumerate}

\subsection{S4: Adjacent Cut-In}

\paragraph{Training paraphrases.}
\begin{enumerate}[leftmargin=1.8em,itemsep=0.15em,topsep=0.2em]
\item Let me know if a car starts cutting into my lane.
\item Heads up when someone's coming over into my lane.
\item Tell me if a car's cutting in on me.
\item Warn me when someone starts merging over into my lane.
\item Give me a heads up if a car's coming over.
\item Let me know the second someone starts drifting into my lane.
\item Tell me if a car next to me starts coming over.
\item Flag it when someone starts cutting in.
\item Let me know if a car's starting to merge into my lane.
\item Give me a shout if someone starts cutting into my lane.
\item Tell me when a car starts pulling into my lane.
\item Heads up if someone's edging into my lane.
\item Watch the next lane --- tell me if anyone starts coming over.
\item Warn me if someone starts cutting in on me.
\item Let me know if a neighbor car's coming over.
\item Give me a heads up when a car's coming in from the side.
\item Tell me the second someone starts cutting in.
\item Keep an eye on the next lane --- tell me if a car's coming over.
\item Heads up if a car's starting to merge into my lane.
\item Let me know if someone's drifting over into my lane.
\item Tell me if a car from the side's starting to cut in.
\item Warn me if someone next to me's moving in on me.
\item Give me a shout when a car's coming over from the side.
\item Flag any car that starts coming into my lane.
\item Let me know right away if someone's cutting into my lane.
\end{enumerate}

\paragraph{Test paraphrases.}
\begin{enumerate}[leftmargin=1.8em,itemsep=0.15em,topsep=0.2em]
\item Tell me if another car starts cutting in on me.
\item Give me a heads up when someone starts coming into my lane.
\item Let me know if a car's coming over into mine.
\item Warn me the second someone starts cutting in.
\item Heads up if a car's pulling into my lane.
\item Tell me if someone's starting to drift into my lane.
\item Let me know if a car next to me starts cutting in.
\item Flag it when someone's coming over into my lane.
\item Give me a shout if a car starts merging into my lane.
\item Warn me if someone's edging over into my lane.
\item Tell me when a car starts coming into my lane from the side.
\item Let me know if someone starts squeezing into my lane.
\item Heads up if a neighbor car's cutting in.
\item Watch my sides --- tell me if anyone's coming over.
\item Give me a heads up the moment someone starts merging in.
\item Tell me if a car's starting to pull into my lane.
\item Let me know if someone starts cutting in on me.
\item Warn me if a car's coming in from the side.
\item Heads up when someone starts moving into my lane.
\item Tell me right away if a car's cutting into my lane.
\end{enumerate}

\subsection{S5: VRU Crossing}

\paragraph{Training paraphrases.}
\begin{enumerate}[leftmargin=1.8em,itemsep=0.15em,topsep=0.2em]
\item Alert me if a pedestrian or cyclist crosses the road ahead.
\item Alert me if someone on foot or on a bicycle crosses the road ahead.
\item Give me a heads-up if a pedestrian or cyclist starts crossing the road ahead.
\item Give me a warning if a pedestrian or cyclist moves into the roadway ahead.
\item Let me know if a pedestrian or cyclist starts crossing the street.
\item Let me know if anyone on foot or on a bike is crossing ahead.
\item Let me know when a pedestrian or cyclist crosses the road in front of me.
\item Tell me if a pedestrian or cyclist crosses the road ahead.
\item Tell me if anyone walking or biking crosses the road ahead.
\item Tell me when someone walking or riding a bike starts crossing ahead of me.
\item Warn me if a pedestrian or cyclist starts moving across the road ahead.
\item Warn me when pedestrians or cyclists are crossing the road ahead.
\item Alert me if a pedestrian or cyclist steps onto the road ahead.
\item Notify me when someone on foot or on a bike crosses the road ahead.
\item Let me know if a pedestrian or cyclist is stepping into the roadway ahead.
\item Watch for pedestrians or cyclists crossing ahead and warn me.
\item Flag it if someone on foot or on a bike enters the road ahead.
\item I need to know when a pedestrian or cyclist is crossing ahead.
\item Please watch for pedestrians or cyclists crossing the road and alert me.
\item Warn me right away if a pedestrian or cyclist enters the road ahead.
\item Give me a warning when someone walks or bikes across the street in front of me.
\item Tell me immediately if a pedestrian or cyclist steps into the road ahead.
\item Alert me whenever someone on foot or on a bike crosses the road ahead.
\item Monitor the road ahead for pedestrians or cyclists crossing and notify me.
\item Keep an eye out for pedestrians or cyclists crossing ahead and let me know.
\end{enumerate}

\paragraph{Test paraphrases.}
\begin{enumerate}[leftmargin=1.8em,itemsep=0.15em,topsep=0.2em]
\item Alert me if a pedestrian or cyclist is crossing the street ahead.
\item Alert me when someone on foot or on a bike is crossing up ahead.
\item Give me a heads-up when someone on foot or on a bike crosses in front of me.
\item Give me a warning when a pedestrian or cyclist is crossing the street in front of me.
\item Let me know if there is a pedestrian or cyclist crossing ahead.
\item Let me know when a pedestrian or cyclist is crossing the street ahead.
\item Tell me whenever a pedestrian or cyclist crosses in front of me.
\item Warn me when a pedestrian or cyclist starts crossing in front of me.
\item Notify me if a pedestrian or cyclist crosses the road ahead.
\item Warn me if someone walking or riding a bicycle moves into the roadway ahead.
\item Tell me if a person on foot or on a bike is crossing the road in front.
\item Give me a heads-up if a pedestrian or cyclist steps onto the street ahead.
\item Let me know the moment a pedestrian or cyclist enters the roadway ahead.
\item Alert me when a pedestrian or cyclist starts crossing ahead.
\item I would like a warning if someone is crossing the road ahead on foot or by bike.
\item Signal me when a pedestrian or cyclist begins crossing ahead.
\item Flag any pedestrians or cyclists crossing the road ahead.
\item Watch for pedestrians or cyclists crossing and alert me immediately.
\item Inform me if anyone on foot or cycling is crossing the road ahead.
\item Please warn me when pedestrians or cyclists step into the road ahead.
\end{enumerate}

\subsection{S6: Sudden Appearance}

\paragraph{Training paraphrases.}
\begin{enumerate}[leftmargin=1.8em,itemsep=0.15em,topsep=0.2em]
\item Please monitor the surroundings and alert me when a pedestrian or vehicle suddenly emerges from an occluded area.
\item Warn me if a pedestrian or vehicle suddenly appears from a blind spot.
\item Let me know when a pedestrian or vehicle suddenly comes out from behind an obstruction.
\item Notify me if a pedestrian or vehicle suddenly emerges from a hidden area.
\item Tell me when a pedestrian or vehicle suddenly appears from a concealed spot.
\item Give me a heads-up if a pedestrian or vehicle suddenly comes out from behind cover.
\item Please watch for pedestrians or vehicles suddenly appearing from occluded areas.
\item Flag it if a pedestrian or vehicle suddenly emerges from a blind zone.
\item Monitor the surroundings and warn me if a pedestrian or vehicle suddenly darts out.
\item Keep an eye out for pedestrians or vehicles suddenly appearing from hidden spots.
\item I need to know when a pedestrian or vehicle suddenly appears from an obscured area.
\item Alert me the moment a pedestrian or vehicle suddenly comes out from behind an obstacle.
\item Watch for pedestrians or vehicles darting out from blind areas and tell me.
\item Give me a warning if a pedestrian or vehicle suddenly emerges from an occluded zone.
\item Heads up if a pedestrian or vehicle suddenly appears from a concealed area.
\item Please inform me whenever a pedestrian or vehicle suddenly darts out from behind cover.
\item Signal me when a pedestrian or vehicle suddenly emerges from a hidden spot.
\item Let me know right away if a pedestrian or vehicle suddenly emerges from a blind area.
\item Warn me as soon as a pedestrian or vehicle suddenly comes out from behind an obstruction.
\item Alert me if a pedestrian or vehicle suddenly appears from an area outside my view.
\item Tell me when a pedestrian or vehicle suddenly emerges from a blocked line of sight.
\item Watch for pedestrians or vehicles suddenly appearing from concealed spots and notify me.
\item Can you alert me when a pedestrian or vehicle suddenly darts out from hiding?
\item I want to be warned when a pedestrian or vehicle suddenly appears from a blind spot.
\item Please monitor for pedestrians or vehicles suddenly appearing from occluded areas ahead.
\end{enumerate}

\paragraph{Test paraphrases.}
\begin{enumerate}[leftmargin=1.8em,itemsep=0.15em,topsep=0.2em]
\item Notify me when a pedestrian or vehicle suddenly comes out from a hidden area.
\item Give me a heads-up if a pedestrian or vehicle suddenly darts out from behind an obstacle.
\item Warn me whenever a pedestrian or vehicle suddenly emerges from a blind spot.
\item Tell me if a pedestrian or vehicle suddenly appears from an obscured zone.
\item Let me know the moment a pedestrian or vehicle suddenly appears from a concealed area.
\item Please flag any sudden appearance of a pedestrian or vehicle from an occluded spot.
\item I would like a warning when a pedestrian or vehicle suddenly darts out from behind cover.
\item Alert me if a pedestrian or vehicle suddenly emerges from a hidden zone.
\item Watch the surroundings and inform me if a pedestrian or vehicle suddenly darts out.
\item Give me a heads-up when a pedestrian or vehicle suddenly appears from a blind area.
\item Inform me right away if a pedestrian or vehicle suddenly shows up from behind an obstruction.
\item Keep me posted if a pedestrian or vehicle suddenly appears from a concealed spot.
\item Can you warn me when a pedestrian or vehicle darts out from an occluded area?
\item Signal me if a pedestrian or vehicle suddenly appears from a hidden spot.
\item I need an alert whenever a pedestrian or vehicle suddenly emerges from a hidden area.
\item Please tell me when a pedestrian or vehicle suddenly appears from a blind zone.
\item Monitor the surroundings and flag it if a pedestrian or vehicle suddenly appears from an obscured area.
\item Heads up when a pedestrian or vehicle suddenly darts out from behind cover.
\item Warn me if you spot a pedestrian or vehicle suddenly appearing from a hidden area.
\item Let me know if a pedestrian or vehicle suddenly emerges from a spot outside my view.
\end{enumerate}

\providecommand{\DSBExampleFigDir}{figures/examples}

\providecommand{\DSBStripImgW}{0.245\linewidth}
\providecommand{\DSBSinglePairImgW}{0.98\linewidth}

\newenvironment{DSBLevelBlock}{%
    \par\smallskip\noindent\begin{minipage}{\linewidth}\small
}{%
    \end{minipage}\par\vspace{0.6em}
}

\newcommand{\DSBPromptHeader}[3]{%
    \noindent
    \begin{minipage}[t]{0.22\linewidth}
        \raggedright
        \textbf{#1}\par
        {\footnotesize #2}
    \end{minipage}\hfill
    \begin{minipage}[t]{0.76\linewidth}
        \raggedright
        #3
    \end{minipage}\par
}

\newcommand{\DSBFourFrames}[8]{%
    \par\vspace{0.35em}
    \begingroup
    \centering
    \setlength{\tabcolsep}{0pt}
    \begin{tabular*}{\linewidth}{@{\extracolsep{\fill}}cccc@{}}
        \includegraphics[width=\DSBStripImgW]{\DSBExampleFigDir/#1} &
        \includegraphics[width=\DSBStripImgW]{\DSBExampleFigDir/#2} &
        \includegraphics[width=\DSBStripImgW]{\DSBExampleFigDir/#3} &
        \includegraphics[width=\DSBStripImgW]{\DSBExampleFigDir/#4} \\
        {\scriptsize #5} & {\scriptsize #6} & {\scriptsize #7} & {\scriptsize #8}
    \end{tabular*}
    \par
    \endgroup
}

\providecommand{\DSBSinglePairPromptH}{5em}
\providecommand{\DSBSinglePromptImageGap}{0.65em}

\newcommand{\DSBOptionGrid}[4]{%
    \par\vspace{0.25em}
    {\scriptsize
    \setlength{\tabcolsep}{0pt}
    \begin{tabular}{@{}p{0.46\linewidth}@{\hspace{0.06\linewidth}}p{0.46\linewidth}@{}}
        A. #1 & B. #2 \\
        C. #3 & D. #4
    \end{tabular}}%
}

\newcommand{\DSBCardText}[3]{%
    {\small\textbf{#1}}\par
    {\footnotesize #2}\par
    \vspace{0.35em}
    {\small
    \raggedright
    \emergencystretch=1.5em
    #3\par}%
}

\newcommand{\DSBSingleCard}[3]{%
    \begin{minipage}[t]{\linewidth}
        \raggedright

        \begin{minipage}[t][\DSBSinglePairPromptH][t]{\linewidth}
            \raggedright
            #1
        \end{minipage}

        \par\vspace{\DSBSinglePromptImageGap}

        \centering
        \includegraphics[width=\DSBSinglePairImgW]{\DSBExampleFigDir/#2}\par
        \vspace{0.25em}
        {\scriptsize #3}
    \end{minipage}%
}

\newcommand{\DSBSinglePair}[6]{%
    \par\vspace{0.2em}\noindent
    \begin{tabularx}{\linewidth}{@{}X@{\hspace{0.04\linewidth}}X@{}}
        \DSBSingleCard{#1}{#2}{#3}
        &
        \DSBSingleCard{#4}{#5}{#6}
    \end{tabularx}
    \par\vspace{0.75em}
}

\clearpage
\section{\texorpdfstring{Event-Linked L1--L4 Task Examples}{Event-Linked L1--L4 Task Examples}}
\label{app:l1l4-examples}

Figures~\ref{fig:l1234-example-lead-start}--\ref{fig:l1234-example-vru-crossing}
show event-linked L1--L4 task examples from the DriveStreamBench core subcategories. Each example groups tasks built around the same monitored driving event. These examples illustrate the task construction and input format. In each example, the L4 block is the streaming watch-and-notify task, the L3 block is the corresponding temporal probe over the same event, and the L1/L2 blocks provide single-frame foundation probes when such probes are available.

\subsection{Example 1: Lead Vehicle Starts Moving}
\label{app:example-lead-start}

\noindent
\begin{minipage}{\textwidth}

\begin{DSBLevelBlock}
\DSBPromptHeader
{L4: watch-and-notify}
{Streaming alert decision}
{{\emph{Please watch the front car and tell me when it starts moving.}}}
\DSBFourFrames
{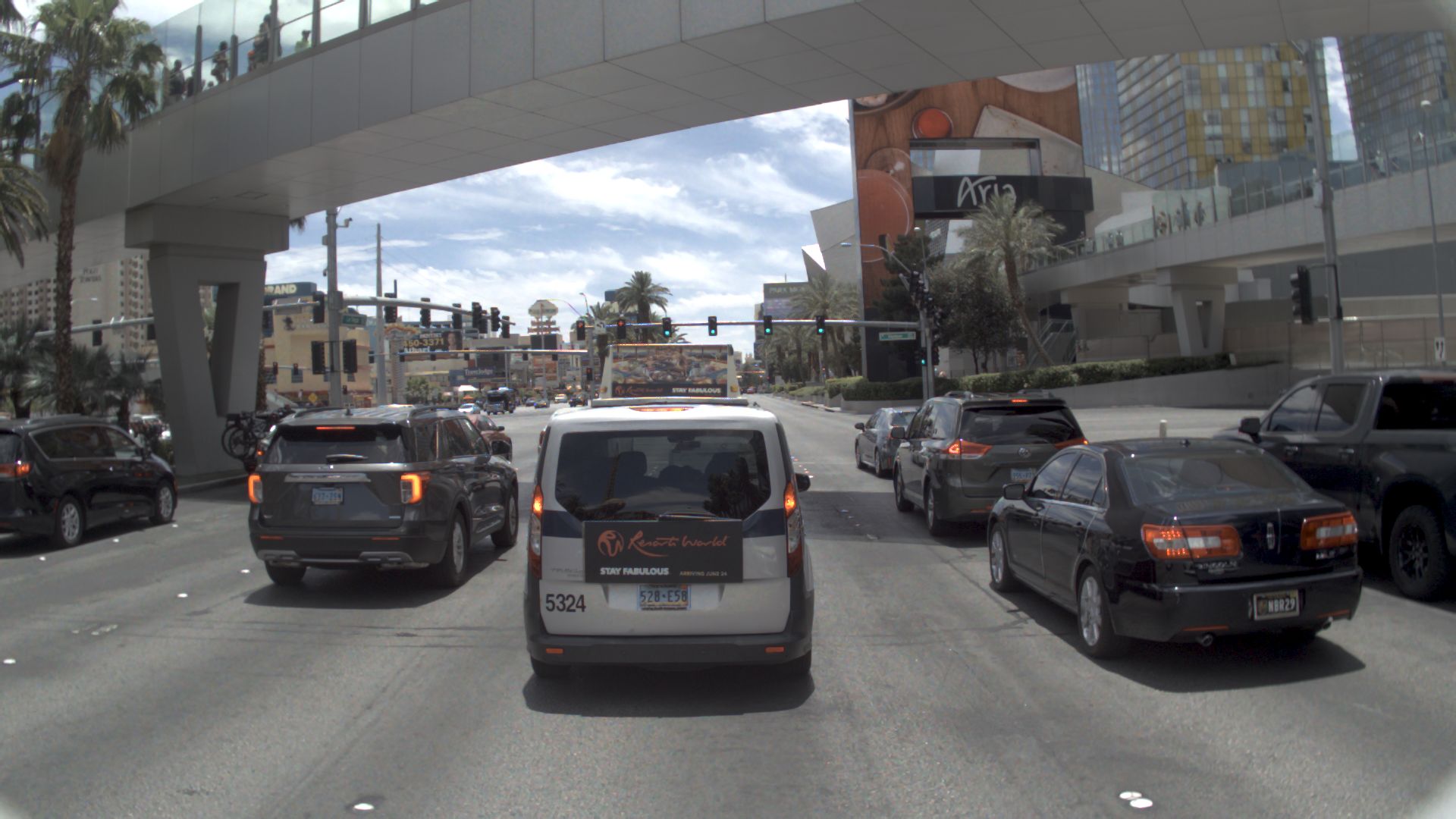}{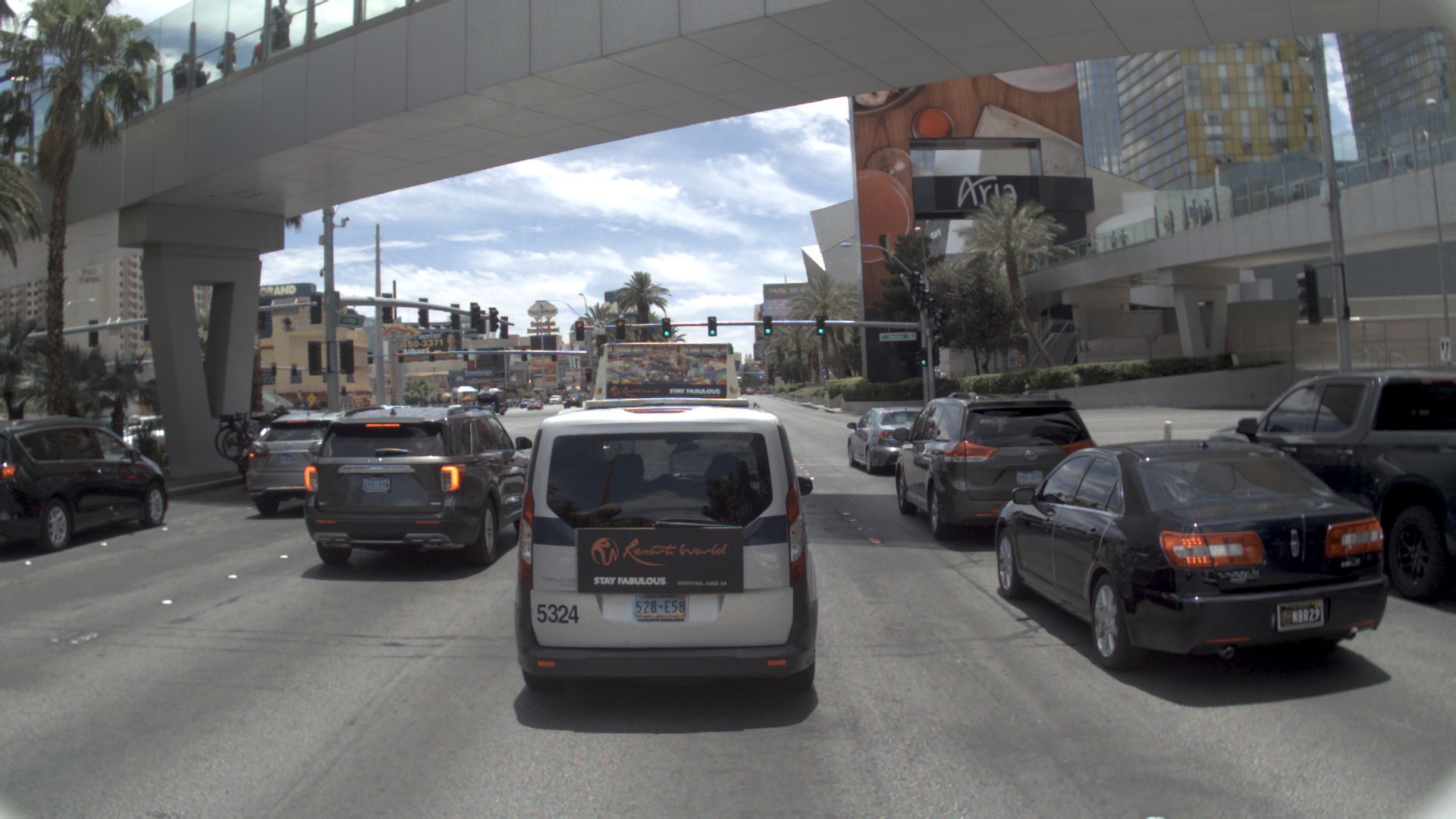}{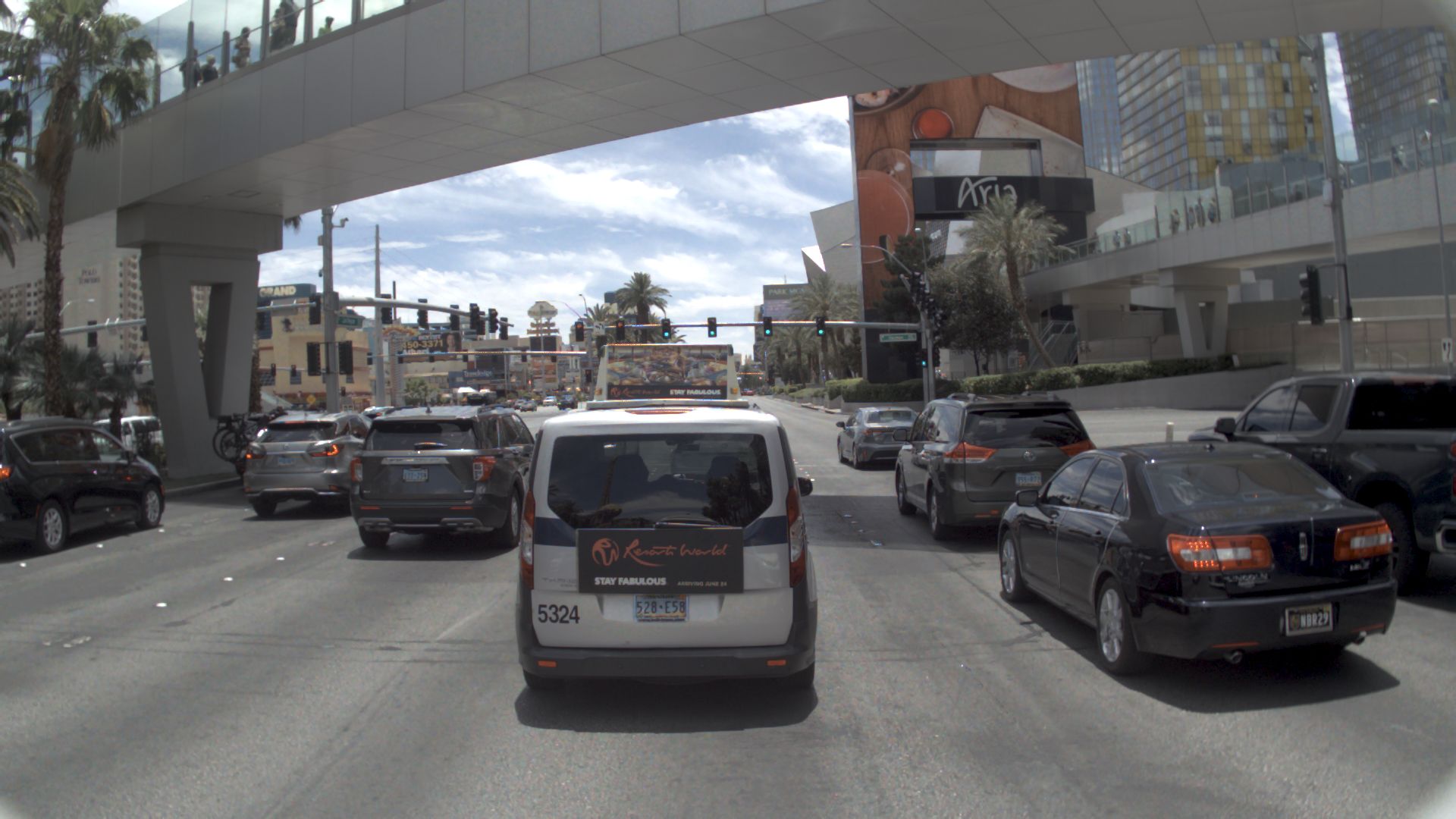}{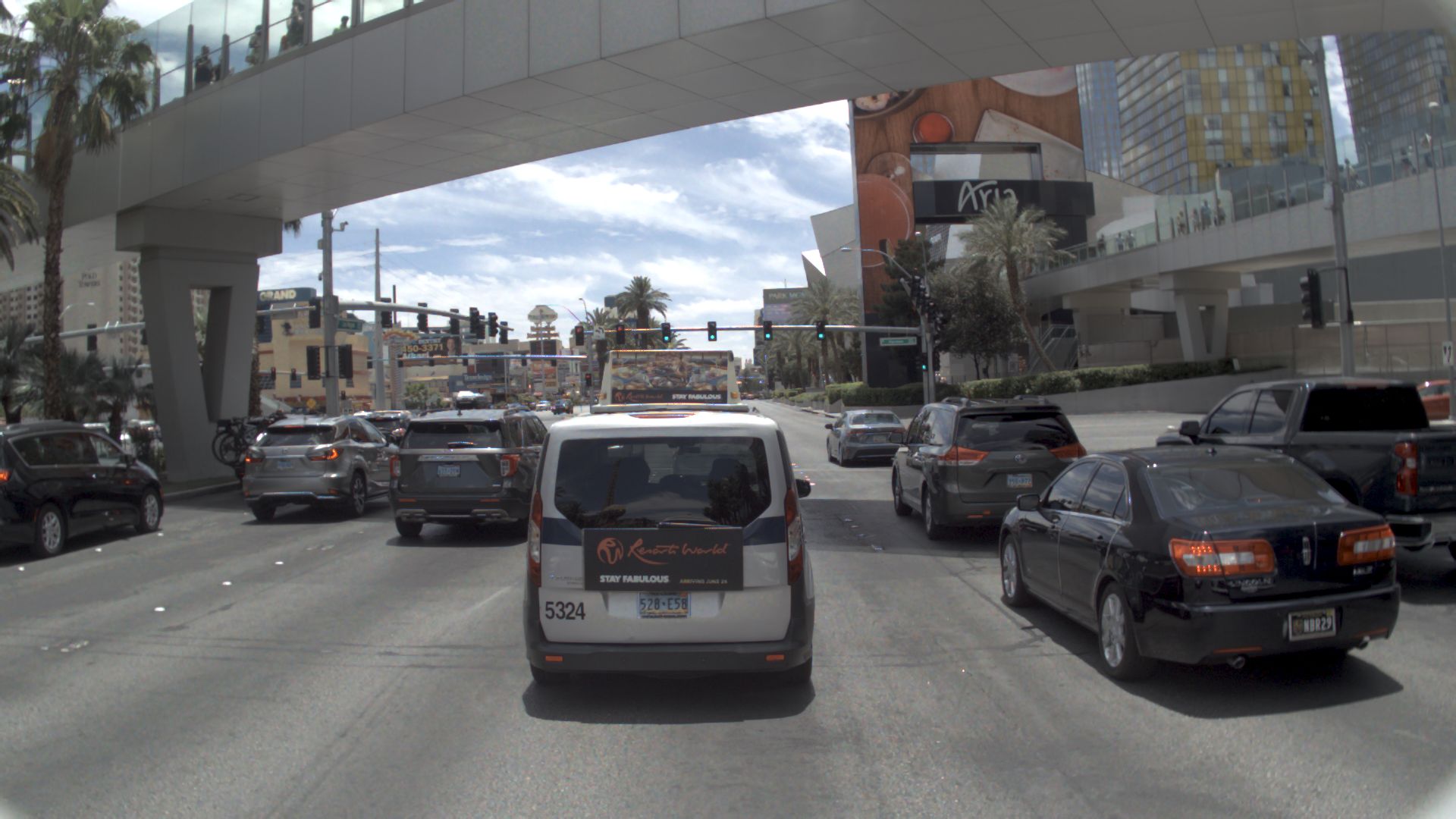}
{stream}{stream}{\textbf{trigger}}{stream}
\end{DSBLevelBlock}

\begin{DSBLevelBlock}
\DSBPromptHeader
{L3: temporal comprehension}
{Corresponding temporal probe}
{{\emph{Did the car ahead just pull away from a stop?}}}
\DSBFourFrames
{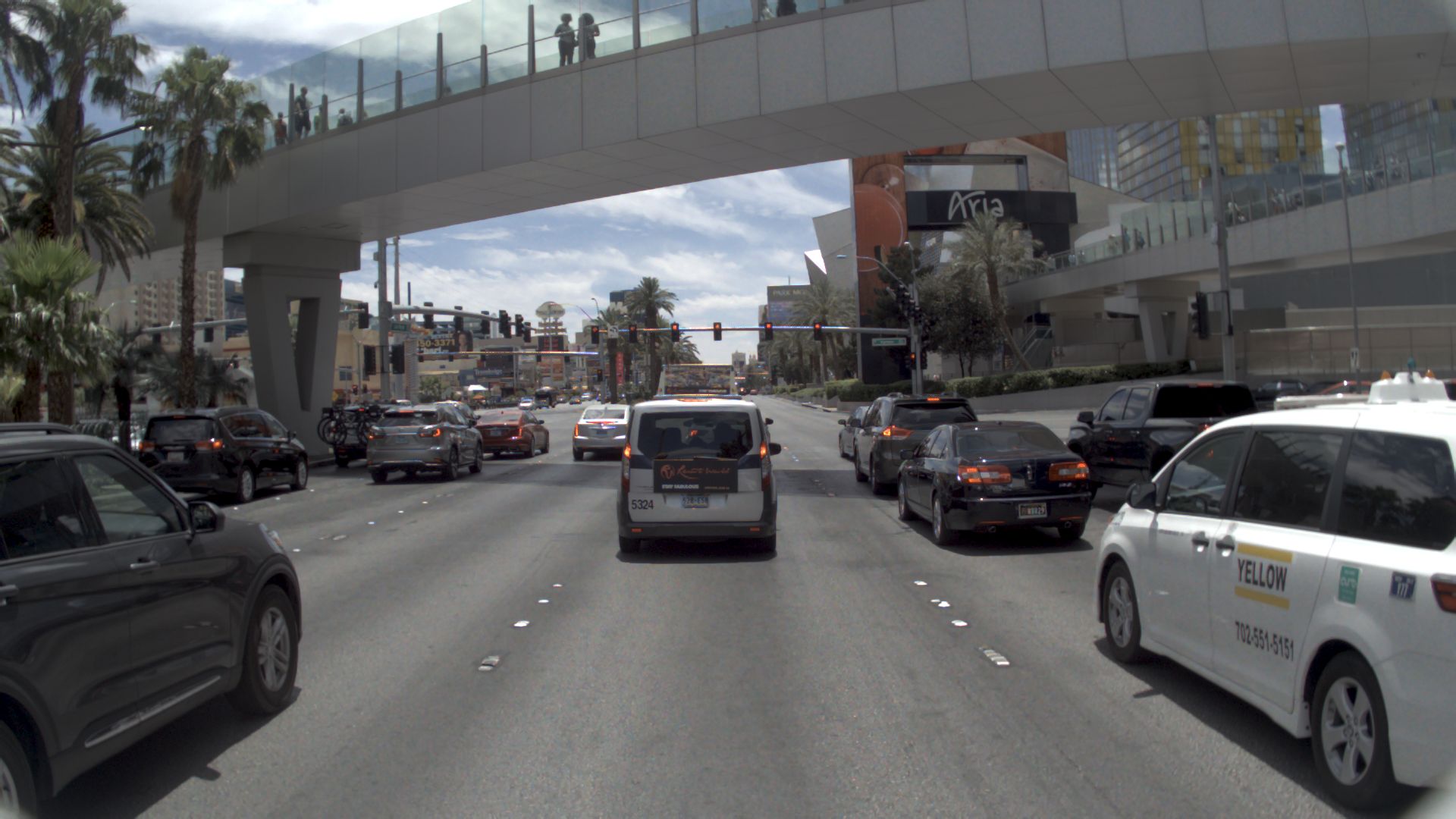}{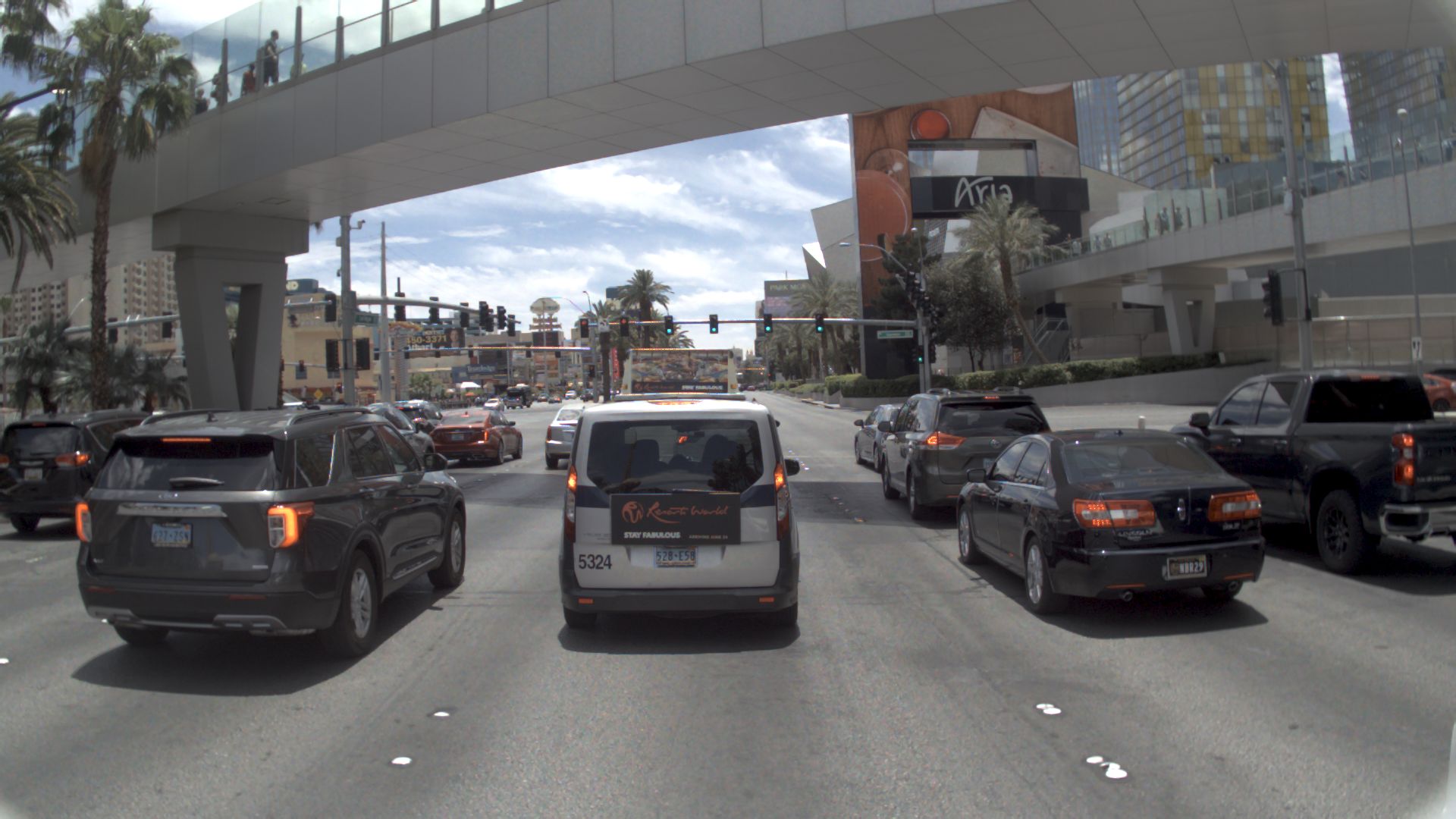}{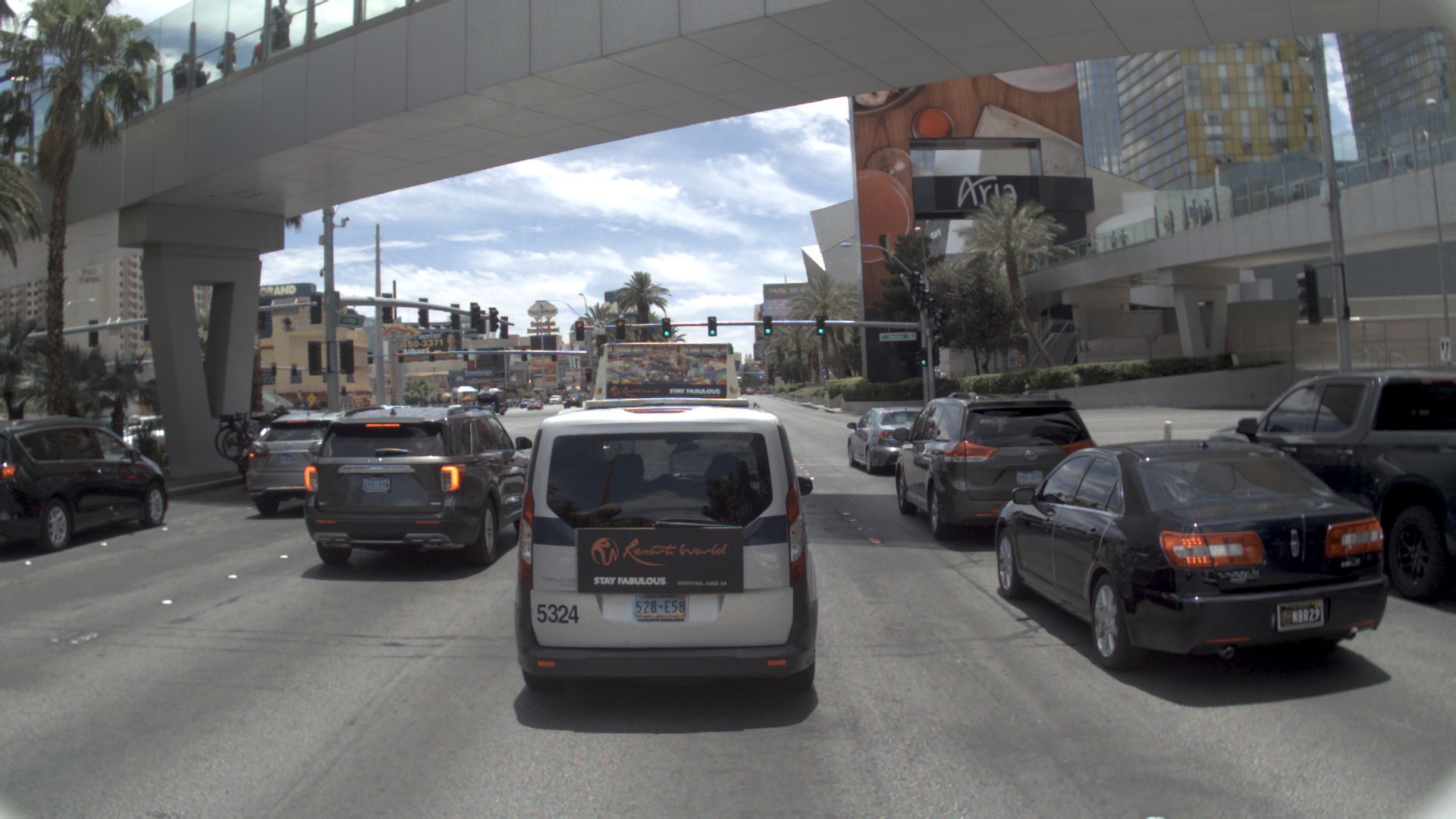}{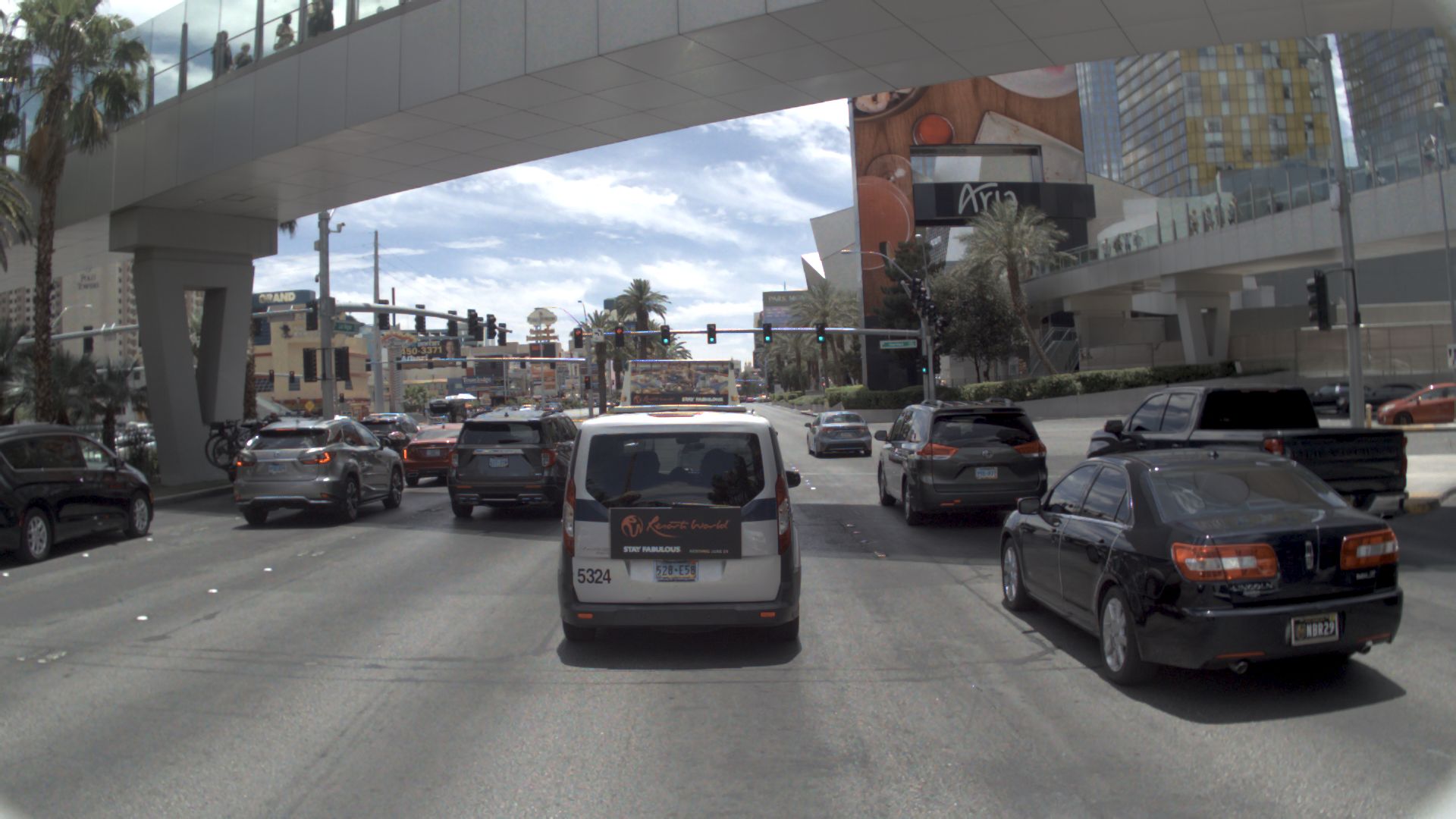}
{$t_0$}{$t_1$}{$t_2$}{$t_3$}
\end{DSBLevelBlock}

\DSBSinglePair
{%
\DSBCardText
{L2: relational understanding}
{Single-frame ego-centric relation probe}
{{\emph{Is there a vehicle in the adjacent lane to the right of the ego vehicle?}}}
}
{ex1_l2.jpg}
{L2 single-image frame}
{%
\DSBCardText
{L1: recognition}
{Single-frame recognition probe}
{{\emph{Can you see pedestrians or cyclists in the scene?}}}
}
{ex1_l1.jpg}
{L1 single-image frame}

\vspace{0.25em}
\captionof{figure}{Event-linked L1--L4 task example for a lead-vehicle-start event.}
\label{fig:l1234-example-lead-start}
\end{minipage}

\clearpage
\Needspace{0.88\textheight}
\subsection{Example 2: Lead Vehicle Brakes}
\label{app:example-lead-brake}

\noindent
\begin{minipage}{\textwidth}

\begin{DSBLevelBlock}
\DSBPromptHeader
{L4: watch-and-notify}
{Streaming alert decision}
{{\emph{Can you alert me when the lead vehicle brakes?}}}
\DSBFourFrames
{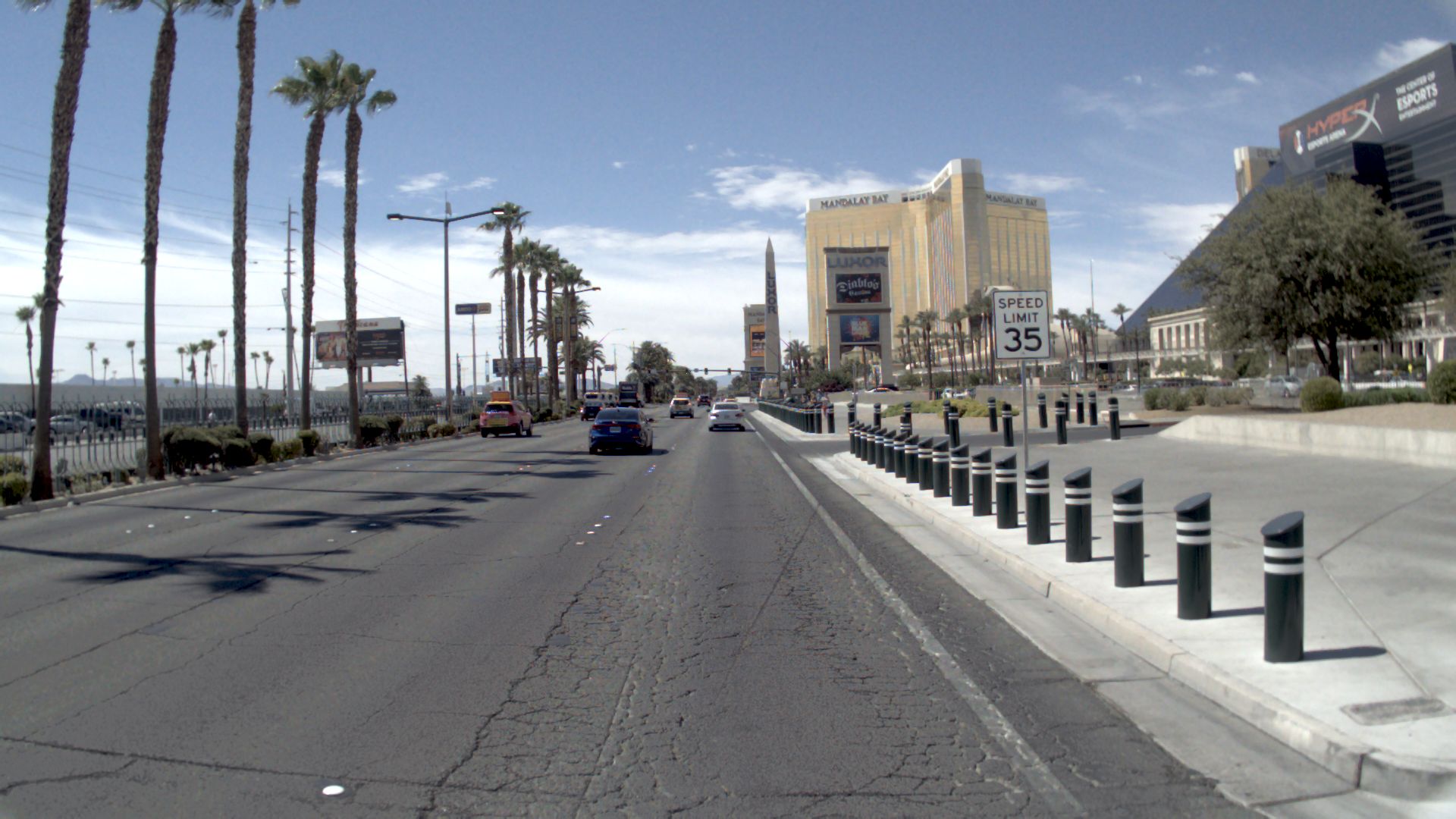}{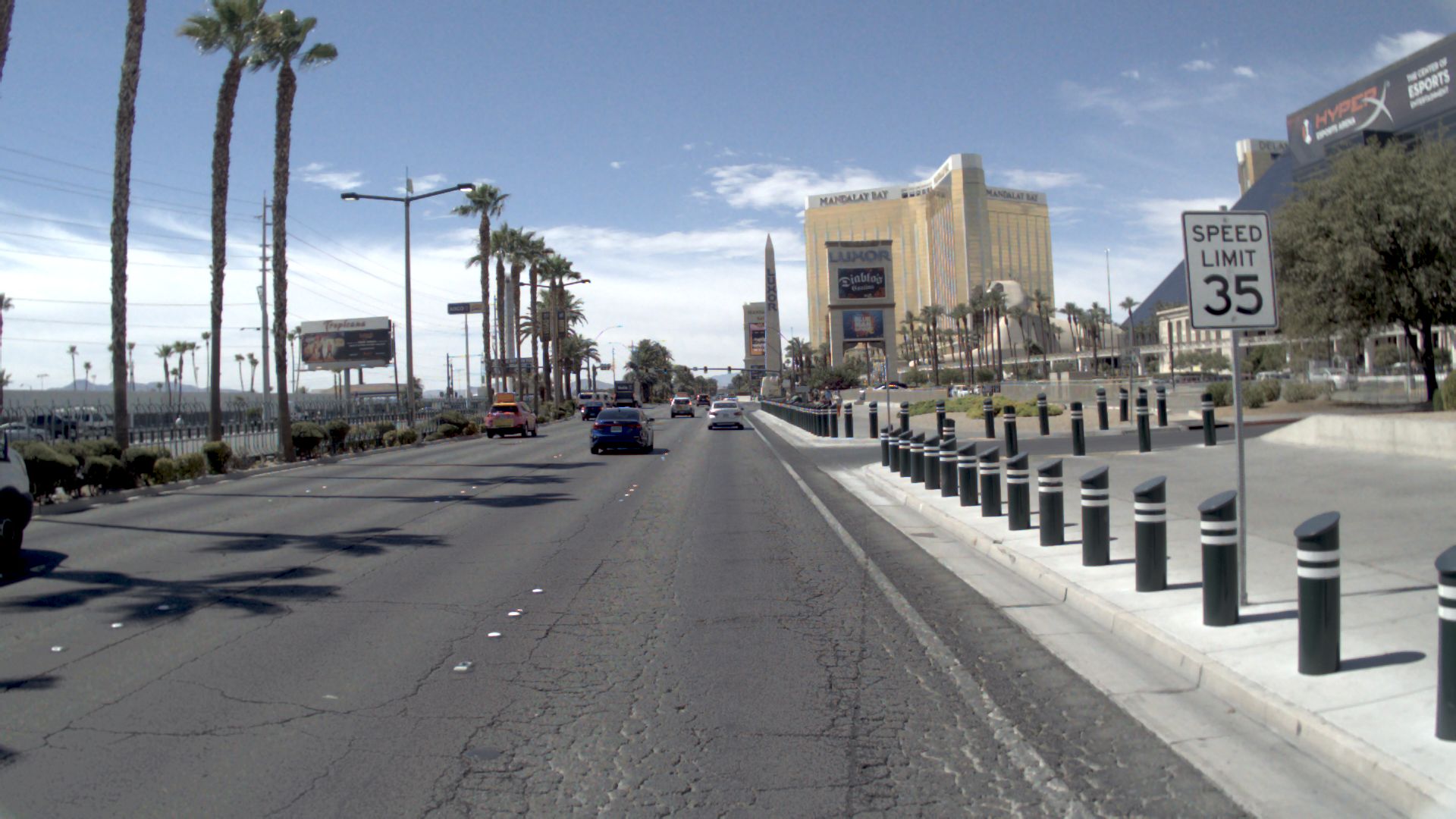}{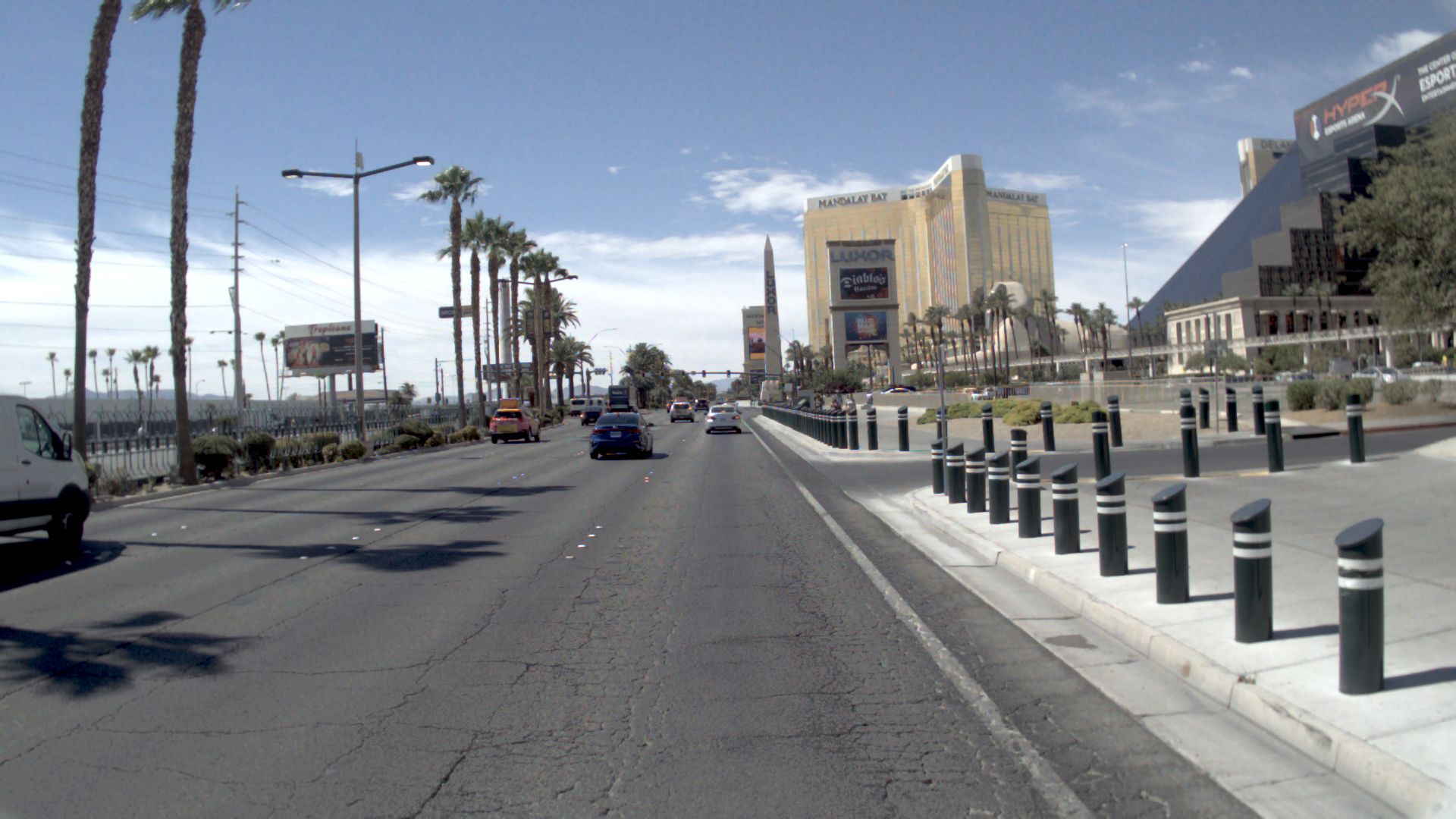}{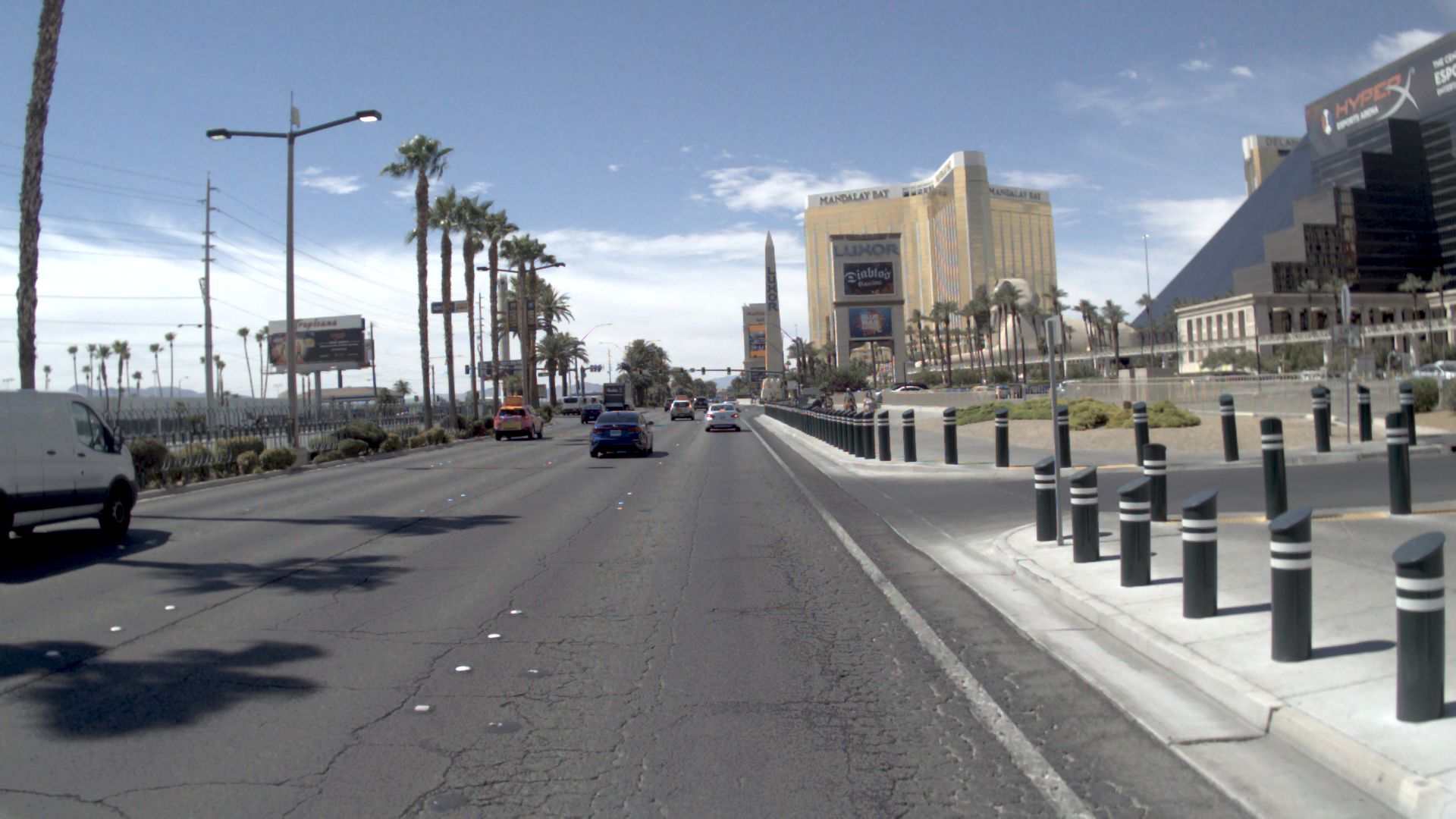}
{stream}{stream}{\textbf{trigger}}{stream}
\end{DSBLevelBlock}

\begin{DSBLevelBlock}
\DSBPromptHeader
{L3: temporal comprehension}
{Corresponding temporal probe}
{{\emph{Did the car ahead just slow down?}}}
\DSBFourFrames
{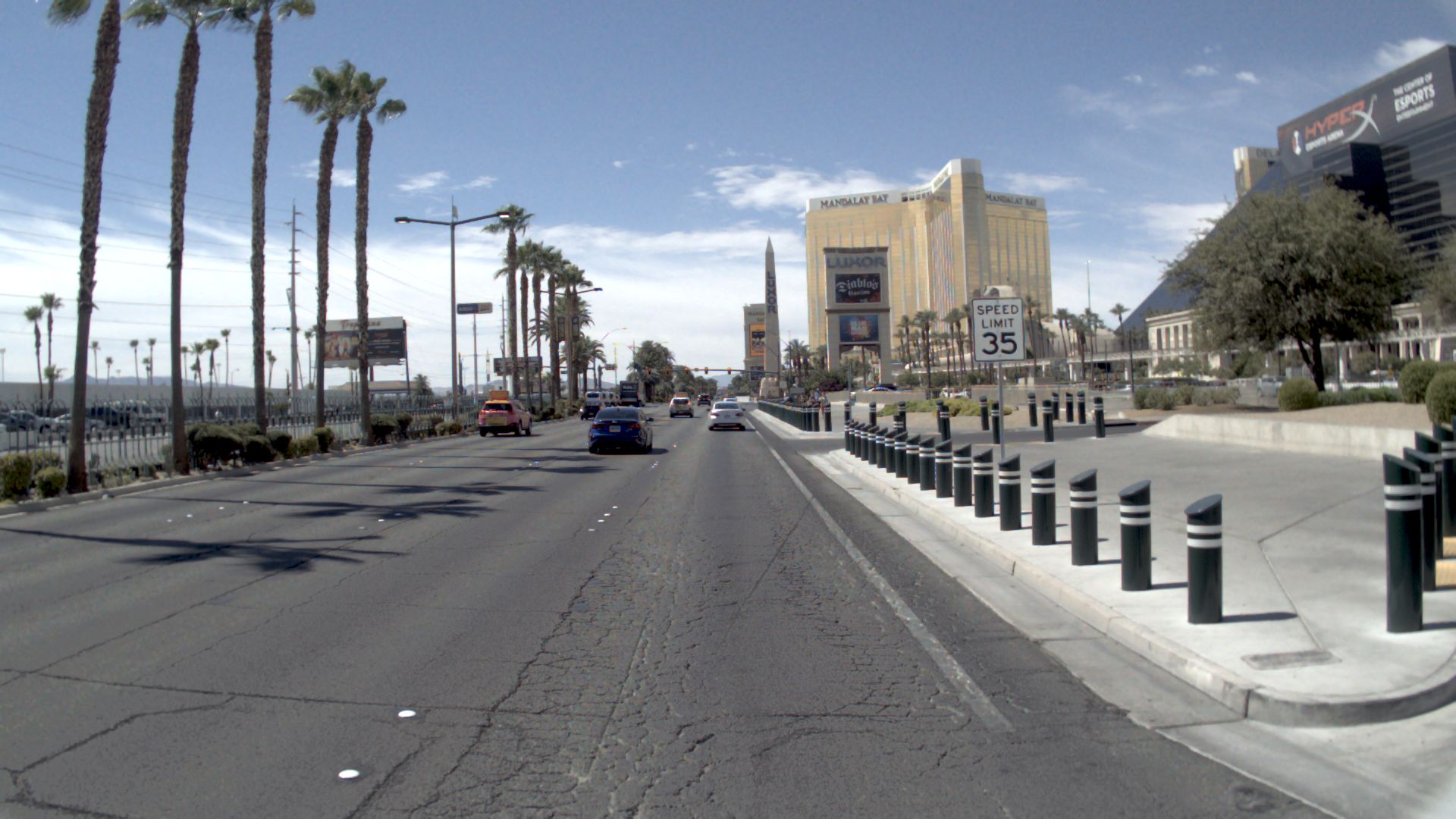}{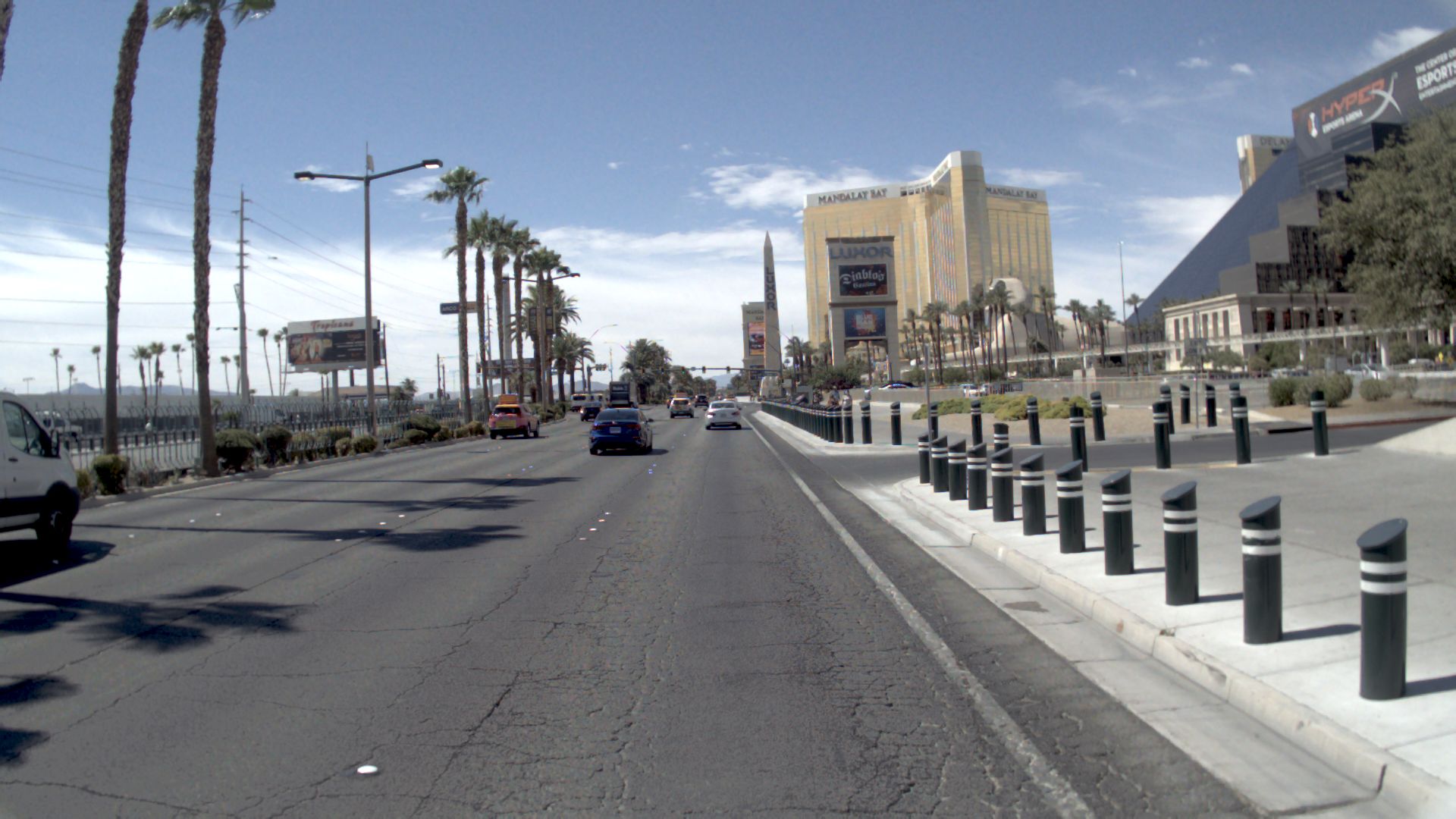}{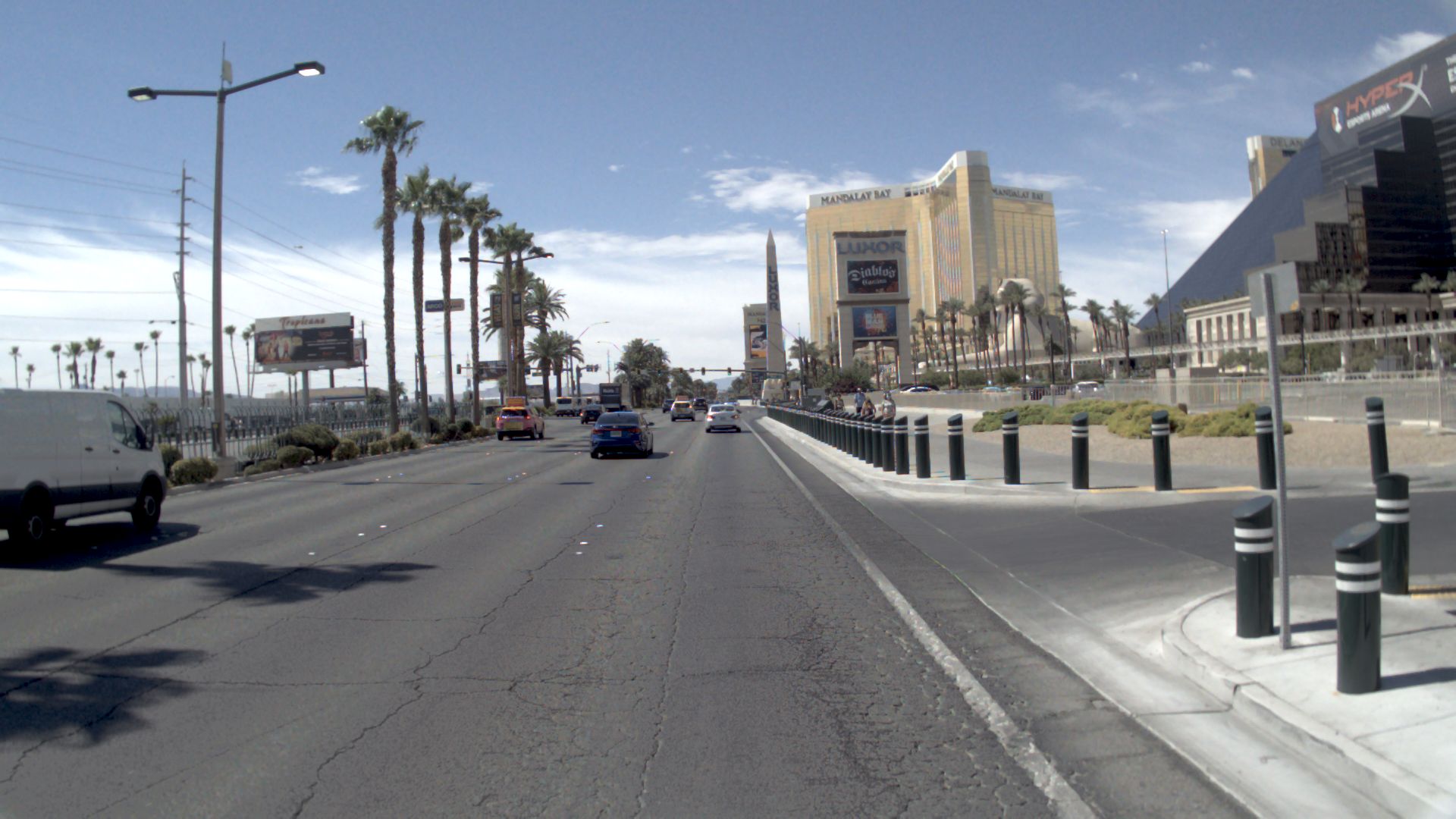}{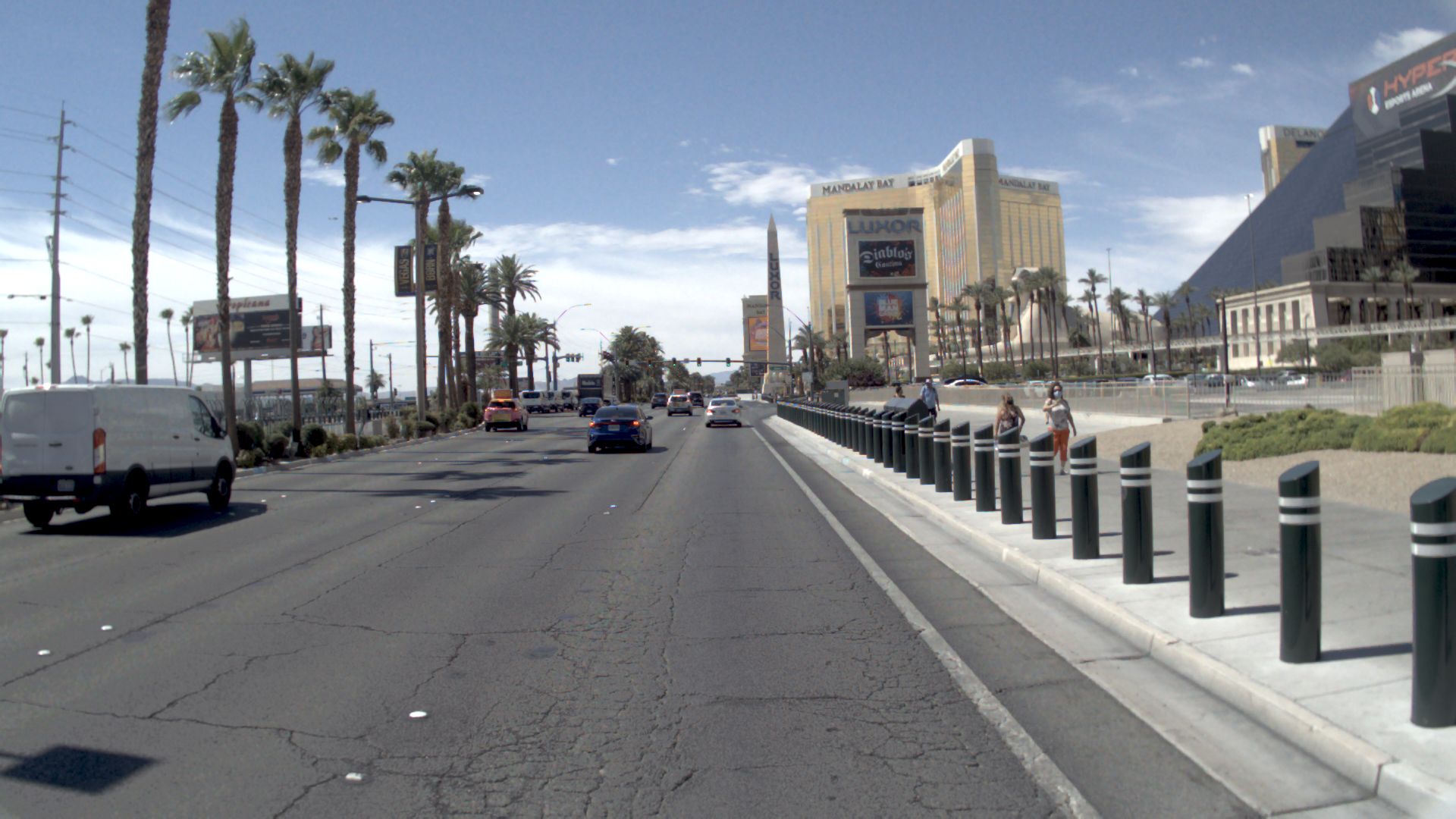}
{$t_0$}{$t_1$}{$t_2$}{$t_3$}
\end{DSBLevelBlock}

\DSBSinglePair
{%
\DSBCardText
{L2: relational understanding}
{Single-frame ego-centric relation probe}
{{\emph{Are there vehicles in my lane ahead?}}}
}
{ex3_l2.jpg}
{L2 single-image frame}
{%
\DSBCardText
{L1: recognition}
{Single-frame recognition probe}
{{\emph{Can you see a blue sedan in the scene?}}}
}
{ex3_l1.jpg}
{L1 single-image frame}

\vspace{0.25em}
\captionof{figure}{Event-linked L1--L4 task example for a lead-vehicle-brake event.}
\label{fig:l1234-example-lead-brake}
\end{minipage}

\clearpage
\Needspace{0.88\textheight}
\subsection{Example 3: Traffic Light Change}
\label{app:example-traffic-light-change}

\noindent
\begin{minipage}{\textwidth}

\begin{DSBLevelBlock}
\DSBPromptHeader
{L4: watch-and-notify}
{Streaming alert decision}
{{\emph{Alert me when the traffic light changes from red to green.}}}
\DSBFourFrames
{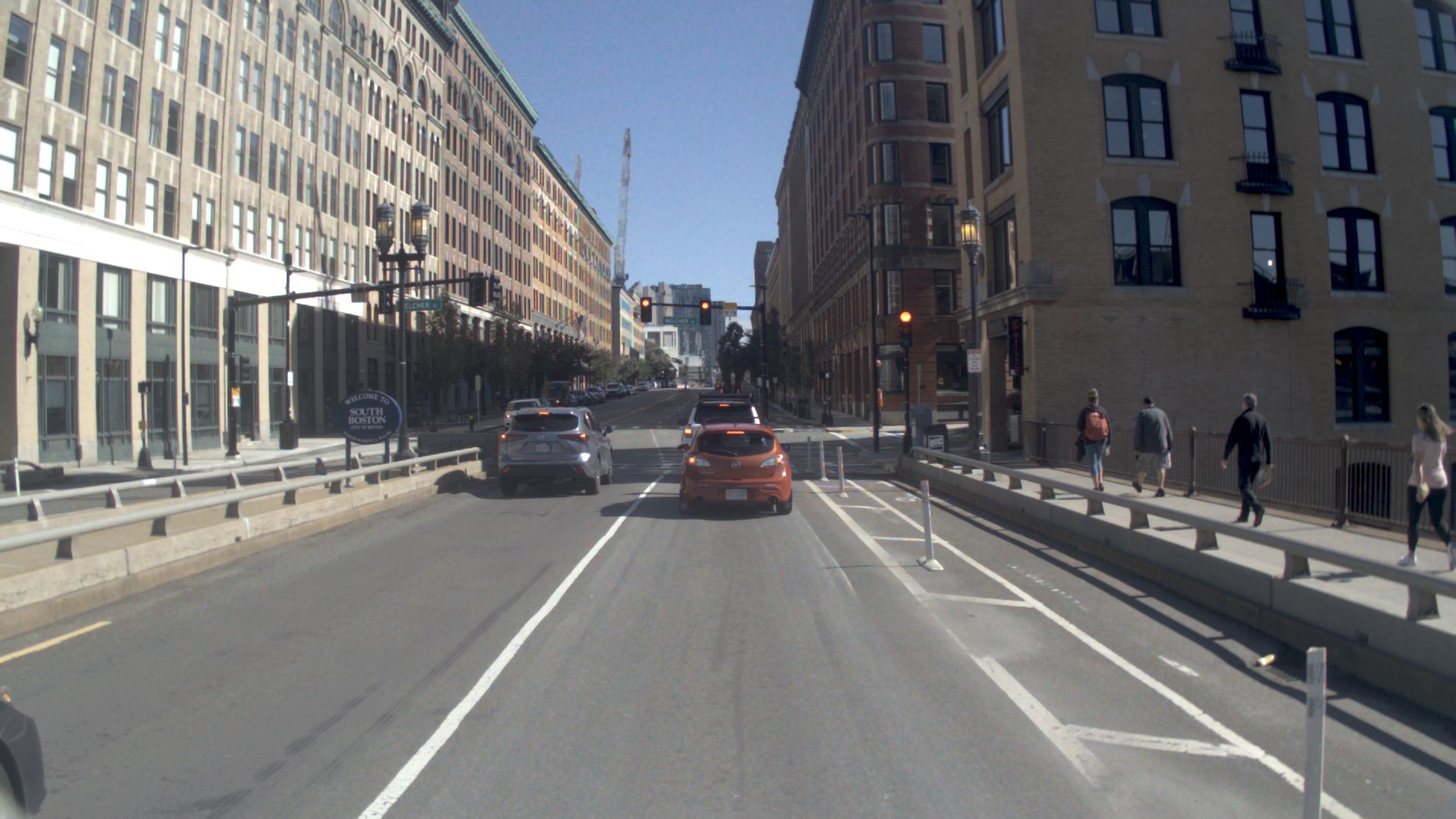}{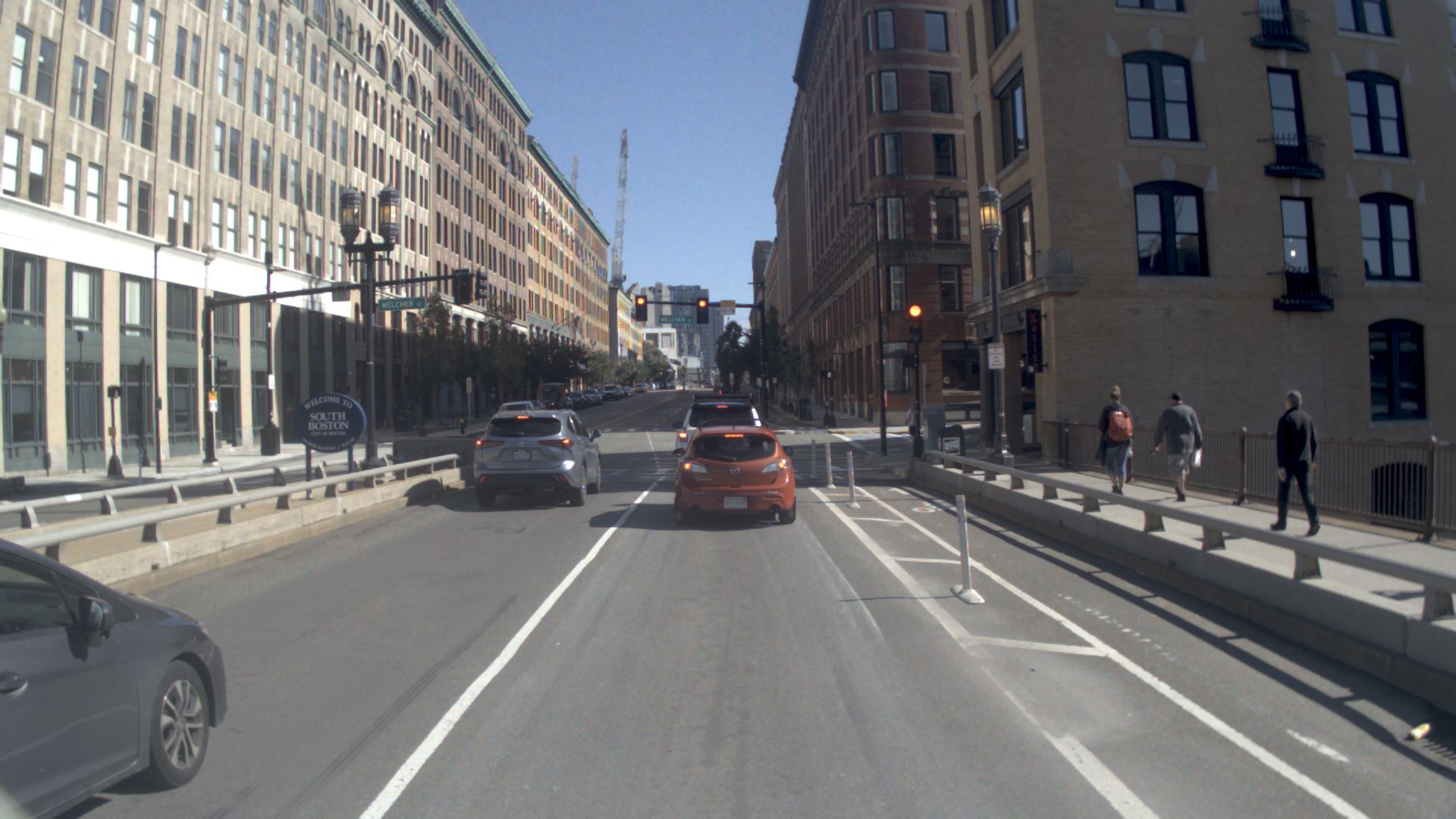}{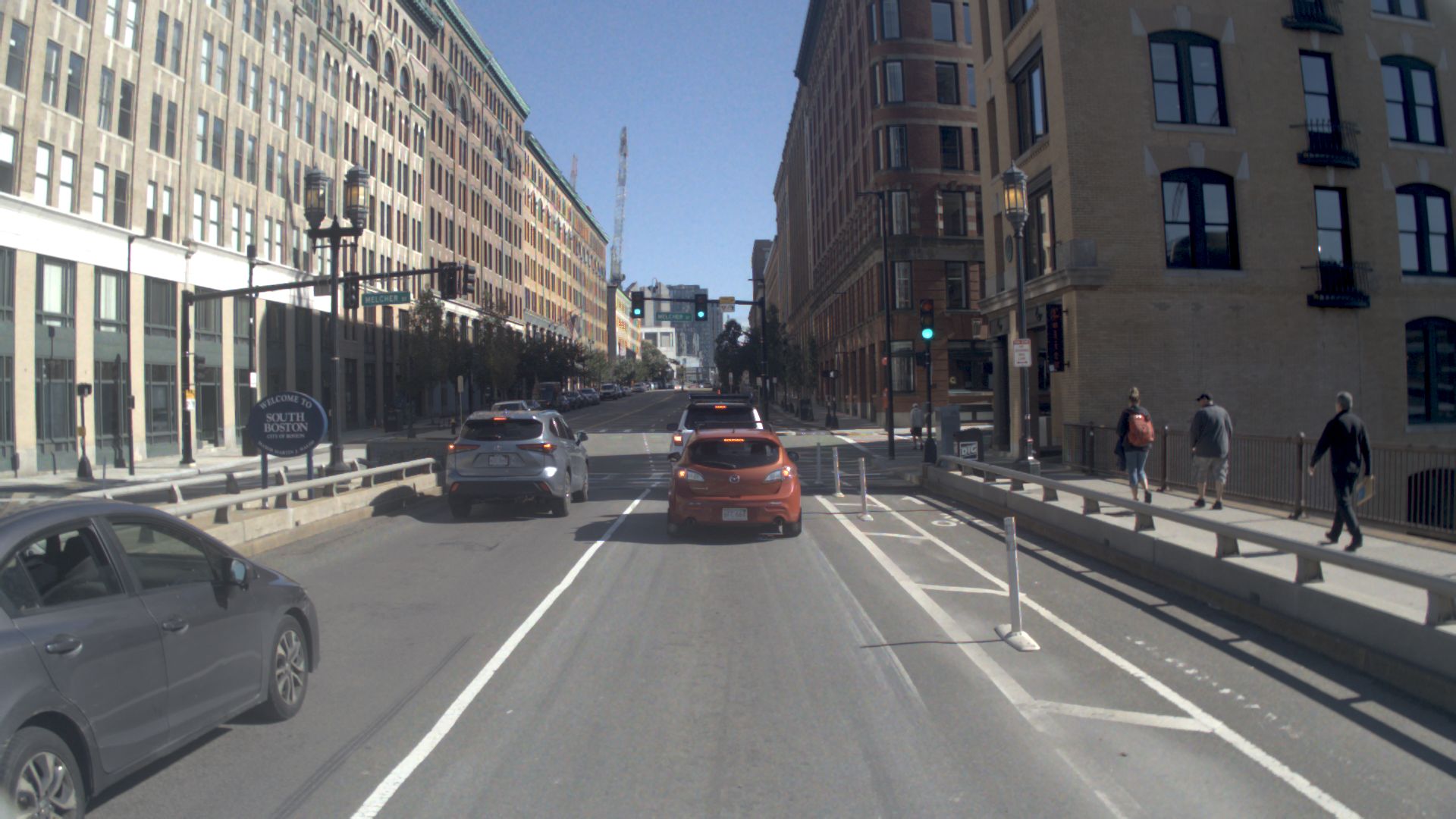}{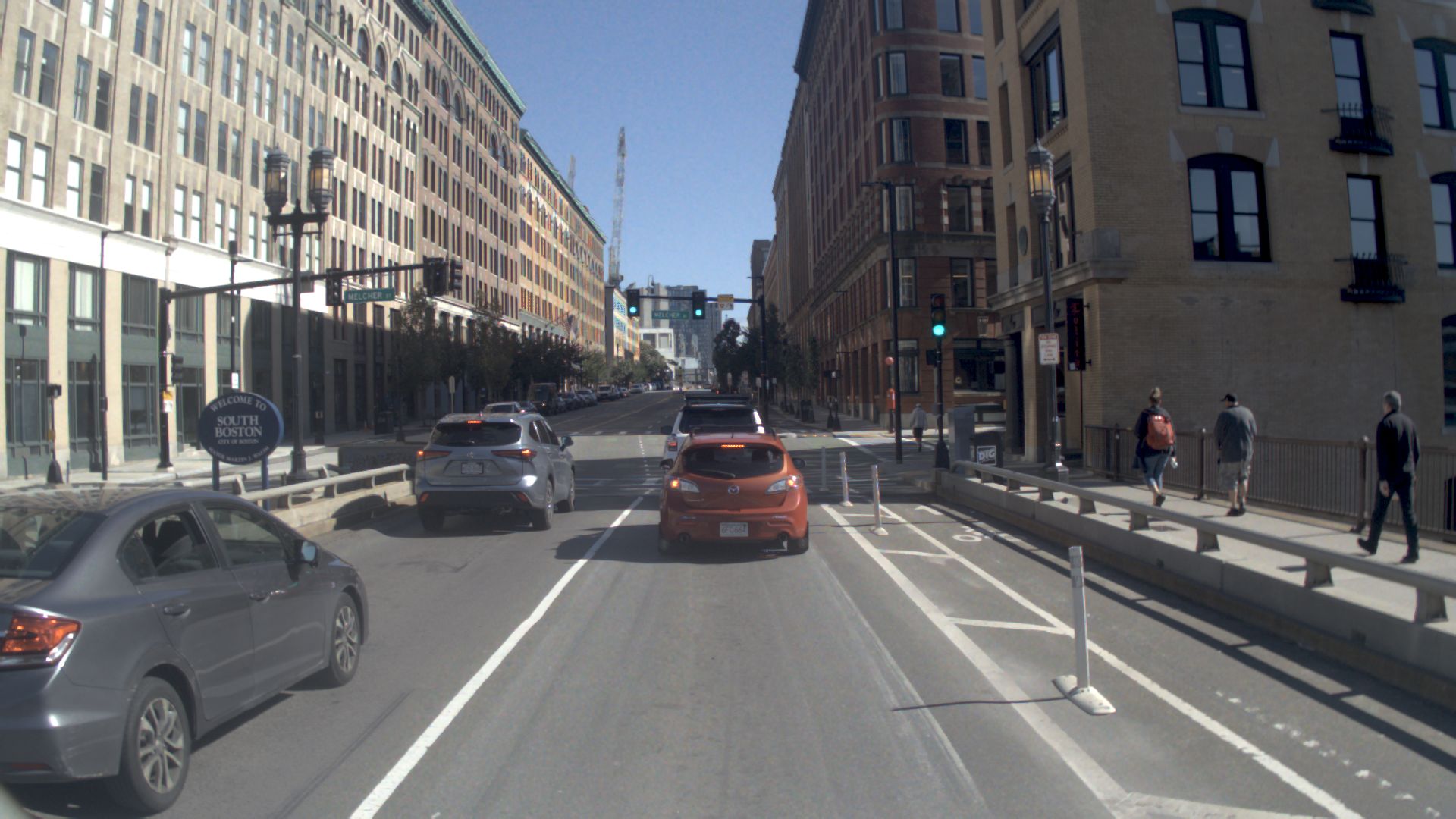}
{stream}{stream}{\textbf{trigger}}{stream}
\end{DSBLevelBlock}

\begin{DSBLevelBlock}
\DSBPromptHeader
{L3: temporal comprehension}
{Corresponding temporal probe}
{{\emph{Did the traffic light just change from red to green?}}}
\DSBFourFrames
{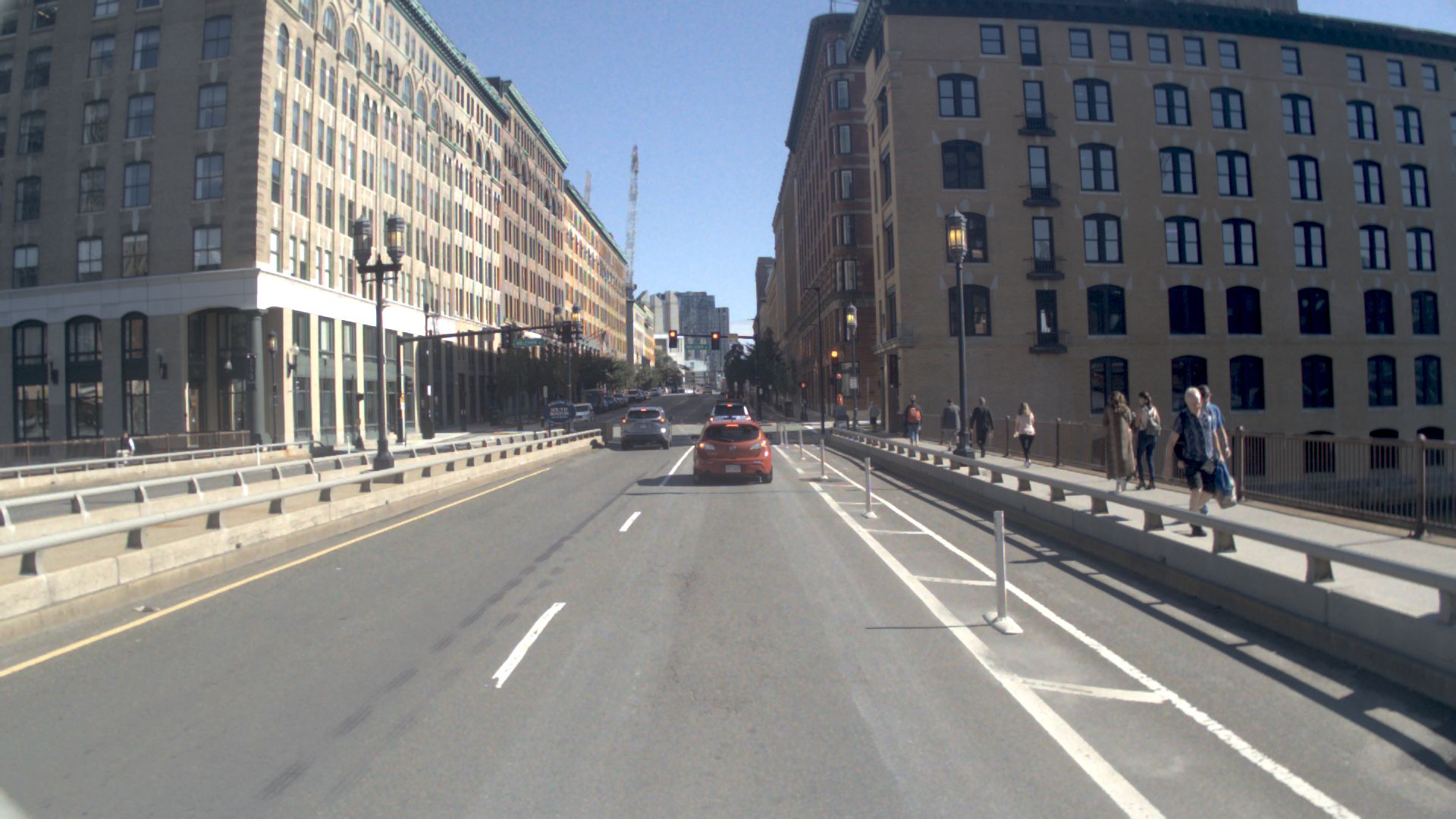}{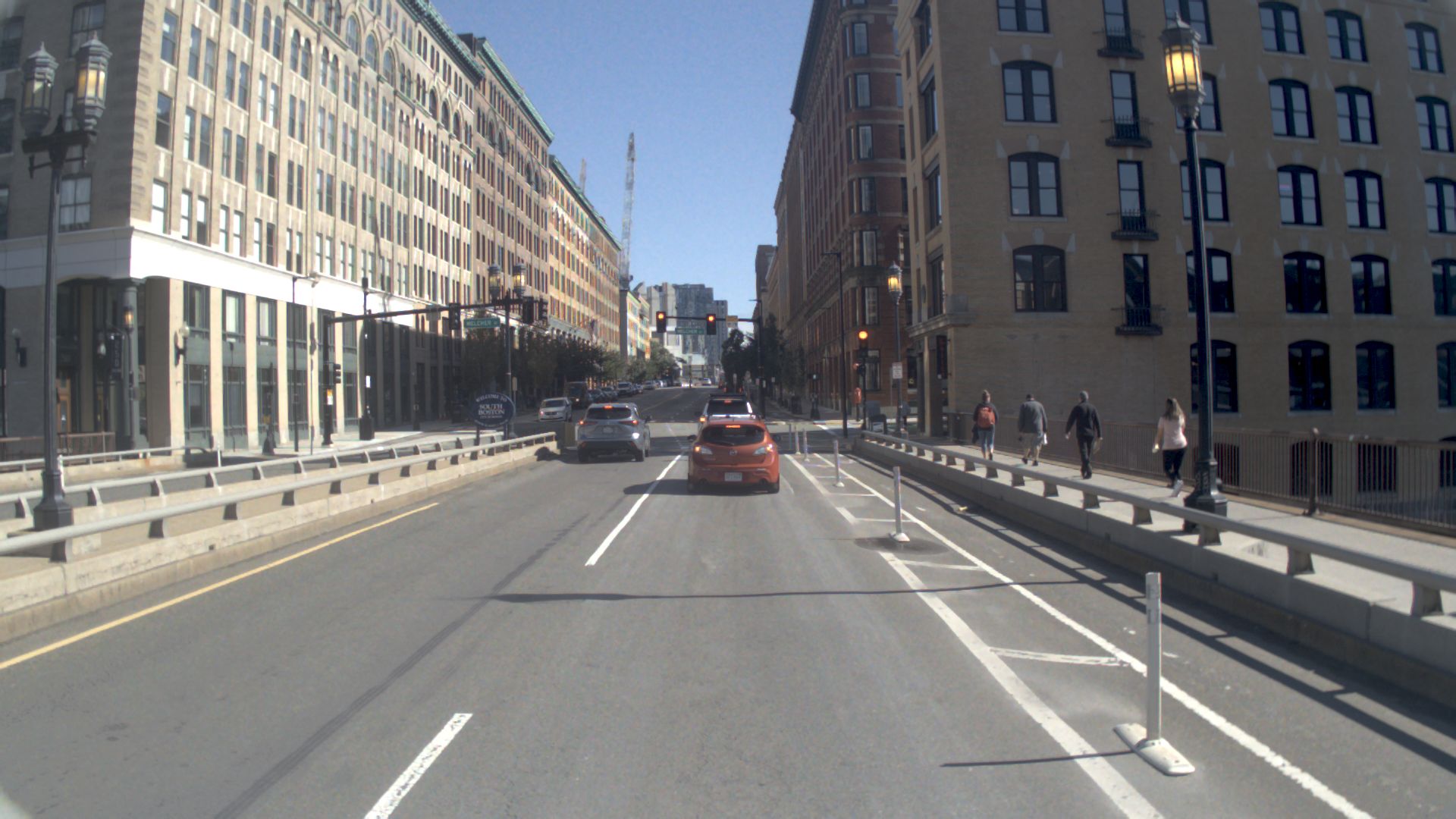}{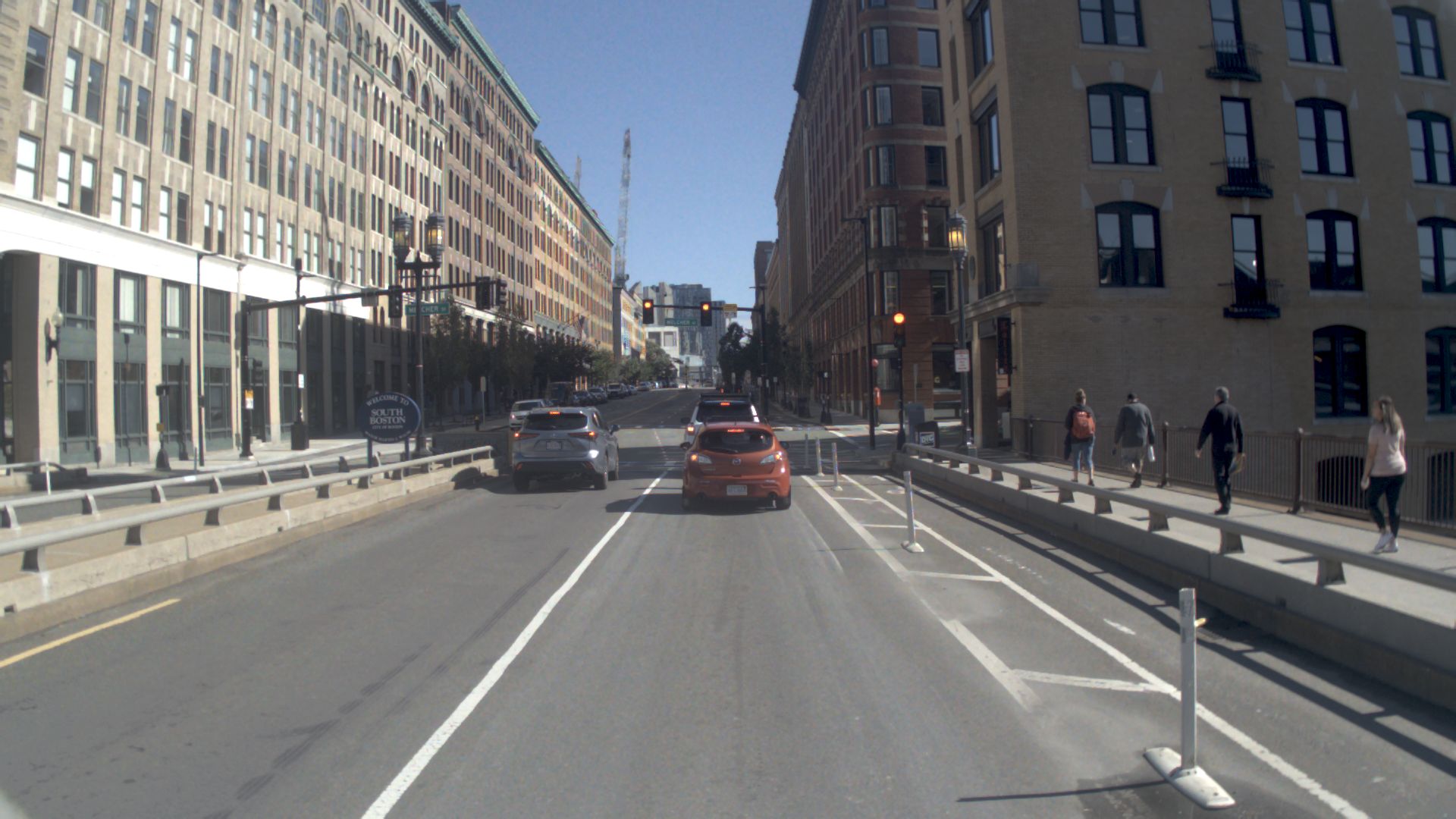}{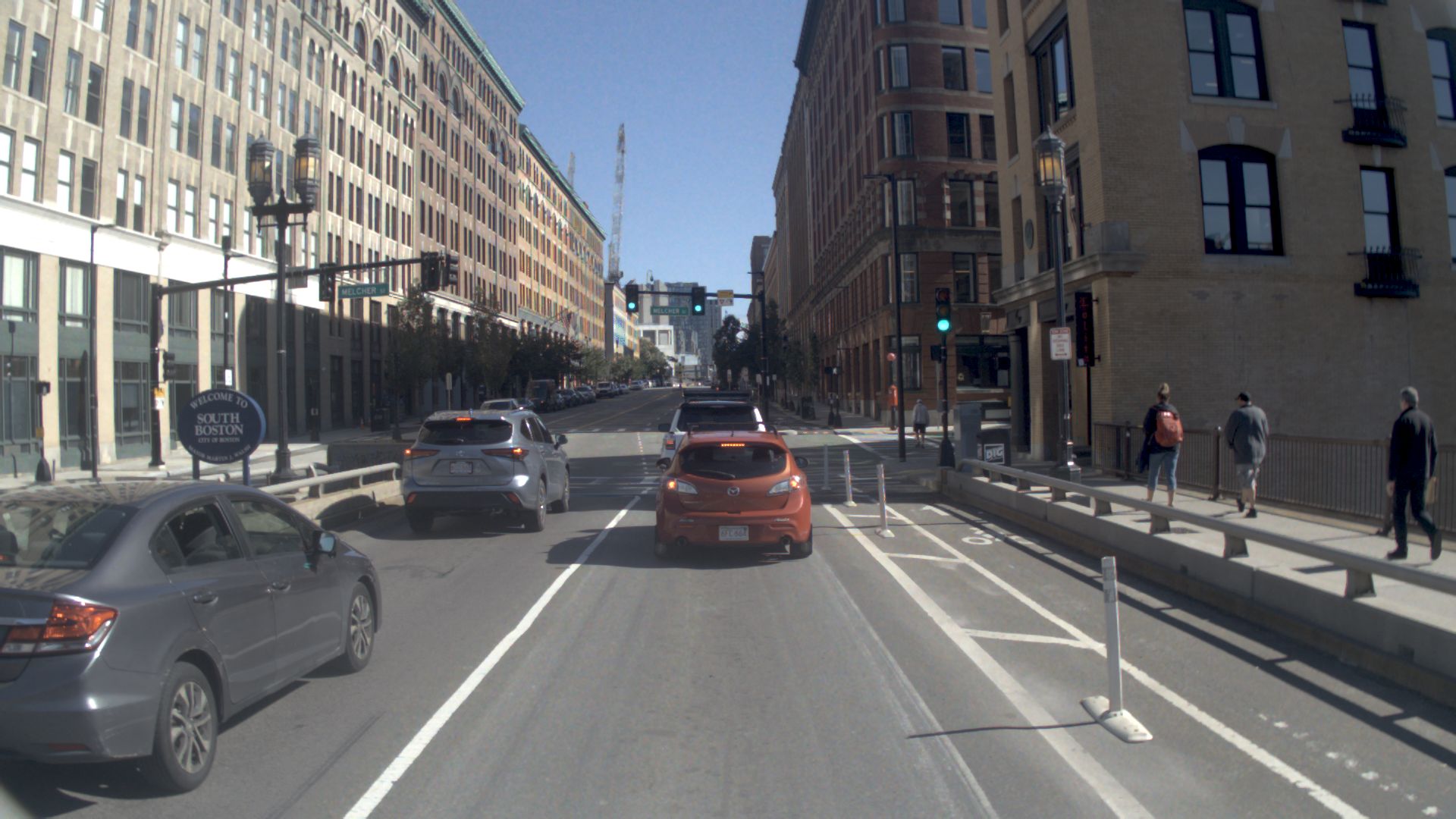}
{$t_0$}{$t_1$}{$t_2$}{$t_3$}
\end{DSBLevelBlock}

\DSBSinglePair
{%
\DSBCardText
{L2: relational understanding}
{Single-frame traffic-control relation probe}
{{\emph{What color is the traffic light for my lane right now?}\par
\DSBOptionGrid{Green}{Red}{Unknown}{Yellow}}}
}
{ex2_l2.jpg}
{L2 single-image frame}
{%
\DSBCardText
{L1: recognition}
{Single-frame recognition probe}
{{\emph{Are there traffic lights visible in the scene?}}}
}
{ex2_l1.jpg}
{L1 single-image frame}

\vspace{0.25em}
\captionof{figure}{Event-linked L1--L4 task example for a traffic-light-change event.}
\label{fig:l1234-example-traffic-light}
\end{minipage}

\clearpage
\Needspace{0.88\textheight}
\subsection{Example 4: Adjacent Vehicle Cuts In}
\label{app:example-adjacent-cut-in}

\noindent
\begin{minipage}{\textwidth}

\begin{DSBLevelBlock}
\DSBPromptHeader
{L4: watch-and-notify}
{Streaming alert decision}
{{\emph{Let me know if a car starts cutting into my lane.}}}
\DSBFourFrames
{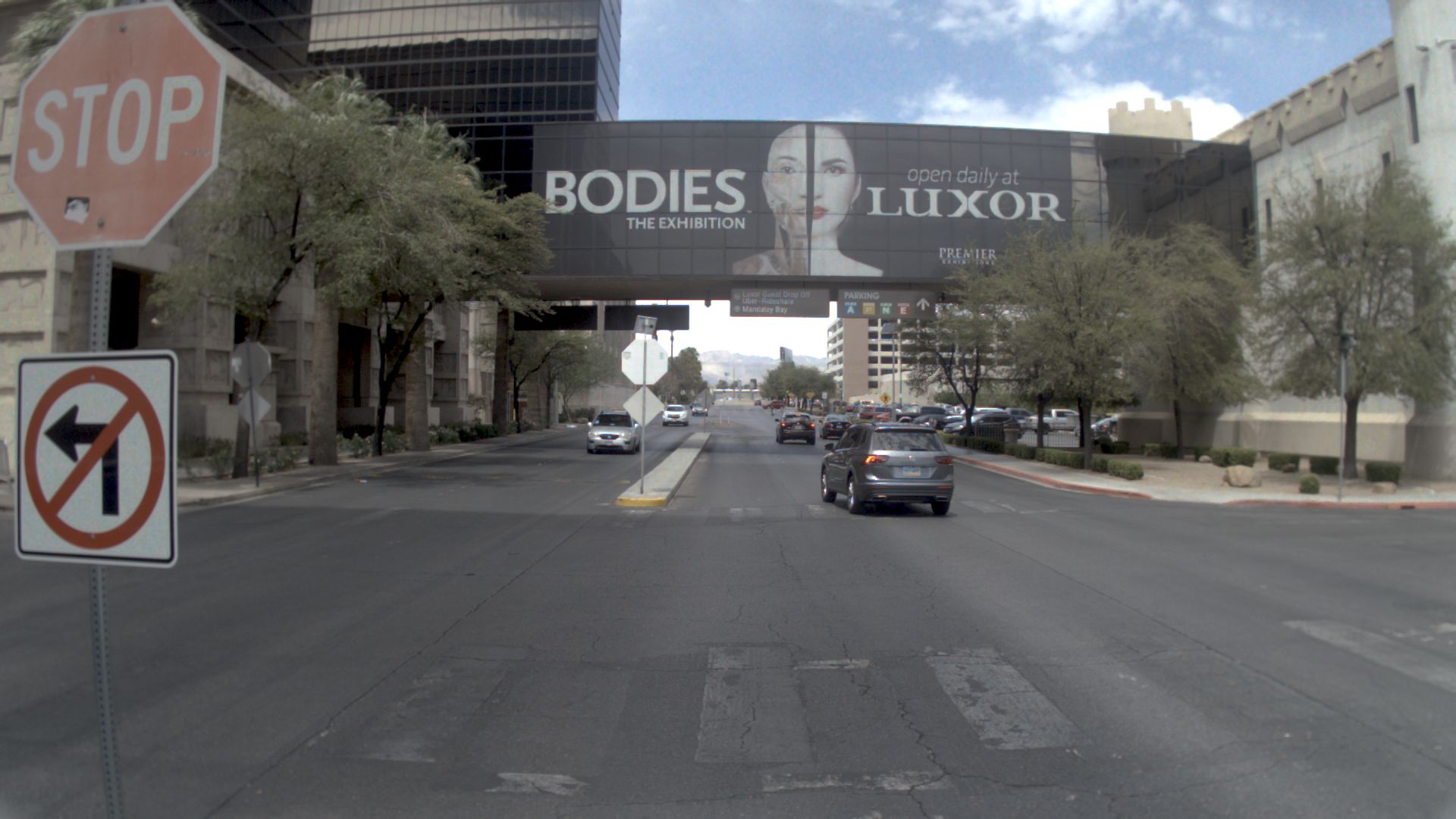}{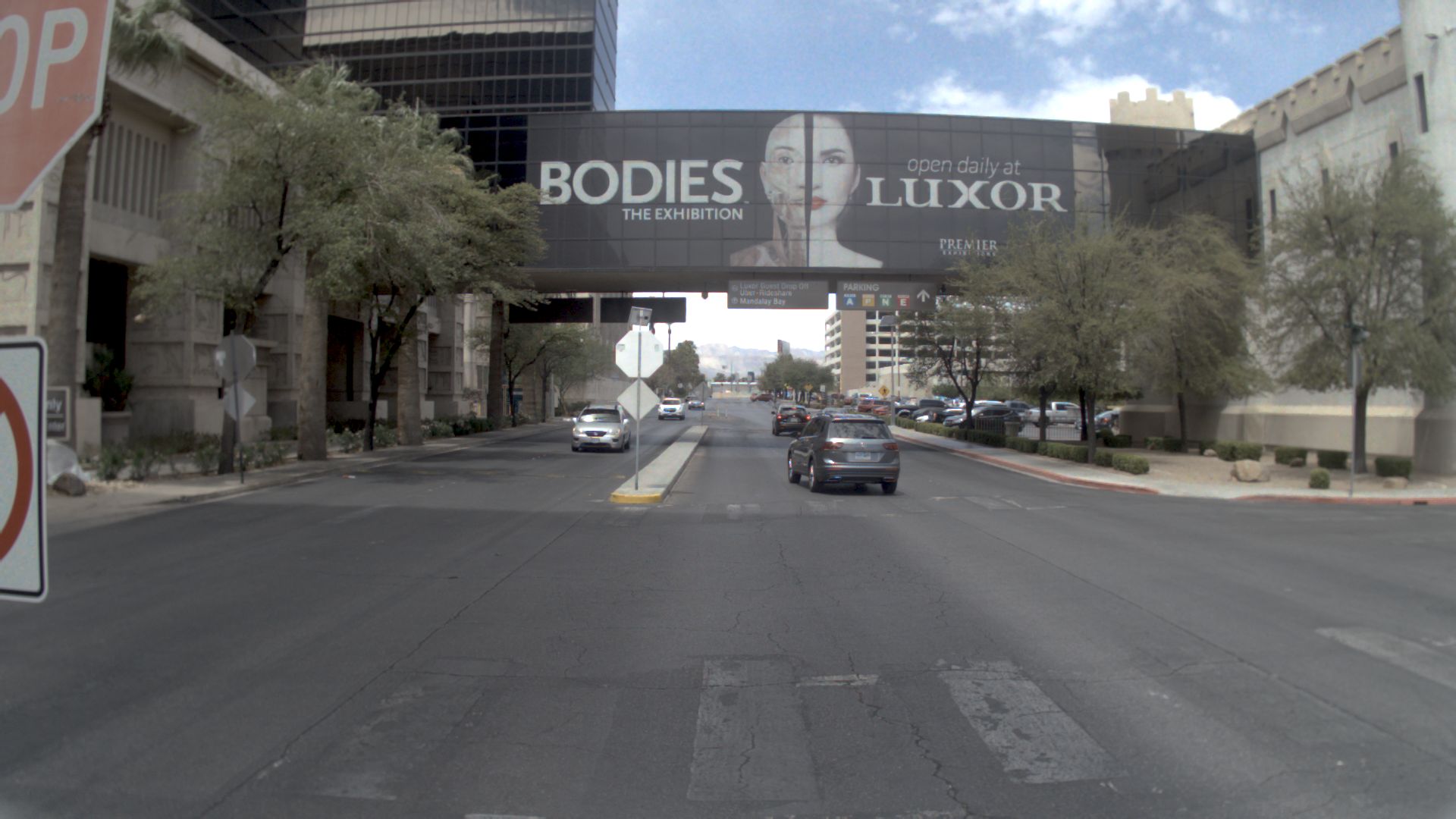}{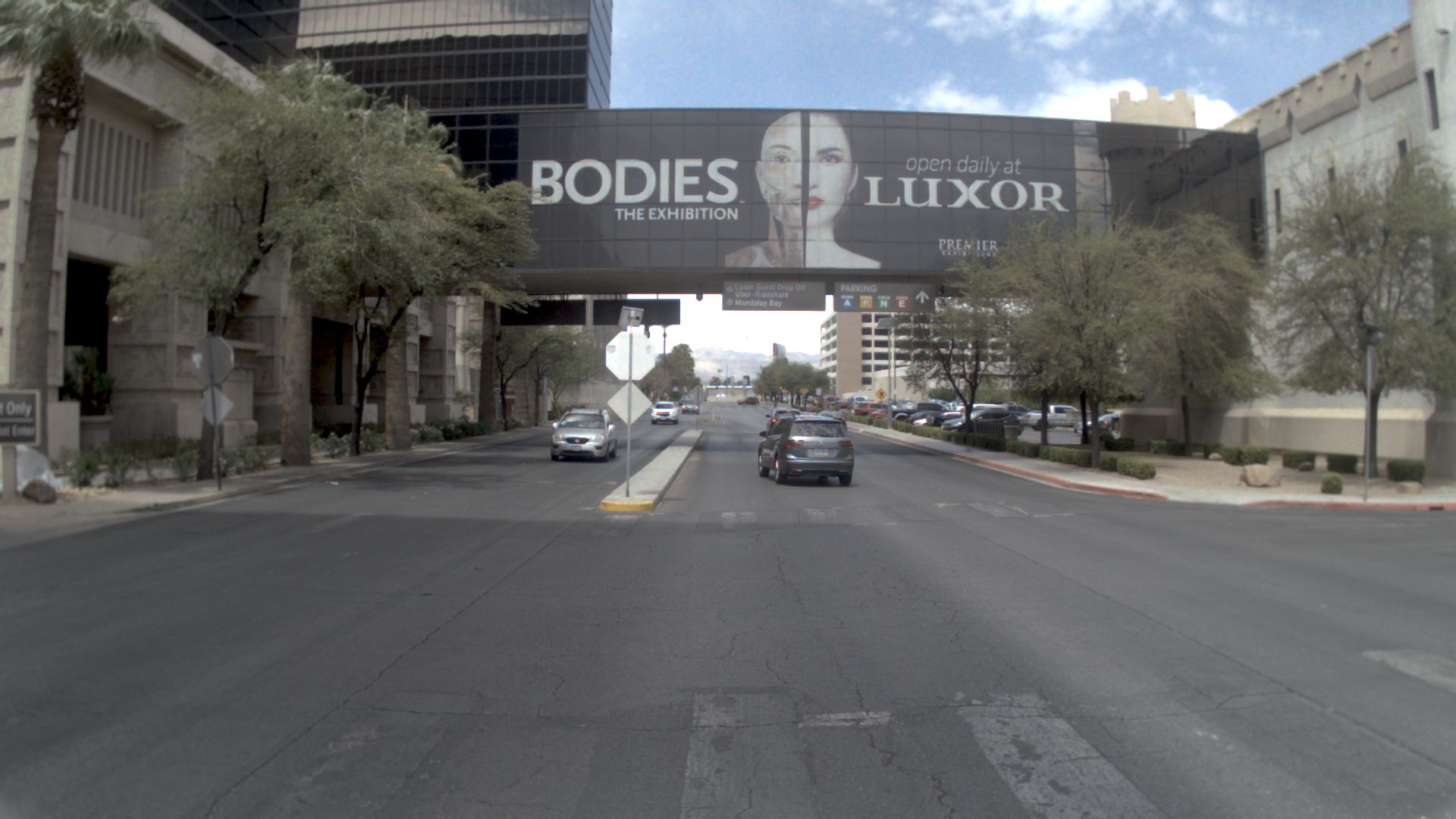}{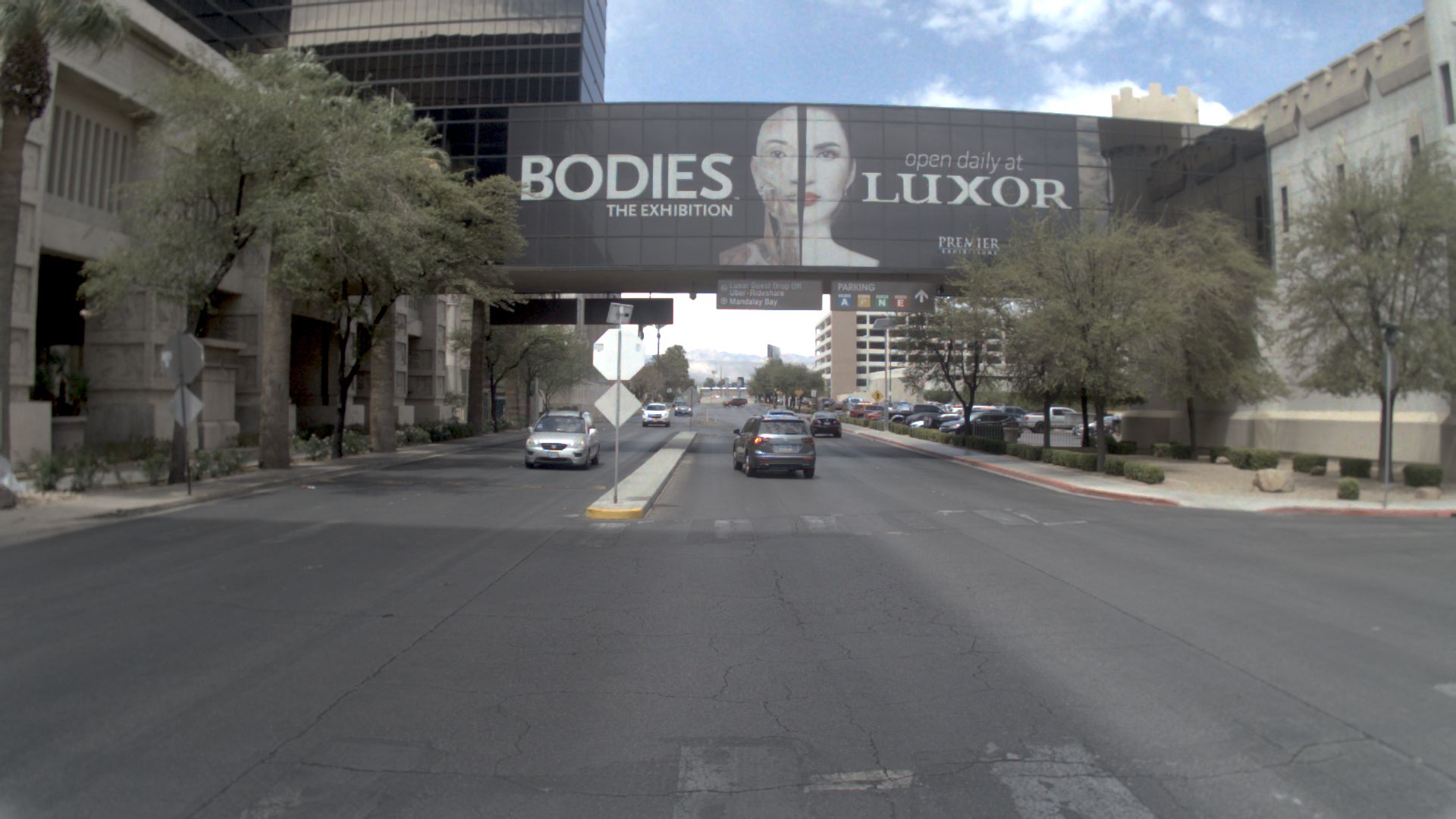}
{stream}{stream}{\textbf{trigger}}{stream}
\end{DSBLevelBlock}

\begin{DSBLevelBlock}
\DSBPromptHeader
{L3: temporal comprehension}
{Corresponding temporal probe}
{{\emph{Is a vehicle cutting in from the adjacent lane?}}}
\DSBFourFrames
{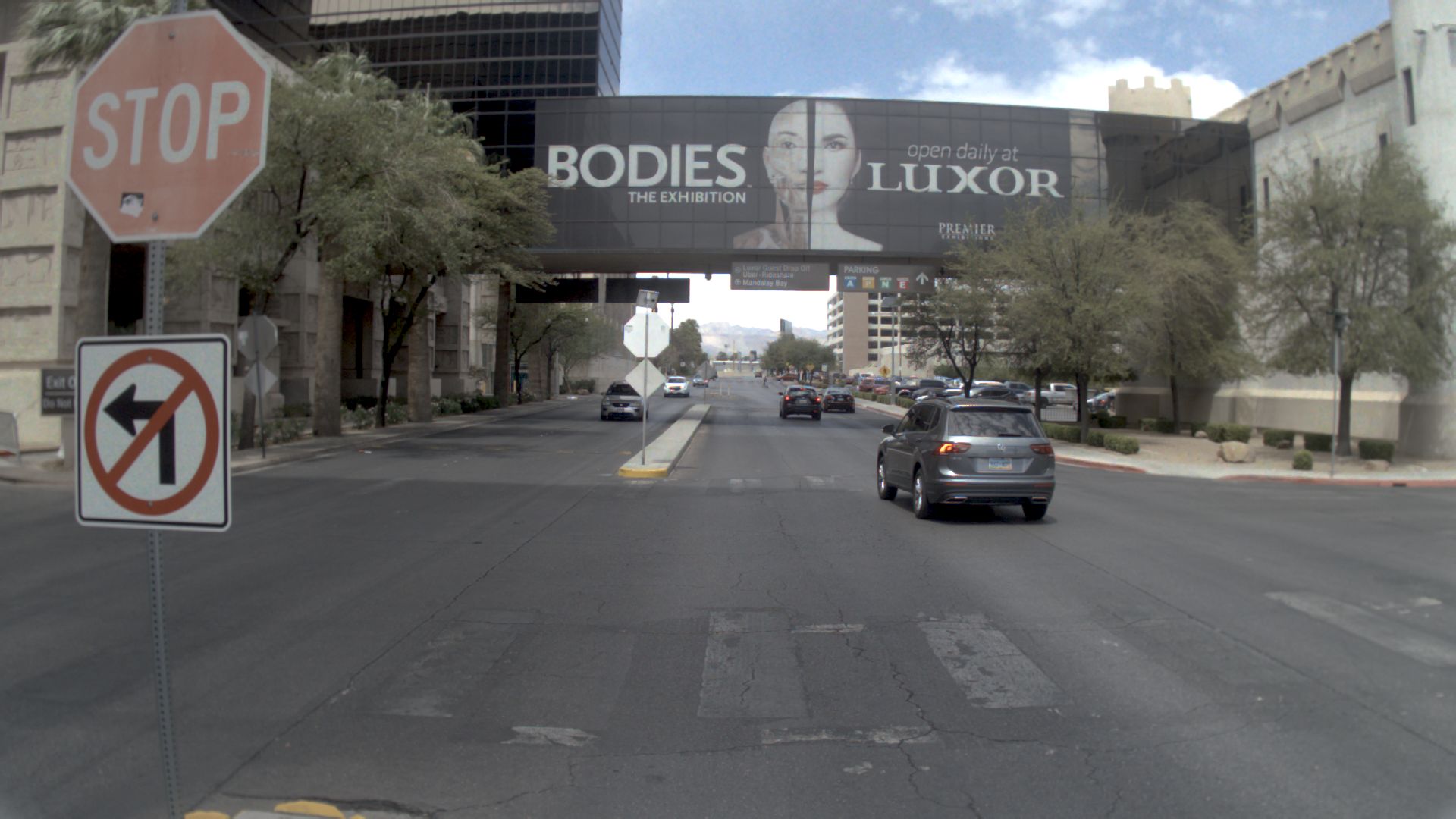}{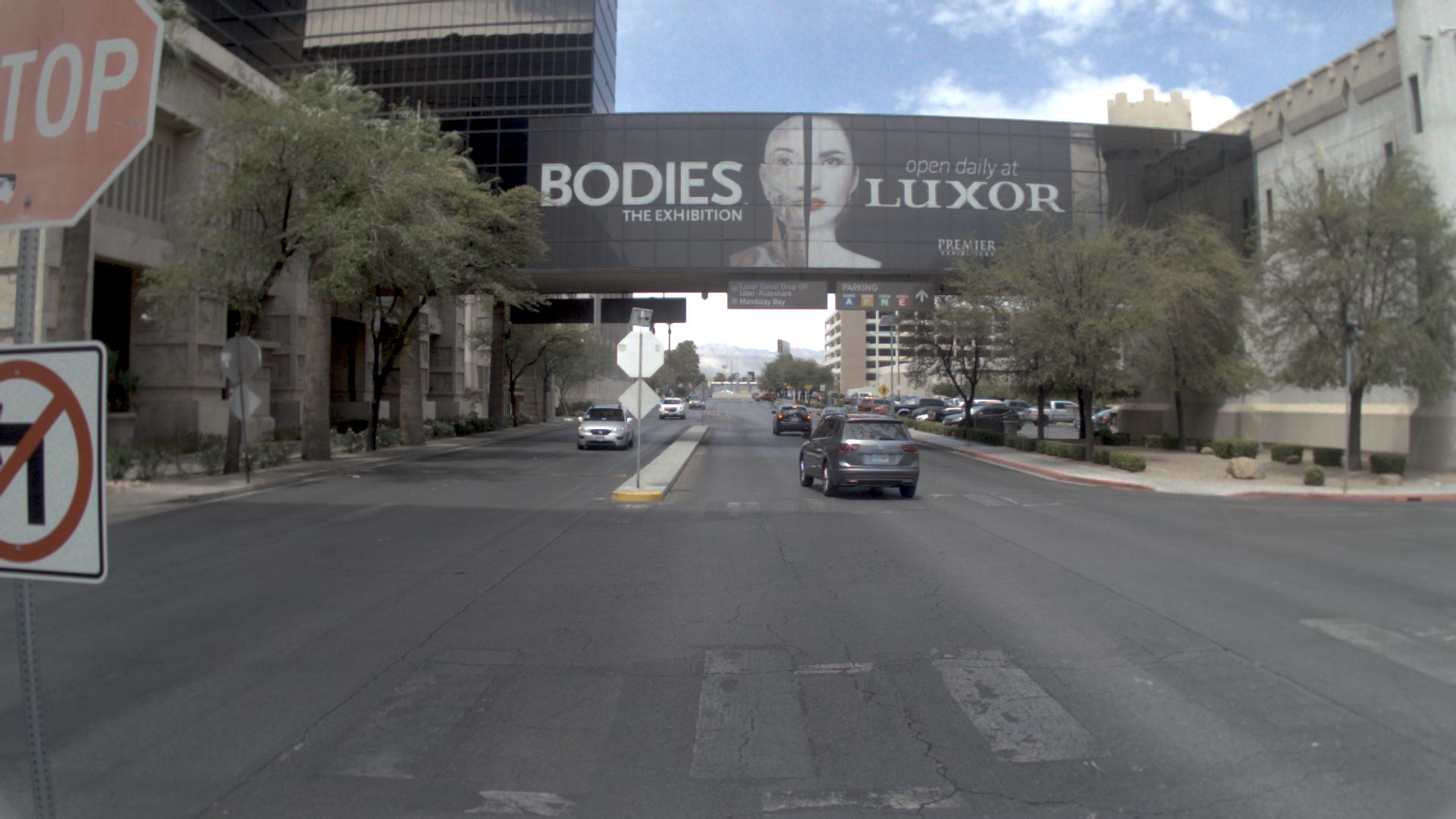}{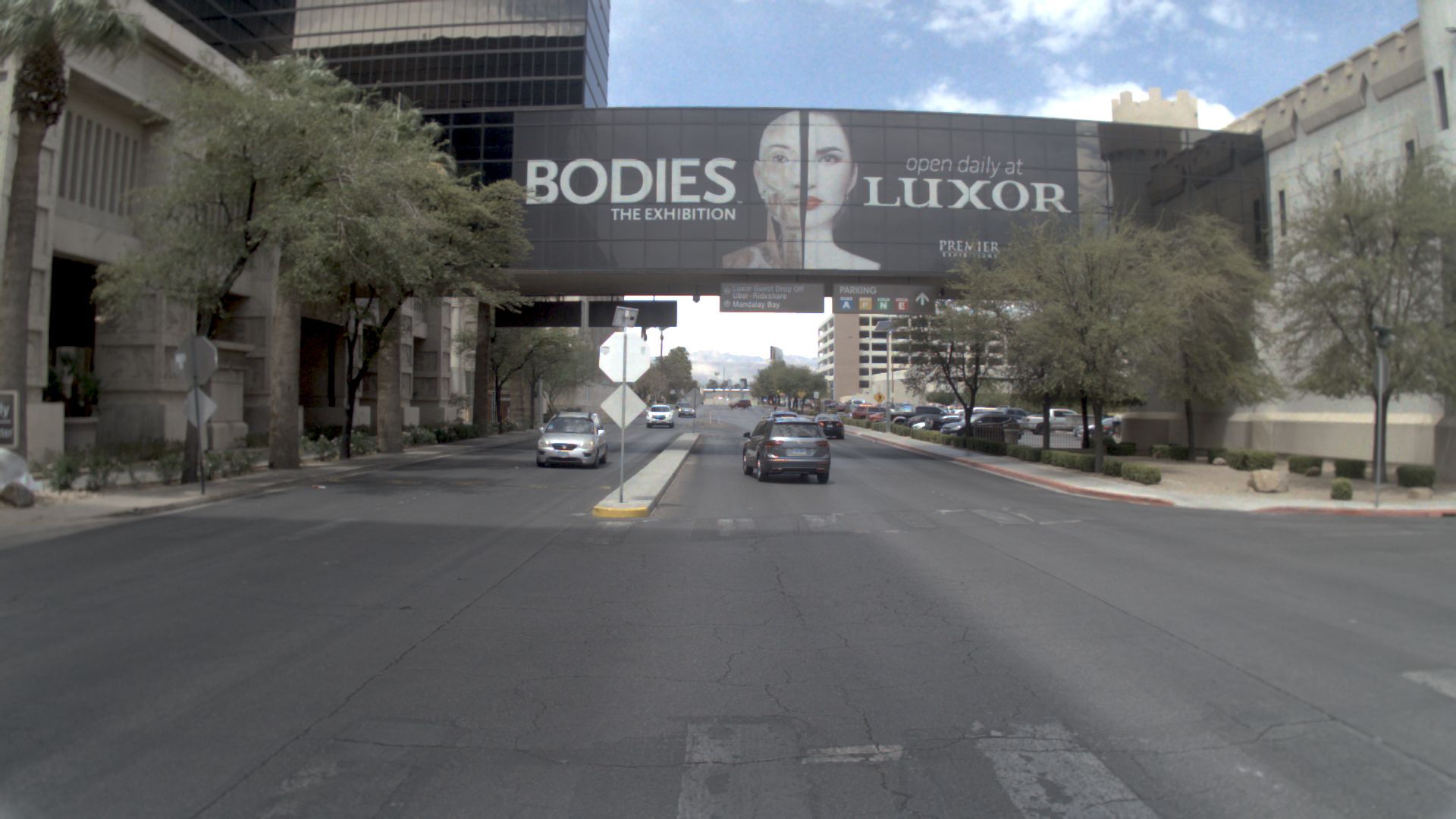}{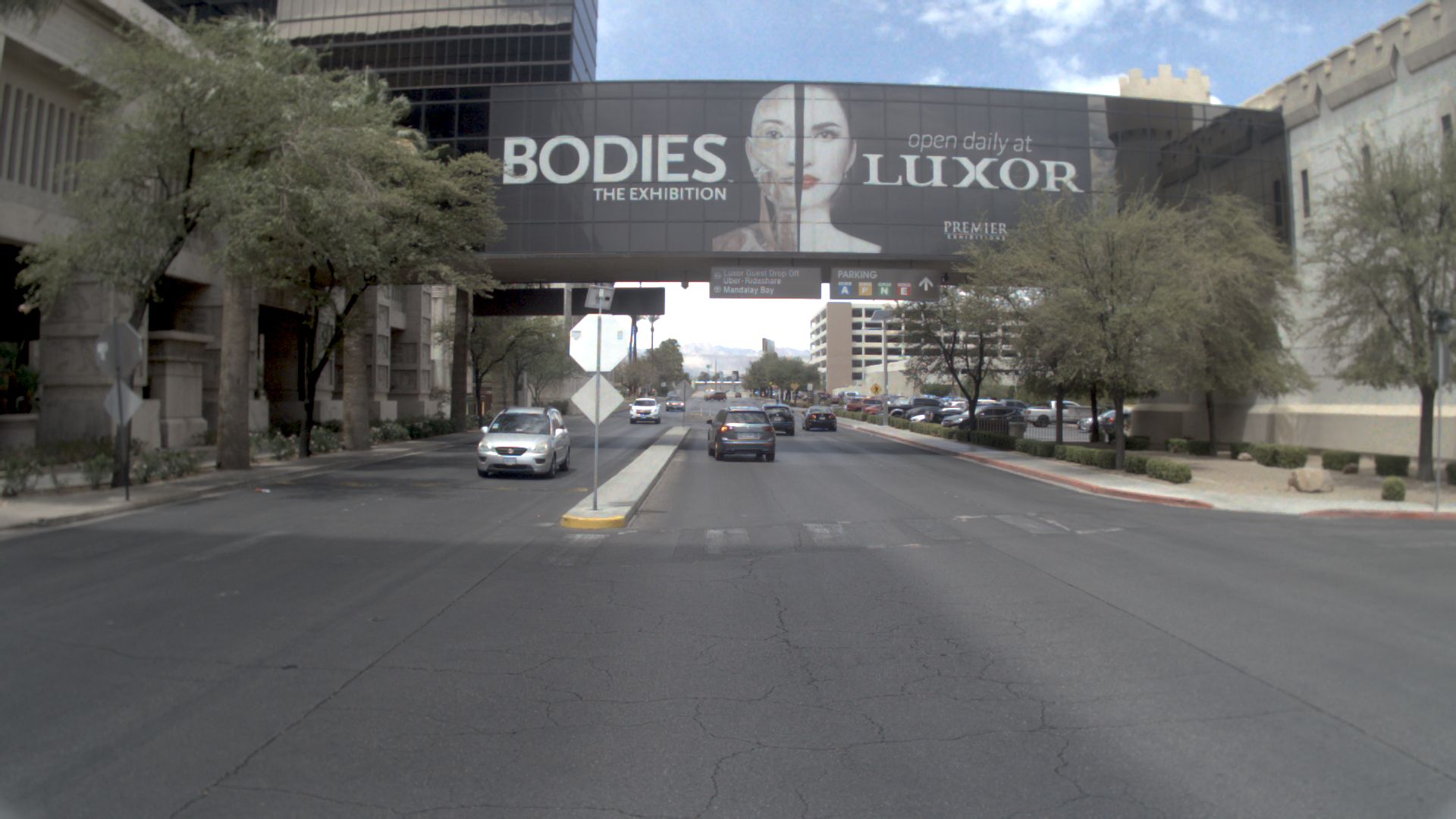}
{$t_0$}{$t_1$}{$t_2$}{$t_3$}
\end{DSBLevelBlock}

\DSBSinglePair
{%
\DSBCardText
{L2: relational understanding}
{Single-frame ego-centric relation probe}
{{\emph{Is there a vehicle in the adjacent lane to the right of the ego vehicle?}}}
}
{ex4_l2.jpg}
{L2 single-image frame}
{%
\DSBCardText
{L1: recognition}
{Single-frame recognition probe}
{{\emph{Can you see a motor vehicle in the scene?}}}
}
{ex4_l1.jpg}
{L1 single-image frame}

\vspace{0.25em}
\captionof{figure}{Event-linked L1--L4 task example for an adjacent-cut-in event.}
\label{fig:l1234-example-adjacent-cut-in}
\end{minipage}

\clearpage
\Needspace{0.88\textheight}
\subsection{Example 5: Vulnerable Road User Crossing}
\label{app:example-vru-crossing}

\noindent
\begin{minipage}{\textwidth}

\begin{DSBLevelBlock}
\DSBPromptHeader
{L4: watch-and-notify}
{Streaming alert decision}
{{\emph{Let me know if a pedestrian or cyclist is stepping into the roadway ahead.}}}
\DSBFourFrames
{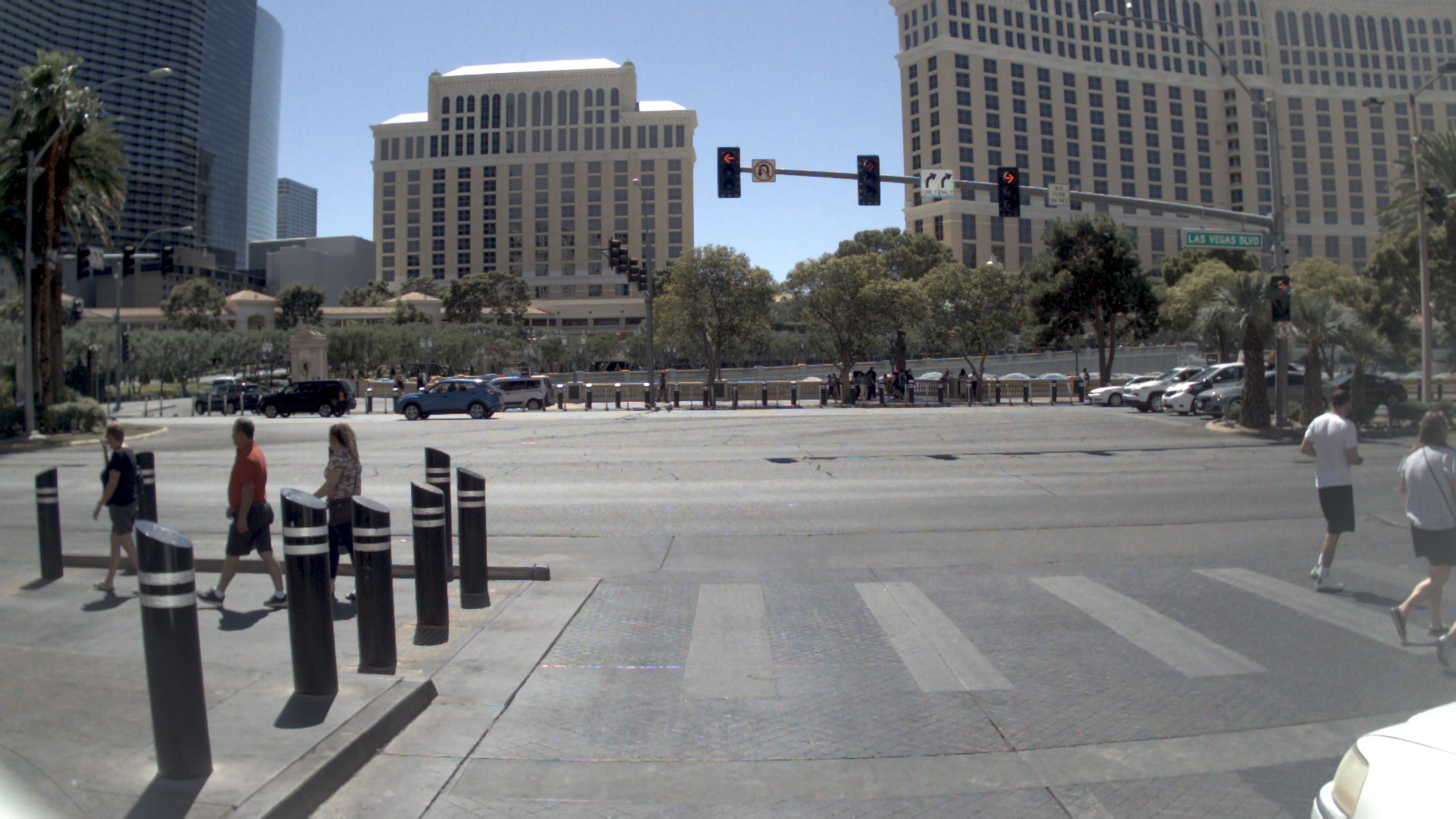}{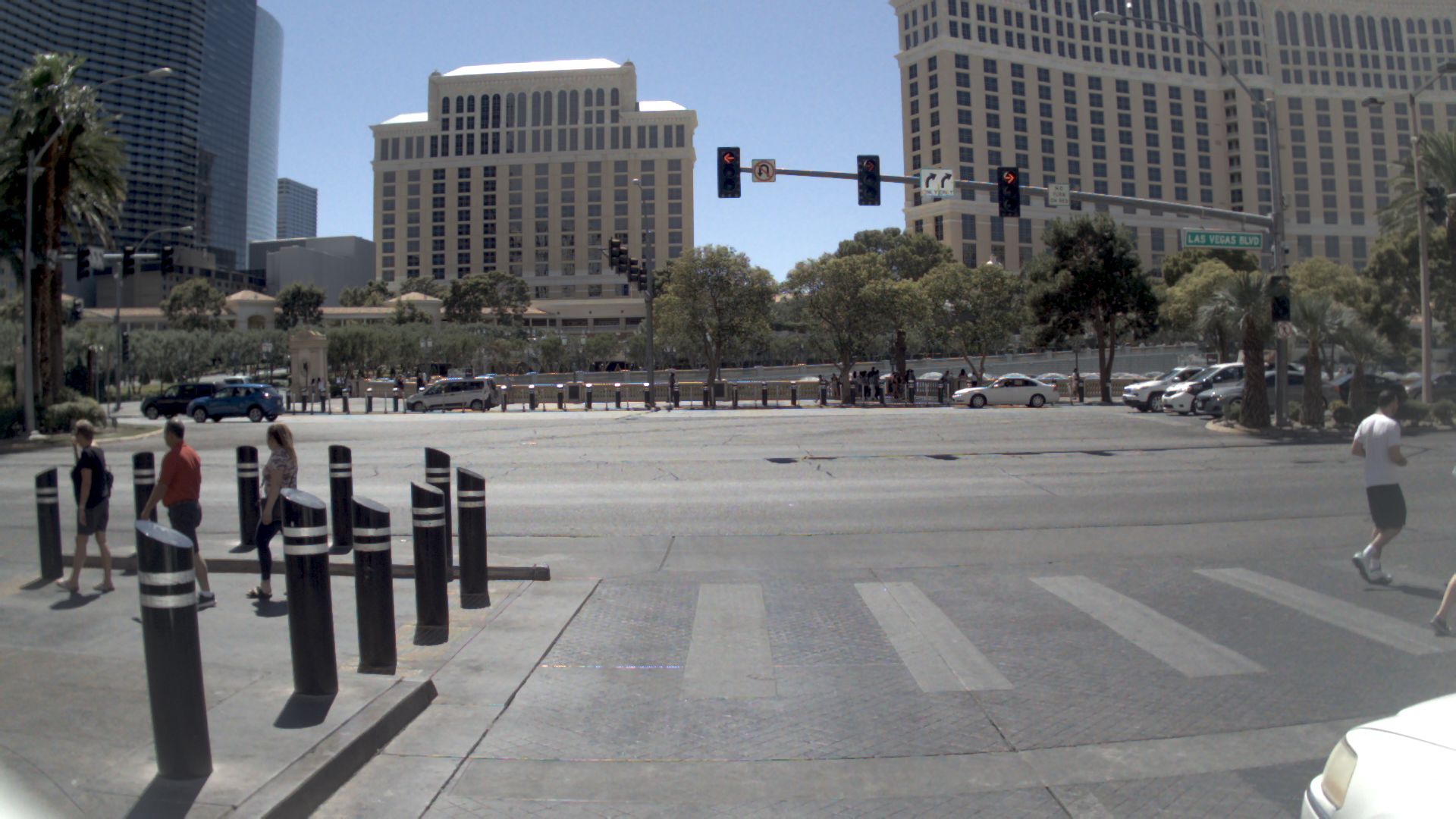}{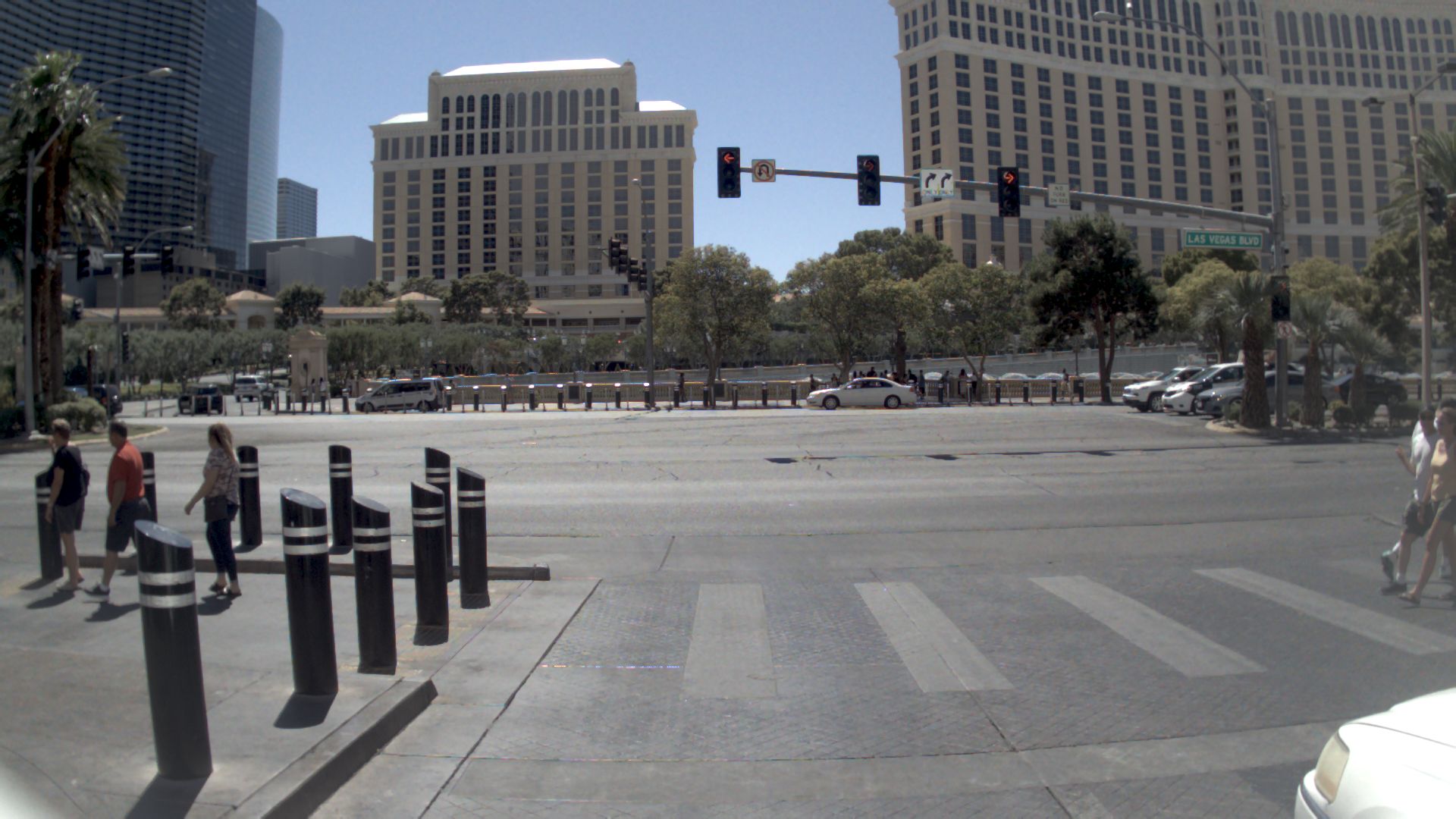}{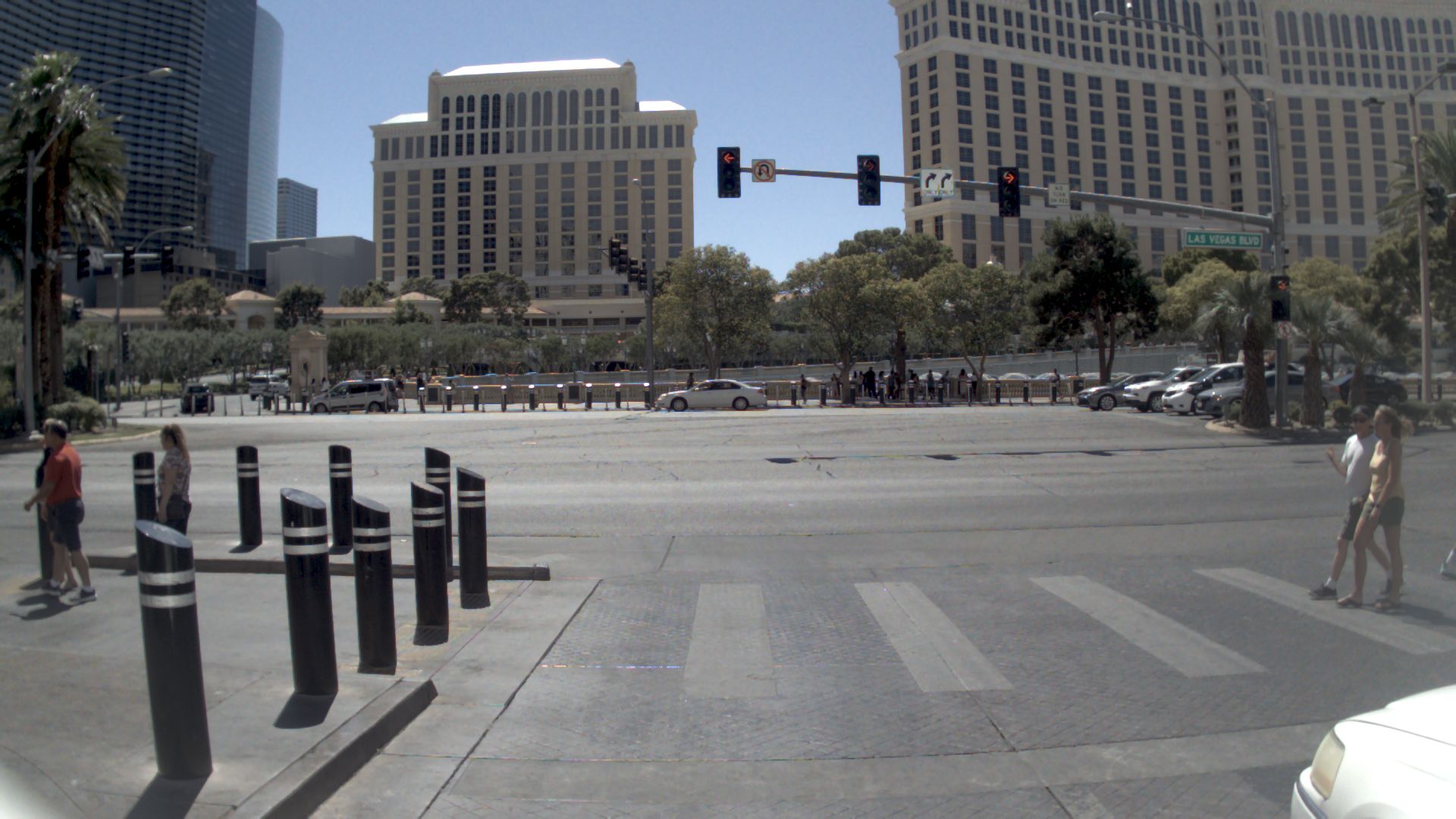}
{stream}{stream}{\textbf{trigger}}{stream}
\end{DSBLevelBlock}

\begin{DSBLevelBlock}
\DSBPromptHeader
{L3: temporal comprehension}
{Corresponding temporal probe}
{{\emph{Did a pedestrian or non-motor vehicle just cross and affect my passage?}}}
\DSBFourFrames
{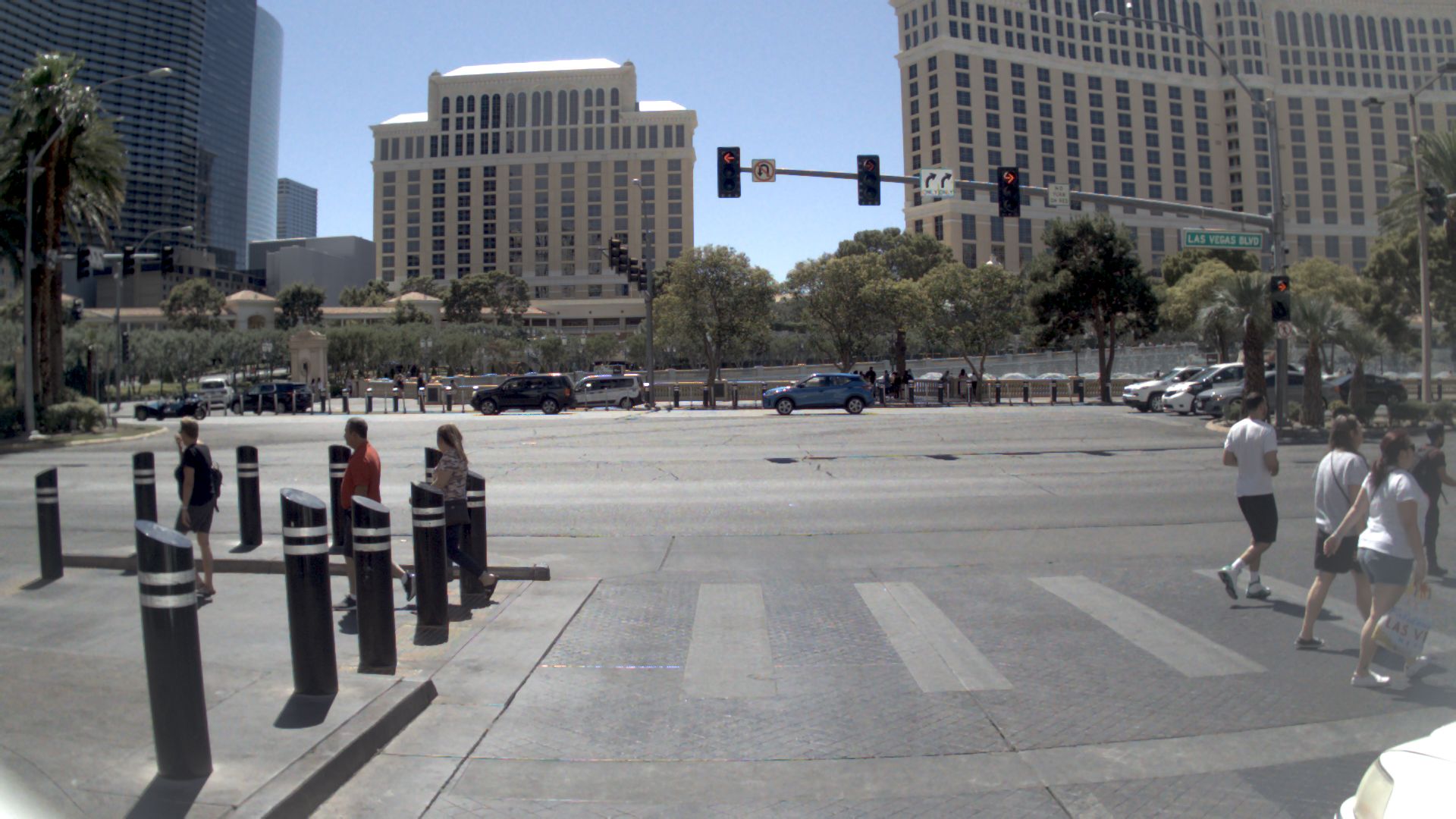}{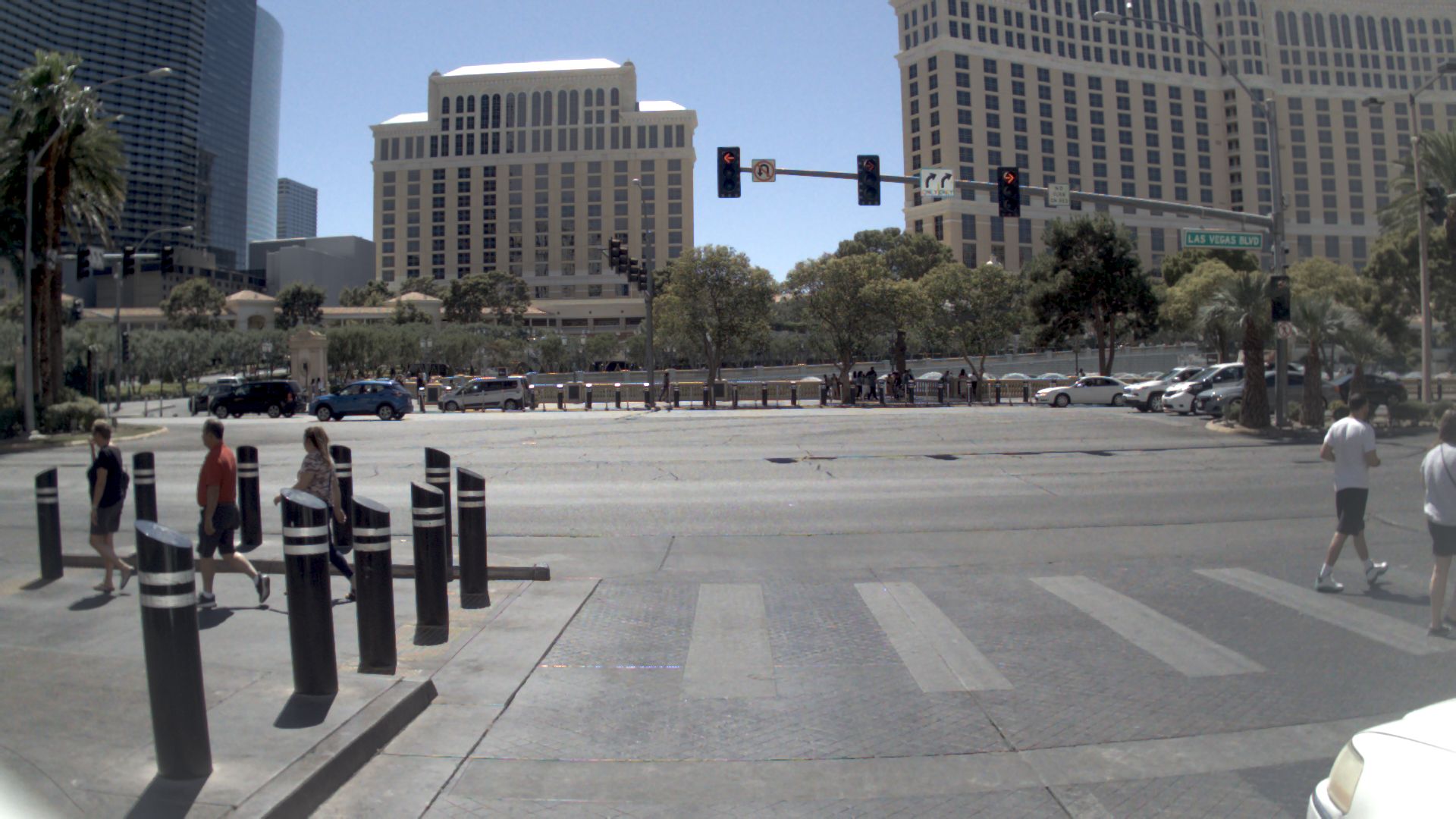}{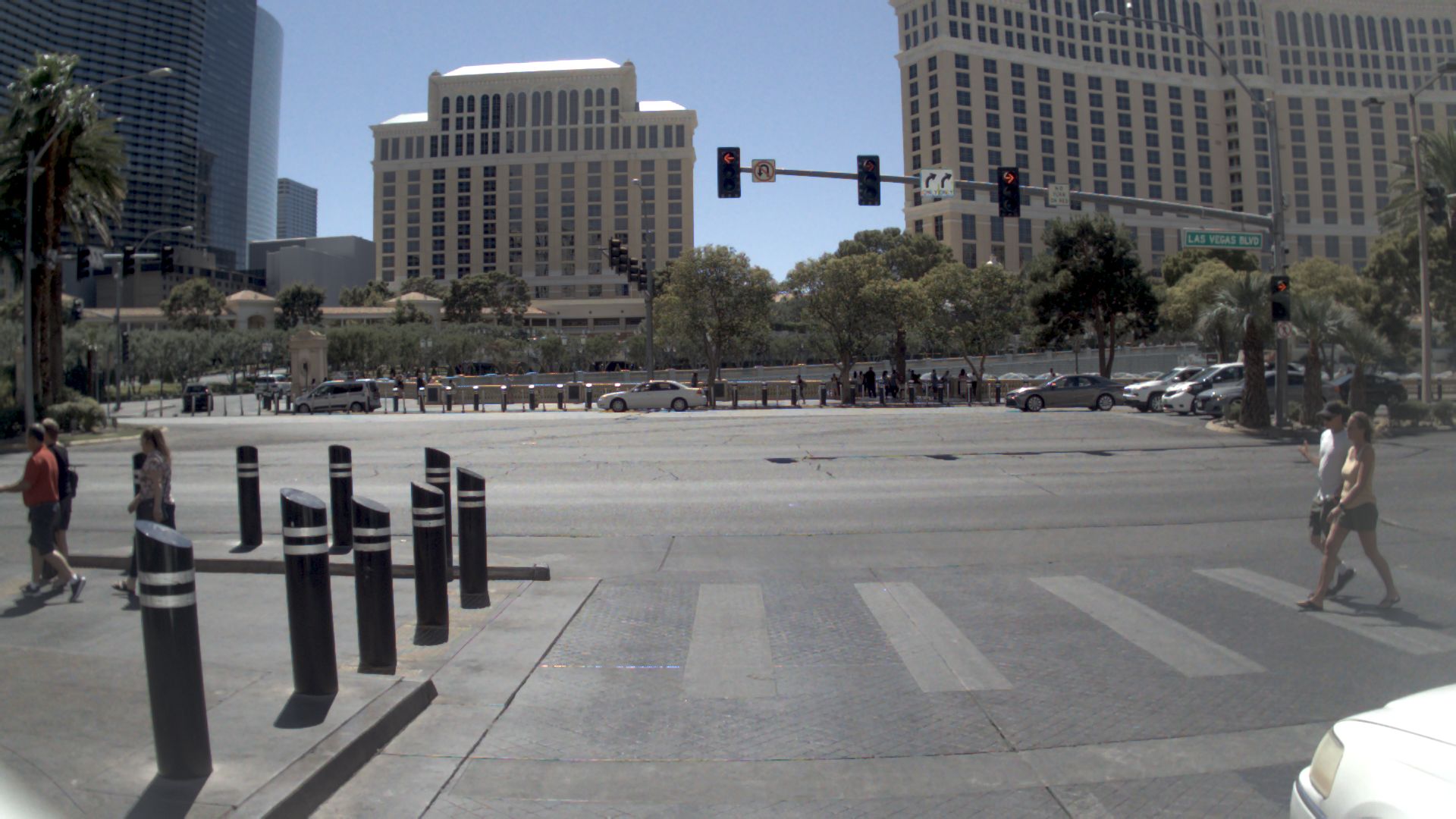}{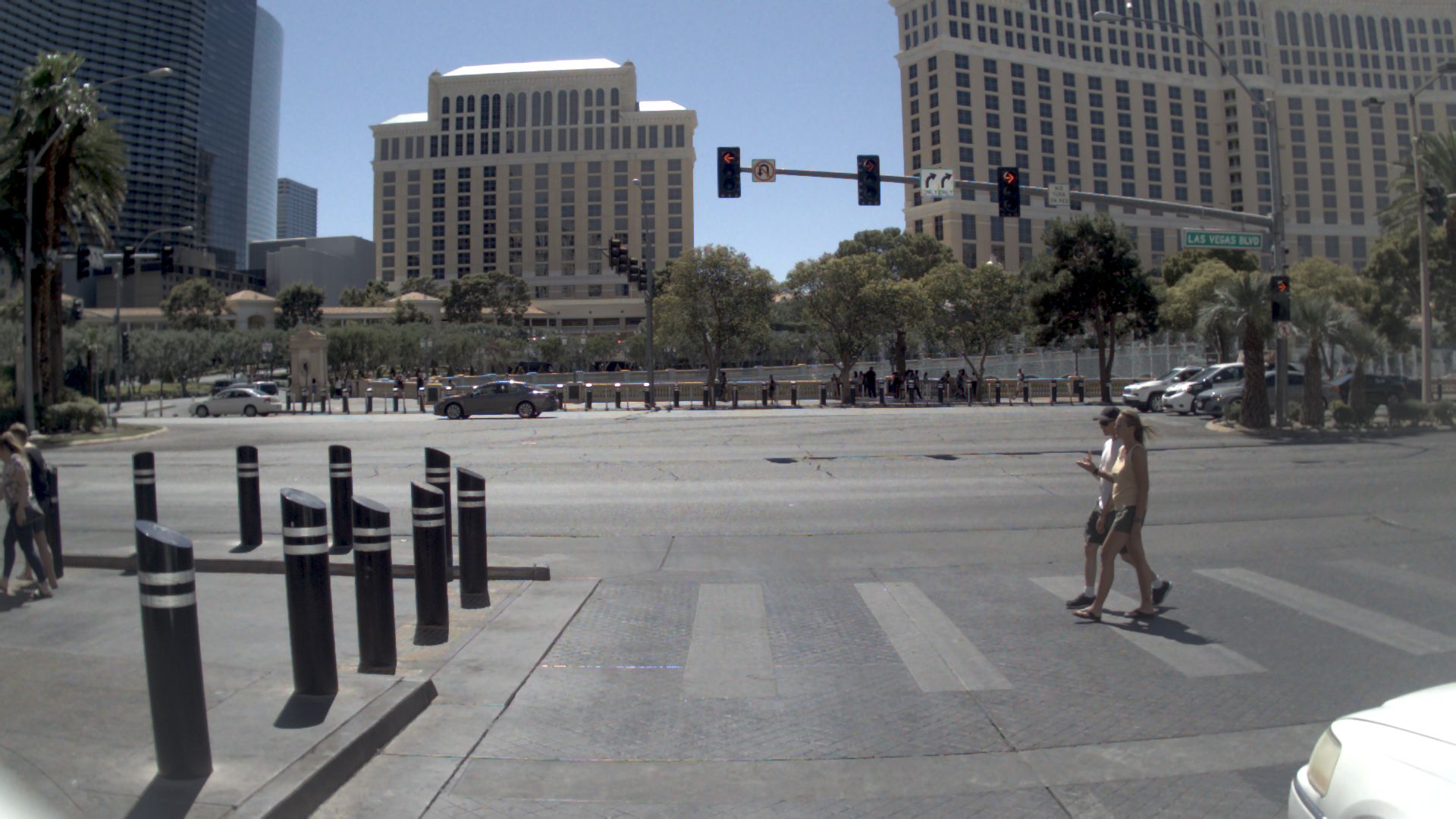}
{$t_0$}{$t_1$}{$t_2$}{$t_3$}
\end{DSBLevelBlock}

\DSBSinglePair
{%
\DSBCardText
{L2: relational understanding}
{Single-frame ego-centric relation probe}
{{\emph{Is there a motor vehicle on the road ahead?}}}
}
{ex5_l2.jpg}
{L2 single-image frame}
{%
\DSBCardText
{L1: recognition}
{Single-frame recognition probe}
{{\emph{Can you see pedestrians in the scene?}}}
}
{ex5_l1.jpg}
{L1 single-image frame}

\vspace{0.25em}
\captionof{figure}{Event-linked L1--L4 task example for a vulnerable-road-user-crossing event.}
\label{fig:l1234-example-vru-crossing}
\end{minipage}

\clearpage
\subsection{Representative alert timelines}
\label{app:qualitative-alert-timelines}
\begin{figure}[H]
\centering
\includegraphics[width=\textwidth]{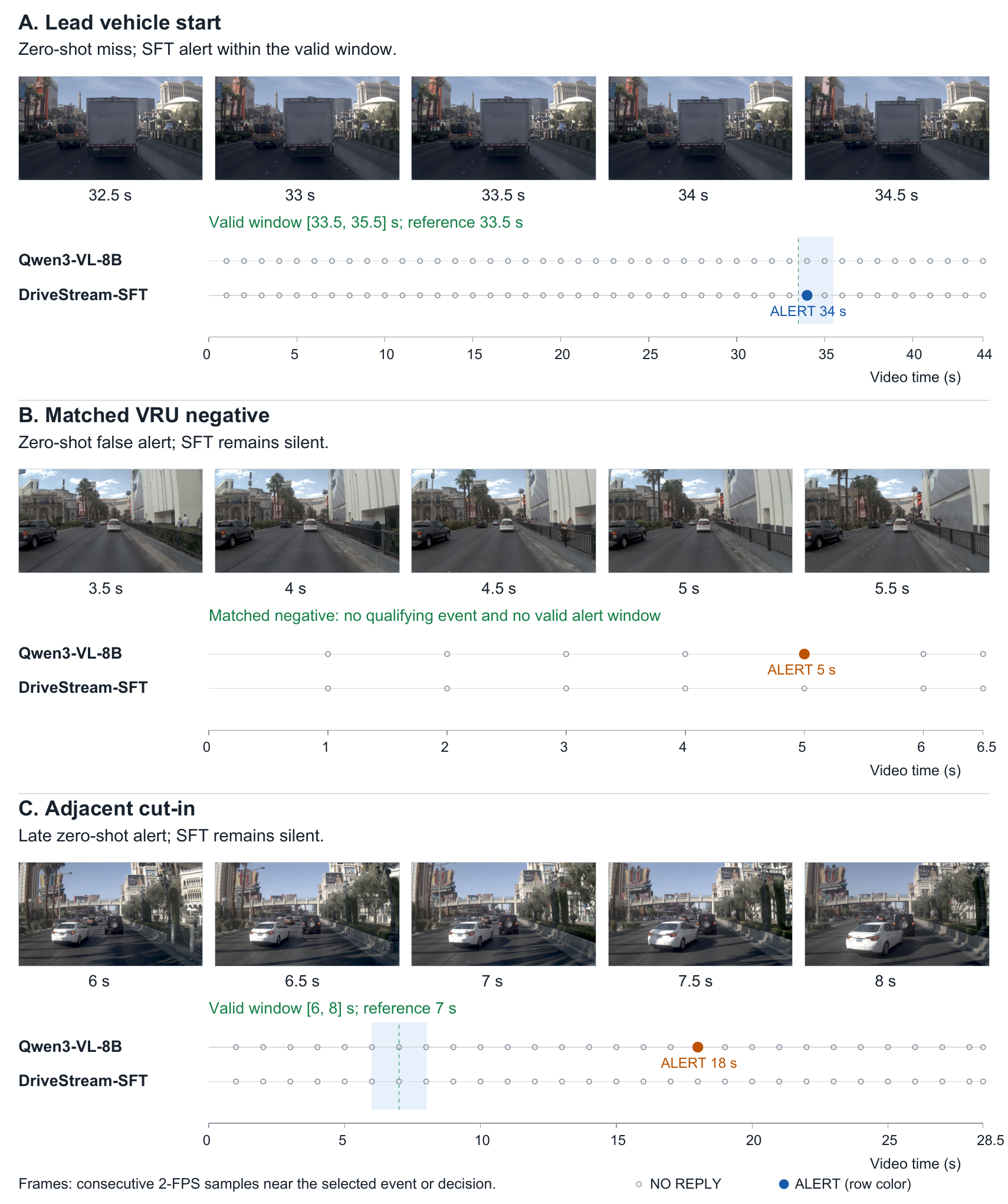}
\caption{Representative causal watch-and-notify timelines. Each row shows every
recorded model decision; hollow circles denote \texttt{NO REPLY}, and filled
circles denote \texttt{ALERT}. Shading marks the valid window for positive
sessions. (A) The reference is 33.5\,s: Qwen3-VL-8B remains silent, while
DriveStream-SFT alerts at 34.0\,s inside [33.5, 35.5]\,s. (B) Pedestrians remain
behind the roadside railing: Qwen3-VL-8B alerts at 5.0\,s, while SFT remains
silent throughout the 6.5-s session. (C) For an adjacent cut-in with reference
7.0\,s and valid window [6.0, 8.0]\,s, Qwen3-VL-8B first alerts at 18.0\,s,
while SFT remains silent throughout the 28.5-s session.}
\label{fig:qualitative-alert-timelines}
\end{figure}

\clearpage
\section{Hyperparameters and Compute Resources}
\label{app:sft-infer-hparams}

This appendix reports the training and inference hyperparameters used for the
DriveStream-SFT variants. The hyperparameters are fixed across runs. All local
training and open-weight inference experiments were conducted on 16 NVIDIA H20
GPUs.

\begin{table}[H]
    \centering
    \small
    \setlength{\tabcolsep}{5pt}
    \renewcommand{\arraystretch}{1.05}
    \caption{
    Shared SFT training hyperparameters for DriveStream-SFT variants.
    All four SFT runs use the same backbone, optimizer, training method, sequence budget,
    learning rate, hardware setting, and effective batch size.
    }
    \label{tab:sft-shared-hparams}
    \begin{tabular}{l l}
        \toprule
        \textbf{Parameter} & \textbf{Value} \\
        \midrule
        Base model & Qwen3-VL-8B-Instruct \\
        Tuning method & Full fine-tuning \\
        Distributed training & DeepSpeed ZeRO-3 \\
        Optimizer & AdamW \\
        Learning rate & \(1\times10^{-5}\) \\
        Adam \(\beta_1\) & 0.9 \\
        Adam \(\beta_2\) & 0.95 \\
        Weight decay & 0.1 \\
        Image max token number & 256 \\
        Maximum sequence length & 40{,}960 \\
        Number of epochs & 3 \\
        GPU type & NVIDIA H20 \\
        Number of GPUs & 16 \\
        Per-GPU batch size & 1 \\
        Gradient accumulation steps & 4 \\
        Effective batch size & 64 \\
        \bottomrule
    \end{tabular}
\end{table}

\begin{table}[H]
    \centering
    \small
    \setlength{\tabcolsep}{5pt}
    \renewcommand{\arraystretch}{1.05}
    \caption{
    Shared L4 inference and evaluation hyperparameters.
    These settings are used for all DriveStream-SFT L4 evaluations unless otherwise stated.
    }
    \label{tab:infer-shared-hparams}
    \begin{tabular}{l l}
        \toprule
        \textbf{Parameter} & \textbf{Value} \\
        \midrule
        Target input frame rate & 2 FPS \\
        Frames per decision turn & 2 \\
        Decision cadence & 1 Hz \\
        Inference backend & vLLM \\
        Evaluation batch size & 64 \\
        Maximum model length & 81{,}920 \\
        Maximum image pixels & 262{,}144 \\
        \bottomrule
    \end{tabular}
\end{table}

\clearpage
\section{Experimental Details and Additional Results}
\label{app:experiments}
\label{app:results}

\subsection{Protocol Details and Release Artifacts}
\label{app:protocol}

\paragraph{Prompting.}
Instruction-following VLMs use the locked task prompts and output policies. Third-party streaming baselines retain their native interfaces with the fixed adapters below. Closed-source API inputs are re-encoded at each call; history handling follows the model-family interface.
For prompt details, please refer to Appendix~\ref{app:prompts}

\phantomsection
\label{app:adapter-details}
\paragraph{Adapter-level output mapping for streaming baselines.}
Streaming and proactive baselines expose different native silence and trigger mechanisms. 
To evaluate all systems under the same DriveStreamBench first-decision rule, we fix a model-family adapter before evaluation and map each turn to the common alerting interface summarized in Table~\ref{tab:adapter-mapping}. 
Instruction-following VLMs use strict literal parsing: normalized \texttt{NO REPLY} and \texttt{ALERT} are accepted, while other non-empty outputs are mapped to \texttt{INVALID}. 
For third-party streaming transfer baselines, we use a more permissive action-level adapter: empty output, native silence, or normalized \texttt{NO REPLY} is mapped to \texttt{NO REPLY}, while other non-empty outputs are mapped to \texttt{ALERT}. 
M4 keeps its native interface trigger adapter. 
All adapters are fixed before evaluation and are not tuned per category or per example.

\begin{table}[H]
\centering
\footnotesize
\setlength{\tabcolsep}{3pt}
\renewcommand{\arraystretch}{1.08}
\caption{
Fixed output adapters used in L4 evaluation.
Instruction-following VLMs use strict literal parsing.
Third-party streaming transfer baselines use a more permissive action-level adapter, where any non-empty non-\texttt{NO REPLY} output is treated as an alert action.
M4 keeps its native interface trigger adapter.
}
\label{tab:adapter-mapping}
\begin{tabularx}{\textwidth}{L{0.24\textwidth} L{0.20\textwidth} L{0.27\textwidth} X}
\toprule
Model family & Parser / adapter & Silence mapping \((\to \texttt{NO REPLY})\) & Alert / invalid mapping \\
\midrule
Qwen3-VL ladder &
Strict literal &
Empty output or normalized \texttt{NO REPLY}. &
Normalized \texttt{ALERT} \(\to\) \texttt{ALERT}; other non-empty outputs \(\to\) \texttt{INVALID}. \\

DriveStream-SFT / SFT-L4 &
Strict literal &
Empty output or normalized \texttt{NO REPLY}. &
Normalized \texttt{ALERT} \(\to\) \texttt{ALERT}; other non-empty outputs \(\to\) \texttt{INVALID}. \\

GPT-5.4 / Gemini 3.1 Pro &
Strict literal &
Empty output or normalized \texttt{NO REPLY}. &
Normalized \texttt{ALERT} \(\to\) \texttt{ALERT}; other non-empty outputs \(\to\) \texttt{INVALID}. \\

\midrule
SimpleStream &
Action-level transfer &
Empty output or normalized \texttt{NO REPLY}. &
Any non-empty output not normalized to \texttt{NO REPLY} \(\to\) \texttt{ALERT}. \\

StreamForest &
Action-level transfer &
Empty output or normalized \texttt{NO REPLY}. &
Any non-empty output not normalized to \texttt{NO REPLY} \(\to\) \texttt{ALERT}. \\

Dispider &
Action-level transfer &
Empty output or normalized \texttt{NO REPLY}. &
Any non-empty output not normalized to \texttt{NO REPLY} \(\to\) \texttt{ALERT}. \\

MMDuet2 &
Action-level transfer &
Native suppression by its response-decision mechanism, empty output, or normalized \texttt{NO REPLY}. &
Any triggered step that emits non-empty text not normalized to \texttt{NO REPLY} \(\to\) \texttt{ALERT}. \\

VideoLLM-online &
Action-level transfer &
Native sub-threshold no-decoding step, empty output, or normalized \texttt{NO REPLY}. &
Any decoded non-empty text not normalized to \texttt{NO REPLY} \(\to\) \texttt{ALERT}. \\

M4 &
Native interface &
Native no-trigger state. &
Native requirement-met trigger string, e.g., \texttt{REQUIREMENT MEET AT $t$s}, is regex-parsed and converted to a single \texttt{ALERT} at the extracted video time \(t\). \\

Human &
Interface action &
Literal \texttt{NO REPLY}. &
Literal \texttt{ALERT}. \\

\bottomrule
\end{tabularx}
\end{table}

\paragraph{L1 and L2 accuracy details.}
Each L1 and L2 example has a single gold label \(y_i^{(\ell)}\) drawn from a fixed answer set, either \(\{\texttt{YES}, \texttt{NO}\}\) for yes/no questions or \(\{\texttt{A}, \texttt{B}, \texttt{C}, \texttt{D}\}\) for multiple-choice questions. The raw model output is parsed by attempting a small fixed sequence of candidates --- the cleaned full string, its first non-empty line, a leading \texttt{YES}/\texttt{NO} prefix match, and a leading option-letter (\texttt{A}--\texttt{D}) prefix match --- and the first candidate that normalizes to a label in the question's answer set is taken as the parsed prediction \(\hat{y}_i^{(\ell)}\). Predictions that fail to normalize for any candidate, that are empty, or that the model omitted entirely are counted as incorrect. We report
\[
\mathrm{Acc}_\ell \;=\; \frac{1}{|\mathcal{T}_\ell|} \sum_{i \in \mathcal{T}_\ell} \mathbf{1}\!\left[\hat{y}_i^{(\ell)} =
y_i^{(\ell)}\right], \qquad \ell \in \{\mathrm{L1}, \mathrm{L2}\},
\]
over the full test set of each level; per-subcategory L1 and L2 accuracies use the same definition restricted to one
subcategory.

\paragraph{L3 binary F1 details.}
Each L3 example is mapped to a binary event-occurrence label \(o_i \in \{1,0\}\) where \(o_i=1\) if the gold answer is \texttt{YES} (event occurred) and \(o_i=0\) otherwise. The parsed prediction is mapped the same way. Predictions that fail to parse are treated as the wrong class: a false negative when \(o_i=1\) and a false positive when \(o_i=0\). We report micro \(\mathrm{F1}_{\mathrm{L3}} = 2\,\mathrm{TP} / (2\,\mathrm{TP}+\mathrm{FP}+\mathrm{FN})\) over all L3 sessions; per-subcategory L3 F1 uses the same definition restricted to one subcategory and is not algebraically decomposable into the overall micro F1.

\subsection{Latency Measurement}
\label{app:latency}
For the open-source Qwen3-VL ladder and DriveStream-SFT, we measure single-session latency under bs\,$=$\,1 and max\_num\_seqs\,$=$\,1 on a frozen 102-sample subset stratified by L4 subcategory. 
All rows use the same 16\,$\times$\,NVIDIA H20 GPU, so per-decision numbers are directly comparable within this table. 
We report per-decision median, p95, and max wall-clock time, together with median RTF, defined as median per-decision wall-clock divided by the 1\,s decision interval. 
Under the unrestricted growing-history setting, all evaluated configurations have median RTF below 1, including Qwen3-VL-32B at fps2 (RTF~$=$~0.93); tail latencies remain visible through the p95 and max columns.
Sliding-window or chunked-attention variants are not evaluated in this version.
Frameworks: vLLM~0.10+~\citep{vllm} (TP\,$=$\,4 on 32B).

\begin{table}[H]
\centering
\small
\setlength{\tabcolsep}{6pt}
\caption{Inference latency. Median / p95 / max are per-decision wall-clock times. RTF median is the median per-decision wall-clock divided by the 1\,s decision interval. \textbf{k$^*$} is the first decision turn at which cumulative-median RTF crosses 1.0, a sustainability boundary where higher means a longer real-time window.}
\label{tab:latency}
\begin{tabular}{l c c c c c}
\toprule
Model & Median (ms) & p95 (ms) & max (ms) & RTF median & k$^*$ \\
\midrule
Qwen3-VL-2B (fps2)     &  709 & 1549 & 2354 & 0.71 & 29 \\
Qwen3-VL-2B (fps1)     &  364 &  782 & 1174 & 0.36 & 61 \\
Qwen3-VL-4B (fps2)     &  740 & 1545 & 2407 & 0.74 & 28 \\
Qwen3-VL-4B (fps1)     &  387 &  808 & 1250 & 0.39 & 57 \\
Qwen3-VL-8B (fps2)     &  797 & 1597 & 2607 & 0.80 & 26 \\
Qwen3-VL-8B (fps1)     &  432 &  833 & 1280 & 0.43 & 56 \\
Qwen3-VL-32B (fps2)    &  934 & 1994 & 3095 & 0.93 & 21 \\
Qwen3-VL-32B (fps1)    &  512 & 1084 & 1585 & 0.51 & 41 \\
DriveStream-SFT (fps2) &  740 & 1595 & 2439 & 0.74 & 27 \\
DriveStream-SFT (fps1) &  421 &  845 & 1341 & 0.42 & 55 \\
\bottomrule
\end{tabular}
\end{table}

\subsection{Frame-Rate Sensitivity}
\label{app:fps-ablation}

The default L4 protocol uses 2~FPS as the visual input density while keeping the decision cadence fixed at 1~Hz. 
To test whether the main conclusions are driven by this input-density choice, we rerun the Qwen3-VL ladder and DriveStream-SFT variants at 1~FPS. 
In both settings, the model still makes one decision per second; each decision turn receives one or two video frames.

\begin{table}[H]
    \centering
    \small
    \setlength{\tabcolsep}{5pt}
    \renewcommand{\arraystretch}{1.02}
    \caption{
    Frame-rate sensitivity on L4-Core and L4-OOD.
    Core@2 denotes \(\mathrm{CoreMacro}@2\mathrm{s}\) over S1--S5.
    S6@2 denotes \(\mathrm{AlertScore}@2\mathrm{s}\) on the S6 sudden-appearance stress split.
    \(\Delta\) is computed as 2~FPS minus 1~FPS.
    All values are percentages except the deltas, which are percentage-point differences.
    }
    \label{tab:fps-sensitivity}
    \begin{tabular}{l c c c c c c}
        \toprule
        Model &
        1~FPS Core@2 &
        2~FPS Core@2 &
        \(\Delta\) &
        1~FPS S6@2 &
        2~FPS S6@2 &
        \(\Delta\) \\
        \midrule
        Qwen3-VL-2B        &  4.3 &  5.1 & \(+0.8\) & 51.9 & 46.9 & \(-5.0\) \\
        Qwen3-VL-4B        & 24.7 & 22.6 & \(-2.1\) & 54.3 & 53.7 & \(-0.6\) \\
        Qwen3-VL-8B        & 30.0 & 30.3 & \(+0.3\) & \underline{59.3} & \underline{62.2} & \(+2.9\) \\
        Qwen3-VL-32B       & 45.0 & 41.7 & \(-3.3\) & \textbf{66.5} & \textbf{63.1} & \(-3.4\) \\
        DriveStream-SFT-L4 & \underline{58.4} & \underline{62.1} & \(+3.7\) & 30.7 & 26.3 & \(-4.4\) \\
        DriveStream-SFT    & \textbf{72.3} & \textbf{72.3} & \(+0.0\) & 47.3 & 46.2 & \(-1.1\) \\
        \bottomrule
    \end{tabular}
\end{table}

Table~\ref{tab:fps-sensitivity} shows that L4 performance is largely insensitive to doubling the visual input density from 1~FPS to 2~FPS. 
On L4-Core, all six models change by at most 3.7 percentage points, and the full DriveStream-SFT model is unchanged. 
The S6 stress split shows a similar pattern, with all changes within 5.0 percentage points. 
These results suggest that, for the current L4 sessions and tested models, 1~FPS already captures most of the visual evidence needed for the alerting decision. 
The default 2~FPS setting should therefore be read as a conservative input-density choice rather than as a source of inflated leaderboard performance. 
The main exception is DriveStream-SFT-L4, which gains 3.7 points on L4-Core at 2~FPS; conversely, Qwen3-VL-4B and Qwen3-VL-32B are slightly better at 1~FPS. 
We treat these small reversals as evaluation variation or possible sensitivity to shorter visual contexts, rather than as evidence that lower frame rate is systematically preferable.

\begin{table}[H]
    \centering
    \small
    \setlength{\tabcolsep}{7pt}
    \renewcommand{\arraystretch}{1.02}
    \caption{
    RTF is the median per-decision wall-clock divided by the 1\,s decision interval; lower is better.
    The ratio column is \(\mathrm{RTF}_{2\mathrm{FPS}} / \mathrm{RTF}_{1\mathrm{FPS}} \).
    }
    \label{tab:fps-rtf}
    \begin{tabular}{l c c c}
        \toprule
        Model & 1~FPS RTF & 2~FPS RTF & Ratio \\
        \midrule
        Qwen3-VL-2B      & 0.36 & 0.71 & \(1.97\times\) \\
        Qwen3-VL-4B      & 0.39 & 0.74 & \(1.90\times\) \\
        Qwen3-VL-8B      & 0.43 & 0.80 & \(1.86\times\) \\
        Qwen3-VL-32B     & 0.51 & 0.93 & \(1.82\times\) \\
        DriveStream-SFT  & 0.42 & 0.74 & \(1.76\times\) \\
        \bottomrule
    \end{tabular}
\end{table}

Table~\ref{tab:fps-rtf} shows the corresponding deployment trade-off. 
Moving from 1~FPS to 2~FPS increases session-level RTF by \(1.76\times\)--\(1.97\times\), close to the expected increase from processing twice as many frames. 
Combined with Table~\ref{tab:fps-sensitivity}, this indicates that 2~FPS roughly doubles the visual processing load while bringing only small changes in L4-Core and S6 alerting scores.

\subsection{Short-Version Protocol for Proprietary API VLMs}
\label{app:short_version}

Proprietary API VLMs are evaluated on a short-version protocol due to API image/context limits and evaluation cost. This setting uses the same L4 watch-and-notify semantics and scoring rules as the main benchmark, but crops every L4 test session to a shorter video segment. For positive sessions, we crop to a window of at most 15 seconds around the annotated reference time \(t_{\mathrm{ref}}\). For negative sessions, we use the first 15 seconds of the session. The cropping is applied to all test samples rather than selecting a subset. These results are reported in Table~\ref{tab:short} and should be interpreted as short-version protocol-transfer results under API-serving constraints. For all proprietary APIs, we set temperature to 0, set top-k to 1, and disable thinking mode.

To calibrate the short-version L4 setting, we also evaluate Qwen3-VL-8B, Qwen3-VL-32B, and DriveStream-SFT on the same cropped L4 test set. Qwen3-VL-8B and Qwen3-VL-32B are evaluated zero-shot, while DriveStream-SFT uses the same trained reference model as in the full-test leaderboard. These rows serve as reference points for interpreting proprietary API results without collapsing the primary benchmark to the short-version setting.

\begin{table}[H]
\centering
\small
\setlength{\tabcolsep}{4pt}
\caption{Short-version L4 evaluation for proprietary API VLMs. Qwen3-VL and DriveStream-SFT rows are included as reference points.}
\label{tab:short}
\begin{tabular}{l c c c}
\toprule
Model & \textbf{Core Macro} & \(\mathrm{Hit}@2\mathrm{s}\) & \(\mathrm{NSR}\) \\
\midrule
GPT-5.4 & \underline{53.3} & \underline{47.7} & \underline{84.6} \\
Gemini 3.1 Pro & 29.3 & 39.0 & 25.3 \\
\midrule
Qwen3-VL-8B & 34.5 & 32.9 & 55.4 \\
Qwen3-VL-32B & 43.1 & 36.5 & 72.9 \\
DriveStream-SFT & \textbf{72.4} & \textbf{61.8} & \textbf{97.6} \\
\bottomrule
\end{tabular}
\end{table}

Table~\ref{tab:short} shows that GPT-5.4 improves over zero-shot open-weight backbones under the short-version L4 protocol, but still does not close the gap to the benchmark-trained reference model. 
GPT-5.4 reaches 53.3 Core Macro, outperforming Qwen3-VL-32B under the same cropped setting by 10.2 points, but remains 19.1 points below DriveStream-SFT (short). 
Gemini 3.1 Pro obtains a higher positive-side hit rate than Qwen3-VL-32B (39.0 vs. 36.5), but its much lower \(\mathrm{NSR}\) (25.3) indicates frequent false alerts on negative sessions, leading to a lower Core Macro. 
These results support the same conclusion as the full-test leaderboard: stronger general-purpose VLMs help, but user-conditioned watch-and-notify still requires calibrated alert timing and no-alert restraint.

\begin{table}[H]
\centering
\small
\setlength{\tabcolsep}{5pt}
\renewcommand{\arraystretch}{1.02}
\caption{
Per-category short-version results for proprietary API VLMs.
The upper block reports \(\mathrm{AlertScore}@2\mathrm{s}\); the lower block reports \(\mathrm{NSR}\).
All values are percentages.
}
\label{tab:short-api-breakdown}
\begin{tabular}{l l c c c c c}
\toprule
Metric & Model & S1 & S2 & S3 & S4 & S5 \\
\midrule
\multirow{2}{*}{\(\mathrm{AlertScore}@2\mathrm{s}\)}
& GPT-5.4        & 66.3 & 56.6 & 83.7 & 2.4 & 57.3 \\
& Gemini 3.1 Pro & 39.6 & 23.1 & 39.8 & 3.4 & 40.7 \\
\midrule
\multirow{2}{*}{\(\mathrm{NSR}\)}
& GPT-5.4        & 58.7 & 94.4 & 92.0 & 95.0 & 82.7 \\
& Gemini 3.1 Pro & 40.7 & 19.3 & 35.3 & 1.9 & 29.0 \\
\bottomrule
\end{tabular}
\end{table}

The per-category breakdown in Table~\ref{tab:short-api-breakdown} shows that S4 adjacent cut-in remains difficult even for proprietary API VLMs: both GPT-5.4 and Gemini 3.1 Pro obtain near-zero S4 AlertScore. 
The two API models differ mainly in negative-session restraint. 
GPT-5.4 maintains high \(\mathrm{NSR}\) on S2--S5 except for S1, whereas Gemini 3.1 Pro has low \(\mathrm{NSR}\) across all five categories, especially S4. 
This suggests that Gemini's lower short-version score is driven less by the absence of positive-side triggering and more by poor no-alert restraint.

\subsection{Timing-only reference baselines}
\label{app:temporal-prior}

Table~\ref{tab:temporal-prior} gives a timing-only reference for the leaderboard. 
These baselines never observe video frames: they either remain silent, alert immediately, or alert at fixed or category-specific times estimated from the training split. 
Their scores indicate how much of the task can be solved from simple alert-position priors alone. 
The positive-oracle temporal prior uses the positive/negative label and is for reference only.

\begin{table}[H]
    \centering
    \small
    \setlength{\tabcolsep}{4pt}
    \caption{
    \textit{Positive-oracle Temporal Prior}, timing-only baselines for L4-Core.
    Models do not observe video frames and alert only by simple rules or training-split timing priors. All values are percentages.
    }
    \label{tab:temporal-prior}
    \begin{tabular}{l l c c c}
        \toprule
        Baseline & Input & \textbf{Core Macro} & \(\mathrm{Hit}@2\mathrm{s}\) & \(\mathrm{NSR}\) \\
        \midrule
        Always Silent                  & none                         & 0.0           & 0.0  & 100.0 \\
        Alert at First Turn            & none                         & 0.0           & 5.3  &   0.0 \\
        Fixed Absolute Time            & train-time prior             & 0.0           & 5.7  &   0.0 \\
        Category Temporal Prior        & category $+$ train prior     & 0.0           & 5.7  &   0.0 \\
        Positive-oracle Temporal Prior & category $+$ positive oracle & \textbf{9.9}  & 5.7  & 100.0 \\
        \bottomrule
    \end{tabular}
\end{table}

\subsection{Bootstrap Confidence Intervals for Core Macro}
\label{app:ci}

We estimate sampling uncertainty for \(\mathrm{CoreMacro}@2\mathrm{s}\) with stratified bootstrap over L4-Core sessions. 
For each model and each bootstrap replicate, we resample positive and negative sessions with replacement independently within each L4-Core subcategory, recompute \(\mathrm{Hit}@2\mathrm{s}\), \(\mathrm{NSR}\), and \(\mathrm{AlertScore}@2\mathrm{s}\) per subcategory, and then average \(\mathrm{AlertScore}@2\mathrm{s}\) over S1--S5. 
We use \(B=10{,}000\) bootstrap replicates with random seed 42 and report percentile 95\% confidence intervals. 
The human calibration row is not included here because it is collected on a balanced pilot subset with a different protocol.

\begin{table}[H]
    \centering
    \small
    \setlength{\tabcolsep}{6pt}
    \caption{
    Bootstrap 95\% confidence intervals for \(\mathrm{CoreMacro}@2\mathrm{s}\).
    The intervals are computed by stratified session-level bootstrap within each L4-Core subcategory, resampling positives and negatives independently. All values are percentages.
    }
    \label{tab:ci}
    \begin{tabular}{l c c}
        \toprule
        Model & Core Macro & 95\% CI \\
        \midrule
        Qwen3-VL-2B          &  5.1 & [\phantom{0}2.7, \phantom{0}7.4] \\
        Qwen3-VL-4B          & 22.6 & [19.9, 25.2] \\
        Qwen3-VL-8B          & 30.3 & [27.4, 33.0] \\
        Qwen3-VL-32B         & 41.7 & [38.4, 44.7] \\
        \midrule
        StreamForest         &  0.0 & [\phantom{0}0.0, \phantom{0}0.0] \\
        SimpleStream         & 17.9 & [14.7, 21.0] \\
        Dispider             &  9.9 & [\phantom{0}6.7, 12.8] \\
        VideoLLM-online      &  0.0 & [\phantom{0}0.0, \phantom{0}0.0] \\
        MMDuet2              & 10.6 & [\phantom{0}7.5, 13.5] \\
        M4         &  5.8 & [\phantom{0}3.1, \phantom{0}8.5] \\
        \midrule
        DriveStream-SFT-L4   & \underline{62.1} & \underline{[59.2, 64.8]} \\
        DriveStream-SFT      & \textbf{72.3} & \textbf{[68.8, 75.5]} \\
        \bottomrule
    \end{tabular}
\end{table}

The confidence intervals support the main leaderboard trends. 
DriveStream-SFT remains well separated from the strongest zero-shot backbone and from all third-party streaming/proactive transfer baselines, even after the latter are evaluated with action-level adapters. 
Its interval also does not overlap with the L4-only reference baseline. 
By contrast, differences among weak zero-shot or transfer baselines should not be over-interpreted when their confidence intervals overlap.

\subsection{Hit-Only Positive-Side Timing Analysis}
\label{app:hit-only}

The primary L4 metric, \(\mathrm{AlertScore}@\tau\), combines two requirements: timely alerting on positive sessions and restraint on negative sessions. 
A possible concern is that some general streaming or proactive streaming baselines may receive low \(\mathrm{AlertScore}@\tau\) mainly because the negative-restraint term is too strict. 
To isolate this effect, we report a hit-only analysis that removes the negative-session \(\mathrm{NSR}\) term and keeps only the positive-side timing term \(\mathrm{Hit}@\tau\).

This diagnostic keeps the same positive-session first-decision rule as the main metric. 
A positive session is counted as a hit only if the first non-\texttt{NO REPLY} decision is mapped to the literal \texttt{ALERT} action and its video time lies inside the correctness window \(W_\tau\). 
Negative sessions are ignored in this analysis. 
Thus, the hit-only score should be read as a positive-side detection-and-timing diagnostic, not as a deployable alerting metric: a useful watch-and-notify system must still avoid false alerts on negatives, which is why \(\mathrm{AlertScore}@\tau\) remains the primary ranking metric.

For compactness, we define
\[
\mathrm{CoreHit}@\tau =
\frac{1}{5}
\sum_{c\in\{\mathrm{S1},\mathrm{S2},\mathrm{S3},\mathrm{S4},\mathrm{S5}\}}
\mathrm{Hit}_{c}@\tau ,
\]
where \(\mathrm{Hit}_{c}@\tau\) is computed within L4-Core subcategory \(c\).

\begin{table}[H]
    \centering
    \small
    \setlength{\tabcolsep}{4pt}
    \renewcommand{\arraystretch}{1.02}
    \caption{
    Hit-only positive-side timing results on L4-Core at \(\tau=2\)s.
    Unlike \(\mathrm{AlertScore}@2\mathrm{s}\), this table removes the negative-session \(\mathrm{NSR}\) term and reports only \(\mathrm{Hit}@2\mathrm{s}\) on positives.
    \textbf{CoreHit@2} is the unweighted mean over S1--S5.
    All values are percentages.
    }
    \label{tab:hit-only-subcat}
    \begin{tabular}{l c c c c c c c}
        \toprule
        Model & fps & S1 & S2 & S3 & S4 & S5 & \textbf{CoreHit@2s} \\
        \midrule
        \multicolumn{8}{@{}l}{\textit{Open-source VLMs, zero-shot}} \\
        Qwen3-VL-2B          & 2 &  0.0 &  0.0 &  1.9 &  0.0 & 12.2 &  2.8 \\
        Qwen3-VL-4B          & 2 &  5.9 &  0.0 & 29.1 &  1.2 & 64.3 & 20.1 \\
        Qwen3-VL-8B          & 2 & 23.8 & 10.6 & 46.6 &  2.5 & 63.3 & 29.4 \\
        Qwen3-VL-32B         & 2 & 32.7 & 11.7 & 72.8 &  4.9 & 59.2 & 36.3 \\
        \midrule
        \multicolumn{8}{@{}l}{\textit{General streaming transfer baselines}} \\
        StreamForest         & 1 &  0.0 &  0.0 &  0.0 &  0.0 &  0.0 &  0.0 \\
        SimpleStream         & 1 &  5.0 &  0.0 & 39.8 &  0.0 & 14.3 & 11.8 \\
        Dispider             & 1 &  1.0 &  1.1 &  5.8 &  \underline{8.6} & 12.2 &  5.8 \\
        VideoLLM-online      & 2 &  0.0 &  0.0 &  0.0 &  4.9 & 21.4 &  5.3 \\
        \midrule
        \multicolumn{8}{@{}l}{\textit{Proactive-trained streaming transfer baselines}} \\
        MMDuet2              & 1 &  0.0 &  0.0 & 15.5 &  0.0 & 15.3 &  6.2 \\
        M4         & 1 &  7.9 &  0.0 &  1.0 &  4.9 &  2.0 &  3.2 \\
        \midrule
        \multicolumn{8}{@{}l}{\textit{Reference baselines trained on DriveStreamBench}} \\
        DriveStream-SFT-L4   & 2 & \textbf{85.2} & \underline{34.0} & \underline{77.7} &  0.0 & \underline{69.4} & \underline{53.3} \\
        DriveStream-SFT      & 2 & \underline{84.2} & \textbf{53.2} & \textbf{81.5} & \textbf{18.5} & \textbf{72.5} & \textbf{62.0} \\
        \bottomrule
    \end{tabular}
\end{table}

\begin{table}[H]
    \centering
    \small
    \setlength{\tabcolsep}{7pt}
    \renewcommand{\arraystretch}{1.02}
    \caption{
    Hit-only tolerance curve on L4-Core.
    Each entry is \(\mathrm{CoreHit}@\tau\), the macro average of positive-side \(\mathrm{Hit}@\tau\) over S1--S5.
    The negative-session \(\mathrm{NSR}\) term is omitted.
    All values are percentages.
    }
    \label{tab:hit-only-tau}
    \begin{tabular}{l c c c c c}
        \toprule
        Model & fps & \(\mathrm{Hit}@1\mathrm{s}\) & \(\mathrm{Hit}@2\mathrm{s}\) & \(\mathrm{Hit}@3\mathrm{s}\) & \(\mathrm{Hit}@5\mathrm{s}\) \\
        \midrule
        \multicolumn{6}{@{}l}{\textit{Open-source VLMs, zero-shot}} \\
        Qwen3-VL-2B          & 2 &  1.6 &  2.8 &  3.2 &  4.1 \\
        Qwen3-VL-4B          & 2 & 13.6 & 20.1 & 23.5 & 26.3 \\
        Qwen3-VL-8B          & 2 & 21.0 & 29.4 & 34.2 & 38.8 \\
        Qwen3-VL-32B         & 2 & 24.9 & 36.3 & 41.6 & 48.5 \\
        \midrule
        \multicolumn{6}{@{}l}{\textit{General streaming transfer baselines}} \\
        StreamForest         & 1 &  0.0 &  0.0 &  0.0 &  0.0 \\
        SimpleStream         & 1 &  6.1 & 11.8 & 12.6 & 19.7 \\
        Dispider             & 1 &  2.5 &  5.8 &  6.3 &  9.4 \\
        VideoLLM-online      & 2 &  1.7 &  5.3 &  7.6 & 10.5 \\
        \midrule
        \multicolumn{6}{@{}l}{\textit{Proactive-trained streaming transfer baselines}} \\
        MMDuet2              & 1 &  1.8 &  6.2 &  6.2 & 12.5 \\
        M4         & 1 &  2.5 &  3.2 &  4.0 &  5.4 \\
        \midrule
        \multicolumn{6}{@{}l}{\textit{Reference baselines trained on DriveStreamBench}} \\
        DriveStream-SFT-L4   & 2 & \underline{42.1} & \underline{53.3} & \underline{57.3} & \underline{60.2} \\
        DriveStream-SFT      & 2 & \textbf{49.7} & \textbf{62.0} & \textbf{67.1} & \textbf{75.3} \\
        \bottomrule
    \end{tabular}
\end{table}

The hit-only analysis supports three conclusions. 
First, removing the \(\mathrm{NSR}\) term shows that some third-party transfer baselines contain limited recoverable positive-side signal under the action-level adapter: Dispider, VideoLLM-online, and MMDuet2 rise from near-zero strict-literal Hit to 5--6\% CoreHit@2. 
However, these gains remain far below DriveStream-SFT at 62.0 CoreHit@2 and far below Qwen3-VL-32B at 36.3. 
This indicates that the transfer-baseline gap is not only a negative-restraint artifact; positive-side timely alerting remains weak even under the more permissive action interpretation.

Second, scaling improves positive-side hit rate for the Qwen3-VL ladder, but does not close the gap. 
CoreHit@2 rises from 2.8 for Qwen3-VL-2B to 36.3 for Qwen3-VL-32B, while the 8B DriveStream-SFT reference reaches 62.0. 
The DriveStream-SFT advantage over the strongest zero-shot Qwen3-VL backbone is also stable across tolerance windows: \(+24.8\), \(+25.7\), \(+25.5\), and \(+26.8\) points at \(\tau=1,2,3,5\) seconds, respectively.

Third, multi-task supervision improves the L4-only reference in specific categories. 
The largest gains are on S2 lead-vehicle brake and S4 adjacent cut-in: S2 Hit@2 rises from 34.0 to 53.2, while S4 rises from 0.0 to 18.5. 
The S4 result is especially informative because the L4-only checkpoint never hits the target-timing window on adjacent cut-in, whereas adding L1--L3 supervision makes the low-resource cut-in category partially recover.

\subsection{Summary of Robustness Checks}
\label{app:robustness-sensitivity}

We report three sensitivity analyses for the L4 leaderboard. 
Appendix~\ref{app:first-decision-sensitivity} tests whether the first-decision rule drives the ranking. 
Appendix~\ref{app:parser-sensitivity} reports a strict-literal diagnostic for third-party streaming baselines, showing how much the action-level transfer adapter changes their scores. 
Appendix~\ref{app:s4-excluded} tests whether the low-resource S4 slice drives the headline Core Macro. 
Together, these checks evaluate whether the main results are artifacts of alert-decision policy, output-format compliance, or subcategory weighting.

\subsection{Sensitivity to the First-Decision Rule}
\label{app:first-decision-sensitivity}

The primary L4 metric uses a first-decision rule: the first non-\texttt{NO REPLY} action determines the alert time on positive sessions. 
This is operationally motivated by the interruptive nature of alerts, but it may appear strict if a model emits an early alert before a later correct alert. 
We therefore rescore the same generated logs under a more permissive positive-side variant for instruction-following VLMs: \emph{Any-valid-\texttt{ALERT}} gives hit credit if any strict \texttt{ALERT} falls inside the timing window, while keeping the same negative-side \(\mathrm{NSR}\).

\begin{table}[H]
    \centering
    \small
    \setlength{\tabcolsep}{5pt}
    \renewcommand{\arraystretch}{1.02}
    \caption{
    Sensitivity to positive-side alert-decision rules on L4-Core for instruction-following VLMs.
    \textbf{Official first} is the main first-decision score.
    \textbf{Any-valid} gives hit credit if any strict \texttt{ALERT} falls inside the timing window.
    \textbf{NSR} is unchanged because the negative-side rule is not modified.
    All values are percentages.
    }
    \label{tab:first-decision-sensitivity}
    \begin{tabular}{l c c c}
        \toprule
        Model & Official first Core & Any-valid Core & NSR \\
        \midrule
        Qwen3-VL-2B        &  5.1 &  6.1 & \textbf{98.7} \\
        Qwen3-VL-4B        & 22.6 & 31.6 & 63.6 \\
        Qwen3-VL-8B        & 30.3 & 34.1 & 47.9 \\
        Qwen3-VL-32B       & 41.7 & 48.5 & 63.1 \\
        DriveStream-SFT-L4 & \underline{62.1} & \underline{62.1} & \underline{98.6} \\
        DriveStream-SFT    & \textbf{72.3} & \textbf{72.3} & 96.5 \\
        \bottomrule
    \end{tabular}
\end{table}

The any-valid rule raises several zero-shot Qwen3-VL scores, indicating that these models sometimes emit a mistimed alert before a later in-window alert. 
However, DriveStream-SFT is unchanged at 72.3\%, and its gap over Qwen3-VL-32B remains 23.8 points even under this permissive positive-side rule. 
Thus the main ranking among instruction-following VLMs is not driven by the first-decision rule.

\subsection{Strict Literal Parser Diagnostic}
\label{app:parser-sensitivity}

The official leaderboard uses action-level transfer adapters for third-party streaming baselines, while keeping strict literal parsing for instruction-following VLMs. 
For completeness, Table~\ref{tab:strict-parser-diagnostic} compares the official action-level transfer scores against the all-strict literal diagnostic for streaming/proactive baselines. 
This diagnostic measures output-format compliance rather than the primary watch-and-notify score.

\begin{table}[H]
    \centering
    \small
    \setlength{\tabcolsep}{4pt}
    \renewcommand{\arraystretch}{1.04}
    \caption{
    Strict-literal diagnostic for third-party streaming/proactive baselines.
    \textbf{Official} is the score used in the main leaderboard: action-level transfer adapters for StreamForest, SimpleStream, Dispider, VideoLLM-online, and MMDuet2, and native interface parsing for M4.
    \textbf{Strict diagnostic} treats only literal \texttt{ALERT} as an alert action for the five action-level transfer baselines; M4 keeps its native interface adapter in both columns.
    All values are percentages.
    }
    \label{tab:strict-parser-diagnostic}
    \resizebox{\textwidth}{!}{%
    \begin{tabular}{l c c c c c c c}
        \toprule
        Model &
        Official Core & Official Hit & Official NSR &
        Strict-diagnostic Core & Strict-diagnostic Hit & Strict-diagnostic NSR &
        \(\Delta\) Core \\
        \midrule
        StreamForest     &  0.0 &  0.0 & \textbf{100.0} &  0.0 & 0.0 & \textbf{100.0} & \(+0.0\) \\
        SimpleStream     & \textbf{17.9} & \textbf{11.8} &  92.4 & \textbf{17.9} & \textbf{11.8} & \underline{92.4} & \(+0.0\) \\
        Dispider         &  9.9 &  5.8 &  37.9 &  0.0 & 0.0 & \textbf{100.0} & \(+9.9\) \\
        VideoLLM-online  &  0.0 &  5.3 &   0.0 &  0.0 & 0.0 & \textbf{100.0} & \(+0.0\) \\
        MMDuet2          & \underline{10.6} &  \underline{6.2} &  \underline{97.2} &  1.5 & 0.8 & \textbf{100.0} & \(+9.1\) \\
        M4     &  5.8 &  3.2 &  69.9 &  \underline{5.8} & \underline{3.2} &  69.9 & \(+0.0\) \\
        \bottomrule
    \end{tabular}%
    }
\end{table}

Table~\ref{tab:strict-parser-diagnostic} shows that the action-level adapter recovers some free-form alert behavior for Dispider and MMDuet2, while also exposing poor negative restraint for Dispider and VideoLLM-online. 
Thus the low transfer-baseline scores in the main leaderboard are not caused merely by requiring the literal \texttt{ALERT} token. 
Even after relaxing the interface for these third-party systems, the strongest streaming/proactive transfer baseline remains far below DriveStream-SFT.

\subsection{Turn-Level Failure Cases for Streaming Transfer Baselines}
\label{app:streaming-failure-cases}

The official leaderboard uses action-level transfer adapters for third-party streaming and proactive baselines, while instruction-following VLMs use strict literal parsing. 
This section provides turn-level qualitative examples that explain the score changes in Table~\ref{tab:strict-parser-diagnostic} and the low transfer scores in Table~\ref{tab:leaderboard}. 
The examples are not additional evaluation data; they are representative cases drawn from the frozen L4 inference logs and are included to make the adapter behavior and failure modes concrete. 
Timing windows follow the same category-specific rule as the official \(\mathrm{AlertScore}@2\mathrm{s}\) metric.

\paragraph{Dispider: descriptive substitution instead of \texttt{ALERT}.}
On a typical \texttt{lead\_vehicle\_start} positive session, Dispider produces the same descriptive sentence at every decision step:

\begin{PromptBlock}
qid:     ..._lead_vehicle_start_test_positive_0301
cat:     lead_vehicle_start    t_ref: 4.0    window: [4.0, 6.0]

t = 1.0  "The event happens in zero - five seconds."
t = 2.0  "The event happens in zero - five seconds."
t = 3.0  "The event happens in zero - five seconds."   <- before window
t = 4.0  "The event happens in zero - five seconds."   <- in window
t = 5.0  "The event happens in zero - five seconds."   <- in window
t = 6.0  "The event happens in zero - five seconds."   <- in window
\end{PromptBlock}

Under a strict literal diagnostic, no turn equals \texttt{ALERT}, so the session receives no literal-trigger hit. 
Under the action-level adapter, the first non-\texttt{NO REPLY} output at \(t=1.0\) is mapped to \texttt{ALERT}; it precedes the valid window, so this session is a miss.
This example illustrates a free-form temporal description rather than the requested binary alert action.

\paragraph{MMDuet2: native suppression followed by free-form description.}
MMDuet2's native response-decision mechanism suppresses generation until the model is confident; once it fires, the output can be a description rather than the literal trigger string:

\begin{PromptBlock}
qid:     ..._traffic_light_change_test_positive_0300
cat:     traffic_light_change    t_ref: 5.0    window: [5.0, 7.0]

t = 1.0--6.0   "NO REPLY"            (native suppression)
t = 7.0        "The traffic light has turned yellow,
                indicating the start of the yellow phase."   <- in window
t = 8.0--9.0   "NO REPLY"
\end{PromptBlock}

A strict literal diagnostic rejects the description as non-\texttt{ALERT}. 
The action-level adapter accepts it as an in-window alert action. 
This is the dominant gain pattern for MMDuet2 on traffic-light and VRU cases and accounts for its increase from \(1.5\) to \(10.6\) Core Macro. 
MMDuet2 occasionally emits the literal trigger string, explaining its small but nonzero strict-diagnostic score.

\paragraph{VideoLLM-online: first-turn acknowledgement breaks restraint.}
VideoLLM-online often emits a free-form acknowledgement at the beginning of a session and then remains silent. 
A representative positive session is:

\begin{PromptBlock}
qid:     ..._lead_vehicle_start_test_positive_0302
cat:     lead_vehicle_start    t_ref: 7.5    window: [7.5, 9.5]

t = 0.5    "I can help you with that! Based on the
            current video frames, it appears that the
            front car's left wheel is moving. Would..."   <- before window
t = 1.0--9.5    "NO REPLY"
\end{PromptBlock}

Under the strict-diagnostic parser for transfer baselines, this acknowledgement is not a literal \texttt{ALERT} and therefore gives no positive hit. 
Under the official action-level adapter, it is treated as an alert action, but it still misses the positive session because it occurs before the timing window. 
The same first-turn acknowledgement also appears on negative sessions, where it breaks negative silence. 
This explains why VideoLLM-online has nonzero \(\mathrm{Hit}@2\mathrm{s}\) under the action-level adapter but collapses to \(0.0\%\) \(\mathrm{NSR}\) in the main leaderboard.

\paragraph{M4 / OmniMMI: native trigger-time mismatch.}
M4 / OmniMMI uses its native requirement-met trigger string rather than free-form text at the metric layer. 
The adapter extracts the model's self-reported alert time from the native trigger string:

\begin{PromptBlock}
qid:     ..._lead_vehicle_start_test_positive_0308
cat:     lead_vehicle_start    t_ref: 7.0    window: [7.0, 9.0]

parsed_output:
t = 0.0--8.0   "NO REPLY"
t = 9.0        "REQUIREMENT MEET AT 0.0s"   <- decoded trigger turn
t = 10.0--11.0 "NO REPLY"

Adapter session-level: predicted_alert_times = [0.0]
\end{PromptBlock}

This case illustrates a native time-reference mismatch. 
The decoded trigger turn appears late in the session, while the model's self-reported internal timestamp points to \(0.0\) seconds, which the native adapter uses for scoring. 
We retain the native adapter in Table~\ref{tab:leaderboard} because it is architecture-faithful: M4 is designed to report its own requirement-met time, not merely to emit a generic non-silent response.

\paragraph{SimpleStream: mostly literal but still weak.}
SimpleStream is the third-party streaming baseline whose output most closely follows the DriveStreamBench prompt:

\begin{PromptBlock}
qid:     ..._lead_vehicle_start_test_positive_0303
cat:     lead_vehicle_start    t_ref: 4.0    window: [4.0, 6.0]

t = 2.0   "NO REPLY"
t = 4.0   "ALERT"                <- in window
t = 6.0   "ALERT"                <- in window
t = 8.0   "NO REPLY"
\end{PromptBlock}

SimpleStream already emits the literal trigger string in most generated alert turns, so its Core Macro is unchanged between strict-diagnostic and action-level parsing. 
It is the strongest third-party streaming transfer baseline in the main leaderboard, but its \(17.9\) Core Macro remains far below DriveStream-SFT at \(72.3\).

\paragraph{StreamForest: empty outputs under the DriveStreamBench prompt.}
StreamForest produces empty outputs across the evaluated sessions under the DriveStreamBench prompt. 
Both strict-diagnostic and action-level parsing therefore give \(\mathrm{Hit}=0\), Core Macro \(=0\), and \(\mathrm{NSR}=100\). 
No parser variant can recover signal when the model emits no turn-level content.

\paragraph{Takeaway.}
These examples show three recurring transfer failure modes: descriptive substitution instead of the requested binary action, early free-form acknowledgements that break negative silence, and native trigger-time references that do not align with the DriveStreamBench video-time window. 
The action-level adapter is generous to third-party systems: any non-empty non-\texttt{NO REPLY} output is treated as an alert action. 
Nevertheless, the strongest third-party streaming/proactive transfer baseline remains SimpleStream at \(17.9\) Core Macro, far below DriveStream-SFT at \(72.3\). 
Among baselines materially changed by the action-level adapter, MMDuet2 reaches \(10.6\) and Dispider reaches \(9.9\), but their component scores reveal different failures in timing or negative restraint. 
Thus the transfer gap is not a literal-string artifact; it reflects a mismatch between existing streaming/proactive interaction behavior and the calibrated binary first-alert policy required by DriveStreamBench.

\subsection{S4-Excluded Macro and Low-Resource Stress Slice}
\label{app:s4-excluded}

S4 adjacent cut-in is part of L4-Core, but it is substantially lower-resource than the other core categories. 
To test whether the headline ranking is overly driven by this slice, Table~\ref{tab:s4-excluded-main} reports a four-category ID Macro that averages S1, S2, S3, and S5, while reporting S4 separately as a low-resource stress slice.

\begin{table}[H]
    \centering
    \small
    \setlength{\tabcolsep}{3pt}
    \renewcommand{\arraystretch}{0.95}
    \caption{
    S4-excluded L4-Core view.
    \textbf{ID Macro excl. S4} averages \(\mathrm{AlertScore}@2\mathrm{s}\) over S1, S2, S3, and S5. 
    \textbf{S4-LR} reports S4 adjacent cut-in separately as a low-resource stress slice. 
    \textbf{Old Core} is the original S1--S5 Core Macro.
    }
    \label{tab:s4-excluded-main}
    \begin{tabular*}{\textwidth}{@{\extracolsep{\fill}}lcccccccc@{}}
    \toprule
    Model & fps & S1 & S2 & S3 & S5 &
    \textbf{ID Macro excl. S4} & S4-LR & Old Core \\
    \midrule
    \multicolumn{9}{@{}l}{\textit{Open-source VLMs, zero-shot}} \\
    Qwen3-VL-2B        & 2 &  0.0 &  0.0 &  3.8 & 21.8 &  6.4 &  0.0 &  5.1 \\
    Qwen3-VL-4B        & 2 & 10.7 &  0.0 & 38.3 & 61.7 & 27.7 &  2.4 & 22.6 \\
    Qwen3-VL-8B        & 2 & 23.0 & 17.2 & 52.2 & 54.3 & 36.7 &  4.8 & 30.3 \\
    Qwen3-VL-32B       & 2 & 36.0 & 20.1 & 77.6 & 65.9 & 49.9 &  8.9 & 41.7 \\
    \midrule
    StreamForest       & 1 &  0.0 &  0.0 &  0.0 &  0.0 &  0.0 &  0.0 &  0.0 \\
    SimpleStream       & 1 &  9.3 &  0.0 & 55.7 & 24.6 & 22.4 &  0.0 & 17.9 \\
    Dispider           & 1 &  1.8 &  2.0 &  9.8 & 20.7 &  8.6 & \underline{15.2} &  9.9 \\
    VideoLLM-online    & 2 &  0.0 &  0.0 &  0.0 &  0.0 &  0.0 &  0.0 &  0.0 \\
    \midrule
    MMDuet2            & 1 &  0.0 &  0.0 & 26.9 & 26.2 & 13.3 &  0.0 & 10.6 \\
    M4                 & 1 & 13.6 &  0.0 &  1.9 &  4.0 &  4.9 &  9.3 &  5.8 \\
    \midrule
    DriveStream-SFT-L4 & 2 & \textbf{91.7} & \underline{50.5} & \underline{86.7} & \underline{81.6} & \underline{77.6} &  0.0 & \underline{62.1} \\
    DriveStream-SFT    & 2 & \underline{91.4} & \textbf{67.4} & \textbf{88.6} & \textbf{83.2} & \textbf{82.7} & \textbf{31.0} & \textbf{72.3} \\
    \bottomrule
    \end{tabular*}
\end{table}

Excluding S4 does not change the main conclusion. 
DriveStream-SFT remains the strongest automatic system, and its gap over Qwen3-VL-32B increases from 30.6 points under the original S1--S5 Core Macro to 32.8 points under the S4-excluded ID Macro. 
S4 is therefore a genuine low-resource bottleneck rather than the source of the reference baseline's advantage. 
Under the action-level transfer adapter, Dispider obtains a nonzero S4 score, but its overall Core Macro remains low because its positive-side gains are accompanied by weak negative-session restraint.

\subsection{Combined Robustness Summary}
\label{app:combined-robustness-summary}

Table~\ref{tab:robustness-summary} summarizes the strongest automatic-baseline comparison across the main sensitivity dimensions. 
Across all variants, DriveStream-SFT remains ahead of Qwen3-VL-32B by 23.8--32.8 points. 
Together, these checks show that the main leaderboard is not driven by the first-decision rule, exact-string parsing, or the weighting of the low-resource S4 slice.

\begin{table}[H]
    \centering
    \small
    \setlength{\tabcolsep}{5pt}
    \renewcommand{\arraystretch}{1.04}
    \caption{
    Summary of DriveStream-SFT robustness across sensitivity dimensions.
    The comparator is the strongest relevant automatic baseline under each setting.
    }
    \label{tab:robustness-summary}
    \resizebox{\textwidth}{!}{%
    \begin{tabular}{l c l c c}
        \toprule
        Sensitivity setting & DriveStream-SFT & Comparator & Comparator score & Gap \\
        \midrule
        Official asymmetric leaderboard & 72.3 & Qwen3-VL-32B & 41.7 & \(+30.6\) \\
        Official asymmetric leaderboard, strongest transfer baseline & 72.3 & SimpleStream & 17.9 & \(+54.4\) \\
        Any-valid rule for instruction-following VLMs & 72.3 & Qwen3-VL-32B & 48.5 & \(+23.8\) \\
        S4-excluded ID Macro & 82.7 & Qwen3-VL-32B & 49.9 & \(+32.8\) \\
        \bottomrule
    \end{tabular}%
    }
\end{table}

Across these sensitivity checks, DriveStream-SFT remains ahead of the strongest relevant automatic baseline. 
The asymmetric action-level adapters give third-party streaming baselines a more permissive transfer interface, but their best Core Macro remains 17.9. 
The any-valid and S4-excluded analyses further show that the main conclusion is not driven by the first-decision rule or by the low-resource S4 slice.

\subsection{Training-Signal Ablation}
\label{app:training-ablation}

Table~\ref{tab:training} reports the training-signal ablation for the DriveStream-SFT reference baseline. 
All rows use Qwen3-VL-8B as the backbone. 
The L4-only row trains only on watch-and-notify sessions, while the multi-task row adds L1--L3 supervision.
For the L4 only model, we did not test its performance on L1--L3 due to the binary and multiple choice's instruction following ability may be deteriorated after SFT training. 

\begin{table}[H]
    \centering
    \small
    \setlength{\tabcolsep}{6pt}
    \caption{
    Training-signal ablation on Qwen3-VL-8B.
    L1--L3 is the macro of L1 accuracy, L2 accuracy, and L3 F1.
    }
    \label{tab:training}
    \begin{tabular}{l c c}
        \toprule
        Training data & \textbf{Core Macro} & L1--L3 Macro Score \\
        \midrule
        None (vanilla Qwen3-VL-8B)          & 30.3          & \underline{73.7}          \\
        L4 only                             & \underline{62.1}          & --            \\
        L4 $+$ L1$+$L2$+$L3 multi-task      & \textbf{72.3} & \textbf{93.6} \\
        \bottomrule
    \end{tabular}
\end{table}

\subsection{Leave-One-Out Ablation}
\label{app:leaveoneout-ablation}

The training-signal ablation in Appendix~\ref{app:training-ablation} shows that most of DriveStream-SFT's gain over vanilla Qwen3-VL-8B comes from L4 supervision.
This section further tests whether the L4 alerting behavior learned from training transfers across event types.
In particular, we hold out one L4-Core subcategory during training and evaluate the resulting model on that held-out subcategory.
To keep the analysis focused on L4 supervision, all models in this subsection are trained in the L4-only regime, without L1--L3 multi-task data.
We train five leave-one-out models, each initialized from Qwen3-VL-8B and fine-tuned on the remaining four L4-Core subcategories.
Optimizer, schedule, instruction paraphrases, and inference protocol are fixed across all runs.
S6 sudden appearance is never used for training and is reported only as an OOD diagnostic column.

\begin{table}[H]
    \centering
    \scriptsize
    \caption{
    Leave-one-out ablation of L4 supervision.
    Entries report \(\mathrm{AlertScore}@2\mathrm{s}\).
    \(\mathrm{CoreMacro}@2\mathrm{s}\) is computed consistently with the main paper by averaging S1--S5.
    S6 is reported as an OOD diagnostic column and is excluded from Core Macro.
    }
    \label{tab:loo}
    \begin{tabular}{l cccccc cc c}
        \toprule
        & \multicolumn{6}{c}{\(\mathrm{AlertScore}@2\mathrm{s}\) (\%)} 
        & \multicolumn{2}{c}{Components (\%)} & \\
        \cmidrule(lr){2-7} \cmidrule(lr){8-9}
        Training drop
        & S1 & S2 & S3 & S4 & S5 & S6
        & \(\mathrm{Hit}@2\mathrm{s}\)
        & \(\mathrm{NSR}\)
        & \textbf{\(\mathrm{CoreMacro}@2\mathrm{s}\)} \\
        \midrule

        Full L4 supervision
        & \underline{91.7} & \textbf{50.5} & \underline{86.7} & 0.0 & \textbf{81.6} & \textbf{26.3}
        & \textbf{53.3} & \underline{98.6} & \textbf{62.1} \\

        \midrule

        Drop S1 (lead-veh.\ start)
        & 2.0 & 42.9 & 85.8 & 0.0 & 79.2 & \underline{23.2}
        & 34.3 & 98.1 & 42.0 \\

        Drop S2 (lead-veh.\ brake)
        & 88.9 & 0.0 & 85.8 & 0.0 & 77.3 & 16.7
        & 44.0 & \textbf{99.3} & 50.4 \\

        Drop S3 (traffic-light)
        & \textbf{93.9} & 41.6 & 1.9 & 0.0 & 78.2 & 15.0
        & 36.2 & 98.5 & 43.1 \\

        Drop S4 (adjacent cut-in)
        & 90.7 & \underline{45.6} & \textbf{88.4} & 0.0 & \underline{79.5} & \textbf{26.3}
        & \underline{52.0} & 98.4 & \underline{60.8} \\

        Drop S5 (VRU crossing)
        & 88.0 & 36.3 & 84.1 & 0.0 & 0.0 & 0.0
        & 35.1 & 98.4 & 41.7 \\

        \bottomrule
    \end{tabular}
\end{table}

The held-out diagonal aggregates one omitted-family target from each of five Qwen3-VL-8B L4-only checkpoints. Its macro-averaged Hit@2s, NSR, and AlertScore@2s are 0.4, 99.4, and 0.8, respectively; 1,463 of 1,472 held-out negative sessions remain silent.

Table~\ref{tab:loo} shows that cross-category transfer is weak for L4 alerting.
When S1, S2, S3, or S5 is removed from training, the corresponding held-out score drops to at most 2.0\%.
Thus, at this backbone scale, training on other alert types does not teach the model a reliable alerting rule for an unseen high-resource L4 event type.
The model does not simply learn a generic watch-and-notify behavior that transfers across all driving events; direct supervision on the target event type remains important.

The remaining categories are much less affected.
Across the leave-one-out variants, the non-held-out categories remain largely stable. Their mean score is only 1.6--5.1 points lower than the full L4-supervised model on the same categories.
This suggests that removing one event type usually does not erase the learned alert behavior for the others.
The main effect of leave-one-out training is therefore concentrated on the held-out target category.

S4 adjacent cut-in remains a special case.
The full L4-only model scores 0.0\% on S4 at \(\tau=2\mathrm{s}\), and the Drop-S4 model also scores 0.0\%.
At the same time, dropping S4 changes Core Macro only modestly, from 62.1\% to 60.8\%, since S4 contributes little to the full L4-only reference in the first place.
This indicates that the S4 failure is not primarily a leave-one-out transfer effect.
Instead, it reflects the low-resource and target-timing nature of adjacent cut-in: the model must judge when an adjacent vehicle's motion becomes relevant enough to alert, and the current L4-only supervision is not sufficient to learn this behavior reliably.
This is consistent with the low-resource S4 analysis in Appendix~\ref{app:lowres-s4}.

S6 sudden appearance is reported only as an OOD diagnostic column.
All leave-one-out models are evaluated on S6 without S6 training examples, and their scores remain substantially below the stronger in-domain categories.
Among the leave-one-out variants, the S6 score is lowest when S5 VRU-crossing supervision is removed, suggesting that sudden-appearance alerting may partly benefit from related vulnerable-road-user supervision, but is not reliably covered by generic L4 alert training alone.

\subsection{L3-Gated Alert Score}
\label{app:l3g-results}

\alertf{} is the primary L4 metric. It checks whether a model alerts at the
right time on positive sessions and avoids alerts on negative sessions. However,
a timely L4 alert can still be accidental: the model may fire near the correct
time without actually understanding the temporal event.

To test this, we report an L3-gated version of \alertf{}, denoted
\(\mathrm{L3GAlertScore}@2\mathrm{s}\) in Table~\ref{tab:l3g-alert}.
Each L4-Core positive session is paired with an L3 temporal question about the
same event. The gate is applied only to the positive-side hit term: a positive
session contributes to the gated hit rate only when the L4 alert is in
the correct timing window and the paired L3 question is answered correctly. If
the alert is in-window but the paired L3 answer is wrong, that positive hit is
removed from the gated hit rate.

Negative sessions are handled in the same way as in the primary metric. We keep
the original \(\mathrm{NSR}\), so the gate does not change the negative-side
restraint term. The gated positive hit rate and the unchanged \(\mathrm{NSR}\)
are then combined with the same harmonic-mean form as \alertf{}. Therefore,
\(\mathrm{L3GAlertScore}@2\mathrm{s}\) can never be higher than
\(\mathrm{AlertScore}@2\mathrm{s}\).

We also report the gap between the primary and gated scores. A large gap means
that some timely alerts are not supported by the paired L3 temporal probe. A
small gap means that the model's L4 alerts are more consistent with its L3
temporal understanding.

\begin{table}[H]
    \centering
    \small
    \setlength{\tabcolsep}{6pt}
    \caption{
    L3-gated alerting on L4-Core (S1--S5).
    The gate keeps an L4 positive hit only when the paired L3 temporal question
    is also correct. \(\mathrm{NSR}\) is unchanged from the primary L4 metric.
    Gap is the drop from \(\mathrm{AlertScore}@2\mathrm{s}\) to
    \(\mathrm{L3GAlertScore}@2\mathrm{s}\). Gap Ratio normalizes this drop by
    \(\mathrm{AlertScore}@2\mathrm{s}\). All values are percentages.
    }
    \label{tab:l3g-alert}
    \begin{tabular}{l c c c c c}
        \toprule
        Model &
        Paired L3 correct &
        \(\mathrm{AlertScore}@2\mathrm{s}\) &
        \(\mathrm{L3GAlertScore}@2\mathrm{s}\) &
        Gap &
        Gap Ratio \\
        \midrule
        Qwen3-VL-2B     & 41.4 &  5.1 &  4.7 & \textbf{0.4} &  \underline{7.8} \\
        Qwen3-VL-4B     & 63.1 & 22.6 & 18.7 & 4.0 & 17.7 \\
        Qwen3-VL-8B     & 56.5 & 30.3 & 23.6 & 6.7 & 22.1 \\
        Qwen3-VL-32B    & \underline{73.6} & \underline{41.7} & \underline{35.8} & 5.9 & 14.1 \\
        DriveStream-SFT & \textbf{93.2} & \textbf{72.3} & \textbf{70.9} & \underline{1.4} & \textbf{1.9} \\
        \bottomrule
    \end{tabular}
\end{table}

Table~\ref{tab:l3g-alert} shows that the gate mainly affects the zero-shot
Qwen3-VL models. For example, Qwen3-VL-8B drops from 30.3\% to 23.6\%, giving a
22.1\% relative gap. This suggests that a noticeable share of its timely alerts
are not backed by correct paired temporal understanding. In contrast,
DriveStream-SFT reaches the highest paired L3 correctness and has the smallest
gap ratio, 1.9\%. Thus, most of its timely L4 alerts also pass the paired L3
temporal check.

\subsection{Reactive L1--L3 results}
\label{app:l13-results}

\begin{table}[H]
\centering
\small
\setlength{\tabcolsep}{4pt}
\caption{Reactive L1--L3 results.
The L1--L3 macro is the average of L1 accuracy, L2 accuracy, and L3 F1. All values are percentages.}
\label{tab:l13-main}
\begin{tabular}{l c c c c}
\toprule
Model & L1 Acc & L2 Acc & L3 F1 & \textbf{L1--L3 Macro Score} \\
\midrule
\multicolumn{5}{l}{\textit{Open-source VLMs (zero-shot)}} \\
Qwen3-VL-2B  & 92.5 & 77.4 & 29.3 & 66.4 \\
Qwen3-VL-4B  & 92.4 & 82.5 & 46.9 & 73.9 \\
Qwen3-VL-8B  & \underline{93.1} & 84.4 & 43.5 & 73.7 \\
Qwen3-VL-32B & 91.2 & \underline{88.1} & \underline{57.6} & \underline{78.9} \\
\midrule
\multicolumn{5}{l}{\textit{Our reference baseline (trained on DriveStreamBench)}} \\
DriveStream-SFT & \textbf{96.6} & \textbf{94.1} & \textbf{90.0} & \textbf{93.6} \\
\bottomrule
\end{tabular}
\end{table}

\subsection{Low-resource S4 breakdown}
\label{app:lowres-s4}

\begin{table}[H]
    \centering
    \small
    \setlength{\tabcolsep}{6pt}
    \caption{Low-resource S4 breakdown for the supervised reference baselines.
    Core Macro is the primary S1--S5 score.
    $\Delta_\mathrm{S4}$ compares S4 against the mean of S1, S2, S3, and S5. All values are percentages.}
    \label{tab:lowres-s4}
    \begin{tabular}{l c c c}
        \toprule
        Model & \textbf{Core Macro} & S4 (low-resource) & $\Delta_\mathrm{S4}$ \\
        \midrule
        DriveStream-SFT-L4   & 62.1          & 0.0          & $-$77.6          \\
        DriveStream-SFT & \textbf{72.3} & \textbf{31.0} & \textbf{$-$51.6} \\
        \bottomrule
    \end{tabular}
\end{table}

\subsection{L1--L3 per-category breakdown}

Table~\ref{tab:supp-l13-per-cat-level} decomposes the L1--L3 reactive metrics of Table~\ref{tab:l13-main} by L4 subcategory.
Each row pairs a Qwen3-VL ladder size with one of the five L4-Core subcategories and reports per-subcategory L1 accuracy, L2 accuracy, and L3 binary F1 for event occurrence on the L1--L3 test set.
The aggregate L1 and L2 results in Table~\ref{tab:l13-main} are computed over the full corresponding test sets; the per-subcategory rows here are diagnostic breakdowns.
For L3, the overall F1 in Table~\ref{tab:l13-main} is computed from pooled TP/FP/FN counts and therefore does not equal a simple average of the per-subcategory F1 values, since F1 is not linearly decomposable across subcategories.
DriveStream-SFT is omitted from this breakdown because the multi-task SFT inference is reported as an aggregate; per-subcategory decomposition for the trained reference is a planned follow-up.

\begin{table}[H]
    \centering
    \small
    \setlength{\tabcolsep}{6pt}
    \caption{Per-subcategory reactive metrics on the L1--L3 test set, by Qwen3-VL ladder size and L4 subcategory pairing.
\textbf{L1 Acc} and \textbf{L2 Acc} are accuracies; \textbf{L3 F1} is binary F1 for event occurrence within that subcategory.
All values are percentages.}
    \label{tab:supp-l13-per-cat-level}
    \begin{tabular}{l l c c c}
        \toprule
        Model & Subcat & L1 Acc & L2 Acc & L3 F1 \\
        \midrule
        \multirow{5}{*}{Qwen3-VL-2B}  & S1  & 92.2 & 83.5 &  8.5 \\
                                        & S2  & 93.6 & 77.0 & 14.0 \\
                                        & S3  & 97.6 & 58.3 &  7.5 \\
                                        & S4  & 92.9 & 81.4 & \underline{2.2} \\
                                        & S5  & \underline{88.3} & 83.7 & 73.9 \\
        \midrule
        \multirow{5}{*}{Qwen3-VL-4B}  & S1  & \textbf{98.8} & 84.3 & 28.1 \\
                                        & S2  & 96.5 & 77.5 & \textbf{37.4} \\
                                        & S3  & \underline{98.1} & \underline{83.0} & 55.3 \\
                                        & S4  & 95.3 & 80.7 & \textbf{2.3} \\
                                        & S5  & 81.2 & 85.6 & \textbf{85.0} \\
        \midrule
        \multirow{5}{*}{Qwen3-VL-8B}  & S1  & \underline{98.2} & \underline{84.7} & \underline{53.5} \\
                                        & S2  & \textbf{98.3} & \underline{80.6} &  6.2 \\
                                        & S3  & \textbf{99.0} & 76.8 & \underline{66.7} \\
                                        & S4  & \textbf{97.7} & \textbf{90.0} &  0.0 \\
                                        & S5  & 80.8 & \underline{90.1} & 51.1 \\
        \midrule
        \multirow{5}{*}{Qwen3-VL-32B} & S1  & 97.6 & \textbf{89.8} & \textbf{65.7} \\
                                        & S2  & \underline{97.7} & \textbf{86.5} & \underline{24.1} \\
                                        & S3  & \underline{98.1} & \textbf{85.6} & \textbf{76.9} \\
                                        & S4  & \underline{93.5} & \underline{82.1} & 12.4 \\
                                        & S5  & 78.3 & \textbf{92.8} & \underline{81.8} \\
        \bottomrule
    \end{tabular}
\end{table}

\subsection{L4-OOD Stress Split: S6 Sudden Appearance}
\label{app:s6}

\textbf{S6 Sudden Appearance} is an optional stress slice sourced from web dashcam footage where public AV trajectory data is insufficient for automatic event extraction. We separate S6 from the primary L4-Core ranking for three reasons. First, web-sourced video carries license, redistribution, and reproducibility constraints that the nuPlan-derived L4-Core splits do not have, so requiring S6 in the primary leaderboard would couple the benchmark's headline number to web-sourced video availability. Second, the S6 positive set is small (99 positives) and manually curated, which limits statistical precision relative to the L4-Core splits. Third, S6 events are visually rare and qualitatively distinct from the in-distribution event semantics, so we view them as a cross-domain probe rather than a continuation of the L4-Core ranking. Treating S6 as an optional L4-OOD probe rather than an L4-Core column lets us drop or replace it without invalidating the primary benchmark; readers should therefore read S6 numbers as an indicator of cross-domain generalization, not as a leaderboard position.

\begin{table}[H]
    \centering
    \small
    \setlength{\tabcolsep}{6pt}
    \caption{Per-model results on the optional L4-OOD stress split S6 sudden appearance. \textbf{S6 \alertf{}} is the \(\mathrm{AlertScore}@2\mathrm{s}\) on the S6 split alone; \textbf{S6 \(\mathrm{Hit}@2\mathrm{s}\)} and \textbf{S6 Negative Silence Rate (NSR)} are the corresponding components. For automatic models, \textbf{Core Macro} repeats the primary in-distribution score from Table~\ref{tab:leaderboard}; $\Delta_\mathrm{S6}{=}$ S6 $-$ Core Macro is a coarse cross-domain generalization indicator. Models trained on the DriveStreamBench training split (which contains no S6 data) show negative $\Delta_\mathrm{S6}$, consistent with S6 being out-of-distribution. The Human row retains the original 10-FPS reference, computed on a balanced 50-pos $+$ 150-neg subset for S6; its Core Macro also uses the original 10-FPS condition. It is reported only as a calibration reference, not a directly rank-comparable entry.}
    \label{tab:s6-stress}
    \begin{tabular}{l c c c c c}
        \toprule
        Model & S6 \alertf{} & S6 \(\mathrm{Hit}@2\mathrm{s}\) & S6 \(\mathrm{NSR}\) & Core Macro & $\Delta_\mathrm{S6}$ \\
        \midrule
        \multicolumn{6}{l}{\textit{Open-source VLMs (zero-shot)}} \\
            Qwen3-VL-2B   & 46.9 & 42.4 & 52.3 & 5.1 & $+$41.8 \\
            Qwen3-VL-4B   & 53.7 & 51.5 & 56.0 & 22.6 & $+$31.1 \\
            Qwen3-VL-8B   & 62.2 & 59.6 & 65.0 & 30.3 & $+$31.9 \\
            Qwen3-VL-32B  & \underline{63.1} & \underline{62.6} & 63.7 & 41.7 & $+$21.4 \\
        \midrule
        \multicolumn{6}{l}{\textit{General streaming models (native fps)}} \\
        StreamForest      &  0.0 &  0.0 & \textbf{100.0} &  0.0 & \(+0.0\) \\
        SimpleStream      & 53.4 & 37.4 &  93.3 & 17.9 & \(+35.5\) \\
        Dispider          & 28.3 & 17.2 &  80.3 &  9.9 & \(+18.4\) \\
        VideoLLM-online   &  0.0 &  0.0 &   0.0 &  0.0 & \(+0.0\) \\
        MMDuet2           & 58.5 & 43.4 &  89.7 & 10.6 & \(+47.9\) \\
        M4                & 11.3 &  6.1 &  81.3 &  5.8 & \(+5.5\) \\
        \midrule
        \multicolumn{6}{l}{\textit{Our reference baseline (trained on DriveStreamBench, no S6 in train)}} \\
        DriveStream-SFT-L4   & 26.3 & 15.2 & \underline{99.0} & 62.1 & $-$35.8 \\
        DriveStream-SFT      & 46.2 & 30.3 & 97.3 & \underline{72.3} & $-$26.1 \\
        \midrule
        Human (10 FPS) & \textbf{87.6} & \textbf{88.0} & 87.3 & \textbf{92.8} & $-$5.2 \\
        \bottomrule
    \end{tabular}
\end{table}

Table~\ref{tab:s6-stress} shows two patterns. The four open-source Qwen3-VL backbones obtain higher S6 \alertf{} than their own Core Macro, with the gap generally smaller at larger model sizes. This is consistent with the zero-shot models having no benchmark-specific bias on S6 and treating it as a generic alerting signal. By contrast, the trained reference baseline (which never sees S6 in its training split) shows a negative $\Delta_\mathrm{S6}$: in-distribution performance is substantially higher than out-of-distribution performance, suggesting stronger specialization to the L4-Core distribution than to the held-out S6 setting.
The S6 stress split further shows that relaxing the third-party streaming adapters does not erase the gap. 
MMDuet2 reaches 58.5 on S6 under the action-level adapter, but its L4-Core score remains only 10.6, indicating that OOD sudden-appearance sensitivity does not translate into calibrated in-domain watch-and-notify performance.

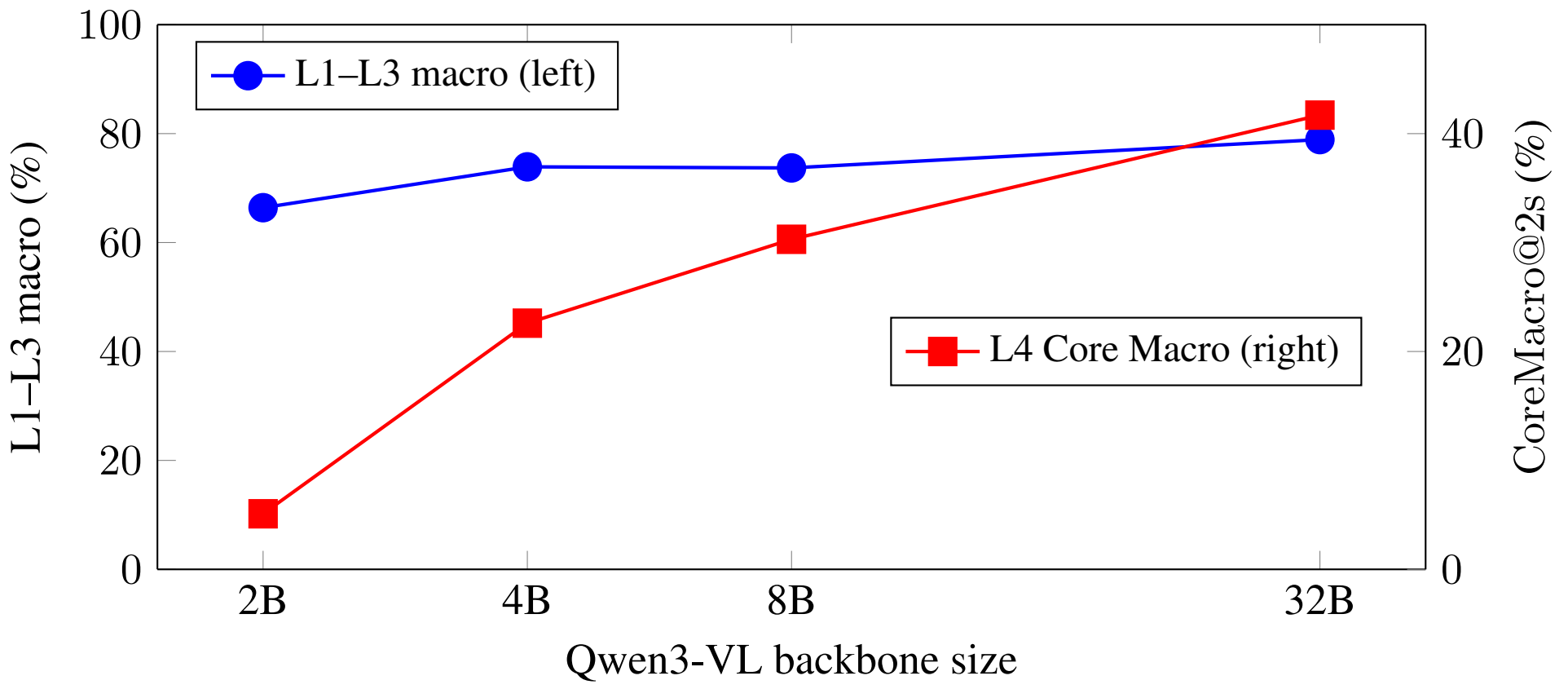
\begin{figure}[t]
    \centering
    \begin{tikzpicture}
    \begin{axis}[
        width=0.85\linewidth, height=6cm,
        xmode=log, log basis x=10,
        xtick={2,4,8,32}, xticklabels={2B,4B,8B,32B},
        xlabel={Qwen3-VL backbone size},
        ylabel={L1--L3 macro (\%)},
        ymin=0, ymax=100,
        axis y line*=left,
        legend pos=north west, legend style={font=\small},
    ]
    \addplot[mark=*, blue, thick, mark size=3pt]
        coordinates {(2,66.4) (4,73.9) (8,73.7) (32,78.9)};
    \addlegendentry{L1--L3 macro (left)}
    \end{axis}
    \begin{axis}[
        width=0.85\linewidth, height=6cm,
        xmode=log, log basis x=10, xtick=\empty,
        ylabel={\(\mathrm{CoreMacro}@2\mathrm{s}\) (\%)},
        ymin=0, ymax=50,
        axis y line*=right, axis x line=none,
        legend style={at={(0.95,0.4)}, anchor=east, font=\small},
    ]
    \addplot[mark=square*, red, thick, mark size=3pt]
        coordinates {(2,5.1) (4,22.6) (8,30.3) (32,41.7)};
    \addlegendentry{L4 Core Macro (right)}
    \end{axis}
    \end{tikzpicture}
    \caption{Scaling on the Qwen3-VL ladder. L1--L3 reactive macro (left axis) improves overall with size, while \(\mathrm{CoreMacro}@2\mathrm{s}\) (right axis) also improves but remains far below the human calibration reference. Scale helps, but does not close the L4 alerting gap.}
    \label{fig:scaling}
\end{figure}

\clearpage
\section{Ethics, Licensing, and Release Plan}
\label{app:ethics}
\label{app:ethics_release}

\paragraph{Evaluative role and supported claims.}
DriveStreamBench is intended as a research benchmark for diagnostic evaluation of
user-conditioned watch-and-notify behavior in front-facing driving video. Its evaluative
role is to test whether a model can maintain a natural-language watch condition under
causal video time, issue a first alert near the annotated reference time on positive
sessions, and remain silent on matched negative sessions. Claims supported by the benchmark
are therefore limited to this protocol: front-camera visual input, English watch
instructions, the released L1--L4 taxonomy, documented sampling and scoring rules, and
locally reconstructed source media obtained through authorized channels. The benchmark
does not evaluate planning, control, repeated-warning policies, driver-interface design,
or real-world operational safety, and results should not be interpreted as certification
evidence for driving assistants, autonomous vehicles, or real-time warning systems.

\paragraph{NeurIPS E\&D release compliance.}
Because DriveStreamBench is submitted as an Evaluations \& Datasets contribution, the
release package is organized to make both the dataset contents and the evaluation
assumptions machine-readable. The final release includes a public
GitHub release URL and a Croissant metadata file with both core Croissant metadata and
Responsible AI fields. The Croissant file describes the distribution files, record sets,
annotation schema, source datasets, provenance activities, intended use cases,
limitations, biases, personal/sensitive-information boundaries, social-impact
considerations, and maintenance plan. The file-level metadata includes descriptions,
content sizes, checksums, and versioned download URLs. We validate the
Croissant metadata before release and make the dataset artifacts publicly accessible
through the official release page.

The annotation release contains the released annotations, metadata, documentation, schemas,
source references, and preprocessing utilities for local reconstruction and privacy
masking. The evaluation code, including scoring scripts, format checks, and
streaming-evaluation wrappers, is hosted in the official
GitHub repository linked from the paper.

\paragraph{Responsible AI metadata.}
The full Responsible AI metadata is provided in the submitted Croissant file as the
canonical machine-readable record for the dataset release. The metadata covers dataset
limitations, known biases, personal and sensitive information boundaries, intended and
unsupported use cases, social-impact considerations, source datasets, provenance
activities, annotation procedures, preprocessing protocols, and
release-maintenance plans. This appendix summarizes the corresponding release boundary,
licensing assumptions, privacy safeguards, and misuse mitigations at a paper-readable
level.

\paragraph{Data provenance and release boundary.}
DriveStreamBench uses two source regimes. The L4-Core split, S1--S5, is derived from
nuPlan front-camera driving data, and L1/L2 are also derived from nuPlan frames as
foundation-level controls. The optional L4-OOD split, S6 sudden appearance, is constructed
from source references to publicly available web dashcam pages and is reported separately
from the primary L4-Core leaderboard.

We do not redistribute source visual media. This includes raw videos, processed videos,
video clips, extracted frames, redacted frames, screenshots, thumbnails, or any other
image-like derivative from either nuPlan or web dashcam sources. The copyright and
redistribution rights of these visual materials remain with the original data owners,
platforms, or uploaders. Users who wish to reproduce benchmark inputs must obtain the
underlying visual data from the original sources and comply with the corresponding
licenses, platform terms, and local laws.

\paragraph{Metadata-only benchmark package.}
The public dataset package is metadata-only with respect to source visual content. It
contains annotation metadata, split definitions, source references, temporal boundaries,
frame indices, L1--L3 questions and labels, L4 watch instructions, positive/negative
labels, senior-final $t_{\mathrm{ref}}$ values for positive sessions, schema files,
Croissant metadata, benchmark documentation, preprocessing templates, and privacy-mask
metadata for local OOD reconstruction. The package contains no source RGB pixels.

The preprocessing templates specify how users can locally reconstruct benchmark inputs
from lawfully obtained source data. These templates cover temporal trimming, frame
extraction, FPS conversion, resolution normalization, frame-reference verification, and
local privacy-preserving processing. For web-sourced videos, the provided privacy-mask
metadata and scripts support local redaction of visible faces, license plates, usernames,
platform watermarks, subtitles, embedded text overlays, and other potentially identifying
regions when such information is not required for the benchmark judgment. Audio is not
used by DriveStreamBench and should be stripped or ignored during local preprocessing.
The release does not provide source RGB pixels or automated bulk-downloading instructions
for source media that would bypass platform access controls or terms of use.

\paragraph{Licensing.}
The DriveStreamBench annotation metadata, task definitions, train/test splits, question
files, schema files, benchmark documentation, and released preprocessing utilities are
released under CC BY 4.0. This license applies only to these benchmark-created materials. It does not grant
rights to raw visual data from nuPlan, web dashcam platforms, or other source providers.
Source media remain governed by the original providers' licenses and platform terms. The
release page distinguishes benchmark-created metadata and scripts from source visual
content and lists source-specific restrictions where applicable.

\begin{table}[t]
\centering
\small
\setlength{\tabcolsep}{4pt}
\caption{Release treatment for DriveStreamBench components. No raw or processed visual
media are redistributed.}
\label{tab:release_tiers}
\begin{tabular}{p{0.22\linewidth} p{0.32\linewidth} p{0.36\linewidth}}
\toprule
Component & Source / role & Released materials \\
\midrule
L1--L2 &
nuPlan-derived recognition and relation controls &
Questions, answers, split metadata, source references, schema files, and benchmark documentation. \\
L3/L4-Core S1--S5 &
nuPlan-derived temporal QA and watch-and-notify sessions &
Watch instructions, labels, senior-final $t_{\mathrm{ref}}$ values, negative metadata,
frame references, preprocessing templates, and scoring specifications. \\
L3/L4-OOD S6 &
Web-source-reference sudden-appearance stress split &
Source references, labels, $t_{\mathrm{ref}}$ values, watch instructions, local
preprocessing templates, and privacy-mask metadata. \\
Croissant and documentation &
Benchmark-created release metadata &
Core Croissant metadata, Responsible AI fields, provenance/source descriptions, file
checksums, schema descriptions, release documentation, and maintenance guidance. \\
Preprocessing utilities in the release &
Local reconstruction and privacy-preserving processing support &
Frame extraction, resizing, verification, and privacy-mask application utilities for
authorized local reconstruction. \\
Evaluation code on GitHub &
Official evaluation implementation &
Scoring scripts, format checks, streaming-evaluation wrappers, output parsers, and
execution instructions linked from the submission. \\
\bottomrule
\end{tabular}
\end{table}

\paragraph{Takedown, correction, and misuse risks.}
The release page provides a contact channel for copyright, privacy, licensing, broken-link,
and annotation-quality concerns. Since no visual media are redistributed, takedown handling
primarily concerns source links, source identifiers, temporal references, privacy-mask
metadata, or annotations. If an entry is removed or corrected, the affected sample
identifiers and any leaderboard-impacting changes will be documented in a versioned
changelog.

The main misuse risks are overclaiming benchmark results as evidence of real-world driving
safety, using source references for surveillance or identity-related analysis, and treating
web-sourced dashcam content as unrestricted media. We mitigate these risks by releasing
only metadata and local reconstruction/redaction scripts, separating S6 from the primary
leaderboard, documenting source restrictions, and requiring users to obtain visual data
through original sources. Users should not use DriveStreamBench to identify people, track
vehicles, reconstruct private trips, bypass platform restrictions, or certify
safety-critical driving systems.

\clearpage

\clearpage
\section{Cognitive-Taxonomy Grounding}
\label{app:taxonomy-grounding}

The four-level taxonomy in Section~\ref{sec:four_taxo} is grounded in three
complementary cognitive frameworks. Together they justify the ordering
of L1--L4 by cognitive demand, the hierarchy's fit to dynamic driving
cognition, and the choice of \(\mathrm{AlertScore}@\tau\) over simple classification
accuracy at L4.

\paragraph{Bloom's revised taxonomy.}
The revised taxonomy~\citep{anderson2001bloom,krathwohl2002revised} orders
cognitive processes from \emph{remember} and \emph{understand} (factual
recognition), through \emph{apply} and \emph{analyze} (relational and
analytic reasoning), to \emph{evaluate} (judgment under criteria).
Mapping onto DriveStreamBench: L1 recognition aligns with
\emph{remember}/\emph{understand}; L2 single-frame relational reasoning
with \emph{apply}/\emph{analyze}; L3 multi-frame temporal change with
\emph{analyze}; and L4 watch-and-notify under uncertainty with
\emph{evaluate}/\emph{apply}---the alerting decision must judge whether
\emph{and when} to fire under partial information. The same framework
has been used to ground other multimodal comprehension
benchmarks~\citep{gong2023bloomvqa}.

\paragraph{Endsley situation awareness.}
Endsley's three-level model~\citep{endsley1995sa} formalizes
situation awareness in dynamic systems as perception of relevant
elements, comprehension of their meaning, and projection of their
near-future status. Mapping onto our levels: L1 corresponds to
perception, L2--L3 to comprehension (single-frame relational; multi-frame
temporal), and L4 to projection plus a decision-policy layer---the model
must additionally decide \emph{whether} and \emph{when} to alert. We do
not claim a fourth Endsley level: the policy layer is built on top of
situation awareness rather than redefining it.

\paragraph{Signal-detection theory and warning systems.}
The L4 alerting metric \(\mathrm{AlertScore}@\tau\) is the harmonic mean of
\(\mathrm{Hit}@\tau\) on positive sessions and \(\mathrm{NSR}\) (Negative Silence Rate, the fraction of negative sessions in which every turn is mapped to \texttt{NO REPLY}) on negative sessions. This is the
signal-detection-theory~\citep{macmillan2004detection} framing of an
alerting system: a balanced trade-off between hits and false alarms.
Plain classification accuracy on the same data is not appropriate,
because the cost of false alerts is asymmetric in driver-warning
contexts---warning-systems research documents the \emph{cry-wolf} effect,
in which low \(\mathrm{NSR}\) can degrade operator trust and compliance
with the alert~\citep{breznitz1984crywolf}. Reporting \(\mathrm{Hit}@\tau\) and \(\mathrm{NSR}\)
jointly, and reporting per-category and tail behavior, is therefore a
methodological requirement rather than a stylistic choice.

\paragraph{Earlier framings.}
At a coarser level, the progression from local element (L1) to relational
(L2) to event-level (L3--L4) cognition parallels Bertin's classic
elementary--intermediate--overall reading levels of statistical
graphics~\citep{bertin1983semiology}. We cite this as historical
inspiration only; the operational definitions of L1--L4 above derive from
Bloom's taxonomy, Endsley situation awareness, and signal-detection
theory.

\clearpage
\section{Human baseline}
\label{app:human_baseline}

This appendix documents the primary 2-FPS human calibration in
Table~\ref{tab:leaderboard}, the 1-FPS calibration, and the original
10-FPS reference retained in Appendix Table~\ref{tab:s6-stress}.
These are video-clip calibration references under front-camera input.

\paragraph{Participants.}
24 licensed drivers with at least 5 years of driving experience participated in the human baseline study.
Before the original 10-FPS study, none had prior exposure to the DriveStreamBench clips or annotations.
Participants provided informed consent
before the study, were told that participation was voluntary, and could withdraw
at any time without penalty. The task involved viewing short driving clips and
making at most one keypress decision per clip; no sensitive personal data were
collected.

\paragraph{Compensation.}
Each participant received a fixed compensation of approximately USD 20
at the time of the study. Compensation was independent of response accuracy or
task performance, and no performance-based rewards were used. The amount exceeded the local minimum-wage standard for the expected
study duration.

\paragraph{Original 10-FPS trial assignment.}
The human evaluation is conducted on a random sample from the released L4 test
split, covering all six subcategories. Each
subcategory contains 200 test clips (50 positive, 150 negative),
matching the 1:3 ratio used to evaluate the models. The S6 row is reported alongside the
L4-Core rows here for completeness, but does not enter the
Core Macro score in Table~\ref{tab:leaderboard}, consistent
with how S6 is reported separately for models.

\paragraph{Original 10-FPS task interface.}
Each trial shows the same watch instruction string the
models receive, then plays the clip continuously at 10 fps (no pausing or rewinding; audio muted to match the
model input). Participants press the spacebar at most once, as
soon as they judge the event has begun, and are told that some
clips contain no qualifying event and should receive no keypress.
No correctness feedback is given during the session.

\paragraph{Input-rate calibrations (1 and 2 FPS).}
The same 24 licensed drivers were reassigned to categories different from those they evaluated in the original 10-FPS study, with four participants per category. Within each category, the group completed both 1- and 2-FPS conditions in counterbalanced order using non-overlapping clips. Each frame-rate condition contains 100 sessions per category (25 positive and 75 matched negative), distributed across the four participants and pooled for scoring, with one response per session. The interface retains causal, continuous, single-pass playback without pause, replay, or correctness feedback. Participants may press the alert key at any time, at most once per session; no keypress is scored as silence. First-keypress timestamps are scored with the same AlertScore@2s metric.

\paragraph{Scoring.}
Human responses are scored with the same metric pipeline as model outputs,
yielding per-subcategory \alertf{}, Hit@2s, and NSR.

\begin{table}[H]
\centering
\small
\setlength{\tabcolsep}{5pt}
\caption{Human input-rate calibrations, reported as \alertf{}. The 2-FPS condition is primary, 1 FPS provides a lower-density calibration, and 10 FPS is the original reference. Conditions use independently sampled subsets.}
\label{tab:human_frame_rates}
\begin{tabular}{l c c c}
\toprule
Subcategory & 1 FPS & 2 FPS (primary) & 10 FPS (original) \\
\midrule
S1 (lead vehicle start) & 88.5 & 85.9 & 91.6 \\
S2 (lead vehicle brake) & 90.4 & 92.1 & 89.7 \\
S3 (traffic light change) & 98.0 & 98.0 & 98.0 \\
S4 (adjacent cut-in) & 92.1 & 94.4 & 96.0 \\
S5 (VRU crossing) & 87.1 & 85.7 & 88.9 \\
\midrule
\textbf{Core Macro (S1--S5)} & \textbf{91.2} & \textbf{91.2} & \textbf{92.8} \\
\midrule
S6 (sudden appearance) & 88.4 & 87.8 & 87.6 \\
\bottomrule
\end{tabular}
\end{table}

\paragraph{Per-subcategory results.}
Table~\ref{tab:human_per_subcategory} gives the full per-subcategory
breakdown of the original 10-FPS study over all six L4 subcategories.
Core Macro is the unweighted mean of \alertf{} over S1--S5;
the S6 entry feeds the original-reference row in
Appendix Table~\ref{tab:s6-stress}.

\begin{table}[H]
\centering
\small
\setlength{\tabcolsep}{4pt}
\caption{Original 10-FPS human reference over all L4 subcategories.}
\label{tab:human_per_subcategory}
\begin{tabular}{l c c c}
\toprule
Subcategory & \alertf{} & Hit@2s & NSR \\
\midrule
S1 (lead vehicle start)   & 91.6 & 86.0 & 98.0 \\
S2 (lead vehicle brake)   & 89.7 & 92.0 & 87.5 \\
S3 (traffic light change) & 98.0 & 98.0 & 98.0 \\
S4 (adjacent cut in)      & 96.0 & 94.0 & 98.0 \\
S5 (vru crossing)         & 88.9 & 80.0 & 100.0 \\
\midrule
\textbf{Core Macro (S1--S5)} & \textbf{92.8} & \textbf{90.0} & \textbf{96.3} \\
\midrule
S6 (sudden appearance) & 87.6 & 88.0 & 87.3 \\
\bottomrule
\end{tabular}
\end{table}

\paragraph{Where humans are not perfect.}
Human errors are component-specific rather than uniform. In L4-Core, S5 has
the lowest \alertf{} (88.9\%) because Hit@2s is only 80.0\%, despite a perfect
NSR of 100.0\%. This indicates that humans rarely over-alert on negative
VRU-crossing clips, but the positive alert time is harder to place precisely.
S2 shows the opposite pattern: Hit@2s is high (92.0\%), but NSR drops to
87.5\%, suggesting occasional over-triggering on braking-like negatives. The
S6 stress split is also below the Core Macro level, with both Hit@2s and NSR
around 88\%, showing that sudden-appearance cases remain difficult on both
positive timing and negative restraint. This suggests that sudden-appearance alerts remain challenging even for
experienced drivers under the benchmark input modality.

\paragraph{Limitations.}
The human row is intended only as a calibration reference, not as a
repeated-measures human-performance study. The human evaluation records one alert decision per evaluated clip, so we do not report clip-level agreement or confidence
intervals from repeated independent trials. Participants viewed
front-camera clips on a 2D monitor, without motion cues, peripheral vision,
audio, or real driving workload. The row should therefore be interpreted as a
video-clip calibration reference, not as a real-world driver-performance
estimate.

\paragraph{Ethics review and participant risk.}
The human-baseline study was reviewed under the applicable institutional
procedure for minimal-risk human-participant annotation studies and was
determined to require no further IRB-style review. Participants were informed
before the study that the task involved viewing short front-facing driving clips
and making alert decisions. They provided consent, participation was voluntary,
and they could withdraw at any time without penalty. No sensitive personal data
were collected.

\clearpage

\section*{NeurIPS Paper Checklist}

\begin{enumerate}

\item {\bf Claims}
    \item[] Question: Do the main claims made in the abstract and introduction accurately reflect the paper's contributions and scope?
    \item[] Answer: \answerYes{}
    \item[] Justification: The abstract and introduction present DriveStreamBench as a driver-centered streaming benchmark for user-conditioned watch-and-notify. Sections~\ref{sec:task_def}--\ref{sec:metrics} and Section~\ref{sec:experiments} define the task, protocol, metrics, and experiments within this stated scope.

\item {\bf Limitations}
    \item[] Question: Does the paper discuss the limitations of the work performed by the authors?
    \item[] Answer: \answerYes{}
    \item[] Justification: Appendix~\ref{app:ethics_release}, Appendix~\ref{app:s6}, and Appendix~\ref{app:human_baseline} discuss limitations, including non-deployment scope, English-only instructions, front-camera input, low-resource S4, web-sourced S6, metadata-only release constraints, sampled audit limitations, and the video-clip nature of the human calibration row.

\item {\bf Theory assumptions and proofs}
    \item[] Question: For each theoretical result, does the paper provide the full set of assumptions and a complete (and correct) proof?
    \item[] Answer: \answerNA{}
    \item[] Justification: The paper does not make theoretical claims or prove new theorems.

\item {\bf Experimental result reproducibility}
    \item[] Question: Does the paper fully disclose all the information needed to reproduce the main experimental results of the paper to the extent that it affects the main claims and/or conclusions of the paper (regardless of whether the code and data are provided or not)?
    \item[] Answer: \answerYes{}
    \item[] Justification: Sections~\ref{sec:task_def}--\ref{sec:metrics} specify the task format, streaming protocol, output mapping, timing windows, scoring rules, and primary metric. Appendices~\ref{app:data_construction}, \ref{app:protocol}, \ref{app:latency}, \ref{app:fps-ablation}, \ref{app:ci}, and \ref{app:experiments} document construction, parsing, latency, frame-rate sensitivity, confidence intervals, and ablations.
    \item[] Guidelines:
    \begin{itemize}
        \item The answer \answerNA{} means that the paper does not include experiments.
        \item If the paper includes experiments, a \answerNo{} answer to this question will not be perceived well by the reviewers: Making the paper reproducible is important, regardless of whether the code and data are provided or not.
        \item If the contribution is a dataset and\slash or model, the authors should describe the steps taken to make their results reproducible or verifiable.
        \item Depending on the contribution, reproducibility can be accomplished in various ways. For example, if the contribution is a novel architecture, describing the architecture fully might suffice, or if the contribution is a specific model and empirical evaluation, it may be necessary to either make it possible for others to replicate the model with the same dataset, or provide access to the model. In general, releasing code and data is often one good way to accomplish this, but reproducibility can also be provided via detailed instructions for how to replicate the results, access to a hosted model (e.g., in the case of a large language model), releasing of a model checkpoint, or other means that are appropriate to the research performed.
        \item While NeurIPS does not require releasing code, the conference does require all submissions to provide some reasonable avenue for reproducibility, which may depend on the nature of the contribution. For example
        \begin{enumerate}
            \item If the contribution is primarily a new algorithm, the paper should make it clear how to reproduce that algorithm.
            \item If the contribution is primarily a new model architecture, the paper should describe the architecture clearly and fully.
            \item If the contribution is a new model (e.g., a large language model), then there should either be a way to access this model for reproducing the results or a way to reproduce the model (e.g., with an open-source dataset or instructions for how to construct the dataset).
            \item We recognize that reproducibility may be tricky in some cases, in which case authors are welcome to describe the particular way they provide for reproducibility. In the case of closed-source models, it may be that access to the model is limited in some way (e.g., to registered users), but it should be possible for other researchers to have some path to reproducing or verifying the results.
        \end{enumerate}
    \end{itemize}

\item {\bf Open access to data and code}
    \item[] Question: Does the paper provide open access to the data and code, with sufficient instructions to faithfully reproduce the main experimental results, as described in supplemental material?
    \item[] Answer: \answerYes{}
    \item[] Justification: Appendix~\ref{app:ethics_release} describes the released annotations, metadata, documentation, schemas, Croissant file, Responsible AI fields, evaluation code, training code, baseline scripts, and reproduction instructions. For web-sourced videos, the release provides source references, preprocessing utilities, and privacy-mask metadata rather than redistributing source visual media.

\item {\bf Experimental setting/details}
    \item[] Question: Does the paper specify all the training and test details (e.g., data splits, hyperparameters, how they were chosen, type of optimizer) necessary to understand the results?
    \item[] Answer: \answerYes{}
    \item[] Justification: Section~\ref{sec:experiments} and Appendices~\ref{app:data_construction}, \ref{app:data_stats}, \ref{app:protocol}, \ref{app:experiments}, and \ref{app:sft-infer-hparams} report data splits, L4 inference settings, scoring rules, output parsing, baseline adapters, optimizer, and SFT hyperparameters.

\item {\bf Experiment statistical significance}
    \item[] Question: Does the paper report error bars suitably and correctly defined or other appropriate information about the statistical significance of the experiments?
    \item[] Answer: \answerYes{}
    \item[] Justification: Appendix~\ref{app:ci} reports stratified bootstrap 95\% confidence intervals for \(\mathrm{CoreMacro}@2\mathrm{s}\), resampling positive and negative sessions independently within each L4-Core subcategory.
    \item[] Guidelines:
    \begin{itemize}
        \item The answer \answerNA{} means that the paper does not include experiments.
        \item The authors should answer \answerYes{} if the results are accompanied by error bars, confidence intervals, or statistical significance tests, at least for the experiments that support the main claims of the paper.
        \item The factors of variability that the error bars are capturing should be clearly stated (for example, train/test split, initialization, random drawing of some parameter, or overall run with given experimental conditions).
        \item The method for calculating the error bars should be explained (closed form formula, call to a library function, bootstrap, etc.).
        \item The assumptions made should be given (e.g., Normally distributed errors).
        \item It should be clear whether the error bar is the standard deviation or the standard error of the mean.
        \item It is OK to report 1-sigma error bars, but one should state it. The authors should preferably report a 2-sigma error bar than state that they have a 96\% CI, if the hypothesis of Normality of errors is not verified.
        \item For asymmetric distributions, the authors should be careful not to show in tables or figures symmetric error bars that would yield results that are out of range (e.g., negative error rates).
        \item If error bars are reported in tables or plots, the authors should explain in the text how they were calculated and reference the corresponding figures or tables in the text.
    \end{itemize}

\item {\bf Experiments compute resources}
    \item[] Question: For each experiment, does the paper provide sufficient information on the computer resources (type of compute workers, memory, time of execution) needed to reproduce the experiments?
    \item[] Answer: \answerYes{}
    \item[] Justification: Appendix~\ref{app:sft-infer-hparams} reports the GPU type, number of GPUs, training setup, inference backend, batch settings, tensor parallelism, and model-length settings. Appendix~\ref{app:latency} reports per-decision latency and real-time factors for open-source models and DriveStream-SFT.

\item {\bf Code of ethics}
    \item[] Question: Does the research conducted in the paper conform, in every respect, with the NeurIPS Code of Ethics \url{https://neurips.cc/public/EthicsGuidelines}?
    \item[] Answer: \answerYes{}
    \item[] Justification: Appendix~\ref{app:ethics_release} and Appendix~\ref{app:human_baseline} discuss privacy, licensing, release boundaries, participant risk, consent, withdrawal, compensation, takedown handling, unsupported uses, and misuse risks.
    \item[] Guidelines:
    \begin{itemize}
        \item The answer \answerNA{} means that the authors have not reviewed the NeurIPS Code of Ethics.
        \item If the authors answer \answerNo{}, they should explain the special circumstances that require a deviation from the Code of Ethics.
        \item The authors should make sure to preserve anonymity (e.g., if there is a special consideration due to laws or regulations in their jurisdiction).
    \end{itemize}

\item {\bf Broader impacts}
    \item[] Question: Does the paper discuss both potential positive societal impacts and negative societal impacts of the work performed?
    \item[] Answer: \answerYes{}
    \item[] Justification: Appendix~\ref{app:ethics_release} discusses potential benefits for evaluating driver-assistance alerting and potential risks from surveillance misuse, privacy exposure, source-media misuse, and overclaiming benchmark scores as real-world safety readiness.
    \item[] Guidelines:
    \begin{itemize}
        \item The answer \answerNA{} means that there is no societal impact of the work performed.
        \item If the authors answer \answerNA{} or \answerNo{}, they should explain why their work has no societal impact or why the paper does not address societal impact.
        \item Examples of negative societal impacts include potential malicious or unintended uses (e.g., disinformation, generating fake profiles, surveillance), fairness considerations (e.g., deployment of technologies that could make decisions that unfairly impact specific groups), privacy considerations, and security considerations.
        \item The conference expects that many papers will be foundational research and not tied to particular applications, let alone deployments. However, if there is a direct path to any negative applications, the authors should point it out. For example, it is legitimate to point out that an improvement in the quality of generative models could be used to generate Deepfakes for disinformation. On the other hand, it is not needed to point out that a generic algorithm for optimizing neural networks could enable people to train models that generate Deepfakes faster.
        \item The authors should consider possible harms that could arise when the technology is being used as intended and functioning correctly, harms that could arise when the technology is being used as intended but gives incorrect results, and harms following from (intentional or unintentional) misuse of the technology.
        \item If there are negative societal impacts, the authors could also discuss possible mitigation strategies (e.g., gated release of models, providing defenses in addition to attacks, mechanisms for monitoring misuse, mechanisms to monitor how a system learns from feedback over time, improving the efficiency and accessibility of ML).
    \end{itemize}

\item {\bf Safeguards}
    \item[] Question: Does the paper describe safeguards that have been put in place for responsible release of data or models that have a high risk for misuse (e.g., pre-trained language models, image generators, or scraped datasets)?
    \item[] Answer: \answerYes{}
    \item[] Justification: Appendix~\ref{app:ethics_release} describes a metadata-only release for source visual data, source-rights boundaries, local privacy redaction for faces, license plates, usernames, watermarks, subtitles, embedded text, S6 separation from the main leaderboard, takedown handling, and misuse mitigations.

\item {\bf Licenses for existing assets}
    \item[] Question: Are the creators or original owners of assets (e.g., code, data, models), used in the paper, properly credited and are the license and terms of use explicitly mentioned and properly respected?
    \item[] Answer: \answerYes{}
    \item[] Justification: Appendix~\ref{app:ethics_release} and the references credit the datasets, model families, and baselines used. The release applies CC BY 4.0 only to benchmark-created metadata, documentation, and released preprocessing utilities, does not redistribute nuPlan or web-sourced visual media, and requires users to obtain source media from original providers under the corresponding licenses, platform terms, and local laws.

\item {\bf New assets}
    \item[] Question: Are new assets introduced in the paper well documented and is the documentation provided alongside the assets?
    \item[] Answer: \answerYes{}
    \item[] Justification: Sections~\ref{sec:task_def}--\ref{sec:metrics} and Appendices~\ref{app:data_construction}, \ref{app:data_stats}, and \ref{app:ethics_release} document DriveStreamBench, annotations, metadata, evaluation harness, DriveStream-SFT, training scripts, taxonomy, prompts, splits, metrics, construction pipeline, Croissant metadata, Responsible AI fields, and release package.

\item {\bf Crowdsourcing and research with human subjects}
    \item[] Question: For crowdsourcing experiments and research with human subjects, does the paper include the full text of instructions given to participants and screenshots, if applicable, as well as details about compensation (if any)?
    \item[] Answer: \answerYes{}
    \item[] Justification: Appendix~\ref{app:human_baseline} reports the human baseline interface, instructions, trial assignment, scoring protocol, participant breakdown, consent and withdrawal procedure, minimal-risk clip-viewing procedure, and fixed non-performance-based compensation.

\item {\bf Institutional review board (IRB) approvals or equivalent for research with human subjects}
    \item[] Question: Does the paper describe potential risks incurred by study participants, whether such risks were disclosed to the subjects, and whether Institutional Review Board (IRB) approvals (or an equivalent approval/review based on the requirements of your country or institution) were obtained?
    \item[] Answer: \answerYes{}
    \item[] Justification: Appendix~\ref{app:human_baseline} describes the minimal-risk clip-viewing procedure, disclosed participant risk, consent and withdrawal terms, absence of sensitive personal-data collection, and the applicable institutional review status or equivalent determination.

\item {\bf Declaration of LLM usage}
    \item[] Question: Does the paper describe the usage of LLMs if it is an important, original, or non-standard component of the core methods in this research? Note that if the LLM is used only for writing, editing, or formatting purposes and does \emph{not} impact the core methodology, scientific rigor, or originality of the research, declaration is not required.
    \item[] Answer: \answerYes{}
    \item[] Justification: Appendix~\ref{app:l13_construction_splits} declares that VLMs are used during L1--L2 QA construction for candidate proposal and filtering under human-written templates. They are not used to produce final labels automatically, are not exposed to evaluated models, and the resulting QA pool is separately audited; LLMs were also used for writing, editing, and formatting assistance.

\end{enumerate}

\end{document}